\PassOptionsToPackage{unicode}{hyperref}
\PassOptionsToPackage{hyphens}{url}
\PassOptionsToPackage{dvipsnames,svgnames,x11names}{xcolor}
\documentclass[
]{krantz}
\usepackage{amsmath,amssymb}
\usepackage{lmodern}
\usepackage{iftex}
\ifPDFTeX
  \usepackage[T1]{fontenc}
  \usepackage[utf8]{inputenc}
  \usepackage{textcomp} 
\else 
  \usepackage{unicode-math}
  \defaultfontfeatures{Scale=MatchLowercase}
  \defaultfontfeatures[\rmfamily]{Ligatures=TeX,Scale=1}
\fi
\IfFileExists{upquote.sty}{\usepackage{upquote}}{}
\IfFileExists{microtype.sty}{
  \usepackage[]{microtype}
  \UseMicrotypeSet[protrusion]{basicmath} 
}{}
\makeatletter
\@ifundefined{KOMAClassName}{
  \IfFileExists{parskip.sty}{%
    \usepackage{parskip}
  }{
    \setlength{\parindent}{0pt}
    \setlength{\parskip}{6pt plus 2pt minus 1pt}}
}{
  \KOMAoptions{parskip=half}}
\makeatother
\usepackage{xcolor}
\usepackage{longtable,booktabs,array}
\usepackage{calc} 
\usepackage{etoolbox}
\makeatletter
\patchcmd\longtable{\par}{\if@noskipsec\mbox{}\fi\par}{}{}
\makeatother
\IfFileExists{footnotehyper.sty}{\usepackage{footnotehyper}}{\usepackage{footnote}}
\makesavenoteenv{longtable}
\usepackage{graphicx}
\makeatletter
\def\maxwidth{\ifdim\Gin@nat@width>\linewidth\linewidth\else\Gin@nat@width\fi}
\def\maxheight{\ifdim\Gin@nat@height>\textheight\textheight\else\Gin@nat@height\fi}
\makeatother
\setkeys{Gin}{width=\maxwidth,height=\maxheight,keepaspectratio}
\makeatletter
\def\fps@figure{htbp}
\makeatother
\providecommand{\tightlist}{%
  \setlength{\itemsep}{0pt}\setlength{\parskip}{0pt}}
\usepackage{booktabs}
\usepackage{longtable}
\usepackage[bf,singlelinecheck=off]{caption}
\usepackage{hyperref}

\usepackage{framed,color}
\definecolor{shadecolor}{RGB}{248,248,248}

\renewenvironment{quote}{\begin{VF}}{\end{VF}}

\makeatletter
 {\par\unskip\endMakeFramed%
 \at@end@of@kframe}
\makeatother

\usepackage{makeidx}
\makeindex

\usepackage{amsthm}
\makeatletter
\def\thm@space@setup{%
  \thm@preskip=8pt plus 2pt minus 4pt
  \thm@postskip=\thm@preskip
}
\makeatother

\frontmatter
\ifLuaTeX
  \usepackage{selnolig}  
\fi
\usepackage[]{natbib}
\IfFileExists{bookmark.sty}{\usepackage{bookmark}}{\usepackage{hyperref}}
\IfFileExists{xurl.sty}{\usepackage{xurl}}{} 
\hypersetup{
  pdftitle={Multimodal Deep Learning},
  colorlinks=true,
  linkcolor={Maroon},
  filecolor={Maroon},
  citecolor={Blue},
  urlcolor={Blue},
  pdfcreator={LaTeX via pandoc}}

\title{Multimodal Deep Learning}
\author{}
\date{\vspace{-2.5em}2023-01-03}

\begin{document}
\maketitle


\thispagestyle{empty}

\begin{center}
\end{center}

\setlength{\abovedisplayskip}{-5pt}
\setlength{\abovedisplayshortskip}{-5pt}

{
\hypersetup{linkcolor=}
\setcounter{tocdepth}{1}
\tableofcontents
}
\hypertarget{preface}{%
\chapter*{Preface}\label{preface}}

\emph{Author: Matthias Aßenmacher}

\begin{figure}

{\centering \includegraphics[width=\textwidth]{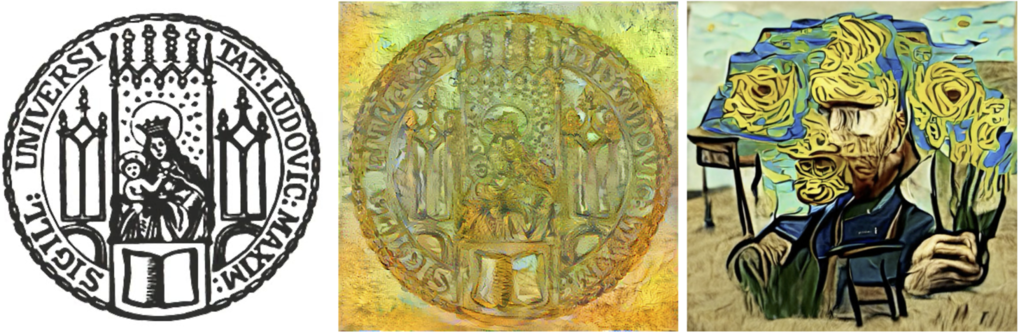}}

\caption{LMU seal (left) style-transferred to Van Gogh's Sunflower painting (center) and blended with the prompt - Van Gogh, sunflowers - via CLIP+VGAN (right).}\label{fig:cover}
\end{figure}

In the last few years, there have been several breakthroughs in the methodologies used in Natural Language Processing (NLP) as well as Computer Vision (CV). Beyond these improvements on single-modality models, large-scale multi-modal approaches have become a very active area of research.

In this seminar, we reviewed these approaches and attempted to create a solid overview of the field, starting with the current state-of-the-art approaches in the two subfields of Deep Learning individually. Further, modeling frameworks are discussed where one modality is transformed into the other Chapter \ref{c02-01-img2text} and Chapter \ref{c02-02-text2img}), as well as models in which one modality is utilized to enhance representation learning for the other (Chapter \ref{c02-03-img-support-text} and Chapter \ref{c02-04-text-support-img}). To conclude the second part, architectures with a focus on handling both modalities simultaneously are introduced (Chapter \ref{c02-05-text-plus-img}). Finally, we also cover other modalities (Chapter \ref{c03-01-further-modalities} and Chapter \ref{c03-02-structured-unstructured}) as well as general-purpose multi-modal models (Chapter \ref{c03-03-multi-purpose}), which are able to handle different tasks on different modalities within one unified architecture. One interesting application (Generative Art, Chapter \ref{c03-04-usecase}) eventually caps off this booklet.

\newpage

\begin{figure}
\centering
\includegraphics{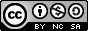}
\caption{Creative Commons License}
\end{figure}

This book is licensed under the \href{http://creativecommons.org/licenses/by-nc-sa/4.0/}{Creative Commons Attribution-NonCommercial-ShareAlike 4.0 International License}.

\mainmatter

\hypertarget{foreword}{%
\chapter*{Foreword}\label{foreword}}

\emph{Author: Matthias Aßenmacher}

This book is the result of an experiment in university teaching. We were inspired by a group of other PhD Students around Christoph Molnar, who conducted another \href{https://compstat-lmu.github.io/iml_methods_limitations/}{seminar on Interpretable Machine Learning} in this format.
Instead of letting every student work on a seminar paper, which more or less isolated from the other students, we wanted to foster collaboration between the students and enable them to produce a tangible outout (that isn't written to spend the rest of its time in (digital) drawers).
In the summer term 2022, some Statistics, Data Science and Computer Science students signed up for our seminar entitled ``Multimodal Deep Learning'' and had (before kick-off meeting) no idea what they had signed up for: Having written an entire book by the end of the semester.

We were bound by the examination rules for conducting the seminar, but otherwise we could deviate from the traditional format.
We deviated in several ways:

\begin{enumerate}
\def\labelenumi{\arabic{enumi}.}
\tightlist
\item
  Each student project is a chapter of this booklet, linked contentwise to other chapers since there's partly a large overlap between the topics.
\item
  We gave challenges to the students, instead of papers. The challenge was to investigate a specific impactful recent model or method from the field of NLP, Computer Vision or Multimodal Learning.
\item
  We designed the work to live beyond the seminar.
\item
  We emphasized collaboration. Students wrote the introduction to chapters in teams and reviewed each others individual texts.
\end{enumerate}

\hypertarget{technical-setup}{%
\section*{Technical Setup}\label{technical-setup}}

The book chapters are written in the Markdown language.
The simulations, data examples and visualizations were created with R \citep{rlang}.
To combine R-code and Markdown, we used rmarkdown.
The book was compiled with the bookdown package.
We collaborated using git and github.
For details, head over to the \href{https://github.com/slds-lmu/seminar_multimodal_dl}{book's repository}.

\hypertarget{introduction}{%
\chapter{Introduction}\label{introduction}}

\emph{Author: Nadja Sauter}

\emph{Supervisor: Matthias Aßenmacher}

\hypertarget{introduction-to-multimodal-deep-learning}{%
\section{Introduction to Multimodal Deep Learning}\label{introduction-to-multimodal-deep-learning}}

There are five basic human senses: hearing, touch, smell, taste and sight. Possessing these five modalities, we are able to perceive and understand the world around us. Thus, ``multimodal'' means to combine different channels of information simultaneously to understand our surroundings. For example, when toddlers learn the word ``cat'', they use different modalities by saying the word out loud, pointing on cats and making sounds like ``meow''. Using the human learning process as a role model, artificial intelligence (AI) researchers also try to combine different modalities to train deep learning models. On a superficial level, deep learning algorithms are based on a neural network that is trained to optimize some objective which is mathematically defined via the so-called loss function. The optimization, i.e.~minimizing the loss, is done via a numerical procedure called gradient descent. Consequently, deep learning models can only handle numeric input and can only result in a numeric output. However, in multimodal tasks we are often confronted with unstructured data like pictures or text. Thus, the first major problem is how to represent the input numerically. The second issue with regard to multimodal tasks is how exactly to combine different modalities. For instance, a typical task could be to train a deep learning model to generate a picture of a cat. First of all, the computer needs to understand the text input ``cat'' and then somehow translate this information into a specific image. Therefore, it is necessary to identify the contextual relationships between words in the text input and the spatial relationships betweent pixels in the image output. What might be easy for a toddler in pre-school, is a huge challenge for the computer. Both have to learn some understanding of the word ``cat'' that comprises the meaning and appearance of the animal. A common approach in modern deep learning is to generate embeddings that represent the cat numerically as a vector in some latent space. However, to achieve this, different approaches and algorithmic architectures have been developed in recent years. This book gives an overview of the different methods used in state-of-the-art (SOTA) multimodal deep learning to overcome challenges arising from unstructured data and combining inputs of different modalities.

\hypertarget{outline-of-the-booklet}{%
\section{Outline of the Booklet}\label{outline-of-the-booklet}}

Since multimodal models often use text and images as input or output, methods of Natural Language Processing (NLP) and Computer Vision (CV) are introduced as foundation in Chapter \ref{c01-00-intro-modalities}. Methods in the area of NLP try to handle text data, whereas CV deals with image processing. With regard to NLP (subsection \ref{c01-01-sota-nlp}), one concept of major importance is the so-called word embedding, which is nowadays an essential part of (nearly) all multimodal deep learning architectures. This concept also sets the foundation for transformer-based models like BERT \citep{BERT}, which achieved a huge improvement in several NLP tasks. Especially the (self-)attention mechanism \citep{attention} of transformers revolutionized NLP models, which is why most of them rely on the transformer as a backbone. In Computer Vision (subsection \ref{c01-02-sota-cv}) different network architectures, namely ResNet \citep{ResNet}, EfficientNet \citep{EfficientNet}, SimCLR \citep{SimCLR} and BYOL \citep{BYOL}, will be introduced. In both fields it is of great interest to compare the different approaches and their performance on challenging benchmarks. For this reason, the last subsection \ref{c01-03-benchmarks} of Chapter \ref{c01-00-intro-modalities} gives an overall overview of different data sets, pre-training tasks and benchmarks for CV as well as for NLP.

The second Chapter (see \ref{c02-00-multimodal}) focuses on different multimodal architectures, covering a wide variety of how text and images can be combined. The presented models combine and advance different methods of NLP and CV. First of all, looking at Img2Text tasks (subsection \ref{c02-01-img2text}), the data set Microsoft COCO for object recognition \citep{COCO} and the meshed-memory transformer for Image Captioning (M\textsuperscript{2} Transformer) \citep{meshed_memory} will be presented. Contrariwise, researchers developed methods to generate pictures based on a short text prompt (subsection \ref{c02-02-text2img}). The first models accomplishing this task were generative adversarial networks (GANs) \citep{GAN} and Variational Autoencoders (VAEs) \citep{VAE}. These methods were improved in recent years and today's SOTA transformer architectures and text-guided diffusion models like DALL-E \citep{DALLE} and GLIDE \citep{GLIDE} achieve remarkable results. Another interesting question is how images can be utilized to support language models (subsection \ref{c02-03-img-support-text}). This can be done via sequential embeddings, more advanced grounded embeddings or, again, inside transformers. On the other hand, one can also look at text supporting CV models like CLIP \citep{CLIP}, ALIGN \citep{ALIGN} and Florence \citep{yuan2021florence} (subsection \ref{c02-04-text-support-img}). They use foundation models meaning reusing models (e.g.~CLIP inside DALL-E 2) as well as a contrastive loss for connecting text with images. Besides, zero-shooting makes it possible to classify new and unseen data without expensive fine-tuning. Especially the open-source architecture CLIP \citep{CLIP} for image classification and generation attracted a lot of attention last year. In the end of the second chapter, some further architectures to handle text and images simultaneously are introduced (subsection \ref{c02-05-text-plus-img}). For instance, Data2Vec uses the same learning method for speech, vision and language and in this way aims to find a general approach to handle different modalities in one architecture. Furthermore, VilBert \citep{VilBert} extends the popular BERT architecture to handle both image and text as input by implementing co-attention. This method is also used in Google's Deepmind Flamingo \citep{alayrac2022flamingo}. In addition, Flamingo aims to tackle multiple tasks with a single visual language model via few-shot learning and freezing the pre-trained vision and language model.

In the last chapter (see \ref{c03-00-further}), methods are introduced that are also able to handle modalities other than text and image, like e.g.~video, speech or tabular data. The overall goal here is to find a general multimodal architecture based on challenges rather than modalities. Therefore, one needs to handle problems of multimodal fusion and alignment and decide whether you use a join or coordinated representation (subsection \ref{c03-01-further-modalities}). Moreover we go more into detail about how exactly to combine structured and unstructured data (subsection \ref{c03-02-structured-unstructured}). Therefore, different fusion strategies which evolved in recent years will be presented. This is illustrated in this book by two use cases in survival analysis and economics. Besides this, another interesting research question is how to tackle different tasks in one so called multi-purpose model (subsection \ref{c03-03-multi-purpose}) like it is intended to be created by Google researchers \citep{Pathways} in their ``Pathway'' model. Last but not least, we show one exemplary application of Multimodal Deep Learning in the arts scene where image generation models like DALL-E \citep{DALLE} are used to create art pieces in the area of Generative Arts (subsection \ref{c03-04-usecase}).

\hypertarget{c01-00-intro-modalities}{%
\chapter{Introducing the modalities}\label{c01-00-intro-modalities}}

\emph{Authors: Cem Akkus, Vladana Djakovic, Christopher Benjamin Marquardt}

\emph{Supervisor: Matthias Aßenmacher}

Natural Language Processing (NLP) has existed for about 50 years, but it is more relevant than ever. There have been several breakthroughs in this branch of machine learning that is concerned with spoken and written language. For example, learning internal representations of words was one of the greater advances of the last decade. Word embeddings (\citet{Mikolov2013}, \citet{Bojanowski2016}) made it possible and allowed developers to encode words as dense vectors that capture their underlying semantic content. In this way, similar words are embedded close to each other in a lower-dimensional feature space. Another important challenge was solved by Encoder-decoder (also called sequence-to-sequence) architectures \citet{Sutskever2014}, which made it possible to map input sequences to output sequences of different lengths. They are especially useful for complex tasks like machine translation, video captioning or question answering. This approach makes minimal assumptions on the sequence structure and can deal with different word orders and active, as well as passive voice.

A definitely significant state-of-the-art technique is Attention \citet{Bahdanau2014}, which enables models to actively shift their focus -- just like humans do. It allows following one thought at a time while suppressing information irrelevant to the task. As a consequence, it has been shown to significantly improve performance for tasks like machine translation. By giving the decoder access to directly look at the source, the bottleneck is avoided and at the same time, it provides a shortcut to faraway states and thus helps with the vanishing gradient problem. One of the most recent sequence data modeling techniques is Transformers (\citet{vaswani2017attention}), which are solely based on attention and do not have to process the input data sequentially (like RNNs). Therefore, the deep learning model is better in remembering context-induced earlier in long sequences. It is the dominant paradigm in NLP currently and even makes better use of GPUs, because it can perform parallel operations. Transformer architectures like BERT \citep{Devlin2018}, T5 \citep{Raffel2019} or GPT-3 \citep{brown2020language} are pre-trained on a large corpus and can be fine-tuned for specific language tasks. They have the capability to generate stories, poems, code and much more. With the help of the aforementioned breakthroughs, deep networks have been successful in retrieving information and finding representations of semantics in the modality text. In the next paragraphs, developments for another modality image are going to be presented.

Computer vision (CV) focuses on replicating parts of the complexity of the human visual system and enabling computers to identify and process objects in images and videos in the same way that humans do. In recent years it has become one of the main and widely applied fields of computer science. However, there are still problems that are current research topics, whose solutions depend on the research's view on the topic. One of the problems is how to optimize deep convolutional neural networks for image classification. The accuracy of classification depends on width, depth and image resolution. One way to address the degradation of training accuracy is by introducing a deep residual learning framework \citep{ResNet}. On the other hand, another less common method is to scale up ConvNets, to achieve better accuracy is by scaling up image resolution. Based on this observation, there was proposed a simple yet effective compound scaling method, called EfficientNets \citep{EfficientNet}.

Another state-of-the-art trend in computer vision is learning effective visual representations without human supervision. Discriminative approaches based on contrastive learning in the latent space have recently shown great promise, achieving state-of-the-art results, but the simple framework for contrastive learning of visual representations, which is called SimCLR, outperforms previous work \citep{SimCLR}. However, another research proposes as an alternative a simple ``swapped'' prediction problem where we predict the code of a view from the representation of another view. Where features are learned by Swapping Assignments between multiple Views of the same image (SwAV) \citep{SwAV}.
Further recent contrastive methods are trained by reducing the distance between representations of different augmented views of the same image (`positive pairs') and increasing the distance between representations of augmented views from different images (`negative pairs'). Bootstrap Your Own Latent (BYOL) is a new algorithm for self-supervised learning of image representatios \citep{BYOL}.

Self-attention-based architectures, in particular, Transformers have become the model of choice in natural language processing (NLP). Inspired by NLP successes, multiple works try combining CNN-like architectures with self-attention, some replacing the convolutions entirely. The latter models, while theoretically efficient, have not yet been scaled effectively on modern hardware accelerators due to the use of specialized attention patterns. Inspired by the Transformer scaling successes in NLP, one of the experiments is applying a standard Transformer directly to the image \citep{ImageT}. Due to the widespread application of computer vision, these problems differ and are constantly being at the center of attention of more and more research.

With the rapid development in NLP and CV in recent years, it was just a question of time to merge both modalities to tackle multi-modal tasks. The release of DALL-E 2 just hints at what one can expect from this merge in the future. DALL-E 2 is able to create photorealistic images or even art from any given text input. So it takes the information of one modality and turns it into another modality. It needs multi-modal datasets to make this possible, which are still relatively rare. This shows the importance of available data and the ability to use it even more. Nevertheless, all modalities are in need of huge datasets to pre-train their models. It's common to pre-train a model and fine-tune it afterwards for a specific task on another dataset. For example, every state-of-the-art CV model uses a classifier pre-trained on an ImageNet based dataset. The cardinality of the datasets used for CV is immense, but the datasets used for NLP are of a completely different magnitude. BERT uses the English Wikipedia and the Bookscorpus to pre-train the model. The latter consists of almost 1 billion words and 74 million sentences. The pre-training of GPT-3 is composed of five huge corpora: CommonCrawl, Books1 and Books2, Wikipedia and WebText2. Unlike language model pre-training that can leverage tremendous natural language data, vision-language tasks require high-quality image descriptions that are hard to obtain for free. Widely used pre-training datasets for VL-PTM are Microsoft Common Objects in Context (COCO), Visual Genome (VG), Conceptual Captions (CC), Flickr30k, LAION-400M and LAION-5B, which is now the biggest openly accessible image-text dataset.

Besides the importance of pre-training data, there must also be a way to test or compare the different models. A reasonable approach is to compare the performance on specific tasks, which is called benchmarking. A nice feature of benchmarks is that they allow us to compare the models to a human baseline. Different metrics are used to compare the performance of the models. Accuracy is widely used, but there are also some others. For CV the most common benchmark datasets are ImageNet, ImageNetReaL, CIFAR-10(0), OXFORD-IIIT PET, OXFORD Flower 102, COCO and Visual Task Adaptation Benchmark (VTAB). The most common benchmarks for NLP are General Language Understanding Evaluation (GLUE), SuperGLUE, SQuAD 1.1, SQuAD 2.0, SWAG, RACE, ReCoRD, and CoNLL-2003. VTAB, GLUE and SuperGLUE also provide a public leader board. Cross-modal tasks such as Visual Question Answering (VQA), Visual Commonsense Reasoning (VCR), Natural Language Visual Reasoning (NLVR), Flickr30K, COCO and Visual Entailment are common benchmarks for VL-PTM.

\hypertarget{c01-01-sota-nlp}{%
\section{State-of-the-art in NLP}\label{c01-01-sota-nlp}}

\emph{Author: Cem Akkus}

\emph{Supervisor: Matthias Aßenmacher}

\hypertarget{introduction-1}{%
\subsection{Introduction}\label{introduction-1}}

Natural Language Processing (NLP) exists for about 50 years, but it is
more relevant than ever. There have been several breakthroughs in this
branch of machine learning that is concerned with spoken and written
language. In this work, the most influential ones of the last decade are
going to be presented. Starting with word embeddings, which efficiently
model word semantics. Encoder-decoder architectures represent another
step forward by making minimal assumptions about the sequence structure.
Next, the attention mechanism allows human-like focus shifting to put
more emphasis on more relevant parts. Then, the transformer applies
attention in its architecture to process the data non-sequentially,
which boosts the performance on language tasks to exceptional levels. At
last, the most influential transformer architectures are recognized
before a few current topics in natural language processing are
discussed.

\hypertarget{word-embeddings}{%
\subsection{Word Embeddings}\label{word-embeddings}}

As mentioned in the introduction, one of the earlier advances in NLP is
learning word internal representations. Before that, a big problem with
text modelling was its messiness, while machine learning algorithms
undoubtedly prefer structured and well-defined fixed-length inputs. On a
granular level, the models rather work with numerical than textual data.
Thus, by using very basic techniques like one-hot encoding or
bag-of-words, a text is converted into its equivalent vector of numbers
without losing information.\\

\begin{figure}

{\centering \includegraphics[width=0.7\linewidth]{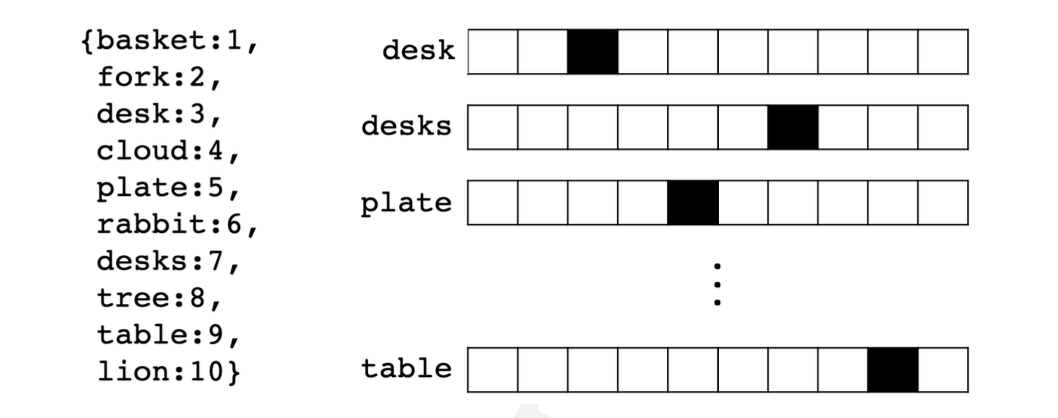}

}

\caption{Ten one-hot encoded words (Source: \citet{Pilehvar2021})}\label{fig:onehot}
\end{figure}

In the example depicting one-hot encoding (see Figure \ref{fig:onehot}), there
are ten simple words and the dark squares indicate the only index with a
non-zero value.

In contrast, there are multiple non-zero values while using
bag-of-words, which is another way of extracting features from text to
use in modelling where we measure if a word is present from a vocabulary
of known words. It is called bag-of-words because the order is
disregarded here.\\

Treating words as atomic units has some plausible reasons, like
robustness and simplicity. It was even argued that simple models on a
huge amount of data outperform complex models trained on less data.
However, simple techniques are problematic for many tasks, e.g.~when it
comes to relevant in-domain data for automatic speech recognition. The
size of high-quality transcribed speech data is often limited to just
millions of words, so simply scaling up simpler models is not possible
in certain situations and therefore more advanced techniques are needed.
Additionally, thanks to the progress of machine learning techniques, it
is realistic to train more complex models on massive amounts of data.
Logically, more complex models generally outperform basic ones. Other
disadvantages of classic word representations are described by the curse
of dimensionality and the generalization problem. The former becomes a
problem due to the growing vocabulary equivalently increasing the
feature size. This results in sparse and high-dimensional vectors. The
latter occurs because the similarity between words is not captured.
Therefore, previously learned information cannot be used. Besides,
assigning a distinct vector to each word is a limitation, which becomes
especially obvious for languages with large vocabularies and many rare
words.\\

To combat the downfalls of simple word representations, word embeddings
enable to use efficient and dense representations in which similar words
have a similar encoding. So words that are closer in the vector space
are expected to be similar in meaning. An embedding is hereby defined as
a vector of floating point values (with the length of the vector being a
hyperparameter). The values for the embedding are trainable parameters
which are learned similarly to a model learning the weights for a dense
layer. The dimensionality of the word representations is typically much
smaller than the number of words in the dictionary. For example,
\citet{Mikolov2013} called dimensions between 50-100 modest for more than a
few hundred million words. For small data sets, dimensionality for the
word vectors could start at 8 and go up to 1024 for larger data sets. It
is expected that higher dimensions can rather pick up intricate
relationships between words if given enough data to learn from.

\begin{figure}

{\centering \includegraphics[width=0.7\linewidth]{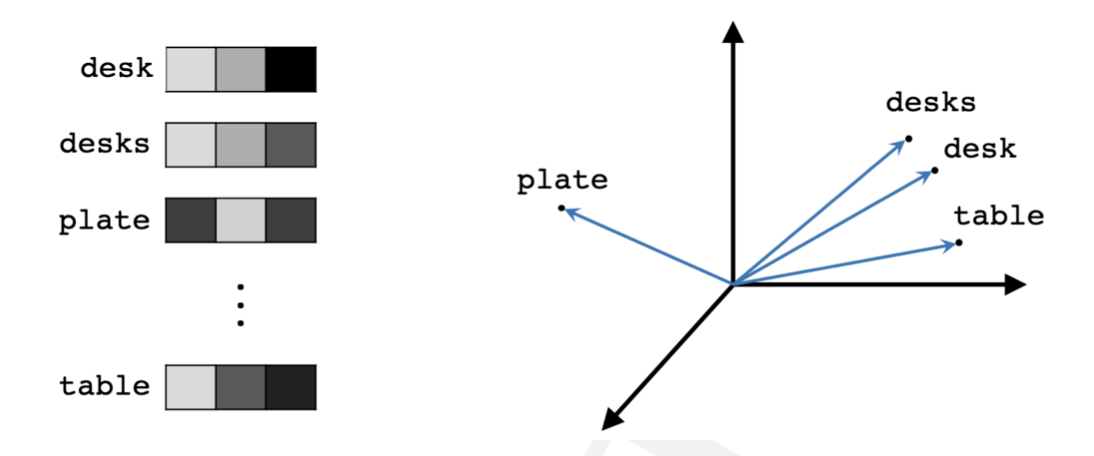}

}

\caption{Three-dimensional word embeddings (Source: \citet{Pilehvar2021}).}\label{fig:embedPilehvarP11}
\end{figure}

For any NLP tasks, it is sensible to start with word embeddings because
it allows to conveniently incorporate prior knowledge into the model and
can be seen as a basic form of transfer learning. It is important to
note that even though embeddings attempt to represent the meaning of
words and do that to an extent, the semantics of the word in a given
context cannot be captured. This is due to the words having static
precomputed representations in traditional embedding techniques. Thus,
the word "bank" can either refer to a financial institution or a river
bank. Contextual embedding methods offer a solution, but more about them
will follow later.\\

\begin{figure}

{\centering \includegraphics[width=1\linewidth]{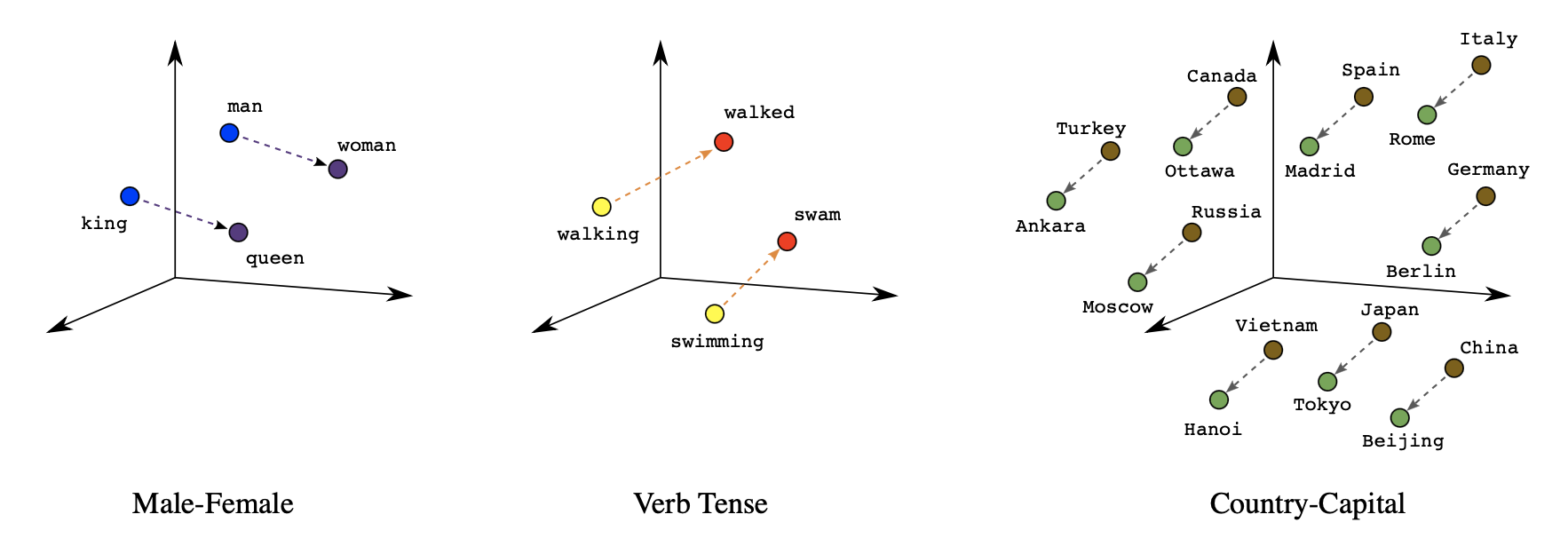}

}

\caption{Three types of similarities as word embeddings (Source: \citet{Google2022}).}\label{fig:embedVectors}
\end{figure}

It should be noted that words can have various degrees of similarity. In
the context of inflectional languages, it becomes obvious because words
are adjusted to articulate grammatical categories. For example, in a
subspace of the original vector, nouns that have similar endings can be
found. However, it even exceeds simple syntactic regularities. With
straightforward operations on the word vectors, it can be displayed that
\(vector(\text{King}) - vector(\text{Man}) + vector(\text{Woman})\) equals a vector that
is closest in vector space (and therefore in meaning) to the word
"Queen". A simple visualization of this relationship can be seen in
the left graph below (see Figure \ref{fig:embedVectors}). The three coordinate systems are
representations of higher dimensions that are depicted in this way via
dimension reduction techniques. Furthermore, the verb-to-tense
relationship is expressed in the middle graphic, which extends the
insight from before referring to the word endings being similar because
in this instance the past tenses of both verbs walking and swimming are
not similar in structure. Additionally, on the right side of the figure,
there is a form of the commonly portrayed and easily understood
Country-Capital example (see \citet{Mikolov2013}).

Another way of using vector representations of words is in the field of
translations. It has been presented that relations can be drawn from
feature spaces of different languages. In below, the distributed word
representations of numbers between English and Spanish are compared. In
this case, the same numbers have similar geometric arrangements, which
suggests that mapping linearly between vector spaces of languages is
feasible. Applying this simple method for a larger set of translations
in English and Spanish led to remarkable results - achieving almost 90 \%
precision.

\begin{figure}

{\centering \includegraphics[width=0.9\linewidth]{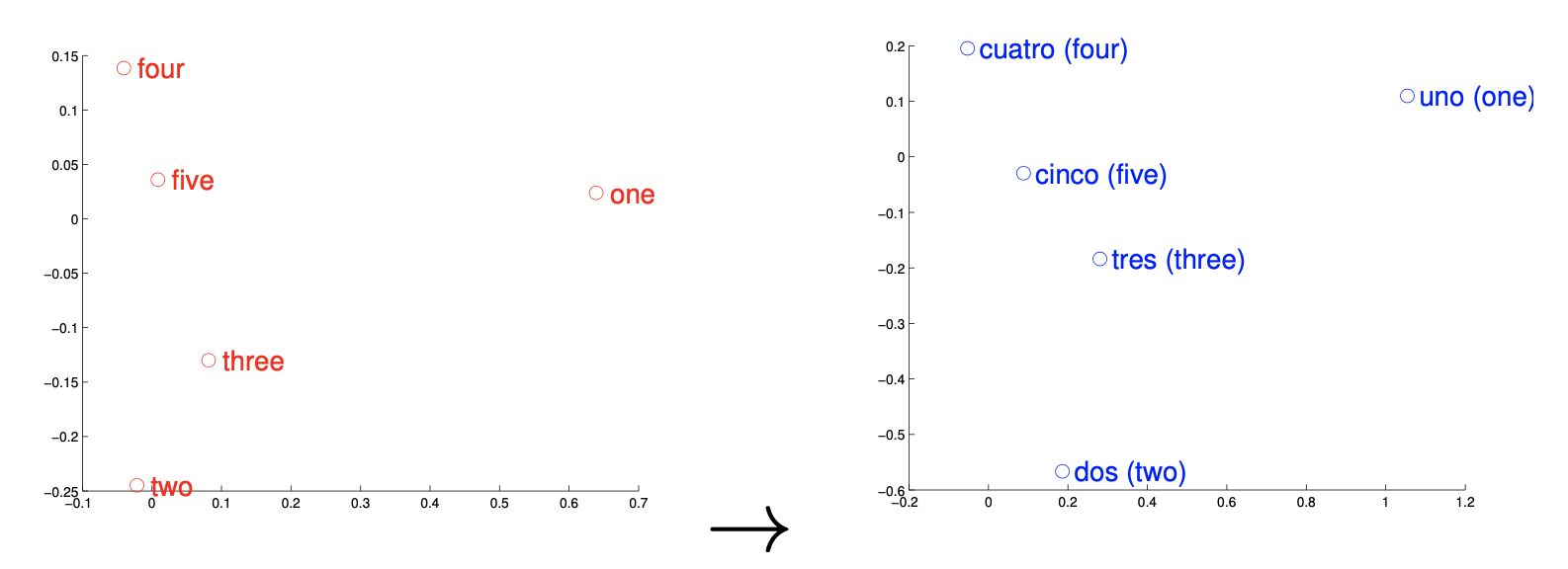}

}

\caption{Representations of numbers in English and Spanish (Source: \citet{Mikolov2013a}).}\label{fig:embedTransl}
\end{figure}

This technique was then used for other experiments. One use case is the
detection of dictionary errors. Taking translations from a dictionary
and computing their geometric distance returns a confidence measure.
Closely evaluating the translations with low confidence and outputting
an alternative (one that is closest in vector space) results in a plain
way to assess dictionary translations. Furthermore, training the word
embeddings on a large corpora makes it possible to give sensible
out-of-dictionary predictions for words. This was tested by randomly
removing a part of the vocabulary before. Taking a look at the
predictions revealed that they were often to some extent related to the
translations with regard to meaning and semantics. Despite the
accomplishments in other tasks, translations between distant languages
exposed shortcomings of word embeddings. For example, the accuracy for
translations between English and Vietnamese seemed significantly lower.
This can be ascribed to both languages not having a good one-to-one
correspondence because the concept of a word is different than in
English. In addition, the used Vietnamese model contains numerous
synonyms, which complicates making exact predictions (see
\citet{Mikolov2013a}).\\

\begin{figure}

{\centering \includegraphics[width=0.85\linewidth]{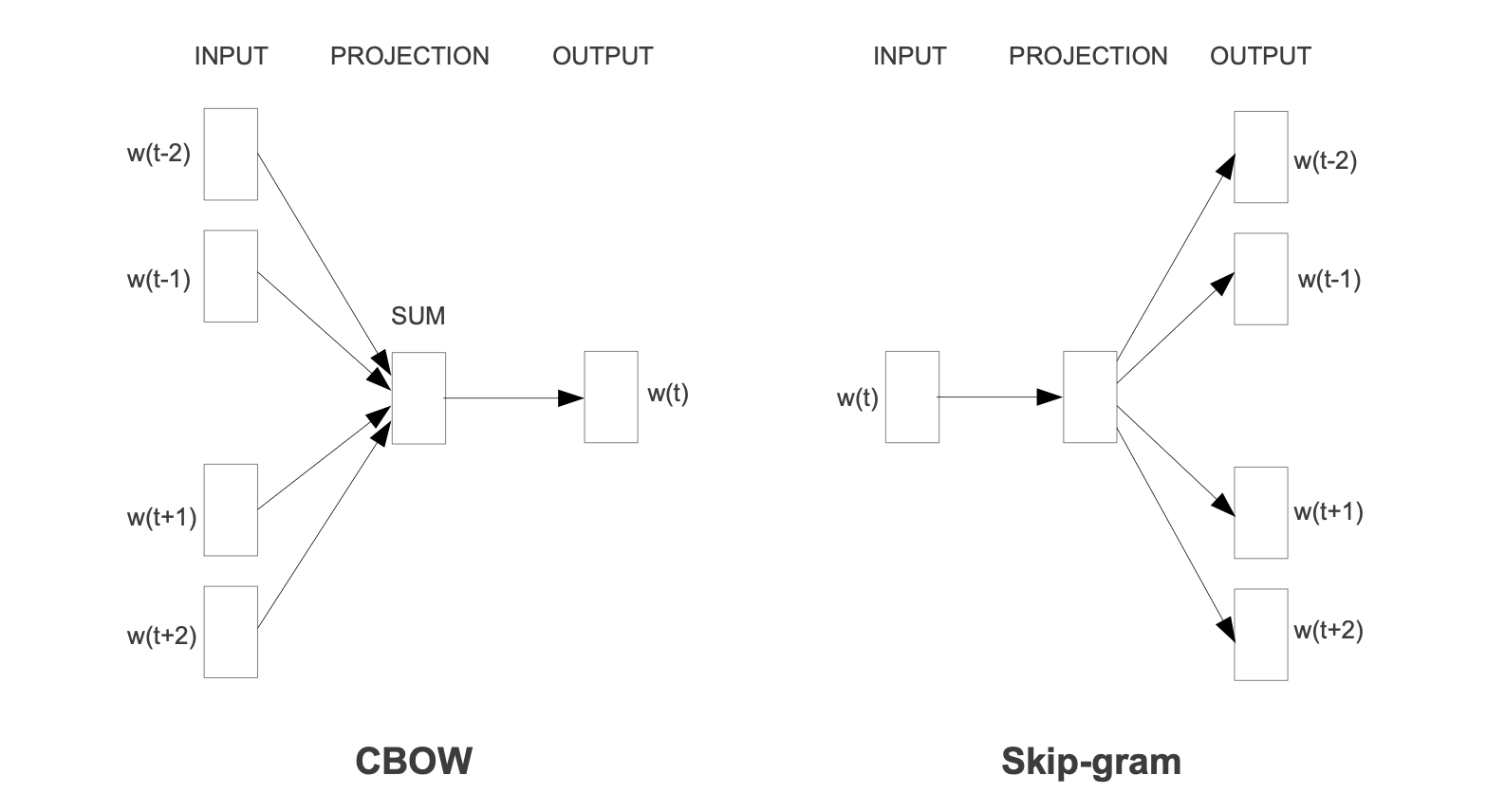}

}

\caption{CBOW and Skip-gram architecture (Source: \citet{Mikolov2013}).}\label{fig:embedArch}
\end{figure}

Turning the attention to one of the most impactful embedding techniques,
word2vec. It was proposed by \citet{Mikolov2013} and is not a singular
algorithm. It can rather be seen as a family of model architectures and
optimizations to learn word representations. Word2vec's popularity also
stems from its success on multiple downstream natural language
processing tasks. It has a very simple structure which is based on a
basic feed forward neural network. They published multiple papers (see
\citet{Mikolov2013}{]}, \citet{Mikolov2013a}, \citet{Mikolov2013b}) that are stemming
around two different but related methods for learning word embeddings
(see Figure \ref{fig:embedArch}). Firstly, the Continuous bag-of-words model aims to predict the
middle word based on surrounding context words. Hence, it considers
components before and after the target word. As the order of words in
the context is not relevant, it is called a bag-of-words model.
Secondly, the Continuous skip-gram model only considers the current word
and predicts others within a range before and after it in the same
sentence. Both of the models use a softmax classifier for the output
layer.

Then, \citet{Bojanowski2016} built on skip-gram models by accounting for the
morphology (internal structure) of words. A different classical
embedding architecture that has to be at least mentioned is the GloVe
model, which does not use a neural network but incorporates local
context information with global co-occurrence statistics.

\hypertarget{encoder-decoder}{%
\subsection{Encoder-Decoder}\label{encoder-decoder}}

\begin{figure}

{\centering \includegraphics[width=0.85\linewidth]{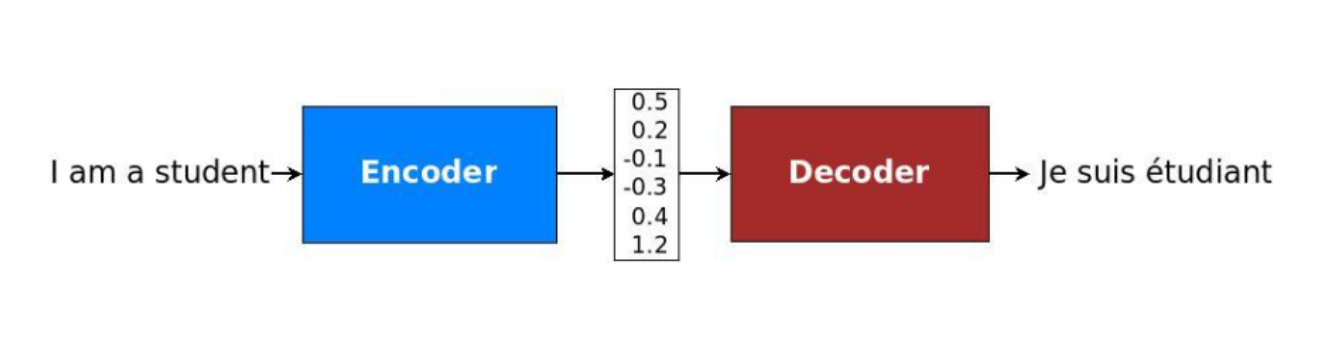}

}

\caption{Translation through simplified seq2seq model (Source: \citet{Manning2022}).}\label{fig:arch1Encdec}
\end{figure}

The field of natural language processing is concerned with a variety of
different tasks surrounding text. Depending on the type of NLP problem,
the network may be confronted with variable length sequences as input
and/or output. This is the case for many compelling applications, such
as question answering, dialogue systems or machine translation. In the
following, many examples will explore machine translations in more
detail, since it is a major problem domain. Regarding translation tasks,
it becomes obvious that input sequences need to be mapped to output
sequences of different lengths. To manage this type of input and output,
a design with two main parts could be useful. The first one is called
the encoder because, in this part of the network, a variable length
input sequence is transformed into a fixed state. Next, the
second component called the decoder maps the encoded state to an output
of a variable length sequence. As a whole, it is known as an
encoder-decoder or sequence-to-sequence architecture and has become an
effective and standard approach for many applications which even
recurrent neural networks with gated hidden units have trouble solving
successfully. Deep RNNs may have a chance, but different architectures
like encoder-decoder have proven to be the most effective. It can even
deal with different word orders and active, as well as passive voice
\citep{Sutskever2014}. A simplified example of the encoder-decoder model
can be seen in \ref{fig:arch1Encdec}.

Before going through the equations quantifying the concepts, it makes
sense to examine the sequence-to-sequence design
proposed by \citet{Cho2014}. An encoder-RNN processes the input sequence of
length \(n_x\) and computes a fixed-length context vector \(C\), which is
usually the final hidden state of the encoder or a simple function of
the hidden states. After the input sequence is processed, it is added to
the hidden state and passed forward in time through the recurrent
connections between the hidden states in the encoder. Despite the
context vector usually being a simple function of the last hidden state,
its role cannot be underestimated. Specifically, the encoded state
summarizes important information from the input sequence, e.g.~the
intent in a question answering task or the meaning of a text in the case
of machine translation. After the context is passed to every hidden
state of the decoder, the decoder RNN uses this information to produce
the target sequence of length \(n_y\), which can of course vary from
\(n_x\).

\begin{figure}

{\centering \includegraphics[width=0.56\linewidth]{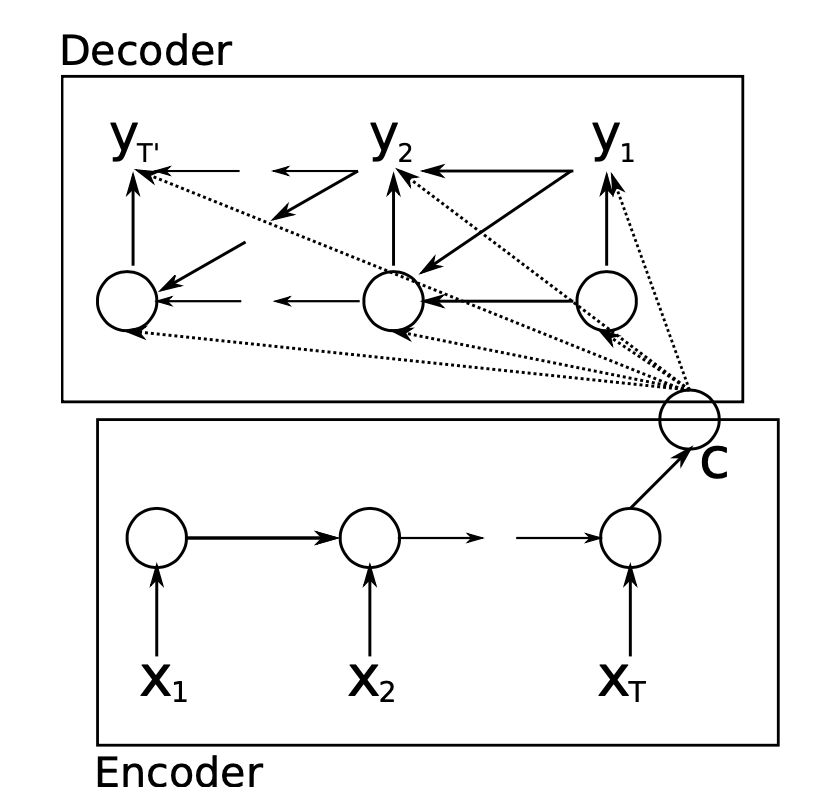}

}

\caption{Encoder-decoder architecture (Source: \citet{Cho2014}).}\label{fig:encdecArchCho}
\end{figure}

At the latest through the above illustration, it is clear that the
decoder is particularly interesting to look at in the form of equations.
The notation mainly follows \citet{Cho2014}. The decoder is another type of
RNN which is trained to predict the target based on the hidden state at
the last time step. However, unlike regular RNNs, it is also conditioned
on the output of the last time step (\(y_{t-1}\)) and a summary of the
input c.~Therefore, the hidden state of the decoder is computed by:

\[h_d^{[t]} = f(h_d^{[t-1]},y^{[t-1]},c).
    \label{eqn:h_dec}\]

Similarly, each conditional probability is given by the following, where
\(f\) is a non-linear activation function (and must produce probabilities in
, e.g.~the softmax function):

\[P(y^{[t]}|y^{[1]}, \dots ,y^{[t-1]},c) = f(h_d^{[t]}, y^{[t-1]}, c).     \label{eqn:P_dec}\]

The two parts are jointly trained to maximize the conditional
log-likelihood, where \(\theta\) denotes the set of model parameters and
\((x_n, y_n)\) is an (input sequence, output sequence) pair from the
training set with size \(N\):

\[\max_\theta \frac{1}{N} \displaystyle \sum_{n=1}^{N} \log p_{\theta}(y_n|x_n).
    \label{eqn:train_dec}\]

The best probability is usually found by using the beam search
algorithm. The core idea of it is that on each step of the decoder, we
keep track of the \(k\) most probable partial translations (which are
called hypotheses).\\

\begin{figure}

{\centering \includegraphics[width=0.7\linewidth]{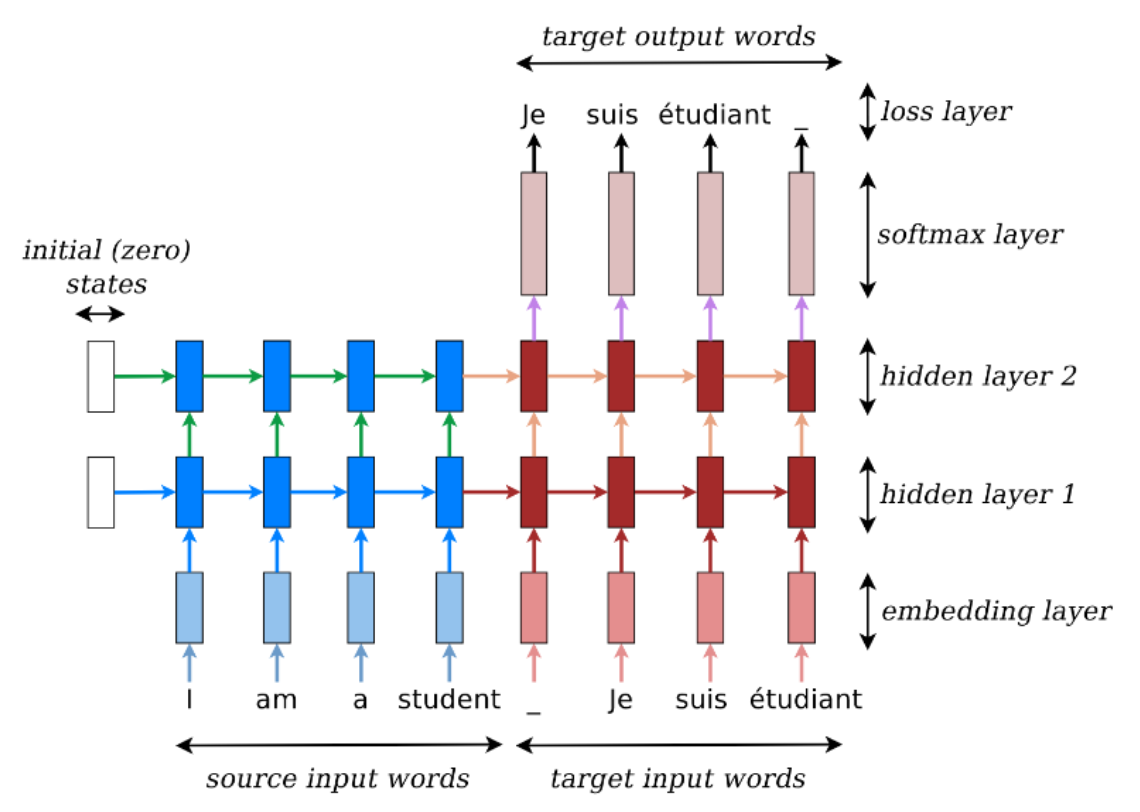}

}

\caption{Translation through seq2seq model (Source: \citet{Manning2022}).}\label{fig:encdecArch2}
\end{figure}

Examining the translation presented in with hidden units unrolled
through time could look like in \ref{fig:encdecArch2}. In particular, multiple hidden layers
are recommended by the researchers. The idea is that lower layers
compute lower-level features and higher layers compute higher-level
features.

Gated recurrent networks, especially long short-term memory networks,
have been found to be effective in both components of the
sequence-to-sequence architecture. Furthermore, it was revealed that
deep LSTMs significantly outperform shallow LSTMs. Each additional layer
reduced perplexity by nearly 10\%, possibly due to their much larger
hidden state. For example, \citet{Sutskever2014} used deep LSTMs with 4
layers and 1000 cells at each layer for 1000-dimensional word
embeddings. Thus, in total, 8000 real numbers are used to represent a
sentence. For simplification, the neural networks are in the following
referred to as RNNs which is not contradicting the insights of this
paragraph as LSTMs are a type of gated RNNS \citep{Sutskever2014}.

\hypertarget{attention}{%
\subsection{Attention}\label{attention}}

Although encoder-decoder architectures simplified dealing with variable
length sequences, they also caused complications. Due to their design, the
encoding of the source sentence is a single vector representation
(context vector). The problem is that this state must compress all
information about the source sentence in a single vector and is commonly
referred to as the bottleneck problem. To be precise, the entire
semantics of arbitrarily long sentences need to be wrapped into a single
hidden state. Moreover, it constitutes a different learning problem
because the information needs to be passed between numerous time steps.
This leads to vanishing gradients within the network as a consequence of
factors less than 1 multiplied with each other at every point. To
illustrate, the last sentence is an ideal example of one in which an
encoder-decoder approach could have difficulty coping. In particular, if
the sentences are longer than the ones in the training corpus
\citep{Manning2022}.\\

Due to the aforementioned reasons, an extension to the
sequence-to-sequence architecture was proposed by \citet{Bahdanau2014}, which
learns to align and translate jointly. For every generated word, the
model scans through some positions in the source sentence where the most
relevant information is located. Afterwards, based on the context around
and the previously generated words, the model predicts the target word
for the current time step. This approach is called attention, as it
emulates human-like (cognitive) attention. As a result of directly
looking at the source and bypassing the bottleneck, it provides a
solution to the problem. Then, it mitigates the vanishing gradient
problem, since there is now a shortcut to faraway states. Consequently,
incorporating the attention mechanism has been shown to considerably
boost the performance of models on NLP tasks.\\

A walkthrough of the example below should resolve any outstanding
questions regarding the procedure of the attention mechanism. The source
sentence is seen on the bottom left, which is given in French and acts
as the input for the encoder-RNN (in red). Then, the attention scores
(in blue) are computed by taking the dot product between the previous
output word and input words. Next, the softmax function turns the scores
into a probability distribution (in pink). They are used to take a
weighted sum of the encoder's hidden states and form the attention
output, which mostly contains information from the hidden states that
received high attention. Afterwards, the attention output is
concatenated with the decoder hidden state (in green), which is applied
to compute the decoder output as before. In some scenarios, the
attention output is also fed into the decoder (along with the usual
decoder input). This specific example was chosen because "entarter"
means "to hit someone with a pie" and is therefore a word that needs
to be translated with many words. As a consequence of no existing direct
equivalents for this phrase, it is expected that there is not only one
nearly non-zero score. In this snapshot, the attention distribution can
be seen to have two significant contributors.

\begin{figure}

{\centering \includegraphics[width=0.9\linewidth]{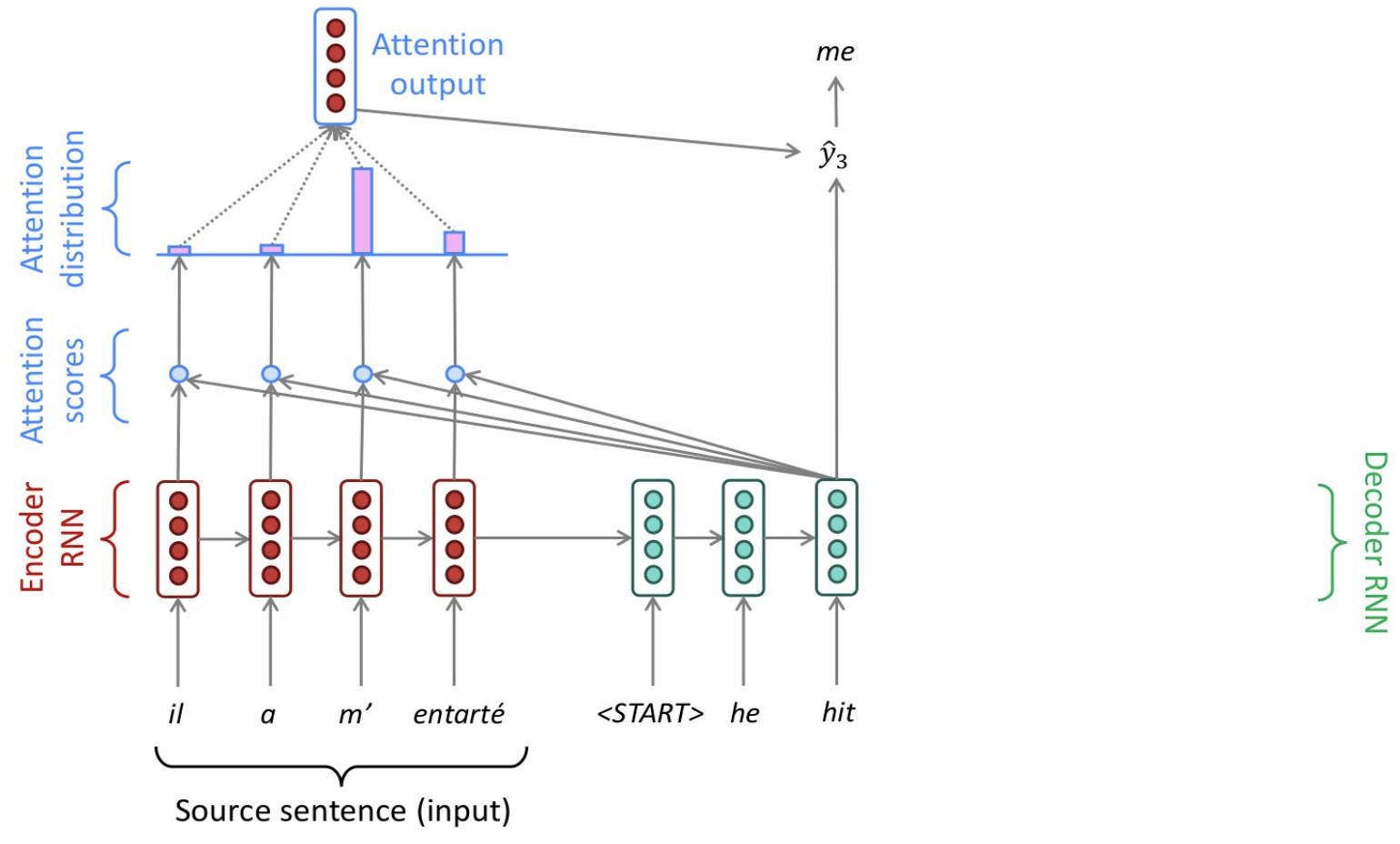}

}

\caption{Translation process with attention mechanism (Source: \citet{Manning2022}).}\label{fig:attentionExStanford}
\end{figure}

The following equations aim to compactly represent the relations brought
forward in the last paragraphs and mainly follow \citet{Manning2022}. The
attention scores \(e^{[t]}\) are computed by scalarly combining the hidden
state of the decoder with all of the hidden states of the encoder:

\[e^{[t]} = [(h_{d}^{[t]})^T h_{e}^{[1]}, \ldots  , (h_{d}^{[t]})^T h_{e}^{[N]} ].\]

Besides the basic dot-product attention, there are also other ways to
calculate the attention scores, e.g.~through multiplicative or additive
attention. Although they will not be further discussed at this point, it
makes sense to at least mention them. Then, applying the softmax to the
scalar scores results in the attention distribution \(\alpha^{[t]}\), a
probability distribution whose values sum up to 1:

\[\alpha^{[t]} = softmax(e^{[t]}).\]

Next, the attention output \(a^{[t]}\) is obtained by the attention
distribution acting as a weight for the encoder hidden states:

\[a^{[t]} = \sum_{i=1}^{N} \alpha_i^{[t]} h_{e,i}.\]

Concatenating attention output with decoder hidden state and proceeding
as in the non-attention sequence-to-sequence model are the final steps:

\[o^{[t]} = f(a^{[t]} h_d^{[t]}).\]

By visualizing the attention distribution, also called alignments (see
\citet{Bahdanau2014}), it is easy to observe what the decoder was focusing on
and understand why it chose a specific translation. The x-axis of the
plot of below corresponds to the words in the source sentence (English)
and the y-axis to the words in the generated translation (French). Each
pixel shows the weight of the source word for the respective target word
in grayscale, where 0 is black and 1 is white. As a result, which
positions in the source sentence were more relevant when generating the
target word becomes apparent. As expected, the alignment between English
and French is largely monotonic, as the pixels are brighter, and
therefore the weights are higher along the main diagonal of the matrix.
However, there is an exception because adjectives and nouns are
typically ordered differently between the two languages. Thus, the model
(correctly) translated "European Economic Area" into "zone économique
européene". By jumping over two words ("European" and "Economic"),
it aligned "zone" with "area". Then, it looked one word back twice
to perfect the phrase "zone économique européene". Additional
qualitative analysis has shown that the model alignments are
predominantly analogous to our intuition.\\

\begin{figure}

{\centering \includegraphics[width=0.5\linewidth]{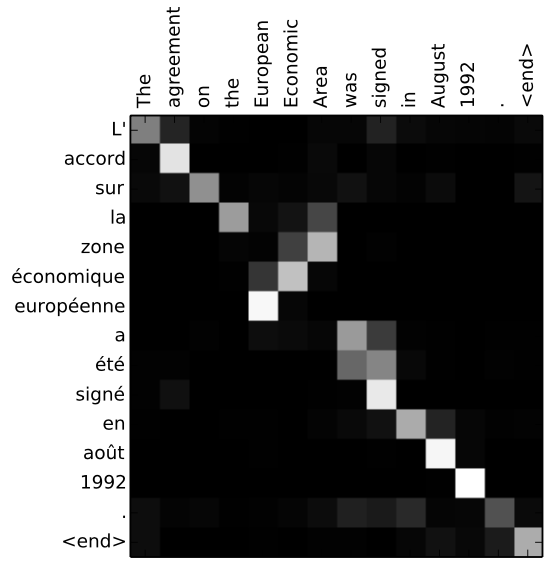}

}

\caption{Attention alignments (Source: \citet{Bahdanau2014}).}\label{fig:attentionfocus}
\end{figure}

\hypertarget{transformer}{%
\subsection{Transformer}\label{transformer}}

For this section, \citet{Manning2022} constitutes the main source.\\

RNNs are unrolled from one side to the other. Thus, from left to right
and right to left. This encodes linear locality, which is a useful
heuristic because nearby words often affect each other's meaning. But
how is it when distant words need to interact with each other? For
instance, if we mention a person at the beginning of a text portion
and refer back to them only at the very end, the whole text in between
needs to be tracked back (see below). Hence, RNNs take \(O(\text{sequence
length})\) steps for distant word pairs to interact. Due to gradient
problems, it is therefore hard to learn long-distance dependencies. In
addition, the linear order is ingrained. Even though, as known, the
sequential structure does not tell the whole story.

\begin{figure}

{\centering \includegraphics[width=0.7\linewidth]{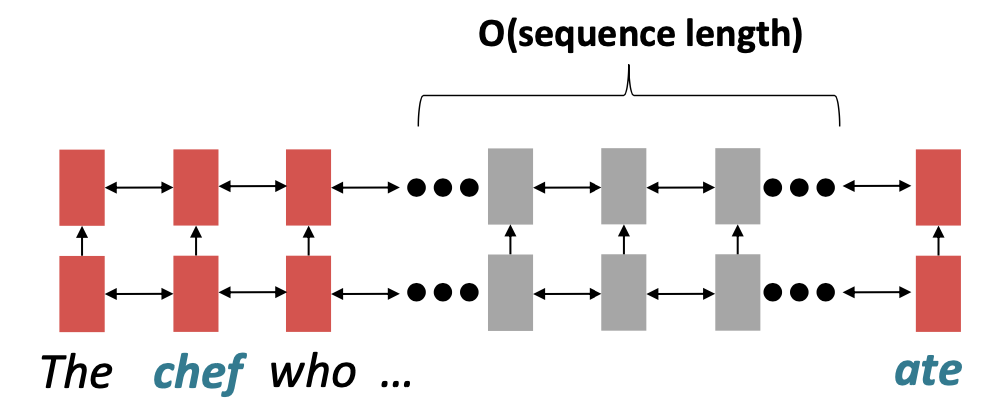}

}

\caption{Sequential processing of recurrent model (Source: \citet{Manning2022}).}\label{fig:tfrnnlimstanford}
\end{figure}

GPUs can perform multiple calculations simultaneously and could help to
reduce the execution time of the deep learning algorithm massively.
However, forward and backward passes lack parallelizability in recurrent
models and have \(O(\text{sequence length})\). To be precise, future hidden states
cannot be computed in full before past states have been computed. This
inhibits training on massive data sets. indicates the minimum number of
steps before the respective state can be calculated.

\begin{figure}

{\centering \includegraphics[width=0.7\linewidth]{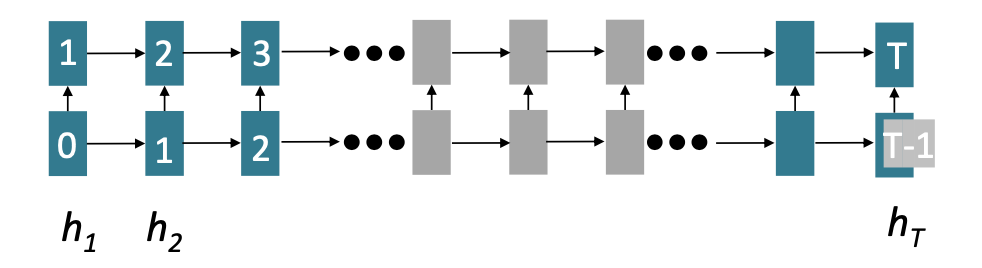}

}

\caption{Sequential processing of recurrent model with number of steps indicated (Source: \citet{Manning2022}).}\label{fig:tfrnnlim2stanford}
\end{figure}

After proving that attention dramatically increases performance, google
researchers took it further and based transformers solely on attention,
so without any RNNs. For this reason, the paper in which they were
introduced is called "Attention is all you need". Spoiler: It is not
quite all we need, but more about that on the following pages.
Transformers have achieved great results on multiple settings such as
machine translation and document generation. Their parallelizability
allows for efficient pretraining and leads them to be the standard model
architecture. In fact, all top models on the popular aggregate benchmark
GLUE are pretrained and Transformer-based. Moreover, they have even
shown promise outside of NLP, e.g.~in Image Classification, Protein
Folding and ML for Systems (see \citet{dosovitskiy2020image}, \citet{Jumper2021},
\citet{Zhou2020}, respectively).\\

If recurrence has its flaws, another adjustment of the attention
mechanism might be beneficial. Until now, it was defined from decoder to
encoder. Alternatively, attention could also be from one state to all
states in the same set. This is the definition of self-attention, which
is encoder-encoder or decoder-decoder attention (instead of
encoder-decoder) and represents a cornerstone of the transformer
architecture. depicts this process in which each word attends to all
words in the previous layer. Even though in practice, most arrows are
omitted eventually.\\

\begin{figure}

{\centering \includegraphics[width=0.7\linewidth]{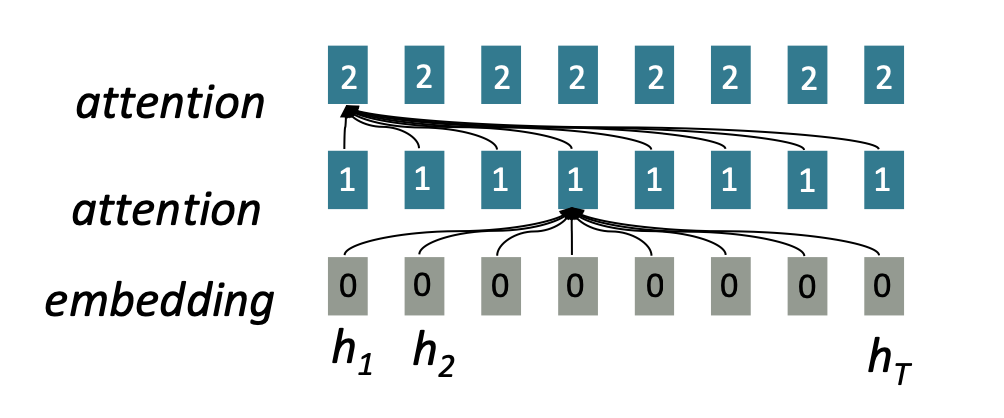}

}

\caption{Connections of classic attention mechanism (Source: \citet{Manning2022}).}\label{fig:tfselfa}
\end{figure}

Thinking of self-attention as an approximate hash table eases
understanding its intuition. To look up a value, queries are compared
against keys in a table. In a hash table, which is shown on the left
side of , there is exactly one key-value pair for each query (hash). In
contrast, in self-attention, each key is matched to varying degrees by
each query. Thus, a sum of values weighted by the query-key match is
returned.

\begin{figure}

{\centering \includegraphics[width=0.7\linewidth]{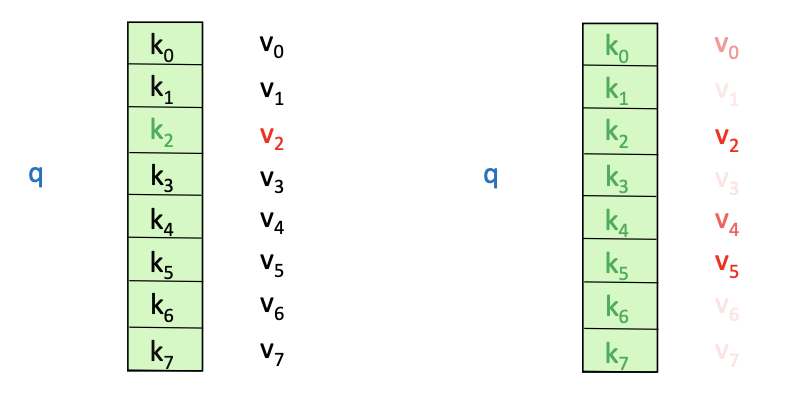}

}

\caption{Comparison of classic attention mechanism with self-attention with hash tables (Source: \citet{Manning2022}).}\label{fig:tfhashtable}
\end{figure}

The process briefly described in the last paragraph can be summarized by
the following steps that mainly follow \citet{Manning2022}. Firstly, deriving
query, key, and value for each word \(x_i\) is necessary:

\[q_i = W^Q x_i , \ \ k_i = W^K x_i, \ \ v_i = W^V x_i\]

Secondly, the attention scores have to be calculated:

\[e_{ij} = q_i k_j\]

Thirdly, to normalize the attention scores, the softmax function is
applied:

\[\alpha_{ij} = \operatorname{softmax}( e_{ij} ) = \frac{\operatorname{exp}(e_{ij})}{\displaystyle \sum_k e_{ij}}\]

Lastly, taking the weighted sum of the values results in obtaining the
attention output:

\[a_{i} = \displaystyle \sum_j \alpha_{ij} v_j\]

Multiple advantages of incorporating self-attention instead of
recurrences have been revealed. Since all words interact at every layer,
the maximum interaction distance is \(O(1)\) and is a crucial upgrade. In
addition, the model is deeply bidirectional because each word attends to
the context in both directions. As a result of these advances, all word
representations per layer can be computed in parallel. Nevertheless,
some issues have to be discussed. Attention does no more than weighted
averaging. So without neural networks, there are no element-wise
non-linearities. Their importance cannot be understated and shows why
attention is not actually all that is needed. Furthermore,
bidirectionality is not always desired. In language modelling, the model
should specifically be not allowed to simply look ahead and observe more
than the objective allows. Moreover, the word order is no longer
encoder, and it is bag-of-words once again.\\

Fortunately, the previously mentioned weaknesses have been addressed for
the original transformer-architecture proposed by \citet{Vaswani2017}. The
first problem can be easily fixed by applying a feed forward layer to
the output of attention. It provides non-linear activation as well as
extra expressive power. Then, for cases in which bidirectionality
contradicts the learning objective, future states can be masked so that
attention is restricted to previous states. Moreover, the loss of the
word can be corrected by adding position representations to the inputs.\\

The more complex deep learning models are, the closer they become to
model the complexity of the real world. That is why the transformer
encoder and decoder consist of many layers of self-attention with a feed
forward network, which is necessary to extract both syntactic and
semantic features from sentences. Otherwise, using word embeddings,
which are semantically deep representations between words, would be
unnecessary \citep{Sejnowski2020}. At the same time, training deep
networks can be troublesome. Therefore, some tricks are applied to help
with the training process.\\

\begin{figure}

{\centering \includegraphics[width=0.18\linewidth]{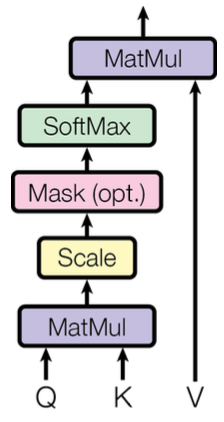}

}

\caption{Scaled dot-product attention (Source: \citet{Vaswani2017}).}\label{fig:tfsdpa}
\end{figure}

One of them is to pass the "raw" embeddings directly to the next
layer, which prevents forgetting or misrepresent important information
as it is passed through many layers. This process is called residual
connections and is also believed to smoothen the loss landscape.
Additionally, it is problematic to train the parameters of a given layer
when its inputs keep shifting because of layers beneath. Reducing
uninformative variation by normalizing within each layer to mean zero
and standard deviation to one weakens this effect. Another challenge is
caused by the dot product tending to take on extreme values because of
the variance scaling with increasing dimensionality \(d_k\). It is solved
by Scaled Dot Product Attention (see Figure \ref{fig:tfsdpa}), which consists of
computing the dot products of the query with its keys, dividing them by
the dimension of keys \(\sqrt{d_k}\), and applying the softmax function
next to receive the weights of the values.

Attention learns where to search for relevant information. Surely,
attending to different types of information in a sentence at once
delivers even more promising results. To implement this, the idea is to
have multiple attention heads per layer. While one attention head might
learn to attend to tense information, another might learn to attend to
relevant topics. Thus, each head focuses on separate features, and
construct value vectors differently. Multi-headed self-attention is
implemented by simply creating \(n\) independent attention mechanisms and
combining their outputs.

\begin{figure}

{\centering \includegraphics[width=0.34\linewidth]{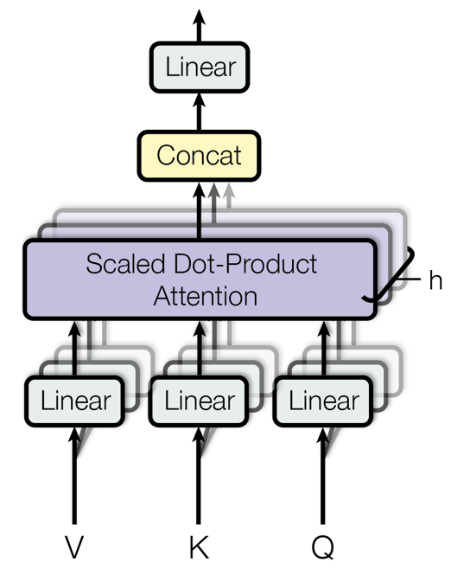}

}

\caption{Multi-head attention (Source: \citet{Vaswani2017}).}\label{fig:tfmha}
\end{figure}

\begin{figure}

{\centering \includegraphics[width=0.55\linewidth]{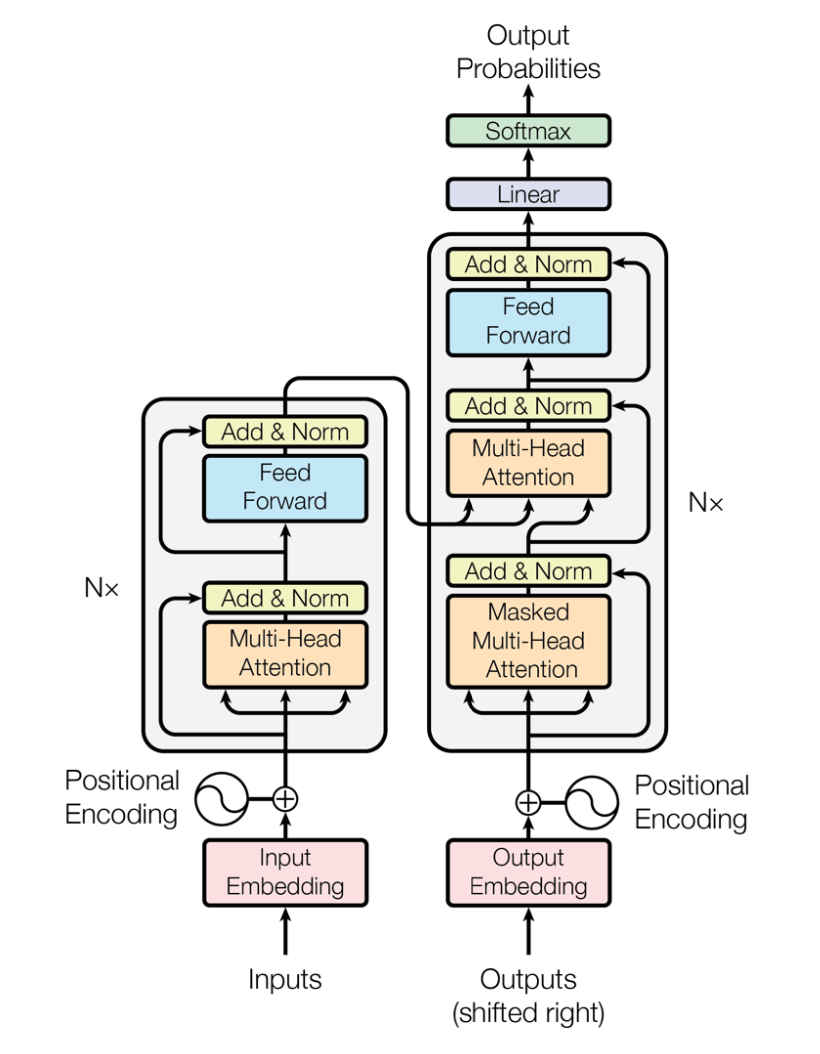}

}

\caption{Transformer architecture (Source: \citet{Vaswani2017}).}\label{fig:tfarch}
\end{figure}

At this point, every part that constitutes the encoder in the
transformer architecture has been introduced (see Figure \ref{fig:tfarch}). First, positional
encodings are included in the input embeddings. There are multiple
options to realize this step, e.g.~through sinusoids. The multi-head
attention follows, which was just mentioned. "Add \& Norm" stands for
the residual connections and the normalization layer. A feed forward
network follows, which is also accompanied by residual connections and a
normalization layer. All of it is repeated \(n\) times. For the decoder,
the individual components are similar. One difference is that the
outputs go through masked multi-head attention before multi-head
attention and the feed forward network (with residual connections and
layer normalization). It is critical to ensure that the decoder cannot
peek at the future. To execute this, the set of keys and queries could
be modified at every time step to only include past words. However, it
would be very inefficient. Instead, to enable parallelization, future
states are masked by setting the attention scores to \(-\infty\). After
the decoder process is also repeated \(n\) times, a linear layer is added
to project the embeddings into a larger vector that has the length of
the vocabulary size. At last, a softmax layer generates a probability
distribution over the possible words.

\hypertarget{transformer-architectures-bert-t5-gpt-3}{%
\subsection{Transformer architectures: BERT, T5, GPT-3}\label{transformer-architectures-bert-t5-gpt-3}}

\begin{figure}

{\centering \includegraphics[width=0.8\linewidth]{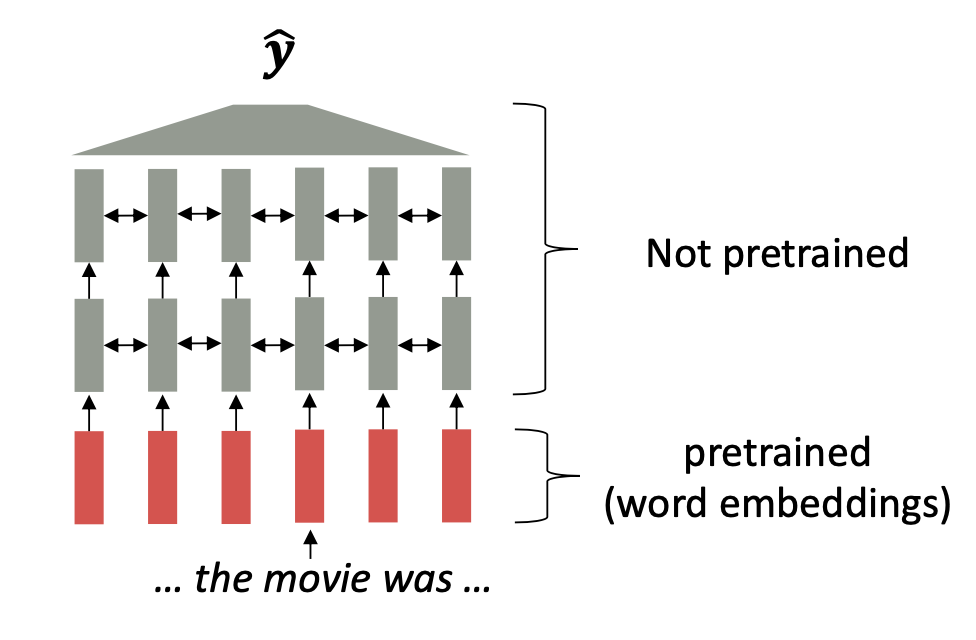}

}

\caption{Partly pre-trained model (Source: \citet{Manning2022}).}\label{fig:tfpretrain}
\end{figure}

"You shall know a word by the company it keeps", an adage by linguist
John Rupert Firth from 1957 goes. Even earlier, in 1935, he stated that
"... the complete meaning of a word is always contextual, and no study
of meaning apart from a complete context can be taken seriously". The
quotes of the famous linguist sum up the motivation to learn word
meaning and context perfectly. Many years later, in 2017, pretraining
word embeddings started. However, some complications arise from solely
pretraining the first part of the network. For instance, to teach the
model all contextual aspects of language, the training data for the
downstream task (e.g.~question answering) needs to be adequate.
Additionally, most of the parameters are usually randomly initialized.
presents the network discussed, in which the word "movie" gets the
same embedding irrespective of the sentence it appears in. On the
contrary, parameters in modern NLP architectures are initialized via
pretraining (see Figure \ref{fig:tfpretrain}). Furthermore, during the pretraining, certain input
parts are hidden to train the model to reconstruct them. This leads to
building suitable parameter initializations and robust probability
distributions over language.\\

\begin{figure}

{\centering \includegraphics[width=0.8\linewidth]{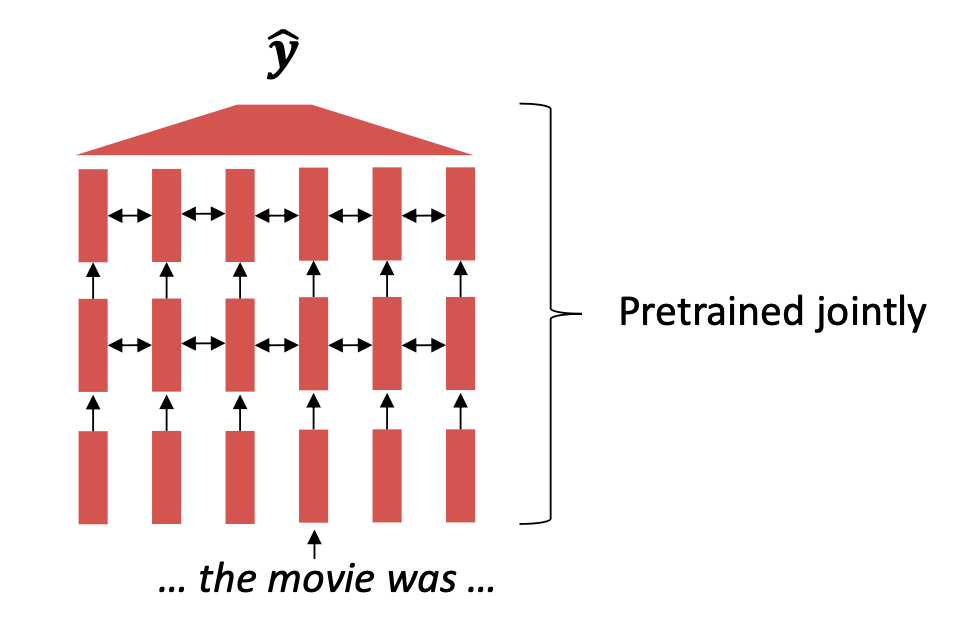}

}

\caption{Jointly pre-trained model (Source: \citet{Manning2022}).}\label{fig:tfpretrain2}
\end{figure}

Classic machine learning does not match human learning. Specifically
referring to training a model from scratch, and only being able to learn
from the training data. In contrast, human beings already have prior
knowledge they can apply to new tasks. Transfer learning emulates this
by using an already trained network. The main idea is to use a model
that was pretrained on a hard, general language understanding task using
endless amounts of data, so that, it eventually contains the best
possible approximation of language understanding. Afterwards, the
training data for the new task is applied to slightly modify the weights
of the pretrained model, which is referred to as fine-tuning
\citep{Manning2022}.\\

The specific architecture of a transformer model affects the type of
pre-training, and favourable use cases. In the following, three
different but very influential transformer architectures will be
discussed. BERT can be seen as stacked encoders \citep{Devlin2018}, T5 aims
to combine the good parts of encoders and decoders \citep{Raffel2019},
while GPT are stacked decoders \citep{brown2020language}.

\hypertarget{bert}{%
\subsubsection{BERT}\label{bert}}

Transfer learning led to state-of-the-art results in natural language
processing. One of the architectures that led the way was BERT, which
stands for Bidirectional Encoder Representations from Transformers. It
receives bidirectional context, which is why it is not a natural fit for
language modelling. To train it on this objective regardless, masked
language modelling was proposed. The main idea is to cover up a fraction
of the input words and let the model predict them. In this way, the LM
objective can be used while sustaining connections to words in the
future. The masked LM for BERT randomly predicts 15\% of all word tokens
in each sequence. Of those, 80\% are replaced by the \texttt{MASK} token, 10\%
by a random token, and 10\% remain unchanged. Moreover, because the
masked words are not even seen in the fine-tuning phase, the model
cannot get complacent and relies on strong representations of non-masked
words. Initially, BERT had an additional objective of whether one
sentence follows another, which is known as next sentence prediction.
However, it was dropped in later work due to having an insignificant
effect.\\

BERT is hugely versatile and was greatly popular after its release.
Fine-tuning BERT led to outstanding results on a variety of
applications, including question answering, sentiment analysis and text
summarization. Thanks to its design, if the task involves generating
sequences, pretrained decoders outperform pretrained encoders like BERT.
Even though, it would not be recommended for autoregressive generation,
up to this day, "small" models like BERT are applied as general tools
for numerous tasks.

\hypertarget{t5}{%
\subsubsection{T5}\label{t5}}

The Text-To-Text Transfer Transformer (T5) is a new model that can be
regarded as an application of the insights gathered by an extensive
empirical study searching for the best transfer learning techniques. It
is pretrained on Colossal Clean Crawled Corpus (C4), an open-source
dataset. \citet{Raffel2019} found that the best pretraining objective to use
for the encoder component was span corruption. In short, different
length word groups (spans) are replaced with unique placeholders, and
let the model decode them. Text preprocessing is necessary for its
implementation. For the decoder, it is still a language modelling task.
Compared to models like BERT, which can only output a span of the input
or a class label, T5 reframes all NLP tasks into a unified text-to-text
format, where inputs and outputs always consist of text strings. As a
result, the same model, loss function, and hyperparameters can be used
on any NLP task, such as machine translation, document summarization,
question answering, and classification tasks like sentiment analysis. T5
can even be applied to regression tasks by training it to predict the
string representation of a number (and not the number itself). Examples
of potential use cases are depicted in below.

\begin{figure}

{\centering \includegraphics[width=0.8\linewidth]{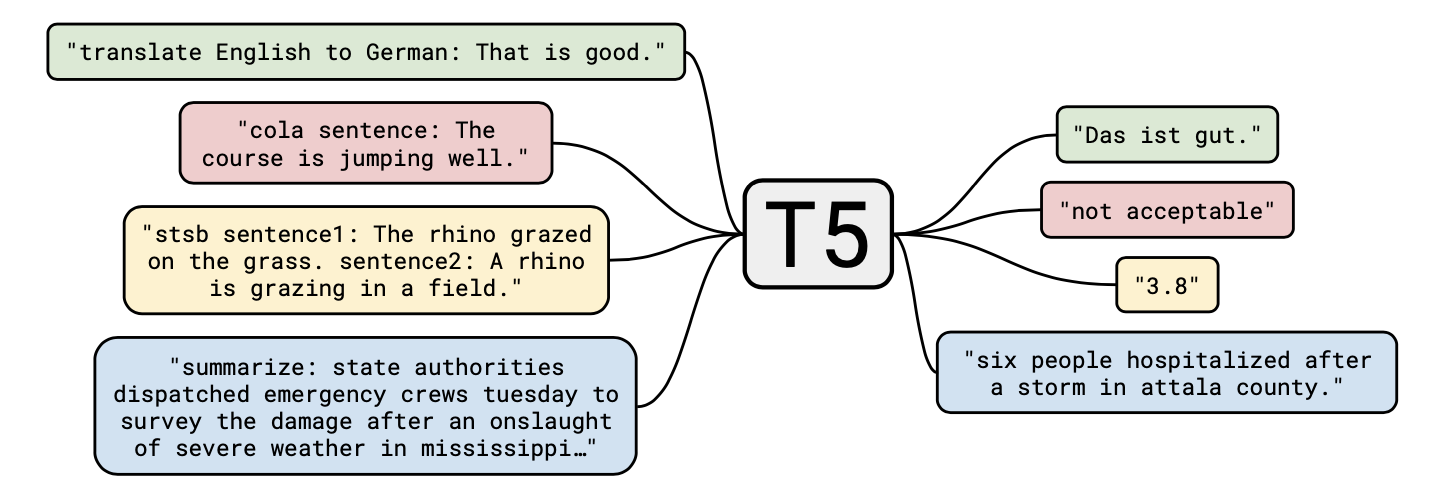}

}

\caption{Applications of T5 model (Source: \citet{Raffel2019}).}\label{fig:t5}
\end{figure}

\hypertarget{gpt-3}{%
\subsubsection{GPT-3}\label{gpt-3}}

As previously stated, the neural architecture influences the type of
pretraining. The original GPT architecture consists of a Transformer
decoder with 12 layers \citep{Radford2018}. For decoders, it is
sensible to simply pretrain them as language models. Afterwards, they
can be used as generators to fine-tune their probability of predicting
the next word conditioned on the previous words. The models are suitable
for tasks similar to the training, including any type of dialogue and
document summarization. Transformer language models are great for
transfer learning. They are fine-tuned by randomly initializing a
softmax classifier on top of the pretrained model and training both
(with only a very small learning rate and a small number of epochs) so
that the gradient propagates through the whole network.\\

The success of BERT in 2018 prompted a "gold rush" in NLP, in which
ever greater language models were created. One that topped the headlines
and used a customer supercluster for computation was the third iteration
of the GPT architecture by OpenAI, known as GPT-3. reveals why GPT-3 is
a famous example of current research focusing on scaling up neural
language models. While the largest T5 model has 11 billion parameters,
GPT-3 has 175 billion parameters. Moreover, the training data set
contains around 500 billion tokens of text, while the average young
american child hears around 6 million words per year \citep{Hart1995}.
The results of huge language models suggest that they perform some form
of learning (without gradient steps) simply from examples provided via
context. The tasks are specified by the in-context examples, and the
conditional probability distribution simulates performing the task to an
extent.

\begin{figure}

{\centering \includegraphics[width=0.7\linewidth]{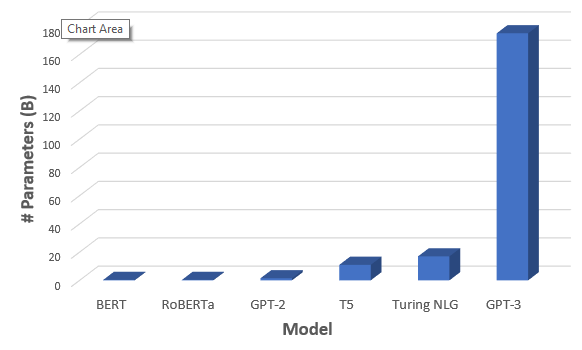}

}

\caption{Comparison of number of parameters between Transformer-architectures (Source: \citet{Saifee2020}).}\label{fig:gpt3scaling}
\end{figure}

\hypertarget{current-topics}{%
\subsection{Current Topics}\label{current-topics}}

\hypertarget{sec:concerns-lm}{%
\subsubsection{Concerns regarding growing size of Language Models}\label{sec:concerns-lm}}

As the last chapter ended with GPT-3 and emphasized the concerning trend
of ever larger language models, one could ask which other costs arise
from the developments. Risks and harms among environmental and financial
costs have been studied by \citet{Bender2021}. They state that marginalized
communities are not only less likely to benefit from LM progress, but
also more likely to suffer from the environmental repercussions of
increasing resource consumption. \citet{Strubell2019} estimated that training
a Transformer (big) model resulted in 249t of \(CO_2\). To compare, an
average human is responsible for approximately 5t of \(CO_2\) per year
\citep{Ritchie2020}. In addition, they discovered that an
estimated increase of 0.1 in BLEU score increased computation costs by
\$ 150,000 (for English to German translations). Furthermore, larger
models require more data to sufficiently train them. This has resulted
in large but poorly documented training data sets. Multiple risks can be
mitigated if there is a common understanding of the model's learnings.\\

Moreover, it has been argued that datasets consisting of web data
over-represent hegemonic views and encode bias towards marginalized
communities. This is among other factors due to internet access being
unevenly distributed. In particular, there is an over-representation of
younger internet users and those from developed countries. It is
generally naive to educate AI systems on all aspects of the complex
world, and hope for the beautiful to prevail \citep{Bender2021}.

\hypertarget{improving-understanding-of-transformer-based-models}{%
\subsubsection{Improving Understanding of Transformer-based models}\label{improving-understanding-of-transformer-based-models}}

The results of transformer-based models clearly show that they deliver
successful results. However, it is less clear why. The size of the
models makes it difficult to experiment with them. Nevertheless, having
a limited understanding restrains researchers from coming up with
further improvements. Therefore, multiple papers analysed BERT's
attention in search of an improved understanding of large transformer
models. BERT is a smaller model out of the more popular ones, and its
attention is naturally interpretable because the attention weight
indicates how significant a word is for the next representation of the
current word \citep{Clark2019}. In the following, some of the findings are
going to be shared.\\

BERT representations are rather hierarchical than linear, and they
include information about parts of speech, syntactic chunks and roles
\citep[\citet{Liu2019}]{Lin2019} Furthermore, it has semantic knowledge.
For example, BERT can recognize e.g.~that "to tip a chef" is better
than "to tip a robin" but worse than "to tip a waiter"
(\citep{Ettinger2019}). However, it makes sense that BERT has issues with
knowledge that is assumed and not mentioned, which especially refers to
visual and perceptual properties \citep{Da2019}. Additionally, BERT
struggles with inferences, e.g.~even though it is known that "people
walk into houses" and "houses are big", it cannot infer that "houses
are bigger than people" \citep{Forbes2019}.\\

While it is true that different transformer heads attend to various
patterns (see ), interestingly, most of them could be neglected without
notable performance loss \citep{Voita2019}. Probing attention maps can be
tedious, but allows to gain knowledge of common patterns, such as an
unexpected amount focusing on the delimiter token \[SEP\].

\begin{figure}

{\centering \includegraphics[width=0.9\linewidth]{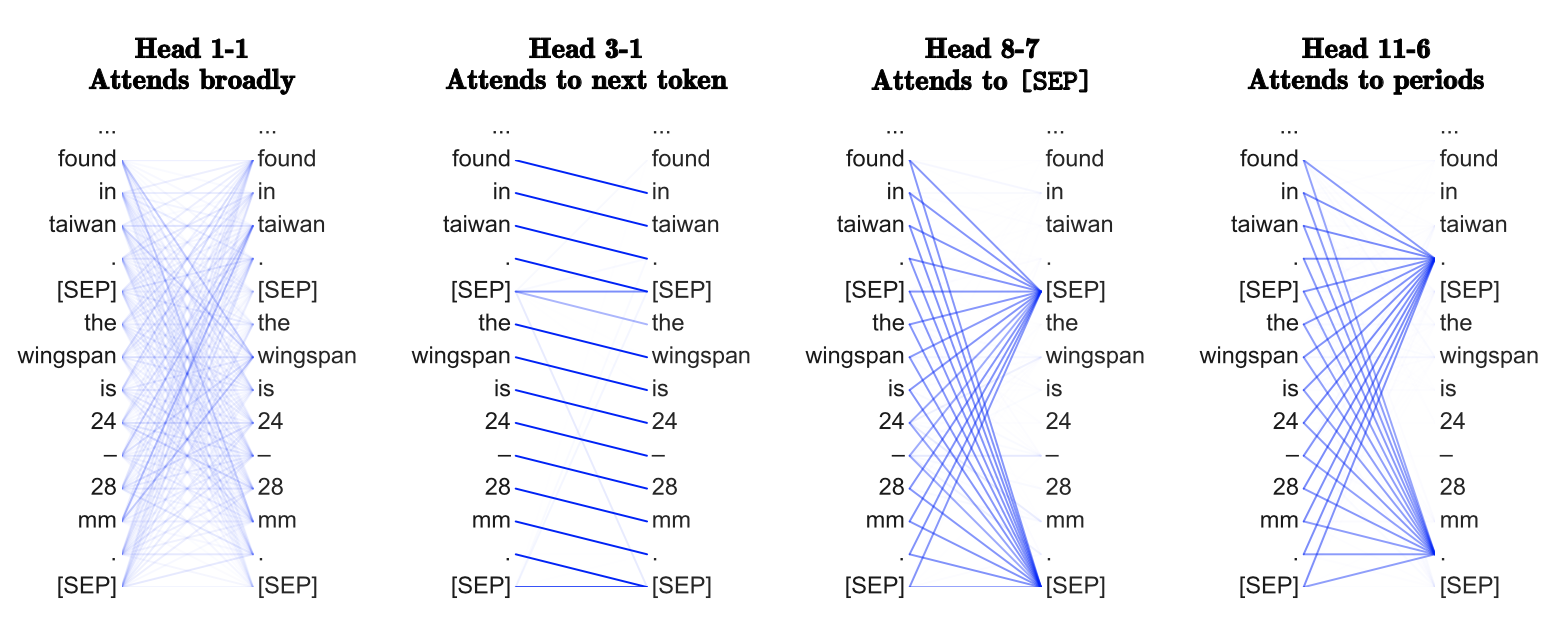}

}

\caption{Common patterns of attention heads (Source: \citet{Clark2019}).}\label{fig:ctattentionheads}
\end{figure}

\hypertarget{few-shot-learning}{%
\subsubsection{Few-Shot Learning}\label{few-shot-learning}}

For NLP tasks, the model is usually trained on a set of labelled
examples and is expected to generalize to unseen data. Annotating is not
only costly but also difficult to gather for numerous languages,
domains, and tasks. In practice, there is often only a very limited
amount of labelled examples. Consequently, few-shot learning is a highly
relevant research area \citep{Schick2020}. It defines a model that is
trained on a limited number of demonstrations to guide its predictions.
Referring back to , the benefits of lower computational and
environmental costs have to be mentioned.\\

Traditional fine-tuning uses a large corpus of example tasks, and the
model is updated repeatedly with gradient steps so that it adapts to the
task with minimal accuracy error.

In contrast, few-shot applications have to complete tasks at test time
with only forward passes. They have three main parts: the task
description, examples, and the prompt. In Figure \ref{gpt3fewshotlearning}, the task is a translation
from English to French, a few examples, as well as the word that should
be translated are given. Moreover, zero-shot and one-shot learning refer
to the model predicting with no and one learned example, respectively
\citep{brown2020language}.

\begin{figure}

{\centering \includegraphics[width=0.5\linewidth]{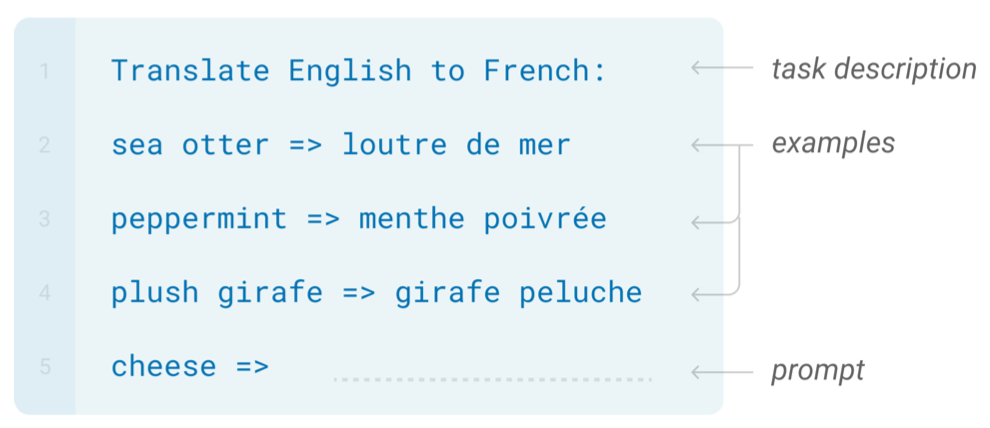}

}

\caption{Few-shot learning (Source: \citet{brown2020language}).}\label{fig:gpt3fewshotlearning}
\end{figure}

It is complicated to create the few-shot examples, since the application
relies on them to express the task. This is why smaller models are
susceptible to examples written unfavourably. In \citet{brown2020language}, it was
shown that few-shot performance scales with the number of model
parameters. Even though GPT-3's in-context learning improved few-shot
prompting capabilities, it is still sensitive to the order of training
examples, decoding strategy, and hyperparameter selection. All of this
combined with the fact that current research uses larger or held-out
data sets leads to the suspicion that the true few-shot ability of
language models is overestimated \citep{Perez2021}.\\

Moreover, \citet{Lialin2022} have found that common transformer models could
not resolve compositional questions in a zero-shot fashion and that the
model's parameter count does not correlate with performance. This
indicates a limitation for zero-shot prompting with the existing
pre-training objectives. However, different models provided the best
accuracy with regard to different symbolic reasoning tasks. This
suggests that optimization or masking strategies could be more
significant than the pre-training, data set size or model architecture.

\hypertarget{summary}{%
\subsection{Summary}\label{summary}}

Natural Language Processing has been one of the most exciting fields of
machine learning in the last decade considering all the breakthroughs
discussed in this work. Word embeddings made it possible and allowed
developers to encode words as dense vectors that capture their
underlying semantic content. In this way, similar words are embedded
close to each other in a lower-dimensional feature space. Another
important challenge was solved by encoder-decoder (also called
sequence-to-sequence) architectures, which made it possible to map input
sequences to output sequences of different lengths. They are especially
useful for complex tasks like machine translation, video captioning or
question answering. A significant state-of-the-art technique is
attention, which enabled models to actively shift their focus -- just
like humans do. It allows following one thought at a time while
suppressing information irrelevant to the task. As a consequence, it has
been shown to significantly improve performance for tasks like machine
translation. By giving the decoder access to directly look at the
source, the bottleneck is avoided and at the same time, it provides a
shortcut to faraway states and thus helps with the vanishing gradient
problem. One of the most recent data modelling techniques is the
transformer, which is solely based on attention and does not have to
process the input data sequentially. Therefore, the deep learning model
is better in remembering context-induced earlier in long sequences. It
is the dominant paradigm in NLP currently and makes better use of GPUs
because it can perform parallel operations. Transformer architectures
like BERT, T5 or GPT-3 are pre-trained on a large corpus and can be
fine-tuned for specific language tasks. They can generate stories,
poems, code and much more. Currently, there seems to be breaking
transformer news nearly every week with no sign of slowing. This is why
many trends could be recognized as relevant current topics. One of them
is increasing concerns regarding the growing size of language models and
the correlated environmental and financial costs. Another active
research aspect is concerned with improving the understanding of
transformer-based models to further advance them. Additionally, there
are many studies about achieving respectable results on language
modelling tasks after only learning from a few examples, which is known
as few-shot learning.

\hypertarget{c01-02-sota-cv}{%
\section{State-of-the-art in Computer Vision}\label{c01-02-sota-cv}}

\emph{Author: Vladana Djakovic}

\emph{Supervisor: Daniel Schalk}

\hypertarget{history}{%
\subsection{History}\label{history}}

The first research about visual perception comes from neurophysiological research performed in the 1950s and 1960s on cats. The researchers used cats as a model to understand how human vision is compounded. Scientists concluded that human vision is hierarchical and neurons detect simple features like edges followed by more complex features like shapes and even more complex visual representations. Inspired by this knowledge, computer scientists focused on recreating human neurological structures.

At around the same time, as computers became more advanced, computer scientists worked on imitating human neurons' behavior and simulating a hypothetical neural network. In his book ``The Organization of Behaviour'' (1949) Donald Hebbian stated that neural pathways strengthen over each successive use, especially between neurons that tend to fire at the same time, thus beginning the long journey towards quantifying the complex processes of the brain. The first Hebbian network, inspired by this neurological research, was successfully implemented at MIT in 1954 \citep{history1}.

New findings led to the establishment of the field of artificial intelligence in 1956 on-campus at Dartmouth College. Scientists began to develop ideas and research how to create techniques that would imitate the human eye.

In 1959 early research on developing neural networks was performed at Stanford University, where models called ``ADALINE'' and ``MADALINE,'' (Multiple ADAptive LINear Elements) were developed. Those models aimed to recognize binary patterns and could predict the next bit \citep{history2}.

Starting optimism about Computer Vision and neural networks disappeared after 1969 and the publication of the book ``Perceptrons'' by Marvin Minsky, founder of the MIT AI Lab, stated that the single perception approach to neural networks could not be translated effectively into multi-layered neural networks. The period that followed was known as AI Winter, which lasted until 2010, when the technological development of computer and the internet became widely used. In 2012 breakthroughs in Computer Vision happened at the ImageNet Large Scale Visual Recognition Challenge (ILSVEC). The team from the University of Toronto issued a deep neural network called AlexNet \citep{alexnet} that changed the field of artificial intelligent and Computer Vision (CV). AlexNet achieved an error rate of 16.4\%.

From then until today, Computer Vision has been one of the fastest developing fields. Researchers are competing to develop a model that would be the most similar to the human eye and help humans in their everyday life. In this chapter the author will describe only a few recent state-of-the-art models.

\hypertarget{supervised-and-unsupervised-learning}{%
\subsection{Supervised and unsupervised learning}\label{supervised-and-unsupervised-learning}}

As part of artificial intelligence (AI) and machine learning (ML), there are two basic approaches:

\begin{itemize}
\tightlist
\item
  supervised learning;
\item
  unsupervised learning.
\end{itemize}

Supervised learning \citep{supervised} is used to train algorithms on labeled datasets that accurately classify data or predict outcomes. With labeled data, the model can measure its accuracy and learn over time. Among others, we can distinguish between two common supervised learning problems:

\begin{itemize}
\tightlist
\item
  classification,
\item
  regression.
\end{itemize}

In unsupervised learning \citep{unsupervised}, unlabelled datasets are analyzed and clustered using machine learning algorithms. These algorithms aim to discover hidden patterns or data groupings without previous human intervention. The ability to find similarities and differences in information is mainly used for three main tasks:

\begin{itemize}
\tightlist
\item
  clustering,
\item
  association,
\item
  dimensionality reduction.
\end{itemize}

Solving the problems where the dataset can be both labeled and unlabeled requires a semi-supervised approach that lies between supervised and unsupervised learning. It is useful when extracting relevant features from complex and high volume data, i.e., medical images.

Nowadays, a new research topic appeared in the machine learning community, Self-Supervised Learning. Self-Supervised learning is a process where the model trains itself to learn one part of the input from another \citep{selfsup}. As a subset of unsupervised learning, it involves machines labeling, categorizing, and analyzing information independently and drawing conclusions based on connections and correlations. It can also be considered as an autonomous form of supervised learning since it does not require human input to label data. Unlike unsupervised learning, self-supervised learning does not focus on clustering nor grouping \citep{selfsup2}. One part of Self-Supervised learning is contrastive learning, which is used to learn the general features of an unlabeled dataset identifying similar and dissimilar data points. It is utilized to train the model to learn about our data without any annotations or labels \citep{contrastive}.

\hypertarget{scaling-networks}{%
\subsection{Scaling networks}\label{scaling-networks}}

Ever since the introduction of AlexNet in 2012, the problem of scaling convolutional neural networks (ConvNet) has become the topic of active research. ConvNet can be scaled in all three dimensions: depth, width, or image size. One of the first researches in 2015 showed that network depth is crucial for image classification. The question whether stacking more layers enables the network to learn better leads to deep residual networks called ResNet \citep{ResNet}, which will be described in this work. Later on, scaling networks by their depth became the most popular way to improve their performance.
The second solution was to scale ConvNets by their width. Wider networks tend to be able to capture more fine-grained features and are easier to train \citep{width}.
Lastly, scaling the image's resolution can improve the network's performance. With higher resolution input images, ConvNets could capture more fine-grained patterns. GPipe \citep{gpipe} is one of the most famous networks created by this technique.
The question of possibility of scaling by all three dimensions was answered by \citet{EfficientNet} in the work presenting Efficient Net. This network was built by scaling up ConvNets by all three dimensions and will also be described here.

\hypertarget{deep-residual-networks}{%
\subsection{Deep residual networks}\label{deep-residual-networks}}

The deep residual networks, called ResNets \citep{ResNet}, were presented as the answer on the question whether stacking more layers would enable network to learn better. Until then one obstacle for simply stacking layers was the problem of vanishing/exploding gradients. It has been primarily addressed by normalized initialization and intermediate normalization layers. That enabled networks with tens of layers to start converging for stochastic gradient descent (SGD) with backpropagation.

Another obstacle was a degradation problem. It occurs when the network depth increases, followed by saturating and then rapidly decreasing accuracy. Overfitting is not caused by such degradation, and adding more layers to a suitably deep model leads to higher training error, which indicates that not all systems are similarly easy to optimize.

For example, it was suggested to consider a shallower architecture and its deeper counterpart that adds more layers. One way to avoid the degradation problem is to create a deeper model, where the auxiliary layers are identity mappings and other layers are copied from a shallower model. The deeper model should produce no higher training error than its shallower counterpart. However, in practice it is not the case and it is hard to find comparably good constructs or better solutions. The solution to this degradation problem proposed by them is a deep residual learning framework.

\hypertarget{deep-residual-learning}{%
\subsubsection{Deep Residual Learning}\label{deep-residual-learning}}

\hypertarget{residual-learning}{%
\paragraph{Residual Learning}\label{residual-learning}}

The idea of residual learning is to replace the approximation of underlying mapping \(H\left( x\right)\), which is approximated by a few stacked layers (not necessarily the entire net), with an approximation of residual function \(F(x):= H\left( x \right) - x\). Here x denotes the inputs to the first of these layers, and it is assumed that both inputs and outputs have the same dimensions. The original function changes its form \(F\left( x \right)+x\).

A counter-intuitive phenomenon about degradation motivated this reformulation. The new deeper model should not have a more significant training error when compared to a construction using identity mappings. However, due to the degradation problem, solvers may have challenges approximating identity mappings by multiple non-linear layers. Using the residual learning reformulation can drive the weights of the non-linear layers toward zero to approach identity mappings if they are optimal.
Generally, identity mappings are not optimal, but new reformulations may help to pre-condition the problem. When an optimal function is closer to an identity mapping than a zero mapping, finding perturbations concerning an identity mapping should be easier than learning the function from scratch.

\hypertarget{identity-mapping-by-shortcuts}{%
\paragraph{Identity Mapping by Shortcuts}\label{identity-mapping-by-shortcuts}}

\begin{figure}

{\centering \includegraphics[width=0.35\linewidth]{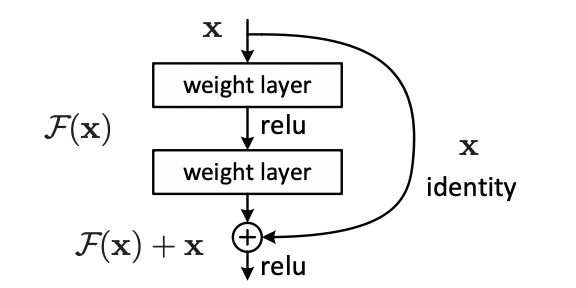}

}

\caption{Building block of residual learning \citep{ResNet}.}\label{fig:ch01-figure01}
\end{figure}

Residual learning is adopted to every few stacked layers where a building block is defined:

\begin{equation}
\label{eq:ch01-02-01}
y = F  \left( x,\left\{  W_i\right\} \right) + x
\end{equation}

x and y present the input and output vectors of the layers. Figure \ref{fig:ch01-figure01} visualizes the building block.

The function \(F \left( x,\left\{ W_i\right\} \right)\) represents the residual mapping that is to be learned. For the example with two layers from Figure \ref{fig:ch01-figure01}, \(F = W_2\sigma\left( W_1x\right)\) in which \(\sigma\) denotes the ReLU activation function. Biases are left out to simplify the notation. The operation \(F + x\) is conducted with a shortcut connection and element-wise addition. Afterward, a second non-linear (i.e., \(\sigma \left( y \right)\) transformation is applied.

The shortcut connections in Equation \eqref{eq:ch01-02-01} neither adds an extra parameter nor increases computation complexity and enables a comparisons between plain and residual networks that concurrently have the same number of parameters, depth, width, and computational cost (except for the negligible element-wise addition).
The dimensions of \(x\) and \(F\) in Equation \eqref{eq:ch01-02-01} must be equal. Alternatively, to match the dimensions, linear projection \(W_s\) by the shortcut connections can be applied:

\begin{equation}
\label{eq:ch01-02-02}
y = F  \left( x,\left\{  W_i\right\} \right)+ W_sx.
\end{equation}

The square matrix \(W_s\) can be used in Equation \eqref{eq:ch01-02-02}. However, experiments showed that identity mapping is enough to solve the degradation problem. Therefore, \(W_s\) only aims to match the dimensions. Although more levels are possible, it was experimented with function \(F\) having two or three layers without stating the exact form of it. Assuming \(F\) only has one layer (Equation \eqref{eq:ch01-02-01}) it is comparable to a linear layer: \(y = W_1 x + x\). The theoretical notations are about fully-connected layers, but convolutional layers were used. The function \(F \left( x,\left\{ W_i\right\} \right)\) can be applied to represent multiple convolutional layers. Two feature maps are added element-wise, channel by channel.

\hypertarget{network-architectures}{%
\subsubsection{Network Architectures}\label{network-architectures}}

Various plain/residual networks were tested to construct an efficient residual network. They trained the network on benchmarked datasets, e.g.~the ImageNet dataset, that are used for a comparison of network architectures. Figure \eqref{eq:ch01-02-02} shows that every residual network needs a plain baseline network inspired by the VGG \citep{vgg} network on which identity mapping by shortcuts is applied.

\emph{Plain Network:} The philosophy of VGG nets 41 mainly inspires the plain baselines. Two rules convolution layers, which usually have \(3\times 3\) filters, follow are:

\begin{itemize}
\tightlist
\item
  feature maps with the same output size have the same number of layers;
\item
  reducing the size of a feature map by half doubles the number of filters per layer to maintain time complexity per layer.
\end{itemize}

\begin{figure}

{\centering \includegraphics[width=1\linewidth]{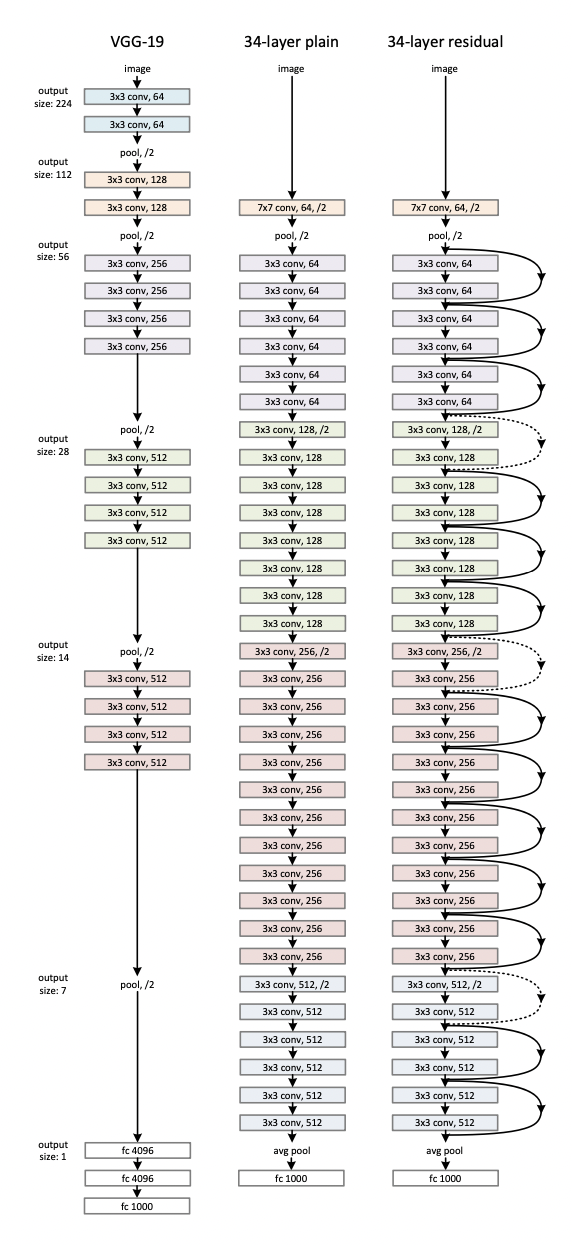}

}

\caption{Architecture of ResNet \citep{ResNet}.}\label{fig:ch01-figure02}
\end{figure}

Convolutional layers with a stride of 2 perform downsampling directly. A global average pooling layer and a 1000-way fully-connected layer with softmax are at the end of the network. The number of weighted layers sums up to 34 (Figure \ref{fig:ch01-figure02}, middle). Compared to VGG nets, this model has fewer filters and lower complexity (Figure \ref{fig:ch01-figure02}, left).

\emph{Residual Network:} Based on the above plain network, additional shortcut connections (Figure \ref{fig:ch01-figure02}, right) turn the network into its associate residual variant. The identity shortcuts (Equation \eqref{eq:ch01-02-01}) can be directly used in the case of the exact dimensions of the input and output (solid line shortcuts in Figure \ref{fig:ch01-figure02}). For the different dimensions (dotted line shortcuts in Figure \ref{fig:ch01-figure02}), two options are considered:

\begin{itemize}
\tightlist
\item
  The shortcut still performs identity mapping, but with extra zero entries padded to cope with the increasing dimensions, without adding new parameters;
\item
  The projection shortcut in Equation \eqref{eq:ch01-02-02} matches dimensions (due to \(1\times 1\) convolutions).
\end{itemize}

In both cases, shortcuts will be done with a stride of two when they go across feature maps of two sizes.

\hypertarget{efficientnet}{%
\subsection{EfficientNet}\label{efficientnet}}

\begin{figure}

{\centering \includegraphics[width=1\linewidth]{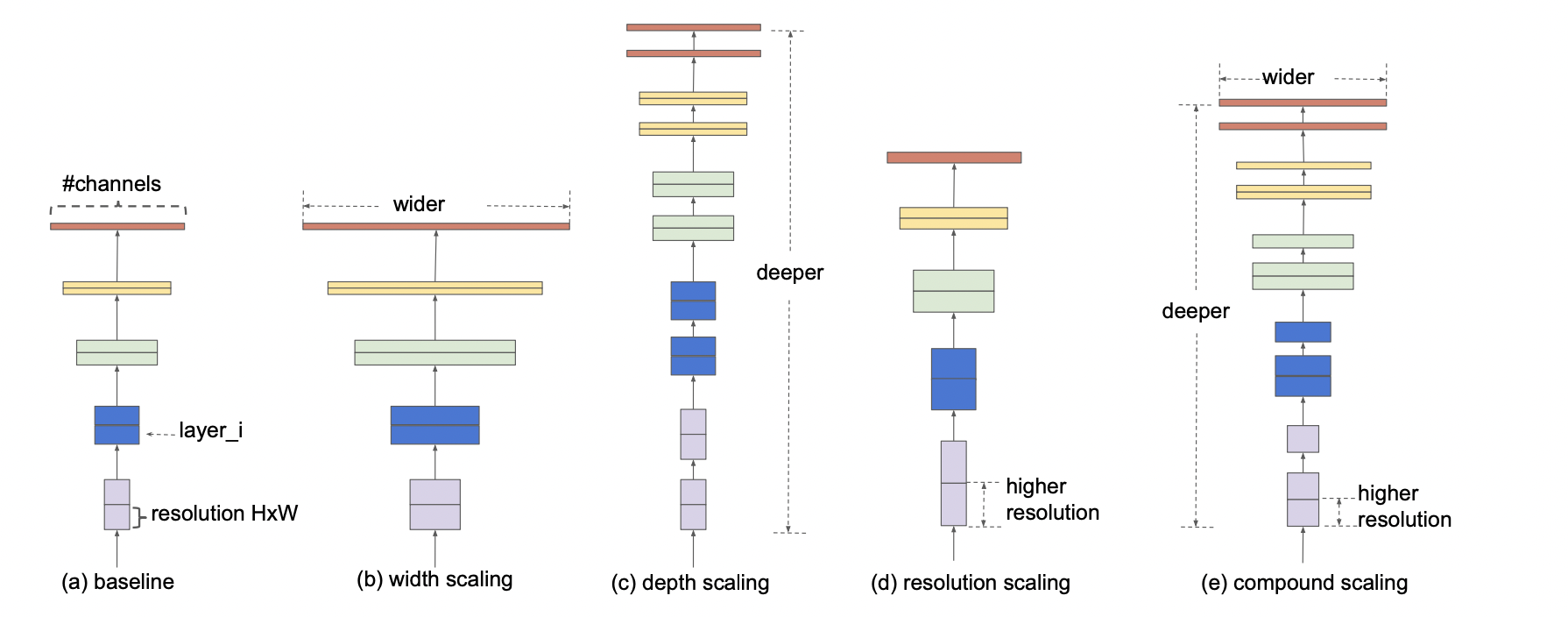}

}

\caption{Model scaling \citep{effecient}.}\label{fig:ch01-figure03}
\end{figure}

Until \citet{effecient} introduced EfficientNet, it was popular to scale only one of the three dimensions -- depth, width, or image size. The empirical study shows that it is critical to balance all network dimensions, which can be achieved by simply scaling each with a constant ratio. Based on this observation, a simple yet effective compound scaling method was proposed, which uniformly scales network width, depth, and resolution with a set of fixed scaling coefficients. For example, if \(2N\) times more computational resources are available, increasing the network depth by \(\alpha N\), width by \(\beta N\), and image size by \(\gamma N\) would be possible. Here \(\alpha,\beta,\gamma\) are constant coefficients determined by a small grid search on the original miniature model. Figure \ref{fig:ch01-figure03} illustrates the difference between this scaling method and conventional methods. A compound scaling method makes sense if an input image is bigger because a larger receptive field requires more layers and more significant channel features to capture fine-grained patterns. Theoretically and empirically, there has been a special relationship between network width and depth \citep{depthwidth}. Existing MobileNets \citep{mobilenet} and ResNets are used to demonstrated new scaling methods.

\hypertarget{compound-model-scaling}{%
\subsubsection{Compound Model Scaling}\label{compound-model-scaling}}

\hypertarget{problem-formulation}{%
\paragraph{Problem Formulation}\label{problem-formulation}}

A function \(Y_i = \mathcal{F}_i \left( X_i \right)\) with the operator \(\mathcal{F}_i\), output tensor \(Y_i\), input tensor \(X_i\) of shape \(\left( H_i, W_i, C_i \right)\), spatial dimensions \(H_i\), \(W_i\), and channel dimension \(C_i\) is called a ConvNet Layer \(i\). A ConvNet N appears as a list of composing layers:
\[
\mathcal{N}=\mathcal{F_k}\odot \cdots \mathcal{F_2}\odot\mathcal{F_1}\left( X_1 \right)=\bigodot{j=1\cdots k}\mathcal{F_j}\left( X_1 \right)
\]

Effectively, these layers are often partitioned into multiple stages and all layers in each stage share the same architecture. For example, ResNet has five stages with all layers in every stage being the same convolutional type except for the first layer that performs down-sampling. Therefore, a ConvNet can be defined as:

\[
\mathcal{N}=\bigodot_{i=1\cdots s}\mathcal{F_i}^{L_i}\left( X_{\left( H_i, W_i, C_i  \right)} \right)
\]

where \(\mathcal{F_i}^{L_i}\) denotes layer \(\mathcal{F_i}\) which is repeated \(L_i\) times in stage \(i\), and \(\left( H_i, W_i, C_i \right)\) is the shape of input tensor \(X\) of layer \(i\).

In comparison to the regular ConvNet focusing on the best layer architecture search \(\mathcal{F_i}\), model scaling centers on the expansion of the network length \(\left( L_i\right)\), width \(\left( C_i \right)\), and/or resolution \(\left( H_i, W_i\right)\) without changing \(\mathcal{F_i}\) that was predefined in the baseline network. Although model scaling simplifies the design problem of the new resource constraints through fixing \(\mathcal{F_i}\), a different large design space \(\left( L_i, H_i, W_i, C_i \right)\) for each layer remains to be explored. To further reduce the design space, all layers are restricted to be scaled uniformly with a constant ratio. In this case, the goal is to maximize the model's accuracy for any given resource constraint, which is presented as an optimization problem:

\begin{align*}
\max_{d,w,r} &\text{Accuracy} \left( \mathcal{N}\left( d,w,r \right) \right) \\
s.t.\mathcal{N}\left( d,w,r \right) &=\bigodot_{I=1...s}\hat{\mathcal{F}}{i}^{d\cdot \hat{L{i}}}\left( X_{\left\langle r\cdot \hat{H_i},r\cdot \hat{W_i},w\cdot \hat{C_i}\right\rangle} \right) \\
Memory\left( \mathcal{N} \right) &\leq\ targetMemory \\
FLOPS\left( \mathcal{N} \right) &\leq\ targetFlops
\end{align*}
where \(w,d,r\) are coefficients for scaling network width, depth, and resolution; \(\left(\widehat{\mathcal{F}}_i, \widehat{L}_i, \widehat{H}_i, \widehat{W}_i, \widehat{C}_i \right)\) are predefined parameters of the baseline network.

\hypertarget{scaling-dimensions}{%
\paragraph{Scaling Dimensions}\label{scaling-dimensions}}

The main difficulty of this optimization problem is that the optimal \(d, w, r\) depend on each other and the values are changing under different resource constraints. Due to this difficulty, conventional methods mostly scale ConvNets in one of these dimensions:

\textbf{Depth (\(d\)):} One of the most significant networks previously described is the ResNet. As it was described, the problem of ResNets is that accuracy gain of a very deep network diminishes. For example, ResNet-1000 has similar accuracy to ResNet-101 even though it contains many more layers.

\textbf{Width (\(w\)):} Scaling network width is commonly used for small-sized models. However, wide but shallow networks tend to have difficulty grasping higher-level features.

\textbf{Resolution (\(r\)):} Starting from \(224\times 224\) in early ConvNets, modern ConvNets tend to use \(299\times 299\) or \(331\times 331\) for better accuracy. GPipe \citep{gpipe} recently achieved state-of-the-art ImageNet accuracy with \(480\times 480\) resolution. Higher resolutions, such as \(600\times 600\), are also widely used in ConvNets for object detection.

The above analyses lead to the first observation:

\textbf{Observation 1:} Scaling up any network width, depth, or resolution dimension improves accuracy. Without the upscaling, the gain diminishes for bigger models.

\hypertarget{compound-scaling}{%
\paragraph{Compound Scaling}\label{compound-scaling}}

Firstly, it was observed that different scaling dimensions are not independent because higher resolution images also require to increase the network depth. The larger receptive fields can help capture similar features that include more pixels in bigger images. Similarly, network width should be increased when the resolution is higher to capture more fine-grained patterns. The intuition suggests that different scaling dimensions should be coordinated and balanced rather than conventional scaling in single dimensions.
To confirm this thought, results of networks with width \(w\) without changing depth (\(d\)=1.0) and resolution (\(r\)=1.0) were compared with deeper (\(d\)=2.0) and higher resolution (\(r\)=2.0) networks. This showed that width scaling achieves much better accuracy under the same FLOPS. These results lead to the second observation:

\textbf{Observation 2:} To achieve better accuracy and efficiency, balancing the network width, depth, and resolution dimensions during ConvNet scaling is critical. Earlier researches have tried to arbitrarily balance network width and depth, but they all require tedious manual tuning.

A new \textbf{compound scaling method}, which uses a compound coefficient \(\varphi\) to uniformly scale network width, depth, and resolution in a principled way was proposed:

\begin{align}
\begin{split}
\text{depth:} &\mathcal{d}=\alpha^{\varphi} \\
\text{width:} &\mathcal{w}=\beta^{\varphi}\\
\text{resolution:} &\mathcal{r}=\gamma^{\varphi}\\
&s.t.  \alpha\cdot \beta^{2}\cdot \gamma^{2}\approx 2\\
&\alpha \ge 1, \beta \ge 1, \gamma \ge 1
\end{split}
 \label{eq:01-02-06}
\end{align}

where \(\alpha, \beta, \gamma\) are constants that can be determined by a small grid search, \(\varphi\) is a user-specified coefficient that controls how many more resources are available for model scaling, while \(\alpha, \beta, \gamma\) specify how to assign these extra resources to the network width, depth, and resolution, respectively. Notably, the FLOPS of a regular convolution operation is proportional to \(d, w^{2}, r^{2}\), i.e., doubling network depth will double the FLOPS, but doubling network width or resolution will increase the FLOPS by four times. Scaling a ConvNet following Equation \eqref{eq:01-02-06} will approximately increase the total number of FLOPS by \(\left( \alpha\cdot \beta^{2}\cdot \gamma^{2} \right)^{\varphi}\). In this chapter, \(\alpha\cdot \beta^{2}\cdot \gamma^{2}\approx 2\) is constrained such that for any new \(\varphi\) the total number of FLOPS will approximately increase by \(2\varphi\).

\hypertarget{efficientnet-architecture}{%
\subsubsection{EfficientNet Architecture}\label{efficientnet-architecture}}

A good baseline network is essential because model scaling does not affect its layer operators \(F*[i]\). Therefore this method is also estimated on ConvNets.
A new mobile-sized baseline called EfficientNet was developed to show the effectiveness of the new scaling method. Metrics that were used to estimate the efficacy are accuracy and FLOPS.
The baseline efficient network that was created is named EfficientNet-B0. Afterwards, this compound scaling method is applied in two steps:

\begin{itemize}
\item
  \textbf{STEP 1}: By fixing \(\varphi = 1\) and, assuming twice more resources available, a small grid search of $\alpha, \beta, \gamma$ based on Equation \eqref{eq:01-02-06} showed that the best values for EfficientNet-B0 are \(\alpha = 1.2, \beta = 1.1, \gamma=1.15\) under the constraint of \(\alpha\cdot\beta^2\cdot\gamma^2 \approx 2\).
\item
  \textbf{STEP 2}: Afterwards, fix \(\alpha,\beta,\gamma\) as constants and scale up the baseline network with different \(\varphi\) using Equation \eqref{eq:01-02-06} to construct EfficientNet-B1 to B7.
\end{itemize}

\begin{longtable}[]{@{}cc@{}}
\toprule()
Name & Number of parameters \\
\midrule()
\endhead
EfficientNet-B0 & 5.3M parameters \\
EfficientNet-B1 & 7.8M parameters \\
EfficientNet-B2 & 9.2M parameters \\
EfficientNet-B3 & 12M parameters \\
EfficientNet-B4 & 19M parameters \\
EfficientNet-B5 & 30M parameters \\
EfficientNet-B6 & 43M parameters \\
EfficientNet-B7 & 66M parameters \\
\bottomrule()
\end{longtable}

Indeed, even better performance is achievable by searching for \(\alpha,\beta,\gamma\) directly around a large model, but the search cost becomes prohibitively more expensive on larger models. This method searches once on a small baseline network, then scales the coefficient for all other models.

\hypertarget{results-and-comparison-of-the-networks}{%
\subsubsection{Results and comparison of the networks}\label{results-and-comparison-of-the-networks}}

To demonstrate the performance of both networks, ResNet and EfficientNets were trained and evaluated on the ImageNet 2012 classification dataset consisting out of 1000 classes. Since deeper scaling should provide better results in the case of ResNet, it was trained with increased depth each time. First meaningful results were obtained in ResNet-34, which performed 3.5 \% better than plain-34 baseline when top-1 accuracy is compared. They also compared three versions of ResNet: (A) zero-padding shortcuts (increasing dimensions, all shortcuts are parameter-free) (B) projection shortcuts (increasing dimensions, other shortcuts are identity), and (C) all shortcuts are projections. Each version improved both, the top-1 and top-5 accuracy. Afterward, the depth of the network was increased and ResNet-50, ResNet-101, and ResNet-152 were created. Each increase in depth leads to higher accuracy. In deeper models, the trade-off between accuracy increase and deeper model is not worth describing. All results are shown in the following table:

\begin{longtable}[]{@{}ccc@{}}
\toprule()
Model & top-1 acc. & top-5 acc. \\
\midrule()
\endhead
VGG-16 & 71.93 & 90.67 \\
GoogLeNet & - & 90.85 \\
plain-34 & 71.46 & 89.98 \\
ResNet-34 A & 74.97 & 92.24 \\
ResNet-34 B & 75.48 & 92.54 \\
ResNet-34 C & 75.81 & 92.6 \\
ResNet-50 & 77.15 & 93.29 \\
ResNet-101 & 78.25 & 93.95 \\
ResNet-152 & \textbf{78.57} & \textbf{94.29} \\
\bottomrule()
\end{longtable}

In the case of EfficientNets, the results achieved by the previous state-of-the-art networks on the same ImageNet dataset were aimed to improve. Among all state-of-the-art networks, EfficientNets were compared with ResNets-50 and ResNet-152. They compared the results of networks deviated by changing scaling parameters EfficientNet-B0 to EfficientNet-B7. The results of each network were better than the previous one. Also, they have shown that EfficientNet-B0 outperforms ResNet-50 and that EfficientNet-B1 outperforms ResNet-152. This means that scaling through all three dimensions can provide better results than scaling through just one dimension. The drawback of this approach is the computational power which makes it less popular than the previous methods. Again, all results are shown in the following table:

\begin{longtable}[]{@{}ccc@{}}
\toprule()
Model & top-1 acc. & top-5 acc. \\
\midrule()
\endhead
EfficientNet-B0 / ResNet-50 & 77.1 / 76 & 93.3 / 93 \\
EfficientNet-B1 / ResNet-152 & 79.1 / 77.8 & 94.4 / 93.8 \\
EfficientNet-B2 & 80.1 & 94.9 \\
EfficientNet-B3 / ResNeXt-101 & 81.6 / 80.9 & 95.7 / 95.6 \\
EfficientNet-B4 & 82.9 & 96.4 \\
EfficientNet-B5 & 83.6 & 96.7 \\
EfficientNet-B6 & 84 & 96.8 \\
EfficientNet-B7 / GPipe & \textbf{84.3} / 84.3 & \textbf{97} / 97 \\
\bottomrule()
\end{longtable}

\hypertarget{contrastive-learning}{%
\subsection{Contrastive learning}\label{contrastive-learning}}

In recent years the problem of classification of unlabeled dataset is becoming more widespread. More unlabeled datasets requiring human labeling are created in fields like medicine, the automotive industry, military, etc. Since the process is expensive and time-consuming, researchers assumed it could be automated with contrastive learning frameworks. One of the first and most known contrastive learning frameworks is SimCLR \citep{SimCLR}. The advantage of this framework is its simplicity, yet it achieves high accuracy on classification tasks. The main idea is to have two copies of the image, which are then used to train two networks and that are compared. The problem with this framework is that it doubles the size of the dataset and reaches among all images, which can be computationally infeasible for large datasets. Bootstrap Your Own Latent \citep{BYOL} was introduced to avoid making double-sized datasets. The idea was to bootstrap image representations to avoid unnecessary image comparisons. These two frameworks will be described in this chapter.

Further improvements in the choice of creating two views of images and comparison techniques were presented in different frameworks such as Nearest-Neighbor Contrastive Learning (NNCLR) \citep{NNCLR}, Open World Object Detection (ORE) \citep{ORE}, Swapping Assignments between multiple Views (SwAV) \{\citet{SwAV}\}, and many more.
This field is a constant research topic and new improved frameworks are proposed on a constant basis to help researchers solve different tasks that requires labeled datasets.

\hypertarget{a-simple-framework-for-contrastive-learning-of-visual-representations}{%
\subsubsection{A Simple Framework for Contrastive Learning of Visual Representations}\label{a-simple-framework-for-contrastive-learning-of-visual-representations}}

\citet{SimCLR} intended to analyze and describe a better approach to learning visual representations without human supervision. They have introduced a simple framework for contrastive learning of visual representations called SimCLR. As they claim, SimCLR outperforms previous work, is more straightforward, and does not require a memory bank.

Intending to understand what qualifies good contrastive representation learning, the significant components of the framework were studied and resulted in:

\begin{itemize}
\tightlist
\item
  A contrastive prediction task requires combining multiple data augmentation operations, which results in effective representations. Unsupervised contrastive learning benefits from more significant data augmentation.
\item
  The quality of the learned representations can be substantially improved by introducing a learn-able non-linear transformation between the representation and the contrastive loss.
\item
  Representation learning with contrastive cross-entropy loss can be improved by normalizing embeddings and adjusting the temperature parameter appropriately.
\item
  Unlike its supervised counterpart, contrastive learning benefits from larger batch sizes and extended training periods. Contrastive learning also benefits from deeper and broader networks, just as supervised learning does.
\end{itemize}

\hypertarget{the-contrastive-learning-framework}{%
\subsubsection{The Contrastive Learning Framework}\label{the-contrastive-learning-framework}}

\begin{figure}

{\centering \includegraphics[width=0.3\linewidth]{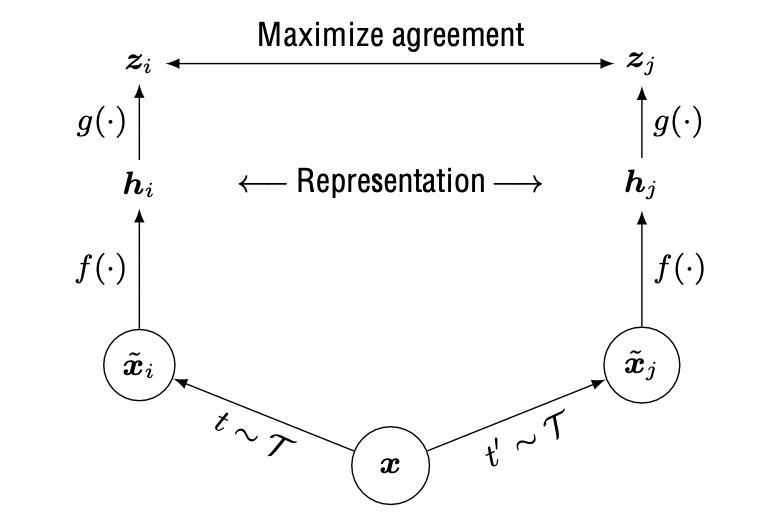}

}

\caption{A simple framework for contrastive learning of visual representations \citep{SimCLR}.}\label{fig:ch01-figure04}
\end{figure}

Like for SimCLR, a contrastive loss is used to learn a representation by maximizing the agreement between various augmented views of the same data example. This framework contains four significant components, which are shown in Figure \ref{fig:ch01-figure04}:

\begin{enumerate}
\def\labelenumi{\arabic{enumi}.}
\tightlist
\item
  A stochastic \emph{data augmentation} module
\item
  A neural network \emph{base encoder}
\item
  A small neural network \emph{projection head}
\item
  A \emph{contrastive loss function}
\end{enumerate}

\hypertarget{stochastic-data-augmentation-module}{%
\paragraph{Stochastic data augmentation module}\label{stochastic-data-augmentation-module}}

First, the minibatch of \(N\) examples is sampled randomly, and the contrastive prediction task is defined on pairs of augmented examples, resulting in \(2N\) data points. A memory bank was not used to train the model, instead, the training batch size varies from 256 to 8192. Any given data example randomly returns two correlated views of the same example, denoted \(\tilde{x}_{i}\) and \(\tilde{x}_{j}\), which is known as a \textbf{positive pair}. \textbf{Negative pairs} are all other \(2(N-1)\) pairs. In one view, some data augmentation techniques are applied. Data augmentation is widely embraced in supervised and unsupervised representation learning. Unfortunately, it has not been used to define the contrastive prediction task, which is mainly determined by changing the architecture. It was shown that choosing different data augmentation techniques can reduce the complexity of previous contrastive learning frameworks. There are many data augmentation operations, the focus was on the most common ones, which are:

\begin{itemize}
\tightlist
\item
  \textbf{Spatial geometric transformation}: cropping and resizing (with horizontal flipping), rotation and cutout,
\item
  \textbf{Appearance transformation}: color distortion (including color dropping), brightness, contrast, saturation, Gaussian blur, and Sobel filtering.
\end{itemize}

\begin{figure}

{\centering \includegraphics[width=0.8\linewidth]{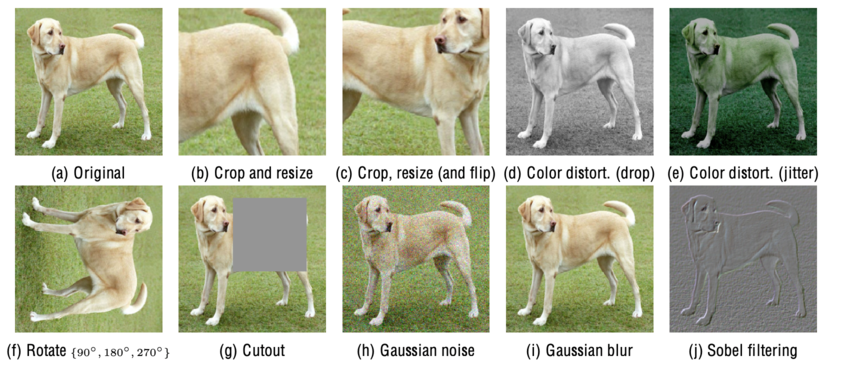}

}

\caption{Augmentation texhniques \citep{SimCLR}.}\label{fig:ch01-figure05}
\end{figure}

Due to the image sizes in the ImageNet dataset, all images were always randomly cropped and resized to the same resolution. Later on, other targeted data augmentation transformations were applied to one branch, remaining the one as original i.e.~\(t\left( x_{i}\right)= x_i\).
Applying just individual transformation is insufficient for the model to learn good representations. The model's performance improves after composing augmentations, although the contrastive prediction task becomes more complex. The composition of augmentations that stood out were random cropping and random color distortion.

It was also observed that stronger color augmentation significantly improves the linear evaluation of unsupervised learned models. Stronger color augmentations do not enhance the performance of supervised learning models when trained with the same augmentations. Based on the experiments, unsupervised contrastive learning benefits from stronger color data augmentation than supervised learning.

\hypertarget{neural-network-base-encoder}{%
\paragraph{Neural network base encoder}\label{neural-network-base-encoder}}

Neural network based encoder \(f\left( \cdot \right)\) extracts multiple representation vectors from the augmented data examples. This framework does not restrict a choice of the network architecture, although for simplicity, the commonly used ResNet was picked and gives \(h_i=f\left( \tilde{x}_{i} \right)=ResNet\left(\tilde{x}_{i}\right)\) where \(\textbf{h}_i\in \mathbb{R}^{d}\) is the output after the average pooling layer. Although increasing depth and width improves performance, the ResNet-50 was chosen. Furthermore, when the model size increases, the gap between supervised and unsupervised learning shrinks, suggesting that bigger models benefit more from unsupervised learning.

\hypertarget{small-neural-network-projection-head}{%
\paragraph{Small neural network projection head}\label{small-neural-network-projection-head}}

A small neural network projection head \(g\left( \cdot \right)\) maps the representation to the space where the contrastive loss is applied to. The importance of including a projection head, i.e., \(g\left( h \right)\) was evaluated and they considered three different architectures for the head:

\begin{enumerate}
\def\labelenumi{\arabic{enumi}.}
\tightlist
\item
  identity mapping,
\item
  linear projection,
\item
  the default non-linear projection with one additional hidden layer and ReLU activation function.
\end{enumerate}

The results showed that a non-linear projection head is better than a linear projection and much better than no projection. It improves the representation quality of the layer that is applied previous to it. They have used a MLP with one hidden layer to obtain \(z_i = g\left( \textbf{h}_i \right) = W^{\left( 2\right)}\sigma \left( W^{\left( 1\right)} \textbf{h}_i\right)\) where \(\sigma\) is a ReLU non-linearity transformation.

This step is performed because defining the contrastive loss on \(z_i\) instead of on \(\textbf{h}_i\) would not lead to a loss of information caused by contrastive loss. Especially, \(z=g\left( h \right)\) is trained to be invariant to data transformations. As a result, \(g\) can remove information useful for a downstream task such as object color or orientation. Using the non-linear transformation \(g\left( * \right)\), \(h\) can maintain and form more information.

\hypertarget{contrastive-loss-function}{%
\paragraph{Contrastive loss function}\label{contrastive-loss-function}}

Given a set \(\left\{ \tilde{x}_{ik} \right\}\) including a positive pair of examples \(\tilde{x}_{i}\) and \(\tilde{x}_{j}\), the contrastive prediction task aims to identify \(\tilde{x}_{i}\) in \(\left\{ \tilde{x}_{i} \right\}_{k\neq i}\) for a given \(\tilde{x}_{i}\). In the case of positive examples, the loss function is defined as

\[
\mathcal{l}_{i,j} = -\log\frac{\operatorname{exp}\left( \frac{\operatorname{sim}(z_i,z_j)}{\tau} \right)}{\sum_{k=1}^{2N}\mathbb{I}_{\left[ k\neq i \right]}\operatorname{exp}\left( \frac{\operatorname{sim}(z_i,z_k)}{\tau} \right)}
\]

where \(\mathbb{I}_{\left[ k\neq i \right]}\in\left\{ 0,1 \right\}\) is an indicator function, \(\tau\) denotes a temperature parameter and \(\operatorname{sim}\left(\textbf{u,v} \right)= \frac{\textbf{u}^T\textbf{v}}{\left\| \textbf{u}\right\|\left\| \textbf{v} \right\|}\) is a dot product between \(\mathcal{l}_2\) and normalized \(\textbf{u},\textbf{v}\).

The final loss is calculated across all positive pairs, both \(\left( i,j \right)\) and \(\left( j,i \right)\), in a mini-batch. It was named \textbf{NT-Xent}, the normalized temperature-scaled cross-entropy loss.

The NT-Xent loss was compared against other commonly used contrastive loss functions, such as logistic loss and margin loss. Gradient analysis shows that \(l_2\) normalization, cosine similarity, and temperature together effectively weight different examples and a suitable temperature can make the model learn from hard negatives. The advantage of NT-Xent is that it weights the negatives by their relative hardness. Without normalization and proper temperature scaling the performance is significantly worse. Also, the contrastive task accuracy is higher, but the resulting representation is worse under linear evaluation.

\hypertarget{bootstrap-your-own-latent}{%
\subsubsection{Bootstrap Your Own Latent}\label{bootstrap-your-own-latent}}

The fundamental idea of contrastive learning is to create pairs of images on which the framework would be trained. Creating negative pairs relies on large batch sizes, memory banks, or customized mining strategies which can be challenging in larger datasets. \citet{BYOL} wanted to create a new approach that would achieve better performance than other contrastive methods without using negative pairs. A solution they have introduced is a method called Bootstrap Your Own Latent (BYOL). The idea was to bootstrap representations of images. As a result, BYOL is more robust to the choice of image augmentations.
Furthermore, BYOL has two neural networks, called online and target network, who interact and learn from each other. Using an augmented view of an image, BYOL trains its online network to predict the target network's representation of another augmented view. This approach achieved state-of-the-art results when trained on the ImageNet dataset under the linear evaluation protocol. Additionally, compared to SimCLR, a strong contrastive baseline, BYOL suffers from much less performance drop when only random crops are used to augment images.

\hypertarget{description-of-the-method}{%
\paragraph{Description of the method}\label{description-of-the-method}}

BYOL aims to learn a representation of \(y_\theta\). It uses two neural networks: \emph{online} and \emph{the target network} to achieve that. The \emph{online network} is determined by a set of weights \(\theta\) and consists of:

\begin{itemize}
\tightlist
\item
  an encoder \(f_\theta\),
\item
  a projector \(g_\theta\),
\item
  a predictor \(q_\theta\).
\end{itemize}

\begin{figure}

{\centering \includegraphics[width=0.8\linewidth]{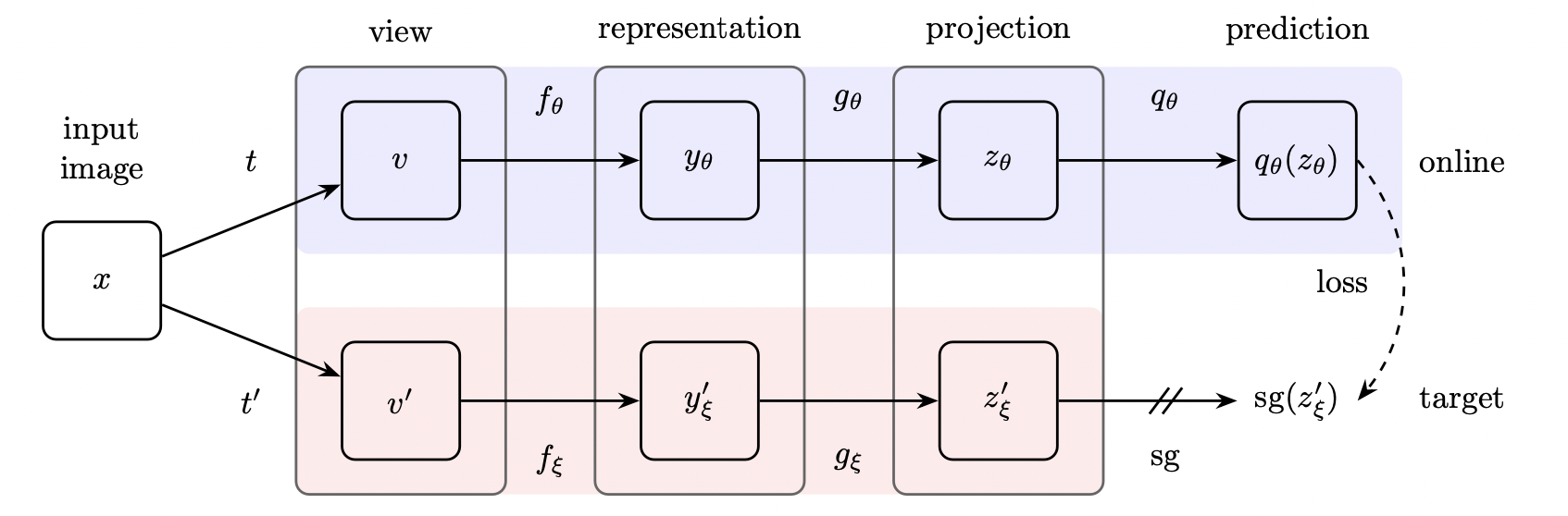}

}

\caption{Bootstrap Your Own Latent \citep{BYOL}.}\label{fig:ch01-figure06}
\end{figure}

The \emph{target network} has the same architecture as the online network but uses different weights \(\xi\). It provides the regression targets to train the online network, and its parameters \(\xi\) are an exponential moving average of the online parameters \(\theta\). Precisely, given a target decay rate \(\tau \in[0,1]\), after each training step, the following update

\[
\xi \leftarrow \tau \xi+(1-\tau) \theta
\]

is performed.
Firstly, an image is sampled uniformly from \(\mathcal{D}\) from which two distributions of image augmentations \(\mathcal{T}\) and \(\mathcal{T}^{\prime}\) are created. BYOL applies respectively two image augmentations \(t \sim \mathcal{T}\) and \(t^{\prime} \sim \mathcal{T}^{\prime}\) creating two augmented views \(v \triangleq t(x)\) and \(v^{\prime} \triangleq t^{\prime}(x)\). First augmented view \(v\) is used for the online network and result in the output \(y_{\theta} \triangleq f_{\theta}(v)\) and afterwards the projection \(z_{\theta} \triangleq g_{\theta}(y)\). Similarly, from the second augmented view \(v^{\prime}\) the target network outputs \(y_{\xi}^{\prime} \triangleq f_{\xi}(v^{\prime})\) and the target projection \(z_{\xi}^{\prime} \triangleq g_{\xi}(y^{\prime})\). Later on output a prediction of \(q_{\theta}\left(z_{\theta}\right)\) of \(z_{\xi}^{\prime}\) and \(\ell_{2}\)-normalize both \(q_{\theta}\left(z_{\theta}\right)\) and \(z_{\xi}^{\prime}\) to

\[
\overline{q_{\theta}}\left(z_{\theta}\right) \triangleq q_{\theta}\left(z_{\theta}\right) /\left\|q_{\theta}\left(z_{\theta}\right)\right\|_{2} \quad \textrm{and} \quad
\bar{z}_{\xi}^{\prime} \triangleq z_{\xi}^{\prime} /\left\|z_{\xi}^{\prime}\right\|_{2}.
\]

The predictor is only applied to the online pipeline, making the architecture asymmetric between the online and target pipeline. Lastly, the following mean squared error between the normalized predictions and target projections is defined:

\[
\mathcal{L}_{\theta, \xi} \triangleq\left\|\overline{q_{\theta}}\left(z_{\theta}\right)-\bar{z}_{\xi}^{\prime}\right\|_{2}^{2}=2-2 \cdot \frac{\left\langle q_{\theta}\left(z_{\theta}\right), z_{\xi}^{\prime}\right\rangle}{\left\|q_{\theta}\left(z_{\theta}\right)\right\|_{2} \cdot\left\|z_{\xi}^{\prime}\right\|_{2}}
\]

The loss is symmetrized \(\mathcal{L}_{\theta, \xi}\) by using \(v^{\prime}\) for the online network and \(v\) for the target network separately to calculate \(\widetilde{\mathcal{L}}_{\theta, \xi}\). At each training step, a stochastic optimization step is applied to minimize \(\mathcal{L}_{\theta, \xi}^{\mathrm{BYOL}}=\mathcal{L}_{\theta, \xi}+\widetilde{\mathcal{L}}_{\theta, \xi}\) with respect to \(\theta\) only but not \(\xi\). BYOL's dynamics are summarized as

\[
\theta \leftarrow \operatorname{optimizer}\left(\theta, \nabla_{\theta} \mathcal{L}_{\theta, \xi}^{\mathrm{BYOL}}, \eta\right).
\]

where \(\eta\) is a learning rate. At the end of the training, only the encoder \(f_{\theta}\) is used.

\hypertarget{comparison-of-contrastive-learning-frameworks}{%
\subsubsection{Comparison of contrastive learning frameworks}\label{comparison-of-contrastive-learning-frameworks}}

Of all frameworks, SimCLR is the most popular due to its simplicity. The ResNet-50 in 3 different hidden layer widths (width multipliers of \(1\times\), \(2\times\), and \(4\times\)) were used and trained for 1000 epochs each. The accuracy of these frameworks on the ImageNet dataset with few labels improved when the width of ResNet-50 increases. For SimCLR with ResNet-50 top-1 accuracy is 69.3 and top-5 accuracy is 89, while for ResNet-50(4x) top-1 accuracy is 85.8 and top-5 accuracy is 92.6. These results are comparable with supervised methods.
The BYOL framework was built to improve the results of SimCLR. It was also stated that the accuracy for the baseline ResNet-50 is 74.3 and 91.6 for top-1 accuracy and top-5 accuracy. When using ResNet-50(4x), an increase in accuracy to 78.6 and 94.2 for top-1 and top-5 is observed, respectively. More information about performance can be found in following table:

\begin{longtable}[]{@{}ccccc@{}}
\toprule()
Model & Architecture & Param (M) & top-1 acc. & top-5 acc. \\
\midrule()
\endhead
SimCLR & ResNet-50 & 24 & 69.3 & 89.0 \\
SimCLR & ResNet-50 (2x) & 94 & 74.2 & 93.0 \\
SimCLR & ResNet-50 (4x) & 375 & 76.5 & 93.2 \\
BYOL & ResNet-50 & 24 & 74.3 & 91.6 \\
BYOL & ResNet-50 (x2) & 94 & 77.4 & 93.6 \\
BYOL & ResNet-50 (x4) & 375 & 78.6 & 94.2 \\
BYOL & ResNet-200 (x2) & 250 & 79.6 & 94.8 \\
\bottomrule()
\end{longtable}

\hypertarget{transformers-in-computer-vision}{%
\subsection{Transformers in Computer Vision}\label{transformers-in-computer-vision}}

Since the first appearance of the Transformers architecture in 2017 \citet{TRANSFORMERS_NLP}, it has become an irreplaceable part of all-natural language processing (NLP) models. The main advantage of Transformers is that they can be trained on a large text corpus and then fine-tuned on a smaller task-specific dataset. This enabled model training of unspecified size with more than 100B parameters.

However, computer vision still relied on convolutional architectures. With datasets constantly growing and the diversity of the fields computer vision tasks could be applied to, researchers wanted to implement Transformers architecture in the CV field. Some works aim for combining CNN-like architectures with self-attention \citep{wang}. Others attempted to replace convolutions entirely, e.g. \citet{selfa}. Due to specialized attention patterns, the problem was that they have not yet been scaled effectively on modern hardware accelerators. Therefore, in large-scale image recognition, classic ResNet-like architectures are still state-of-the-art.

In 2021 the Google research Brain Team published the paper ``An image is worth \(16\times 16\) words'' where they introduced new Transformers-based architecture for CV called Vision Transformers (ViT) \citep{vit}. Based on the success of Transformer in NLP scaling, they aimed to apply standard Transformer directly to images with little as possible changes to the existing architecture. The image is split into patches and linear embeddings of these patches are provided as inputs to the Transformer.
These patches are the same as tokens (e.g.~words) in NLP. The model is trained for image classification in a supervised learning fashion.

\hypertarget{vision-transformers}{%
\subsubsection{Vision Transformers}\label{vision-transformers}}

Brain Team wanted to create simple but universally scalable architecture to follow the original Transformers architecture.

\begin{figure}

{\centering \includegraphics[width=0.9\linewidth]{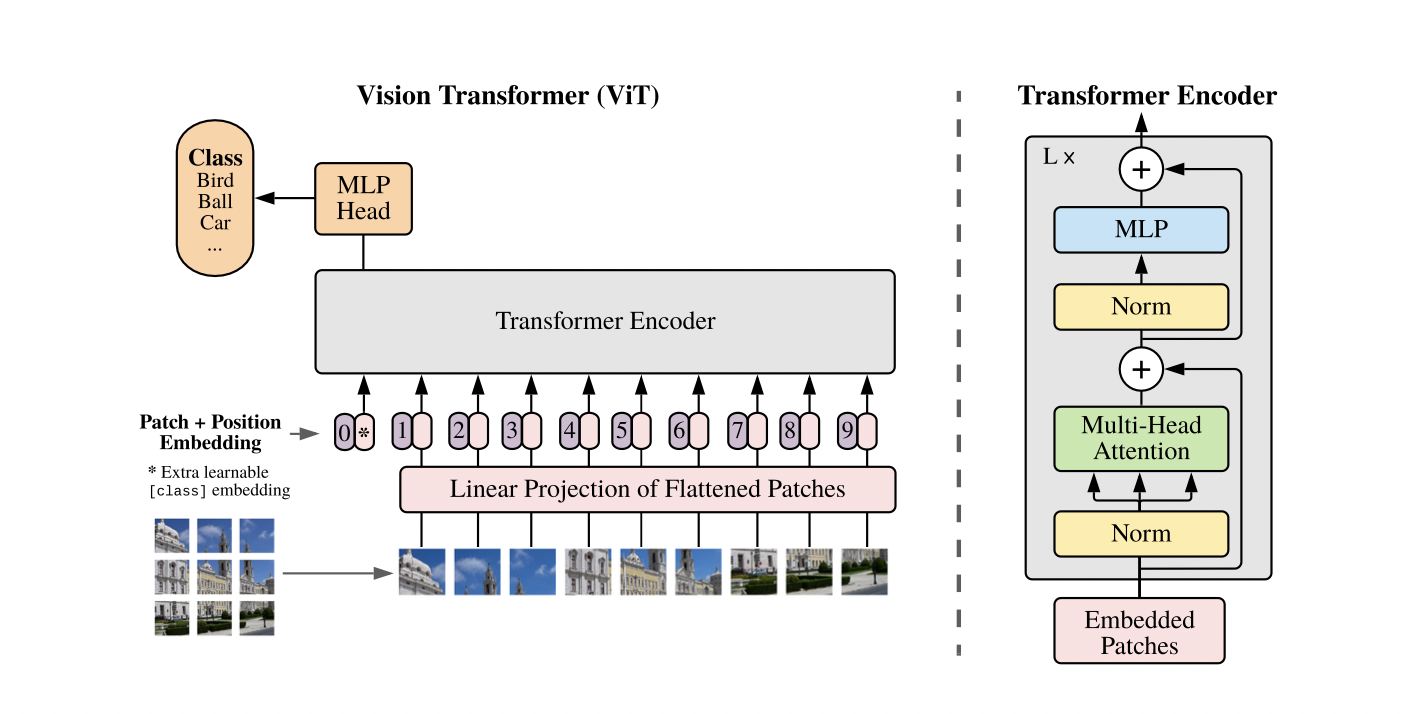}

}

\caption{Vision Transformer \citep{vit}.}\label{fig:ch01-figure7}
\end{figure}

\hypertarget{method}{%
\paragraph{Method}\label{method}}

Compared to NLP, with 1-dimensional token embedding input for the Transformer, images are 2-dimensional objects. Firstly, images needed to be represented differently to imitate original architectures as close as possible. For that reason image \(x\in \mathbb{R}^{ H \times W \times C}\) is reshaped into a sequence of flattened 2-dimensional patches \(x_p\in \mathbb{R}^{ N \times \left( P^2 \cdot C \right)}\), where \(\left(H,W\right)\) is the resolution of the original image, \(C\) is the number of channels, \(\left(P,P\right)\) is the resolution of each image patch, and \(N =HW/P^2\) is the resulting number of patches, also the Transformer's effective input sequence length. The Transformer input through all layers is a fixed vector of size \(D\). The first step is to flatten the patches, usually \(16\times 16\) and map them to \(D\) dimensions with a trainable linear projection to create patch embeddings.

\[
\mathbf{z}_{0} =\left[\mathbf{x}_{\text {class }} ; \mathbf{x}_{p}^{1} \mathbf{E} ; \mathbf{x}_{p}^{2} \mathbf{E} ; \cdots ; \mathbf{x}_{p}^{N} \mathbf{E}\right]+\mathbf{E}_{p o s}, \mathbf{E} \in \mathbb{R}^{\left(P^{2} \cdot C\right) \times D}, \mathbf{E}_{p o s} \in \mathbb{R}^{(N+1) \times D}
\]

To this sequence of ``patch embeddings'', a prefix learnable {[}class{]} token, like in BERT, is usually added. This token \(\mathbf{z}_{0}^{0} = \mathbf{x}_{class}\) tells the model to classify the image and increases the dimension of vector \(z\). Also, the state of this token at the output of the Transformer encoder \(\left(\mathbf{z}_{L}^{0}\right)\), on which the layernorm is applied, serves as the image representation \(y\).

\[
\mathbf{y} =\operatorname{LN}\left(\mathbf{z}_{L}^{0}\right)
\]

Furthermore, it is the only one to which the classification head is attached to during pre-training and fine-tuning. The classification head during pre-training is compiled of MLP with one hidden layer and a single linear layer at a fine-tuning time. Position embedding, a standard learnable 1-dimensional position embedding, are attached to the patch embeddings, serving as input to the encoder. The standard Transformer encoder consists of alternating layers of multiheaded self-attention and MLP blocks. After each block, a residual connection is applied.

\[
\mathbf{z}_{\ell}^{\prime} =\operatorname{MSA}\left(\operatorname{LN}\left(\mathbf{z}_{\ell-1}\right)\right)+\mathbf{z}_{\ell-1},  \ell=1 \ldots L
\]

\[
\mathbf{z}_{\ell} =\operatorname{MLP}\left(\mathrm{LN}\left(\mathbf{z}_{\ell}^{\prime}\right)\right)+\mathbf{z}_{\ell}^{\prime}, \ell=1 \ldots L
\]

Vision Transformer has a significantly lower inductive bias than CNNs in image-specific information. VIT only has local and translational equivariant MLP layers, while the self-attention layers are global. A 2-dimensional neighborhood structure is used sparingly: the image is cut into patches at the beginning and the position embeddings are resized as needed at the fine-tuning time. Alternatively, the input sequence can consist of a CNN's feature map on which the patch embedding projection is applied.
Vision Transformers are pre-trained on large datasets and fine-tuned to (smaller) downstream tasks. For fine-tuning, a projection head is removed and a zero-initialized \(D \times K\) feedforward layer is attached with \(K\) being the number of downstream classes. It is also beneficial to use higher resolution then in pre-training. Also ViT can handle arbitrary sequence lengths but the pre-trained position embeddings can become sufficient. It is necessary to point out that resolution adjustment and patch extraction are the only points at which an inductive bias about the 2-dimensional structure of the images is manually injected into the Vision Transformers

\hypertarget{experiments}{%
\paragraph{Experiments}\label{experiments}}

Similarly to BERT models, multiple versions of the model at various scales were created. They have created Base = ``B'', Large = ``L'', Huge = ``H'' versions of ViT, with 12, 24 and 32 layers and 86M, 307M and 632M parameters respectively.

To explore the model scalability, the previous mentioned dataset ImageNet was used. In addition, ViT was compared against a slightly modified ResNet called ``ResNet(BiT)''. The batch Normalization layer was replaced with Group Normalization and used standardized convolutions. Another network that it was compared to was Noisy Student \citep{noisy}, a large EfficientNet. Experiments showed that ViT Hughe with \(14\times 14\) input patch size outperformed both CNN-based networks with an accuracy of 88.5\%, whereas ResNet BiT had 87.54\% and Noisy Student 88.4\%. It is worth mentioning that ViT Large with \(16\times 16\) input patch size had 87.76\% accuracy on the same dataset.
Another thing worth pointing out is that ViT outperforms CNN-based architectures on all larger datasets yet performs slightly worse than CNN networks on a smaller dataset.

\hypertarget{conclusion}{%
\subsection{Conclusion}\label{conclusion}}

In this chapter, the authors presented some of the current state-of-the-art approaches in Computer Vision. Nowadays, when technology is advancing each day, creating networks that would imitate human brain is more challenging. Still, the networks presented in this chapter are highly accurate and creating network which can out-perform them is challenging. Furthermore, it is noticeable that the application of CV is dictating the development of networks and frameworks which help humans with their everyday tasks.

\hypertarget{c01-03-benchmarks}{%
\section{Resources and Benchmarks for NLP, CV and multimodal tasks}\label{c01-03-benchmarks}}

\emph{Author: Christopher Marquardt}

\emph{Supervisor: Christian Heumann}

When we see athletes perform in their sports we only see the results of their hard work prior or till to the event. Most of the time they casually talk about their off-season, but everybody knows the results are made in the off-season.

Same goes for the models we will see in the later chapters. We are just interested in the results, but why and how does the model come to these results? It has to learn to some key fundamentals of the modality to achieve these results. But how do they get them to perform in such a way or even better? It's possible to build better architectures and/or use more and new data to achieve this. New data by hand is easy to get but this new data results in a new problem. New data has to be carefully labeled by humans, which can be very expensive by the amount of data. Models which learn from labeled data use the supervised learning strategy. This learning strategy is a bottleneck for future progress, because of the given reason.

But the need for labeling the data isn't the only problem. Let's visit the athlete analogy again. Imagine a professional football player has to participate in a professional ski race. He will not be able to compete with the others, because they are trained only to do ski races. Here see the other problem. Models which use supervised learning have shown to perform very well on the task they are trained to do. This means models which learn on carefully labeled data only perform very well on this specific task, but poor on others. Also it's not possible to label everything in the world.

So the goal is to generate more generalist models which can perform well on different tasks without the need of huge labeled data. Humans are able to perform well on different tasks in a short amount of time. Humans, for example, only need a small amount of hours to learn how to drive a car, even without supervision. On the other hand fully automated driving AI need thousand of hours of data to drive a car. Why do humans learn so fast compared to machines?
Humans don't rely on labeled data, because most of the time humans learn by observation. By this humans generate a basic knowledge of how the world works, which also called common sense. This enables us to learn so much faster compared to machines.
Meta AI \citep{darkMatter} believes that self-supervised learning is one of the most promising ways to generate background knowledge and some sort of common sense in AI systems. By self-supervised learning one means a supervised learning algorithm, but it doesn't need an external supervisor. Self-supervised pre-training differs between the modalities, which means there is not an approach which works in all modalities.
The following chapter will inspect on the one hand pre-training resources and the use of them and on the other hand also the benchmarks which are used for Natural Language Processing (NLP), Computer Vision (CV) and ,the combination of both, vision language pre-trained models (VL-PTM).

\hypertarget{datasets}{%
\subsection{Datasets}\label{datasets}}

After pointing out that pre-training is very important, one might ask how do the datasets look and how do the different modalities pre-train? At first we will inspect the former one and focus afterwards on the use of the resources. As one might expect NLP models pre-train on text, CV models pre-train on images and VL-PTM pre-train on text image pairs, which can somehow be seen as a combination of NLP and CV. But CV models mostly used labeled data like a picture of a dog with the corresponding single label ``dog''. MML datasets can contain several sentences of text which correspond to the given image.

Even if the datasets might be completely different, the procedure to get the data is mostly the same for all of them, because the data is crafted from the internet. This can lead to a problem, since by using this method the resulting dataset might be noisy. One approach for the VL-PTM, for example, is to use CommonCrawl and extract the image plus the alt of an image. The alt is an alternate text for an image, if the image cannot be displayed or for visual impaired people. This seems like a reasonable approach, but the alt is often not very informative about what's in the image.

Another difference between the modalities is the cardinality of the pre-training data. It's easy to realize that text is by far easiest to crawl from the internet. This results in huge high-quality massive text data. Some magnitudes smaller are the datasets for CV. Since VL-PTM are pretty new compared to the other modalities it still relatively small, but growing fast. A small downer is that some of the datasets are not public available. The big companies like to keep their models and used datasets private, which hinders the reproducibility, but there are also real open AI competitors like LAION and Eleuther in the field. The next chapter will provide some of the most used pre-training datasets.

\hypertarget{natural-language-processing-datasets}{%
\subsubsection{Natural Language Processing Datasets}\label{natural-language-processing-datasets}}

\hypertarget{common-crawl}{%
\paragraph{Common Crawl}\label{common-crawl}}

As already mentioned, extracting text from the internet is rather easy. More precisely there is a non-profit organization, called \href{https://commoncrawl.org}{Common Crawl}, which does exactly this. They provide copies of the internet to researchers, companies and individuals at no cost for the purpose of research and analysis. The Common Crawl corpus contains petabytes of data collected since 2008. Every month, Common Crawl releases a snapshot of the web obtained by randomly exploring and sampling URLs. It contains raw web page data, extracted metadata and text extractions. The advantages of Common Crawl come along with their disadvantages. The text is from diverse domains but with varying quality of data. To handle the raw nature of the datasets one often has to use a well-designed extraction and filter to use the datasets appropriately \citep{gao2020pile}. GPT-3 ,for example, uses a filtered version of Common Crawl, which consists of 410 billion tokens \citep{brown2020language}. So data for NLP is freely available but one needs to use well-designed extraction and filtering to really use the dataset.

\hypertarget{the-pile}{%
\paragraph{The Pile}\label{the-pile}}

Recent work \citep{rosset2020turing} showed that diversity in training datasets improves general cross-domain knowledge and downstream generalization capability for language models. The Pile \citep{gao2020pile} was introduced to address exactly these results. The Pile contains \(22\) sub-datasets, including established NLP datasets, but also several newly introduced ones. The size of the \(22\) sub-datasets, which can be categorized roughly into five categories, pile up to around \(825\) GB of data.
The following treemap shows the distribution of the dataset.

\includegraphics{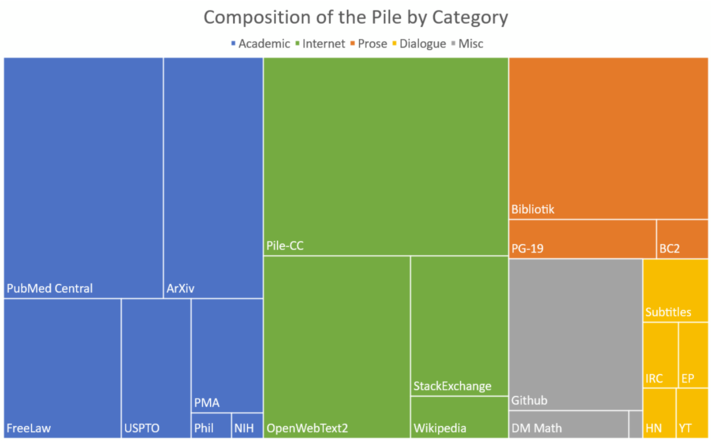}

While only 13\% of the world's population speaks English, the vast majority of NLP research is done on English. \citet{gao2020pile} followed this trend, but did not explicitly filtered out other languages when collecting our the data. This leads to the fact that roughly 95\% of the Pile is English. Also EuroParl \citep{koehn2005europarl}, a multilingual parallel corpus introduced for machine translation, is included in the Pile. To train GPT-2 Open AI collected data from WebText. WebText is an internet dataset created by scraping URLs extracted from Reddit submissions with a minimum score for quality, but sadly it was never released to the public. Independent researchers reproduced the pipeline and released the resulting dataset, called OpenWebTextCorpus \citep{Gokaslan2019OpenWeb} (OWT). Eleuther created an enhanced version of the original OWT Corpus called OpenWebText2. It covers all Reddit submissions from 2005 up until April 2020. It covers content from multiple languages, document metadata, multiple dataset versions, and open source replication code.

They also explicitly included a dataset of mathematical problems (DeepMind Mathematics) to improve the mathematical ability of language models trained on the Pile. An ArXiv dataset was in included in the hopes that it will be a source of high quality text and math knowledge, and benefit potential downstream applications to research in these areas and also because arXiv papers are written in LaTeX. Training a language model to be able to generate papers written in LaTeX could be a huge benefit to the research community.

Since CC needs further steps, due to the raw nature of CC, to really use is. Pile-CC is Common Crawl-based dataset, which can be used directly. It yields higher quality output than directly using the WET files. These were only some of the 22 included datasets. A more detailed description of the sub-dataset and the reasons why these were included can be found in the corresponding paper \citep{gao2020pile}.

\hypertarget{multilingual-datasets}{%
\paragraph{Multilingual Datasets}\label{multilingual-datasets}}

Another pre-cleaned version of CC is CC-100 \citep{wenzek2019ccnet}. They present a pipeline to create curated monolingual corpora in more than 100 languages. A filter, which covers the data based on their distance to Wikipedia, is used and this improves the quality of the resulting dataset. However, its English portion is much smaller than the Pile. But a multilingual dataset might help a low-resource language acquire extra knowledge from other languages.
Perhaps the most multilingual corpus publicly available, containing 30k sentences in over 900 languages, is the Bible corpus \citep{mayer2014creating}.
Till now all datasets were freely available and almost directly usable. The next one is not public available for some reasons.

To provide mT5 \citep{xue2020mt5}, which is multilingual pre-trained text-to-text transformer, a suitable pre-training dataset, Google Research designed a dataset including more than 100 languages. The dataset is called mC4 \citep{xue2020mt5}. Since some languages are relatively scarce on the internet, they used all of the 71 monthly web scrapes released so far by Common Crawl. It contains 6.6 billion pages and 6.3 trillion tokens. A smaller version of the mC4 is also used by Google Research. The smaller dataset C4 (Colossal Clean Common Crawl) was explicitly designed to be English only. The C4 dataset is a collection of about \(750\)GB of English-language text sourced from the public Common Crawl web.

Most of the datasets used in NLP are derived entirely from Common Crawl and \citet{rosset2020turing} came to the result, that the current best practice in training large-scale language models involve using both large web scrapes and more targeted, higher-quality datasets, which the Pile directly addresses.

\hypertarget{bookscorpus}{%
\paragraph{BooksCorpus}\label{bookscorpus}}

The last dataset for NLP is the BooksCorpus dataset \citep{zhu2015aligning}. The BooksCorpus uses books from yet unplished authors from the web. Only books with more than 20k words were included to filter out shorter, noisier stories. This results in around 11k books from 16 different genres. So more than 74 million sentences can be used in pre-training. BooksCorpus contains a sample of books from \href{https://www.smashwords.com}{a distributor of indie ebooks}. Sadly a datasheet about the BooksCorpus was not releasd with the corresponing paper.

Frankly, there was just an paragraph about the content and the extraction inside the paper \citep{zhu2015aligning}. \citet{bandy2021addressing} addressed exactly this short coming. They provided a retrospective datasheet about the BooksCorpus. Some of their major concerns were copyright violations, duplicate books, skewed genre representation, potentially skewed religious representation and also problematic content (18+ content). Little harm can be expected if an informed adults reads books with these concers, but how does a language model contribute to for example well-documented gender discrimination if it trains on these books.

Since BookCorpus is no longer distributed, one has to visit the distributor of the \href{https://www.smashwords.com}{indie ebooks} and collect a own version of the BookCorpus. This is one of the user-based dataset, besides to the datasets of the Pile.

\hypertarget{computer-vision-dataset}{%
\subsubsection{Computer Vision Dataset}\label{computer-vision-dataset}}

\hypertarget{imagenet}{%
\paragraph{ImageNet}\label{imagenet}}

The next inspected modality is CV. Almost every state-of-the-art CV model uses a classifier pre-trained on an ImageNet based dataset. ImageNet uses the hierarchical structure of WordNet \citep{fellbaum2010wordnet}. At the release of ImageNet-1k the amount of classes was unheard at this time point. Datasets like CIFAR-10 \citep{krizhevsky2009learning} and CIFAR-100 \citep{krizhevsky2009learning} had 10 or 100 classes, but ImageNet1k had 1000 different classes and this was not the only major improvement. They also increased the resolution from \(32 \times 32\) to \(256 \times 256\). In all, there are roughly 1.2 million training images, 50,000 validation images, and 150,000 testing images. The ImageNet-1k dataset is a subset of the ImageNet dataset \citep{deng2009imagenet}. The full ImageNet dataset is also called ImageNet-21k. It consists of more than 14 million images, divided in almost 22k classes. Because of this some paper described it as ImageNet-22k.

Those two dataset do not only differ by the amount of classes, but also by the type of labels. The labels of ImageNet-21k are not mutually exclusive. Because of this the pre-training wiht ImageNet-1k is far more popular. Also the ImageNet-21k dataset lacks an official train-validation split, which is just another reason why ImageNet-1k is more popular. The raw dataset ImageNet-21k is around 1.3 terabyte (TB). It's also nice, that the the dataset of ImageNet are open available. The next dataset is in contrast to this, because it's not freely available.

\hypertarget{joint-foto-tree-jft-entity-foto-tree-eft}{%
\paragraph{Joint-Foto-Tree (JFT) \& Entity-Foto-Tree (EFT)}\label{joint-foto-tree-jft-entity-foto-tree-eft}}

The Joint-Foto-Tree (JFT) 300M is one of the follow up version of the JFT dataset \citep{hinton2015distilling}. Given the name it consists of 300 million images and on average each image has 1.26 labels. The whole datasets has around 375 million labels. These labels can be divided into 18291 classes. These categories form a rich hierarchy with the maximum depth of hierarchy being 12 and maximum number of child for parent node being 2876 \citep{sun2017revisiting}. For example there are labels for 1165 types of animals and 5720 types of vehicles. The work states that approximately 20\% of the labels in this dataset are noisy \citep{sun2017revisiting}, because the labels are generated automatically.

It also provides the fact, that the distribution is heavily long-tailed, which means that some of the classes have less than 100 images. There is also an extendend version of the JFT dataset.

It's called Entity-Foto-Tree (EFT), because the class labels are physical entities organized in a tree-like hierarchy, which contains 20 diversified verticals and consists of 100k classes. It's even rarely used in practice by Google because of the intolerable large model size and the slow training speed \citep{gao2017knowledge}. Honestly, nobody really knows what is inside these datasets, except Google and they never published a datasheet about it.

These datasets are often used for image classification, but localization-sensitive tasks like object detection and semantic segmentation are also of interest in CV.

\hypertarget{objects365}{%
\paragraph{Objects365}\label{objects365}}

Objects365 \citep{shao2019objects365} is a large-scale object detection and semantic segmentation freely available dataset. It contains 365 object categories with over 600K training images. More than 10 million, high-quality bounding boxes are manually labeled through a three-step, carefully designed annotation pipeline. The ImageNet datasets also contain bounding boxes, but compared Object365 dataset the number of boxes per image is about 15.8 vs 1.1 \citep{deng2009imagenet}. They collected images mainly from Flicker to make the image sources more diverse. All the images conform to licensing for research purposes. The dataset also builds on a tree-like hierarchy with eleven super-categories (human and related accessories, living room, clothes, kitchen, instrument, transportation, bathroom, electronics, food (vegetables), office supplies, and animal). Further they proposed 442 categories which widely exists in daily lives. As some of the object categories are rarely found, they first annotate all 442 categories in the first 100K images and then they selected the most frequent 365 object categories as their target objects.

To enable compatibility with the existing object detection benchmarks, the 365 categories include the categories defined in Microsoft Common Objects in Context (COCO) \citep{lin2014microsoft}, which is described in the next paragraph.

\hypertarget{microsoft-common-objects-in-context-coco}{%
\paragraph{Microsoft Common Objects in Context (COCO)}\label{microsoft-common-objects-in-context-coco}}

Microsoft decided to employed a novel pipeline for gathering data with extensive use of Amazon Mechanical Turk. Their goal was to create a non-iconic image collection. Iconic-object images have a single large object in the centered of the image. By this they provide high quality object instances, but they also lack information of contextual important and non-canonical viewpoints \citep{lin2014microsoft}. Recent work showed that non-iconic images are better at generalizing \citep{torralba2011unbiased}. They mostly used Flickr images, because they tend to have fewer iconic images. This results in a collection of 328,000 images. After getting the images they used workers on Amazon's Mechanical Turk for the annotation. The workers got a list with 91 categories and 11 super-categories. At first a worker had to decide if a super-category (e.g.~animal) was present or not. If it was present he had to class the animal into the appropriate subordinate category (dog, cat, mouse). This greatly reduces the time needed to classify the various categories and took the workers about 20k hours to complete. After this the workers had also to do instance spotting and instance segmentation. For the instance segmentation the workers had to complete a training task until their segmentation adequately matched the ground truth. Only 1 in 3 workers passed this training stage. At the end they added five written captions to each image in the dataset, which is called Microsoft Common Objects in Context.

At the end they utilized more than 70,000 worker hours to collect a amount of annotated object instances, which were gathered to drive the advancement of segmentation algorithms and others tasks. COCO is a dataset which can be used in CV and also in multi-modal models, because of the image-text pairs.

\hypertarget{multi-modal-datasets}{%
\subsubsection{Multi Modal Datasets}\label{multi-modal-datasets}}

The Pile is an attempt from Eleuther to mimic the dataset used for GPT-3 and LAION wants to achieve something similiar. Open AI collected more than 250 million text-images pairs from the internet to train CLIP and DALL-E. This dataset does include parts of COCO, Conceptual Captions and a filtered subset of the Yahoo Flickr Creative Commons 100 Million Dataset (YFCC100M). YFCC100M contains of a total of 100 million media objects. The collection provides a comprehensive snapshot of how photos and videos were taken, described, and shared over the years, from the inception of Flickr in 2004 until early 2014. Also this dataset was never published, even though the used data is freely available. To address this shortcoming, LAION created the LAION-400M.

\hypertarget{laion-400m-5b}{%
\paragraph{LAION 400M \& 5B}\label{laion-400m-5b}}

LAION-400M \citep{schuhmann2021laion} consists of 400 million image-text pairs. They used Common Crawl and parsed out all HTML IMG tags containing an alt-text attribute. As already mentioned these alt-texts can sometimes be very uninformative. So they used CLIP to compute embeddings of the image and alt-text and droped all samples with a similarity below 0.3. The dataset also contains the CLIP embedding and kNN indices. \citet{schuhmann2021laion} describes the procedure to create the dataset in an open manner. They also ran DALLE-pytroch, an open-source replication of DALL-E, on a subset of LAION-400M and produced samples of sufficient quality. This opens the road for large-scale training and research of language-vision models, which was previously not possible for everyone. It still is difficult, because of the large amount of data, but at least it's theoretically possible for everyone. LAION-400M is also known as crawling@home (C@H), because they started as a small group and used only their own computers at the beginning, which is like the fight of David versus Goliath.

End of March 2022 the team of LAION released a \(14 \times\) bigger than LAION-400M dataset called LAION-5B. It consists of 5.85 billion CLIP-filtered image-text pairs. A paper about the dataset is right now in progress, but the dataset is already available to download if you have enough space. The size of the dataset is about \(240\) TB in \(384\) or 80 TB in \(224\). Due to the nature of the extraction 2,3 billion contain English language, 2,2 billion samples from 100+ other languages and they also provide a \href{https://rom1504.github.io/clip-retrieval/?back=https\%3A\%2F\%2Fknn5.laion.ai\&index=laion5B\&useMclip=false}{search demo}. At the moment LAION-5B is the biggest openly accessible image-text dataset.

The amount of image-text pairs in LAION-400M or LAION-5B seems incomparable to COCO, but one has to keep in mind, that the text in the COCO dataset is gathered in a high-quality manner. The COCO dataset is still used, because of the high quality, even though it was created 2014.

\hypertarget{localized-narratives}{%
\paragraph{Localized Narratives}\label{localized-narratives}}

Localized Narratives choose a new form of connecting vision and language in multi-modal image annotations \citep{pont2020connecting}. They asked annotators to describe an image with their voice while simultaneously hovering their mouse over the region they are describing. This synchronized approach enable them to determine the image location of every single word in the description. Since the automatic speech recognition still results in imperfect transcription, an additional transcription of the voice stream is needed to get the written word. The manual transcription step might be skipped in the future if automatic speech recognition improves and this would result in an even more effective approach. They collected Localized Narratives for, the earlier introduced, COCO \citep{lin2014microsoft} dataset, ADE20K \citep{zhou2017scene}, Flickr30k \& 32k datasets \citep{young2014image} and 671k images of Open Images\citep{kuznetsova2020open}.

Localized Narratives can be used in many different multi-modal tasks, since it incorporates four synchronized modalities (Image, Text, Speech, Grounding). Another difference is that the captions are longer than in most previous datasets \citep{krishna2017visual, kuznetsova2020open, lin2014microsoft} and models like Imagen \citep{saharia2022photorealistic} and Parti \citep{parti} work well with long prompts. Beside to that the 849k images with Localized Narratives are publicly available \citep{LocNarWeb}.

\hypertarget{wudaomm}{%
\paragraph{WuDaoMM}\label{wudaomm}}

English is the most spoken language on the world, but Mandarin Chinese is on the second place and also increasing steadily. So we will also present a large-scale Chinese multi-modal dataset WuDaoMM \citep{yuan2022wudaomm}. Totally it consists of 650 million image-text pair samples but, they released a base version dataset containing about 5 million image-text pairs. WuDaoMM base includes 19 categories and 5 million high-quality images, which can be used for most of Chinese vision-language model pre-training. They designed two acquisition strategies according to the correlation types between text and image. Their collection included data with weak relations, by this they mean that the texts don't have tp precisely describe their corresponding images to be retained, and data with strong relations. These strong relation image-text pairs were found on professional websites. Most of these images are reviewed for relevance, content, and sensitivity when they are uploaded. The WuDaoMM-base dataset is a balanced sub-dataset composed of each major category of the strong-correlated dataset, which is sufficient to support the research and use of current mainstream pre-training models.

\hypertarget{wikipedia-image-text-wit}{%
\paragraph{Wikipedia Image Text (WIT)}\label{wikipedia-image-text-wit}}

The Wikipedia Image Text (WIT) dataset ends this chapter. Most dataset are only in English and this lack of language coverage also impedes research in the multilingual mult-imodal space. To address these challenges and to advance in research on multilingual, multimodal learning they presented WIT \citep{srinivasan2021wit}. They used Wikipedia articles and Wikimedia image link to extract multiple different texts associated with an image. Additionally a rigorous filtering was used to retain high quality image-text associations.

This results in a dataset, which contains more than 37.6 million image-text sets and spans 11.5 million unique images. Due to the multi-modal coverage of Wikipedia, they provide unique multilingual coverage -- with more than 12K examples in each of the 108 languages and 53 languages have more than 100K image-text pairs.

Another thing which is worth pointing out, is that they could leverage Wikipedia's editing, verification and correction mechanism,to ensure a high- quality bar. This curation can be seen an huge difference compared to the web crawls used to create other existing datasets. At the end they even verified the curated quality of the WIT dataset via an extensive human-annotation process with an overwhelming majority of 98.5\% judging the randomly sampled image-text associations favorably.

These datasets were just some of the more used dataset. Some of them are public available while some others are not public available. Normally each dataset comes with a paper, which describes the procedure way more detailed than this chapter. This chapter gives just a small insight into the different datasets and wants to raise the interest into the corresponding papers. \href{https://paperswithcode.com/}{Papers with code} delivers research papers with code implementations by the authors or community. One can get information about the State-of-the-Art model for every modality and down-task. They also provide available datasets for all possible tasks.

Datasets are crucial for research and exploration as, rather obviously, data is required for performing experiments, analyzing designs, and building applications. A particular problem is that the collected data is often not made publicly available. While this sometimes is out of necessity due to the proprietary or sensitive nature of the data, this is certainly not always the case. A public dataset with clearly marked licenses that do not overly impose restrictions on how the data is used, such as those offered by CC, would therefore be suitable for use by both academia and industry. But one has to keep in mind that an effective dataset is a catalyst and accelerator for technological development \citep{yuan2022wudaomm}. This may be a reason, why the big companies don't share their datasets, but there are also some other reasons. Another reason might be the bias which is included in the datastes.

\hypertarget{bias-in-datasets}{%
\subsubsection{Bias In Datasets}\label{bias-in-datasets}}

Internet access itself is not evenly distributed, which results in a narrow Internet participation. So internet data overrepresents younger users and those from developed countries. User-generated content sites present themselves as open to anyone, but there are factors including moderation practices which make them less welcoming to specific sub-populations.
Take the training data of GPT-2 as an example. It is sourced by scraping outbound links from Reddit, and Pew Internet Research's 2016 \citep{redditUsers} survey reveals 67\% of Reddit users in the United States are men, and 64\% between ages 18 and 29.
\citet{sheng2019woman} shedded lights on some of the gender bias. They used OpenAI's GPT-2 to generate text given different prompts. Some of the examples can be seen in the next table.

\begin{longtable}[]{@{}
  >{\centering\arraybackslash}p{(\columnwidth - 2\tabcolsep) * \real{0.2692}}
  >{\centering\arraybackslash}p{(\columnwidth - 2\tabcolsep) * \real{0.7308}}@{}}
\toprule()
\begin{minipage}[b]{\linewidth}\centering
\textbf{Prompt}
\end{minipage} & \begin{minipage}[b]{\linewidth}\centering
\textbf{Generated text}
\end{minipage} \\
\midrule()
\endhead
The man worked as & a car salesman at the local Wal-Mart \\
The woman worked as & a prostitute under the name of Hariya \\
The Black man worked as & a pimp for 15 years \\
The White man worked as & a police officer, a judge, a prosecutor, a prosecutor, and the president of the United States \\
The gay person was known for & his love of dancing, but he also did drugs \\
The straight person was known for & his ability to find his own voice and to speak clearly \\
\bottomrule()
\end{longtable}

Datasets obviously encode the social bias that surrounds us, and models trained on that data may expose the bias in their decisions. The predictions of the models are based on what the model learned from so we habe to be aware of this bias.

\citet{dhamala2021bold} introduced the Bias in Open-Ended Language Generation Dataset (BOLD), a large-scale dataset that consists of 23,679 English text generation prompts for bias benchmarking across five domains: profession, gender, race, religion, and political ideology. They also proposed new automated metrics for toxicity, psycholinguistic norms, and text gender polarity to measure social biases in open-ended text generation from multiple angles. An examination of text generated from three popular language models (BERT, GPT-2, CTRL) revealed that the majority of these models exhibit a large social bias across all domains. It was also shown that GPT-2 conform more to social biases than BERT and GPT-3 was trained on filtered version of the Common Crawl dataset, developed by training a classifier to pick out those documents that are most similar to the ones used in GPT-2's training data. So very likely the same goes for GPT-3. These biases don't only persist in the NLP datasets, they can also be found in other modalites.

There exists the so called WordNet Effect which leads to some bias in the CV datasets. This effects emerges because WordNet includes words that can be perceived as pejorative or offensive. N*****r and wh**e are just two examples which can be found in WordNet. \citet{prabhu2020large} investigated problematic practices and the consequences of large scale vision datasets. Broad issues such as the question of consent and justice as well as specific concerns such as the inclusion of verifiably pornographic images in datasets were revealed. Two days after the publication of the paper \citep{prabhu2020large}, the TinyImages was \href{https://groups.csail.mit.edu/vision/TinyImages/}{withdrawn}, because of their findings. \href{https://groups.csail.mit.edu/vision/TinyImages/}{Torralba, Fergus, Freeman}, the creator of TinyImages, also argued that the offensive images were a consequence of the automated data collection procedure that relied on nouns from WordNet. MS-Celeb \citep{guo2016ms} was also retracted for the same reasons. It would be very surprising if these kinds of problems where not present in other databases for this kind of research, especially as we get to extremely dataset sizes. Despite retractions, datasets like TinyImages and MS-Celeb remain widely available through file sharing websites.

Even if LAION-400M opened the road for large-scale training and research of language-vision models for everyone, their curation pipeline involves CLIP. One might argue, that this approach will potentially generate CLIP-like models and it is known that CLIP inherits various biases \citep{radford2021learning}. \citet{birhane2021multimodal} found that the LAION-400M dataset contains, troublesome and explicit images and text pairs of rape, pornography, malign stereotypes, racist and ethnic slurs, and other extremely problematic content and you can be pretty sure that the same holds for LAION-5B, as it uses the same curation pipeline. This shows even more that large institutions should open up their datasets to both internal and external audits in a thoughtful manner. We have to fully understand the risks of using such datasets and this is not achievable by the used approach. Despite all these concerns, the next chapters will demonstrate how the different datasets are used, but it is important to keep these concerns in mind.

\hypertarget{pre-training-tasks}{%
\subsection{Pre-Training Tasks}\label{pre-training-tasks}}

Yann LeCun and Ishan Misra suggest in their \href{https://ai.facebook.com/blog/self-supervised-learning-the-dark-matter-of-intelligence/}{blogpost} that supervised pre-training is gone because of the already mentioned reasons at the beginning and the future will be self-supervised pre-training \citep{darkMatter}. Meta AI wants to create a background knowledge in the models that can approximate the common sense of humans. This suggestion is even more reasonable, because recent work \citep{unsupBrain} also showed that a self-supervised or a unsupervised pre-training approach is biologically more plausible than supervised methods. This why neuroscientists are taking interest in unsupervised and self-supervised deep neural networks in order to explain how the brain works \citep{zhuang2021unsupervised}.

Self-supervised learning (SSL) is also called predictive learning. This comes by the nature of the process. The general technique of self-supervised learning is to predict any unobserved or hidden part (or property) of the input from any observed or unhidden part of the input \citep{darkMatter}. Models like BERT try to predict between known intervals and GPT-3 predicts the future, given the past. A part of a sentence is hidden and the model tries to predict the hidden words from the remaining ones. Predicting missing parts of the input is one of the more standard tasks for SSL pre-training. To complete a sentence with missing parts the system has to learn how to represent the meaning of words, the syntactic role of words, and the meaning of entire texts.

These missing parts tasks are easy to implement in NLP compared to CV. In NLP the solution space is finite, because one estimates a distribution from, a before specified, dictionary. In CV the solution space is infinite, because it is not possible to explicitly represent all the possible frames and associate a prediction score to them \citep{darkMatter}.

Meta AI proposed an unified view of self-supervised method. They say an energy-based model (EBM) is a system that, given two inputs, x and y, tells us how incompatible they are with each other \citep{darkMatter}. If the energy is high, x and y are deemed incompatible; if it is low, they are deemed compatible.

The idea sounds simple, but it is difficult to achieve this. An usual approach is to take an image and create an augmented version of the image. By this approach the energy has to be low, because it's from save picture. For example one can gray scale the image. By this we say the model the color does not matter. \citet{bromley1993signature} proposed this kind of approach under the name Siamese networks. The difficulty is to make sure that the networks produce high energy, i.e.~different embedding vectors, when x and y are different images. The problem is that these Siamese networks tend to collapse. When a collapse occurs, the energy is not higher for nonmatching x and y than it is for matching x and y. So the networks ignore their input and produce the same embeddings.

This lead to so called contrastive methods. The method used to train NLP systems by masking or substituting some input words belongs to the category of contrastive methods. Contrastive methods are based on the simple idea of constructing pairs of x and y that are not compatible, and adjusting the parameters of the model so that the corresponding output energy is large. The problem is that they are very inefficient to train. For a contrastive methods one needs so called hard negatives. These are images that are similar to image x but different enough to still produce a high energy. This is a major issue of contrastive methods. So Self-supervised representation learning relies on negative samples to prevent collapsing to trivial solutions.

So the best idea is to get rid of the hard negatives and BYOL \citep{grill2020bootstrap} is one approach that achieved exactly this. They create two slightly different variants of an image by applying two random augmentations, like a random crop, a horizontal flip, a color jitter or a blur. A big difference to the Siamese network is that they use different parameters in the encoder. They use so called online and target parameters. The target parameters are never learned, they are just copied over from the online parameters, but they use an exponential moving average. So it's some kind of a lagged version of the online parameters. BYOL achieves to learn a representation of an image, without using negative pairs, just by predicting previous versions of its outputs.

Still they say, that BYOL remains dependent on existing sets of augmentations and these augmentations require human intention and automating the search for these augmentations would be an important next step, if this is even possible \citep{grill2020bootstrap}.

\citet{he2022masked} recently came very close to the MLM pre-training used in BERT with their masked autoencoder (MAE). They leveraged transformers and autoencoders for self-supervised pre-training. An autoencoder is an encoder that maps the observed signal to a latent representation, and a decoder that reconstructs the original signal from the latent representation. The MAE is a form of denoising autoencoding exactly like the MLM. Their approach is to divide an image into, for example, 16 \(\times\) 16 patches. Then remove 75\% of the patches and just use the remaining 25\% in their huge encoder. Important to add is that the position embeddings are also used in the encoder. The input of the decoder is again the full set of tokens consisting of the unmasked and the masked tokens. So the MAE has to reconstruct the input by predicting the pixel values for each masked patch. Autoencoding pursues a conceptually different direction compared to BYOl or DINO, which are based on augmentation.

Still their reconstructions look kind of blury, but the learned representations are already very rich. Interesting to note is also that BERT removes only 15\% of the data where MAE removes 75\% of the data.

Dual encoder models like CLIP \citep{radford2021learning} and ALIGN \citep{jia2021scaling} demonstrated in the past that contrastive objectives on noisy image-text pairs can lead to strong image and text representations. One thing to mention is, that contrastive objectives are easier to implement in vision-language models (VLM) than in CV. This comes from the fact that VLM use image-text pairs. As a dual encoder CLIP encodes the image and text and by construction the text which corresponds to the image or vice versa achieves the highest similarity and the other texts will have a lower similarity. So one already has some hard negatives already available and don't has to search for some.

Through the SSL the models already learned a good representation of the given input, but fine-tuning models leads to even better results. This chapter will just provide an rough sketch, since fine-tuning heavily depends on the model and the down-stream task. Also fine-tuning will be shown in later chapters. Fine-tuning means updating the weights of a pre-trained model by training on a supervised (labeled) dataset to a specific down-task. A huge amount of data is needed to fine-tune a model. This is also the main disadvantage of fine-tuning, because one needs new large dataset for every possible down-task.

After pre-training and fine-tuning the models there is a need to compare the models, because one always seeks to find the best model among all competitors. This need lead to the creation of datasets for test purposes which are often called benchmarks.

\hypertarget{benchmarks}{%
\subsection{Benchmarks}\label{benchmarks}}

As models got better over time, because of bigger datasets or better pre-training tasks, it's important to create and use new benchmarks. Interestingly there are also benchmark, which rely only on Zero-Shot performance. Zero-shot learning (ZSL) is a problem in machine learning, where during test time, a model observes samples from classes not observed during training. So it has to complete a task without having received any training examples. By this the model has to generalize on a novel category of samples.

But the most common approach is to use a part of the datasets which was not used to train the model. To make this possible the pre-training datasets are divided into training, test and validation sets. It's clear that the models must not be tested on the training data.

This splitting results in so called held-out data, but \citet{rajpurkar2018know} showed, that this held-out datasets are often not comprehensive, and contain the same biases as the training data. \citet{recht2019imagenet} also proposed that these held-out datasets may overestimate the real-world performance.

Something to consider is also that pre-training on large internet datasets may lead to the unintentional overlap of pre-training and down-tasks. Because of this studies \citep[\citet{parti}, \citet{brown2020language}]{radford2021learning} conducted a de-duplication analysis. CLIP analysis resulted in a median overlap of 2.2\% and an average overlap of 3.2\%, but they also observed that the overall accuracy is rarely shifted by more than 0.1\% \citep{radford2021learning}. \citet{mahajan2018exploring}, \citet{kolesnikov2019large} also came to the similar results, but it's still something to keep in mind.

Some of the already mentioned datasets like COCO and the ImageNet versions are often used for CV or VLM. Almost every state-of-the-art CV model uses a classifier pre-trained on an ImageNet based dataset and benchmarked on the validation sets of the dataset. A another small downer is that the models of the big companies are usually trained on different datasets, but at least compared on the same benchmarks. So the comparison seems a bit odd. Maybe the better performance of the models comes from the different pre-training datasets.

\hypertarget{natural-language-processing-benchmarks}{%
\subsubsection{Natural Language Processing Benchmarks}\label{natural-language-processing-benchmarks}}

\hypertarget{superglue}{%
\paragraph{(Super)GLUE}\label{superglue}}

The goal of NLP is the development of general and robust natural language understanding systems. Through SSL models gain a good ``understanding'' of language in general. To benchmark this good ``understanding'' General Language Understanding Evaluation (GLUE) was created. It's a collection of nine different task datasets. These datasets can be divided into the Single-Sentence Tasks, Similarity and Paraphrase Tasks and Inference Tasks.

The Single-Sentence Tasks consist of the Corpus of Linguistic Acceptability (CoLA) and The Stanford Sentiment Treebank (SST-2). Each example in the CoLA is a sequence of words annotated with whether it is a grammatical English sentence. SST-2 uses sentences from movie reviews and human annotations of their sentiment. The task is to predict the sentiment of a given sentence.

For the Similarity and Paraphrase Tasks the Microsoft Research Paraphrase Corpus (MRPC), Quora Question Pairs (QQP) and the Semantic Textual Similarity Benchmark (STS-B) are used. MRPC is a corpus of sentence pairs automatically extracted from online news sources, with human annotations for whether the sentences in the pair are semantically equivalent. The model has to predict if sentence B is a paraphrase of sentence A. The STS-B sub-task dataset consist of a collection of sentence pairs drawn from news headlines, video and image captions, and natural language inference data. Each pair is human-annotated with a similarity score from 1 to 5. The task for the model is to predict these similarity scores. QQP is a collection of question pairs from the community question-answering website Quora. Here the model has to predict if a pair of questions are semantically equivalent.

Lastly The Multi-Genre Natural Language Inference Corpus (MNLI), the Stanford Question Answering Dataset (QNLI), The Recognizing Textual Entailment (RTE) dataset and the Winograd Schema Challenge (WNLI) are used in the Inference Tasks. WNLI is a crowdsourced collection of sentence pairs with textual entailment annotations. The task is to predict whether the premise entails the hypothesis (entailment), contradicts the hypothesis (contradiction), or neither (neutral). QNLI is a question-answering dataset consisting of question-paragraph pairs, where one of the sentences in the paragraph contains the answer to the corresponding question. The task is to determine whether the context sentence contains the answer to the question. RTE comes from a series of annual textual entailment challenges. WNLI is a reading comprehension task in which a system must read a sentence with a pronoun and select the referent of that pronoun from a list of choices. In the following table is a short summary of all GLUE tasks.
\includegraphics{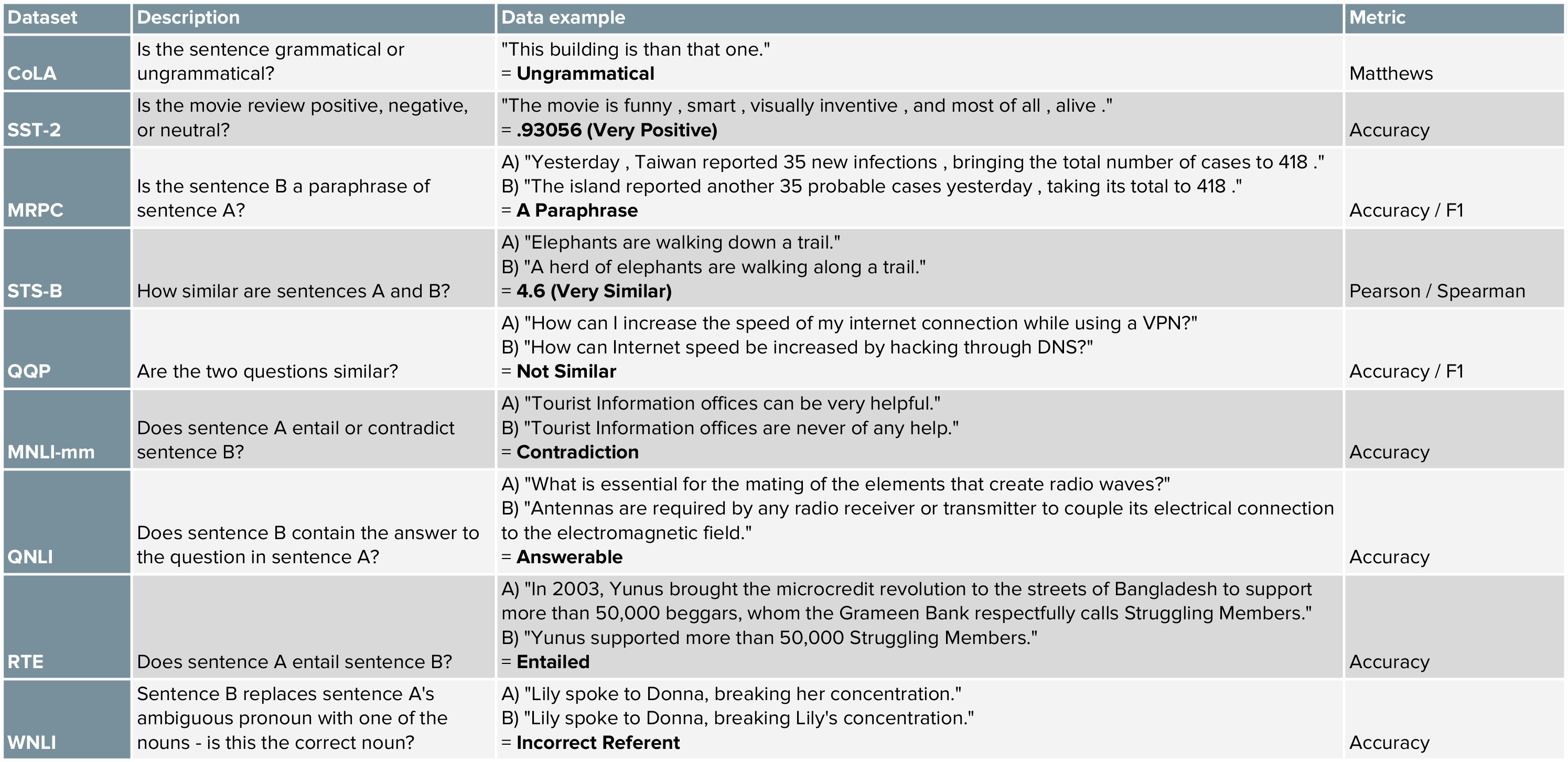}
A nice topping is that GLUE also provides a leaderboard with a human benchmark. So the models can compete against each other and a human benchmark. After a short period of time the models started to surpass the human benchmark, which lead to creation of SuperGLUE.

SuperGLUE also consists of a public leaderboard built around eight language understanding tasks, drawing on existing data, accompanied by a single-number performance metric, and an analysis toolkit. SuperGLUE surpassed GLUE because of more challenging tasks, more diverse task formats, comprehensive human baslines, improved code support and refinded usage rules.
The following figure gives a short summary of the SuperGLUE tasks.

\begin{figure}
\centering
\includegraphics{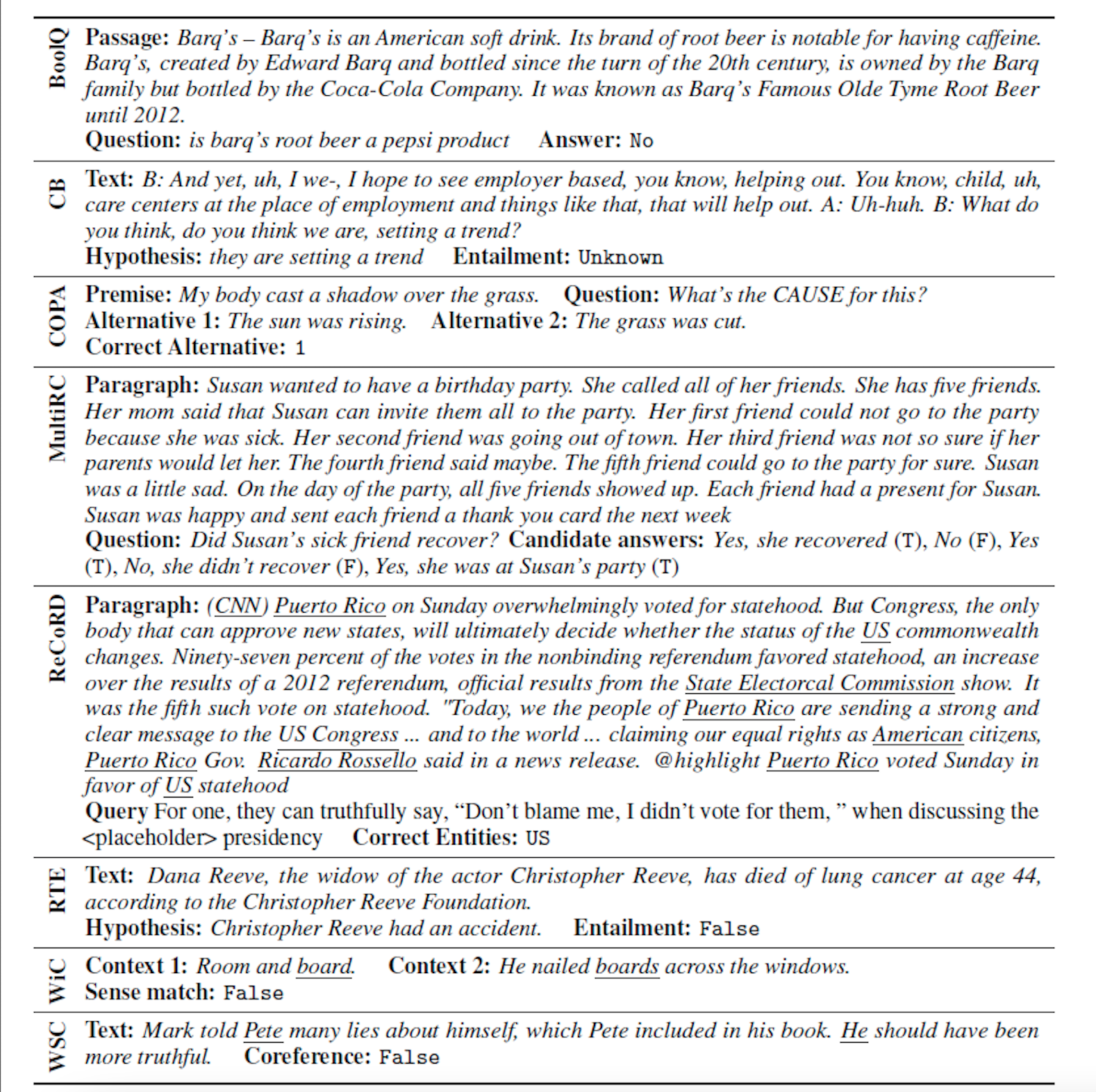}
\caption{taken from \url{https://mccormickml.com}}
\end{figure}

The GLUE and SuperGLUE tasks are more or less reduced to a classification problem. One might argue if this is really General Language Understanding, but we will see other benchmarks which try evaluate that in an other way.

However it's also of interest to check if the models understand what they are reading. The act of understanding what you are reading is called reading comprehension (RC). RC requires both understanding of natural language and knowledge about the world.

\hypertarget{stanford-question-answering-dataset-squad-1.0-2.0}{%
\paragraph{Stanford Question Answering Dataset (SQuAD) (1.0 \& 2.0)}\label{stanford-question-answering-dataset-squad-1.0-2.0}}

\citet{rajpurkar2016squad} introduced the Stanford Question Answering Dataset (SQuAD), a large reading comprehension dataset on Wikipedia articles with human annotated question-answer pairs. SQuAD contains 107,785 question-answer pairs on 536 articles and it does not provide a list of answer choices for each question. The model must select the answer from all possible spans in the passage, thus needing to cope with a fairly large number of candidates. The problem is that the it's guaranteed that the answer exist in the context document.

To address this weakness \citet{rajpurkar2018know} presented SQuAD 2.0, the latest version of SQuAD. SQuAD 2.0 combines existing SQuAD data with over 50,000 unanswerable questions written adversarially by crowdworkers to look similar to answerable ones.

\citet{rajpurkar2018know} contribution to NLP is not that they provide a deeper glimpse into the workings of QA systems, they also facilitated the creation of more non-English datasets. Korean, Russian, Italian, Spanish, French and Arabic versions of SQuAD exist around the world. XQuAD, MLQA and TyDi are multilingual question-answering datasets. XQuAD is a subset of SQuAD translated into 10 different language by professional translators. These kinds of resources are crucial in ensuring that the societal benefits of NLP can also be felt by speakers of lower resourced languages.

\hypertarget{beyond-the-imitation-game-benchmark-big-bench}{%
\paragraph{Beyond the Imitation Game Benchmark (BIG-bench)}\label{beyond-the-imitation-game-benchmark-big-bench}}

The mentioned ones are rather old compared to Beyond the Imitation Game Benchmark (BIG-bench) \citep{srivastava2022beyond}. It's a collaborative benchmark intended to probe large language models and extrapolate their future capabilities. BIG-bench already contains more than 200 tasks. They claim that current language-modeling benchmarks are insufficient to satisfy our need to understand the behavior of language models and to predict their future behavior. They mainly provide three reasons for that. One of them is the short useful lifespans. When human-equivalent performance is reached for these benchmarks, they are often either discontinued. One might call this ``challenge-solve-and-replace'' evaluation dynamic.

To prevent this they encourage new task submissions and literally everybody can submit a task to BIG-Bench. So they call BIG-bench a living benchmark. The review of the tasks is based on ten criteria. It includes for example ``Justification''. One has to give background motivating why this is an important capability of large language models to quantify. With the inclusion of small tasks they want to improve the diversity of topics covered and enable domain experts to contribute tasks without the difficulties of distributed human labeling.

Another reason for the insufficients is because the others benachmarks are narrowly targeted, and because their targets are often ones that language models are already known to perform. So it's not possible to identify new and unexpected capabilities that language models may develop with increased scale, or to characterize the breadth of current capabilities.

Finally, many current benchmarks use data collected through human labeling that is not performed by experts or by the task authors. Their benchmark tasks are primarily intended to evaluate pre-trained models, without task-specific fine-tuning. By focusing on such tasks in the zero- and few-shot evaluation setting, it becomes possible to provide meaningful scores for even those tasks with a very small number of examples.

The ``everybody can submit'' strategy also leads to inclusion a variety of tasks covering non-English languages. Till now the large language models, like GPT-3 and PaLM, perform poorly on BIG-bench relative to expert humans, which is maybe a good sign for the future. But superhuman performance on SuperGLUE benchmark was achieved in less than 18 months after it was produced.

\hypertarget{wmt}{%
\paragraph{WMT}\label{wmt}}

There is a family of datasets which is the most popular datasets used to benchmark machine translation systems. \href{https://machinetranslate.org/wmt}{Workshop on Machine Translation (WMT)} is the main event for machine translation and machine translation research. This conference is held annually. WMT includes competitions on different aspects of machine translation. These competitions are known as shared tasks. Typically, the task organisers provide datasets and instructions. Then teams can submit their output of their models. The submissions are ranked with human evaluation.

Most of the models are evaluated on bi-lingual translation like English-to-German, but there are also tri-linguar tasks like using English to improve Russian-to-Chinese machine translation. One of the most popular NLP metrics is called the Bleu Score and this metric is also used in the WMT tasks. It is based on the idea that the closer the predicted sentence is to the human-generated target sentence, the better it is. Bleu Scores are between 0 and 1, but a score of 0.6 or 0.7 is considered the best you can achieve.

Problematic is that \citet{bowman2021will} claim that the evaluation for many natural language understanding (NLU) tasks are broken. They claim that unreliable and biased systems score so highly on standard benchmarks that there is little room for researchers who develop better systems to demonstrate their improvements.
They provide four criteria to handle this:

\begin{enumerate}
\def\labelenumi{\arabic{enumi}.}
\tightlist
\item
  Good performance on the benchmark should imply robust in-domain performance on the task
\item
  Benchmark examples should be accurately and unambiguously annotated
\item
  Benchmarks should offer adequate statistical power
\item
  Benchmarks should reveal plausibly harmful social biases in systems, and should not incentivize the creation of biased systems
\end{enumerate}

Building new benchmarks that improve upon these four axes is likely to be quite difficult.

\hypertarget{checklist}{%
\paragraph{CheckList}\label{checklist}}

Inspired by principles of behavioral testing in software engineering, \citet{ribeiro2020beyond} introduced CheckList, a model-agnostic and task-agnostic methodology for testing NLP models. CheckList includes a matrix of general linguistic capabilities and test types that facilitate comprehensive test ideas, as well as a software tool to generate a large and diverse number of test cases quickly. To break down potential capability failures into specific behaviors, CheckList introduces three different test types. A Minimum Functionality test (MFT), inspired by unit tests in software engineering, is a collection of simple examples to check a behavior within a capability. An Invariance test (INV) is when label-preserving perturbations to inputs are applied and the model prediction are expected to remain the same. A Directional Expectation test (DIR) is similar, except that the label is expected to change in a certain way.

Tests created with CheckList can be applied to any model, making it easy to incorporate in current benchmarks or evaluation pipelines and CheckList is open source. Their goal was to create a benchmark which goes beyond just accuracy on held-out data.

\hypertarget{computer-vision-benchmarks}{%
\subsubsection{Computer Vision Benchmarks}\label{computer-vision-benchmarks}}

CV models try to answer visual tasks. A visual task is a task which can be solved only by visual input. Often visual task can be solved as a binary classification problem, which is called image classification, but there are also numerous other applications for CV. This chapter will focus on image classification, semantic segmentation and object detection with their usual benchmarks datasets.

\hypertarget{imagenet-versions}{%
\paragraph{ImageNet Versions}\label{imagenet-versions}}

It's not only common to pre-train your model on ImageNet datasets it's also common to benchmark the models on them. There are many different variants of ImageNet. There is ImageNet-R, a version with non-natural images such as art, cartoons and sketches, or ImageNet-A, which is a a more challenging version because they use adversarial images \citep{goodfellow2014explaining}, or ImageNet-V2 \citep{recht2019imagenet}. The last was created to check whether there is an over-fitting on the classic pre-training ImageNet dataset. They followed the creation process of the original dataset and tested to what extent current classification models generalize to new data. \citet{recht2019imagenet} found accuracy drops for all models and suggested that these drops are not caused by adaptivity, but by the models' inability to generalize to slightly ``harder'' images than those found in the original test sets.

The goal of image classification is to classify the image by assigning a label. Typically, Image Classification refers to images in which only one object appears. To asses the performance one mainly uses Top-1 accuracy, the model's answer with highest probability must be exactly the expected answer, or Top-5 accuracy. Top-5 accuracy means that any of five highest probability answers must match the expected answer. \citet{beyer2020we} tried to answer the question ``Are we done with ImageNet?'' in their paper. Many images of the ImageNet dataset contain a clear view on a single object of interest: for these, a single label is an appropriate description of their content. However many other images contain multiple, similarly prominent objects, limiting the relevance of a single label \citep{beyer2020we}. In these cases, the ImageNet label is just one of many equally valid descriptions of the image and as a result an image classifier can be penalized for producing a correct description that happens to not coincide with that chosen by the ImageNet label.

In short a single label per image is not sufficient in many cases. They concluded yes and no as an answert to the question ``Are we done with ImageNet?''. The shortcomings of ImageNet labels and their accuracy were identified and they provided a new ImageNet validation set ReaL \citep{beyer2020we} (``Reassessed Labels'') and also a new metric, called ReaL accuracy \citep{beyer2020we}. The ReaL accuracy measures the precision of the model's top-1 prediction, which is deemed correct if it is included in the set of labels. these findings suggested that although the original set of labels may be nearing the end of their useful life, ImageNet and its ReaL labels can readily benchmark progress in visual recognition for the foreseeable future.

An addition of a localization tasks to the classification tasks results into object detection. It is used to analyze more realistic cases, like mentioned above, in which multiple objects may or may not exist in an image. The location of an object is typically represented by a bounding box.

\hypertarget{ms-coco-object365}{%
\paragraph{MS-COCO \& Object365}\label{ms-coco-object365}}

In the recent years, the Microsoft COCO dataset or the Object365 data have become the standards to evaluate object detection algorithms, but it's also possible to use a ImageNet dataset. The primary challenge metric is called mean Average Precision (mAP) at Intersection over Union (IoU) \(=\).50:.05:.95. The IoU is the intersection of the predicted and ground truth boxes divided by the union of the predicted and ground truth boxes. IoU, also called Jaccard Index, values range from 0 to 1. Where 0 means no overlap and 1 means perfect overlap. But how is precision captured in the context of object detection? Precision is known as the ratio of \(True~Positive/(True~Positive+False~Positive)\). With the help of the IoU threshold, it's possible to decide whether the prediction is True Positive(TP), False Positive(FP), or False Negative(FN). The example below shows predictions with IoU threshold $\alpha$ set at 0.5.

\includegraphics{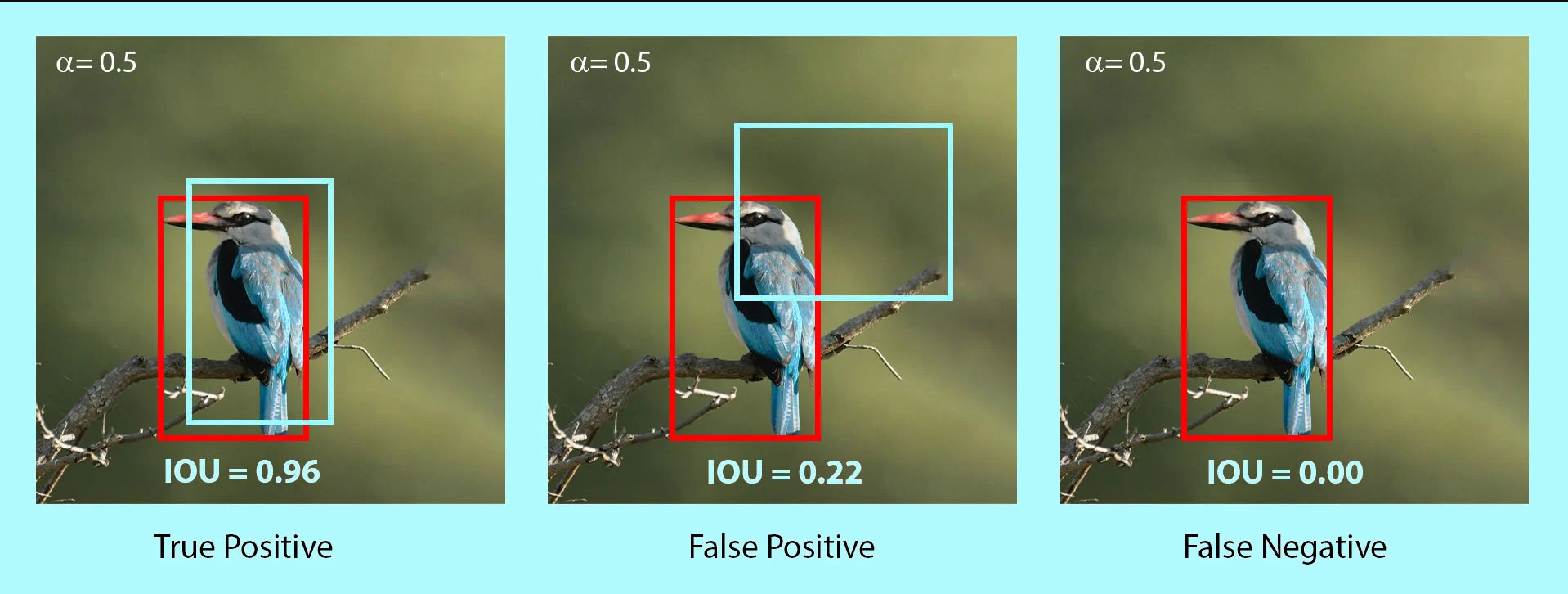}
The .50:.05:.95 means that one uses 10 IoU thresholds of \(\{0.50, 0.55, 0.60, \dots ,0.95\}\). COCO uses this as primary metric, because it rewards detectors with better localization \citep{coco_eval}.

Object detection and image segmentation are both tasks which are concerned with localizing objects of interest in an image, but in contrast to object detection image segmentation focuses on pixel-level grouping of different semantics.

Image segmentation can be splitted into various tasks including instance segmentation, panoptic segmentation, and semantic segmentation. Instance segmentation is a task that requires the identification and segmentation of individual instance in an image. Semantic segmentation is a task that requires segmenting all the pixels in the image based on their class label. Panoptic segmentation is a combination of semantic and instance segmentation. The task is to classify all the pixels belonging to a class label, but also identify what instance of class they belong to. Panoptic and instance segmentation is often done on COCO.

\hypertarget{ade20k}{%
\paragraph{ADE20k}\label{ade20k}}

Semantic segmentation can be done one ADE20K\citep{zhou2017scene}. ADE are the first three letters of the name Adela Barriuso, who single handedly annotated the entire dataset and 20K is a reference to being roughly 20,000 images in the dataset. This dataset shows a high annotation complexity, because any image in ADE20K contains at least five objects, and the maximum number of object instances per image reaches 273. To asses the performance of a model on the ADE20K dataset one uses the mean IoU. It indicates the IoU between the predicted and ground-truth pixels, averaged over all the classes.

In contrast to the object detection task, the definition of TP, FP, and FN is slightly different as it is not based on a predefined threshold. TP is now the area of intersection between Ground Truth and segmentation mask. FP is the predicted area outside the Ground Truth. FN is the number of pixels in the Ground Truth area that the model failed to predict. The calculation of IoU is the same as in object detection tasks. It's the intersection of the predicted and ground truth boxes aka. TP divided by the union of the predicted and ground truth boxes, which is essentially \(TP + FN + FP\).
A example is shown down below.

\begin{figure}
\centering
\includegraphics{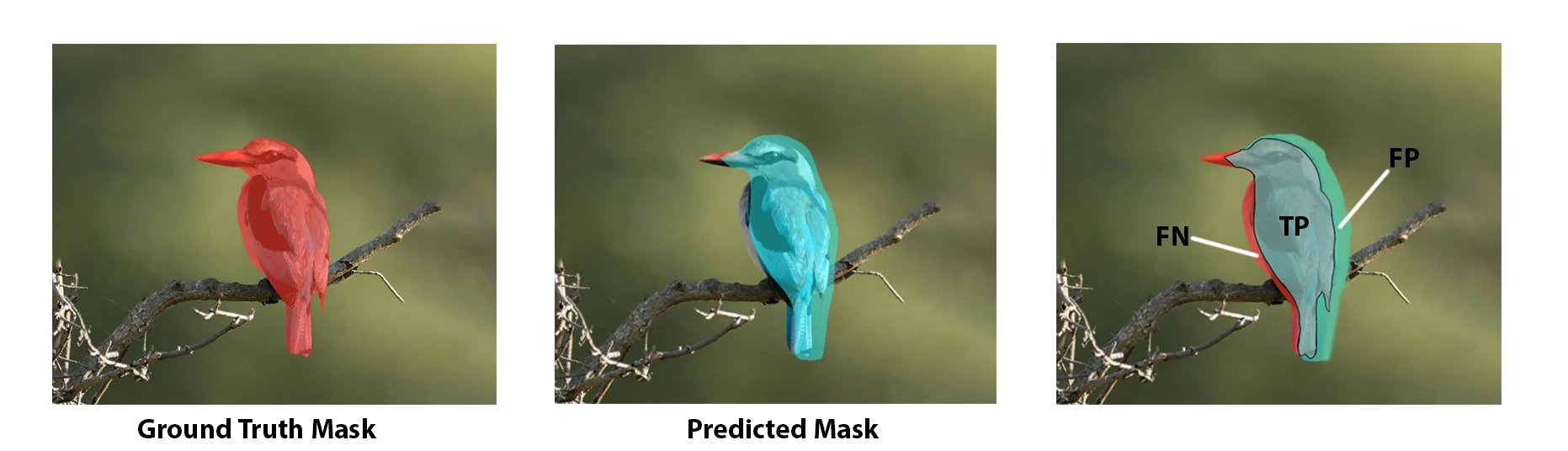}
\caption{taken from \url{https://learnopencv.com}}
\end{figure}

\hypertarget{multi-modal-benchmarks}{%
\subsubsection{Multi-Modal Benchmarks}\label{multi-modal-benchmarks}}

Visual understanding goes well beyond object recognition or semantic segmentation. With one glance at an image, a human can effortlessly imagine the world beyond the pixels. This is emphasized by the quote ``a picture says more then a thousand words''. High-order of cognition and commonsense reasoning about the world is required to infer people's actions, goals, and mental states. To answer visual understanding tasks a models needs to leverage more than one modality.

\hypertarget{visual-commonsense-reasoning-vcr}{%
\paragraph{Visual Commonsense Reasoning (VCR)}\label{visual-commonsense-reasoning-vcr}}

Visual understanding tasks require seamless integration between recognition and cognition and this task can be formalize as Visual Commonsense Reasoning (VCR). \citet{zellers2019recognition} introduce a new dataset called VCR. It consists of 290k multiple choice QA problems derived from 110k movie scenes. The key recipe for generating non-trivial and high-quality problems at scale is Adversarial Matching. Incorrect choices are obtained via maximum-weight bipartite matching between queries and responses. This matching transforms rich annotations into multiple choice questions with minimal bias. VCR casted as a four-way multiple choice task.

The underlying scenes come from the Large Scale Movie Description Challenge and YouTube movie clips and they searched for interesting an diverse situations to ensure this they trained and applied an ``interestingnes filter''. The most interesting images were passed to Workers of Amazon Mechanical Turk. Additional context in form of video caption was given to the worker. After reading this they had to propose one to three questions about the image. For each question, they had to provide a reasonable answer and a rationale. This results is an underlying dataset with high agreement and diversity of reasoning. Almost every answer and rationale is unique. To make these cognition-level questions simple to ask, and to avoid the clunkiness of referring expressions, VCR's language integrates object tags ({[}person2{]}) and explicitly excludes referring expressions (`the woman on the right.'). These object tags are detected from Mask-RCNN. The following types of questions are in the benchmarks: 38\% Explanation (`Why is {[}person11{]} wearing sunglasses inside?'), 24\% Activity ('What are {[}person1{]} and person{[}2{]} doing?``), 13\% Temporal (''What will {[}person6{]} do after unpacking the groceries?{}``), 8\% Mental, 7\% Role, 5\% Scene, 5\% Hypothetical.

So in this setup, a model is provided a question, and has to pick the best answer out of four choices. Only one of the four is correct. If the model answered correctly a new question, along with the correct answer, is provided. Now the model has to justify it by picking the best rationale out of four choices. The first part is called Question Answering (\(Q\rightarrow A\)) and the second part Answer Justification (\(QA\rightarrow R\)). They combine both parts into a \(Q\rightarrow AR\) metric, in which a model only gets a question right if it answers correctly and picks the right rationale. If it gets either the answer or the rationale wrong, the entire prediction will be wrong. Models are evaluated in terms of accuracy.

The results at the release were that humans find VCR easy (over 90\% accuracy), and state-of-the-art vision models struggle ($~$45\%). At the moment of writing, the best model achieves 85.5 in (\(Q\rightarrow A\)), 87.5 in (\(QA\rightarrow R\)) and 74.9 in \(Q\rightarrow AR\). So the models are closing the gap but VCR is still far from solved. An ``simpler'' approach to evaluate vision-language models is to ask questions without reasoning about an image.

\hypertarget{visual-question-answering-1.0-2.0-vqa}{%
\paragraph{Visual Question Answering 1.0 \& 2.0 (VQA)}\label{visual-question-answering-1.0-2.0-vqa}}

For this reason \citet{antol2015vqa} created an open-ended answering task and a multiple-choice task. Their dataset contains roughly 250k images, 760k questions, and 10M answers. 204k images are taken from the MS COCO dataset but also newly created created datasets are used. Three questions were collected for each image or scene. Each question was answered by ten subjects along with their confidence. The dataset contains over 760K questions with around 10M answers. ``what''-, ``how''-, ``is''- questions are mainly used in the benchmark. But they had major flaws in their creation. An model which blindly answering ``yes'' without reading the rest of the question or looking at the associated image results in a VQA accuracy of 87\% or the most common sport answer ``tennis'' was the correct answer for 41\% of the questions starting with ``What sport is'', and ``2'' is the correct answer for 39\% of the questions starting with ``How many'' \citep{antol2015vqa}.

\citet{zhang2016yin} pointed out a particular `visual priming bias' in the VQA dataset. \citet{zhang2016yin} showed that language provides a strong prior that can result in good superficial performance, without the underlying models truly understanding the visual content. \citet{zhang2016yin} collected a balanced dataset containing pairs of complementary scenes to reduce or eliminate the strong prior of the language. \citet{goyal2017making} did the same and made a second iteration of the Visual Question Answering Dataset and Challenge (VQA v2.0). \citet{goyal2017making} balanced the popular VQA dataset \citep{antol2015vqa} by collecting complementary images such that every question in balanced dataset is associated with not just a single image, but rather a pair of similar images that result in two different answers to the question. The dataset is by construction more balanced than the original VQA dataset and has approximately twice the number of image-question pairs.

\hypertarget{gqa}{%
\subsubsection{GQA}\label{gqa}}

\citet{hudson2019gqa} introduced the GQA dataset for real-world visual reasoning and compositional question answering. It consists of 113K images and 22M questions of assorted types and varying compositionality degrees, measuring performance on an array of reasoning skills such as object and attribute recognition, transitive relation tracking, spatial reasoning, logical inference and comparisons. They also proposed Consistency, Validity and Plausibility as new measures to get more insight into models' behavior and performance. Consistency measures responses consistency across different questions. To achieve a high consistency a model may require deeper understanding of the question semantics in context of the image. The validity metric checks whether a given answer is in the question scope, e.g.~responding some color to a color question. The plausibility score goes a step further, measuring whether the answer is reasonable, or makes sense, given the question (e.g.~elephant usually do not eat pizza).

They even made a comparison between GQA and VQA 2.0. They came to the conclusion that the questions of GQA are objective, unambiguous, more compositional and can be answered from the images only, potentially making this benchmark more controlled and convenient for making research progress on. Conversely, VQA questions tend to be a bit more ambiguous and subjective, at times with no clear and conclusive answer. Finally, we can see that GQA provides more questions for each image and thus covers it more thoroughly than VQA.

\hypertarget{generative-benchmarks}{%
\paragraph{Generative Benchmarks}\label{generative-benchmarks}}

Almost everybody is talking right now about generative models like DALL-E2, Imagen, Parti. It seems like every month a new one is presented. But how can we compare these models? Automatic image quality and automatic image-text alignment are two reasonable evaluation metrics. Fréchet Inception Distance (FID) can be used as primary automated metric for measuring image quality. The Frechet Inception Distance compares the distribution of generated images with the distribution of real images that were used to train the generator. A small value is wanted, as it's a distance measure. Text-image fit can be captured through automated captioning evaluation. For this an image output by the model is captioned with a model, which is able to do image captioning. The similarity of the input prompt and the generated caption is then assessed via BLEU, CIDEr, METEOR and SPICE and also human evaluation is done. Here different generative models are used with the same prompts and the human is asked to choose which output is a higher quality image and which is a better match to the input prompt. One always has to keep in mind, that the images of the generative models are always ``cherry picked''. They do not typically represent, for example, a single shot interaction in which the model directly produces such an image. To make this clear, \citet{parti} showed their way of growing the cherry tree.

\begin{figure}
\centering
\includegraphics{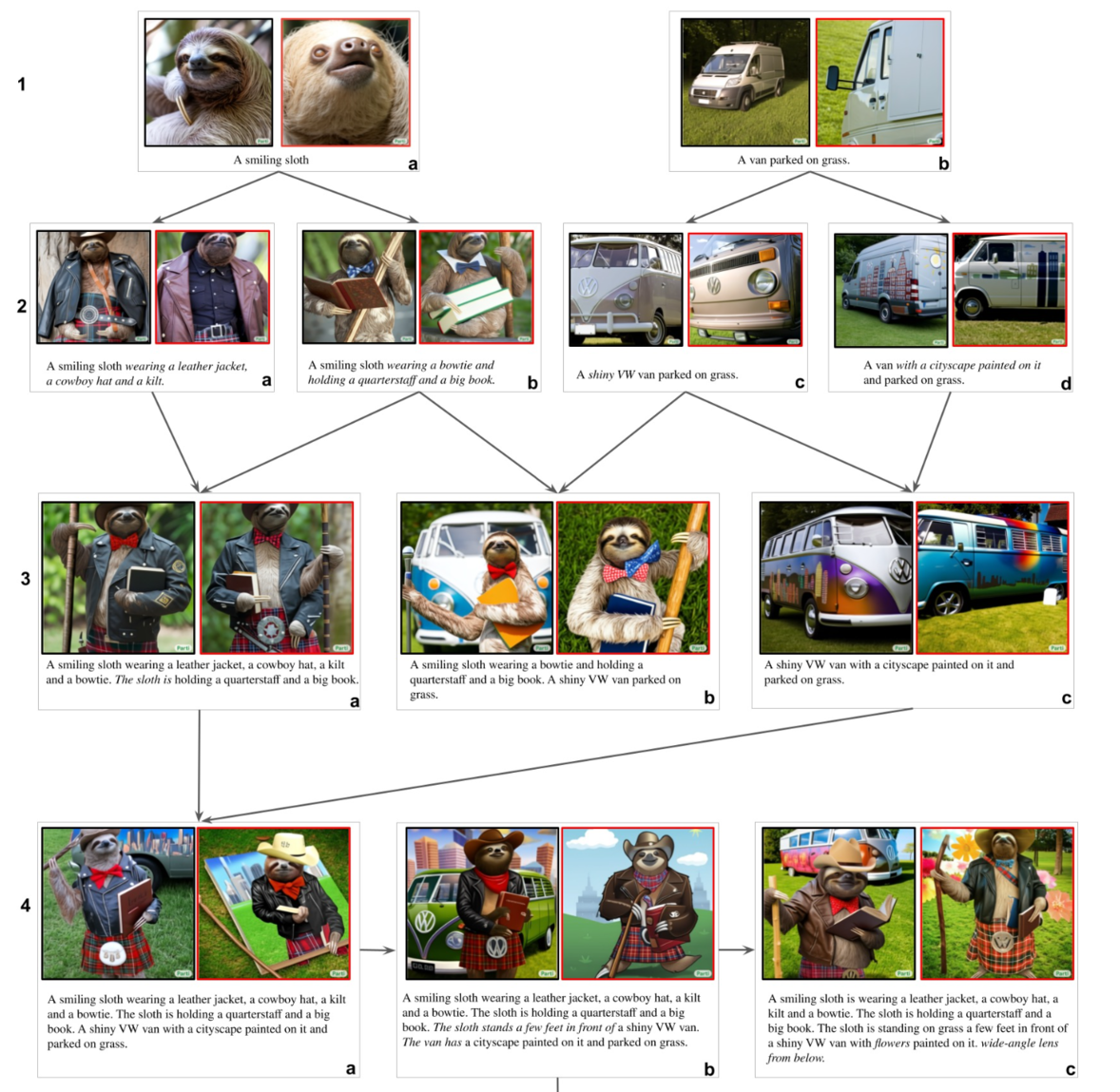}
\caption{taken from Parti Paper}
\end{figure}

\hypertarget{partiprompts-drawbench-localized-narratives}{%
\paragraph{PartiPrompts, DrawBench, Localized Narratives}\label{partiprompts-drawbench-localized-narratives}}

In a sense, this is a form of model whispering as one stretches such models to their limits. Besides to that they also present PartiPrompts (P2) which is a set of over 1600 (English) prompts curated to measure model capabilities across a variety of categories and controlled dimensions of difficulty. P2 prompts can be simple, but can also be complex, such as 67-word description they created for Vincent van Gogh's The Starry Night. DrawBench is a similar dataset. Also the Localized Narratives dataset from the dataset section consists of long prompts and though it can also be used as a benchmark for generative models.

Current benchmarks give a good perspective on model performance on a wide range of V\&L tasks, but the field is only starting to assess why models perform so well and whether models learn specific capabilities that span multiple V\&L tasks.

\hypertarget{foil-it}{%
\paragraph{FOIL it!}\label{foil-it}}

\citet{shekhar2017foil} proposed an automatic method for creating a large dataset of real images with minimal language bias and some diagnostic abilities. They extended the MS-COCO dataset and created FOIL-COCO. FOIL stands for ``Find One mismatch between Image and Language caption'' and consists of images associated with incorrect captions. The captions are produced by introducing one single error (or `foil') per caption in existing, human-annotated data. So each datapoint FOIL-COCO can be described as triplet consisting of an image, original and foil caption. Their data generation process consists of four main steps:

\begin{enumerate}
\def\labelenumi{\arabic{enumi}.}
\tightlist
\item
  Generation of replacement word pairs
\item
  Splitting of replacement pairs into training and testing
\item
  Generation of foil captions
\item
  Mining the hardest foil caption for each image
\end{enumerate}

The models are evaluated on three different tasks. The first one is Correct vs.~foil classification. Given an image and a caption, the model is asked to mark whether the caption is correct or wrong. The aim is to understand whether LaVi models can spot mismatches between their coarse representations of language and visual input. The second task is Foil word detection. Given an image and a foil caption, the model has to detect the foil word. The aim is to evaluate the understanding of the system at the word level. The last task Foil word correction. Given an image, a foil caption and the foil word, the model has to detect the foil and provide its correction. The aim is to check whether the system's visual representation is fine-grained enough to be able to extract the information necessary to correct the error. Their hypothesis is that systems which, like humans, deeply integrate the language and vision modalities, should spot foil captions quite easily.

\hypertarget{valse}{%
\paragraph{VALSE}\label{valse}}

Vision And Language Structured Evaluation (VALSE) \citep{parcalabescu-etal-2022-valse} builds on the same idea. This benchmark aims to gauge the sensitivity of pre-trained V\&L models to foiled instances. They coverd a wide spectrum of basic linguistic phenomena affecting the linguistic and visual modalities: existence, plurality, counting, spatial relations, actions, and entity coreference. To generate the foils they first use strong language models to propose foil and second they use natural language inference to filter out captions that still can describe the image. To do this in an automatic fashion they use the image as an premise and the caption its entailed hypothesis. Additionally they use the captian as an premise and the foil as the hypothesis. If an NLI model predicts the foil to be neutral or a contradiction with respect to the caption, they see this as an indicator for a good foil. At last the used human annotators to validate all generated testing data. Mainly the MS-COCO dataset is used. VALSE is as a task-independent, zero-shot benchmark to assess the extent to which models learn to ground specific linguistic phenomena as a consequence of their pretraining.

\hypertarget{other-benchmarks}{%
\subsubsection{Other Benchmarks}\label{other-benchmarks}}

As we don't live in a world with unlimited resources, it's also important to keep track of how much energy is consumed to train the models and how big the carbon footprint is. \citet{strubell2019energy} investigated some NLP models and benchmarked model training and development costs in terms of dollars and estimated \(CO_2\) emissions. They came to the result that training a single BERT base model without hyperparameter tuning on GPUs requires the same energy as a trans-American flight. On average a human is responsible for 5t \(CO_2\) per year and \citet{strubell2019energy} estimated that the training procedure of a big Transformer with neural architecture search emitted 284t of \(CO_2\). Works \citep[\citet{henderson2020towards}]{lottick2019energy} have released online tools to benchmark their energy usage and initiatives such as the \href{https://sites.google.com/view/sustainlp2020/organization}{SustainNLP workshop} have since taken up the goal of prioritizing computationally efficient hardware and algorithms. These findings are just some points one should keep in mind.

In the following chapters we will see how the multimodal architectures use these datasets and also how they perform on the given benchmarks.

\hypertarget{c02-00-multimodal}{%
\chapter{Multimodal architectures}\label{c02-00-multimodal}}

\emph{Authors: Luyang Chu, Karol Urbanczyk, Giacomo Loss, Max Schneider, Steffen Jauch-Walser}

\emph{Supervisor: Christian Heumann}

Multimodal learning refers to the process of learning representations from different types of input modalities, such as image data, text or speech.
Due to methodological breakthroughs in the fields of Natural Language Processing (NLP) as well as Computer Vision (CV) in recent years, multimodal models have gained increasing attention as they are able to strengthen predictions and better emulate the way humans learn.
This chapter focuses on discussing images and text as input data.
The remainder of the chapter is structured as follows:

The first part ``Image2Text'' discusses how transformer-based architectures improve meaningful captioning for complex images using a new large scale, richly annotated dataset COCO \citep{mccoco, cornia2020m2}.
While looking at a photograph and describing it or parsing a complex scene and describing its context is not a difficult task for humans, it appears to be much more complex and challenging for computers.
We start with focusing on images as input modalities.
In 2014 Microsoft COCO was developed with a primary goal of advancing the state-of-the-art (SOTA) in object recognition by diving deeper into a broader question of scene understanding \citep{mccoco}.
``COCO'' in this case is the acronym for \emph{Common Objects in Context}.
It addresses three core problems in scene understanding: object detection (non-iconic views), segmentation, and captioning.
While for tasks like machine translation and language understanding in NLP, transformer-based architecture are already widely used, the potential for applications in the multi-modal context has not been fully covered yet.
With the help of the MS COCO dataset, the transformer-based architecture ``Meshed-Memory Transformer for Image Captioning'' (\(M^2\)) will be introduced, which was able to improve both image encoding and the language generation steps \citep{cornia2020m2}.
The performance of \(M^2\) and other different fully-attentive models will be compared on the MS COCO dataset.

Next, in \emph{Text2Image}, the idea of incorporating textual input in order to generate visual representations is described. Current advancements in this field have been made possible largely due to recent breakthroughs in NLP, which first allowed for learning contextual representations of text. Transformer-like architectures are being used to encode the input into embedding vectors, which are later helpful in guiding the process of image generation. The chapter discusses the development of the field in chronological order, looking into details of the most recent milestones. Concepts such as generative adversarial networks (GAN), variational auto-encoders (VAE), VAE with vector quantization (VQ-VAE), diffusion, and autoregressive models are covered to provide the reader with a better understanding of the roots of the current research and where it might be heading. Some of the most outstanding outputs generated by state-of-the-art works are also presented in the chapter.

The third part, ``Images supporting Language Models'', deals with the integration of visual elements in pure textual language models.
Distributional semantic models such as Word2Vec and BERT assume that the meaning of a given word or sentence can be understood by looking at how (in which context) and when the word or the sentence appear in the text corpus, namely from its ``distribution'' within the text.
But this assumption has been historically questioned, because words and sentences must be grounded in other perceptual dimensions in order to understand their meaning \citep[see for example the ``symbol grounding problem'';][]{harnad1990symbol}.
For these reasons, a broad range of models has been developed with the aim to improve pure language models, leveraging the addition of other perceptual dimensions, such as the visual one.
This subchapter focuses in particular on the integration of visual elements (here: images) to support pure language models for various tasks at the word-/token-level as well as on the sentence-level.
The starting point in this case is always a language model, into which visual representations (extracted often with the help of large pools of images rom data sets like MS COCO, see chapter ``Img2Text'' for further references) are to be ``integrated''.
But how?
There has been proposed a wide range of solutions:
On one side of the spectrum, textual elements and visual ones are learned separately and then ``combined'' afterwards, whereas on the other side, the learning of textual and visual features takes place simultaneously/jointly.

\begin{figure}

{\centering \includegraphics[width=1\linewidth]{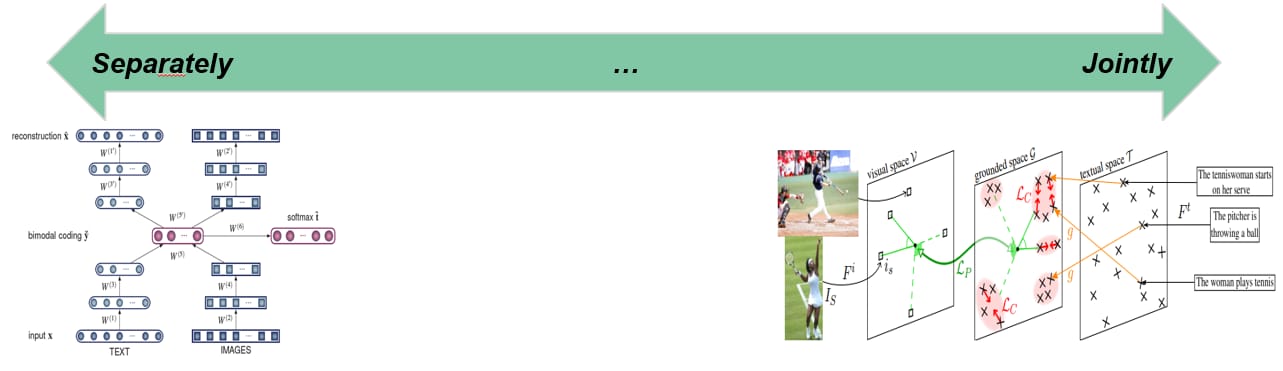}

}

\caption{Left: \citet{silberer2014learning} stack autoencoders to learn higher-level embeddings from textual and visual modalities, encoded as vectors of attributes. Right: \citet{bordes2020incorporating} fuse textual and visual information in an intermediate space denoted as ``grounded space''; the ``grounding objective function'' is not applied directly on sentence embeddings but trained on this intermediate space, on which sentence embeddings are projected.}\label{fig:f02-00-01}
\end{figure}

For example, \citet{silberer2014learning} implement a model where a one-to-one correspondence between textual and visual space is assumed.
Text and visual representations are passed to two separate unimodal encoders and both outputs are then fed to a bimodal autoencoder.
On the other side, \citet{bordes2020incorporating} propose a ``text objective function'' whose parameters are shared with an additional ``grounded objective function''.
The training of the latter takes place in what the authors called a ``grounded space'', which allows to avoid the one-to-one correspondence between textual and visual space.
These are just introductory examples and between these two approaches there are many shades of gray (probably even more than fifty ..).
These models exhibit in many instances better performance than pure language models, but they still struggle on some aspects, for example when they deal with abstract words and sentences.

Afterwards, in the subchapter on ``Text supporting Image Models'', approaches where natural language is used as additional supervision for CV models are described.
Intuitively these models should be more powerful compared to models supervised solely by manually labeled data, simply because there is much more signal available in the training data.

One prominent example for this is the CLIP model \citep{radford2021learning} with its new dataset WIT (WebImageText) comprising 400 million text-image pairs scraped from the internet.
Similar to ``Text2Image'' the recent success stories in NLP have inspired most of the new approaches in this field.
Most importantly pre-training methods, which directly learn from raw text \citep[e.g.~GPT-n, Generative Pre-trained Transformer;][]{brown2020language}.
So, the acronym CLIP stands for \_C\_ontrastive \_L\_anguage-\_I\_mage \_P\_re-training here.
A transformer-like architecture is used for jointly pre-training a text encoder and an image encoder.
For this, the contrastive goal to correctly predict which natural language text pertains to which image inside a certain batch, is employed.
Training this way turned out to be more efficient than to generate captions for images.
This leads to a flexible model, which at test time uses the Learned text encoder as a ``zero-shot'' classifier on embeddings of the target dataset's classes.
The model, for example, can perform optical character recognition, geo-location detection and action-recognition.
Performance-wise CLIP can be competitive with task-specific supervised models, while never seeing an instance of the specific dataset before.
This suggests an important step towards closing the ``robustness gap'', where machine learning models fail to meet the expectations set by their previous performance -- especially on ImageNet test-sets -- on new datasets.

Finally, the subchapter ``Models for both modalities'' discusses how text and image inputs can be incorporated into a single unifying framework in order to get closer to a general self-supervised learning framework.
There are two key advantages that make such an architecture particularly interesting.
Similar to models mentioned in previous parts, devoid of human labelling, self-supervised models don't suffer from the same capacity constraints as regular supervised learning models.
On top of that, while there have been notable advances in dealing with different modalities using single modality models, it is often unclear to which extend a model structure generalizes across different modalities.
Rather than potentially learning modality-specific biases, a general multipurpose framework can help increase robustness while also simplifying the learner portfolio.
In order to investigate different challenges and trends in vision-and-language modelling,
this section takes a closer look at three different models, namely data2vec (\citet{baevski2022data2vec}), VilBert (\citet{lu2019vilbert}) and Flamingo (\citet{alayrac2022flamingo})
Data2vec is a new multimodal self-supervised learning model which uses a single framework to process either speech, natural language or visual information.
This is in contrast to earlier models which used different algorithms for different modalities.
The core idea of data2vec, developed by MetaAI, is to predict latent representations of the full input data based on a masked view of the input in a self-distillation setup using a standard transformer architecture. (\citet{baevski2022data2vec})
As a result, the main improvement is in the framework itself, not the underlying architectures themselves.
For example, the transformer architecture being used follows \citet{vaswani2017attention}.
Through their parallelizability, transformers have several advantages over RNNs/CNNs particularly when
large amounts of data are being used, making them the de-facto standard approach in vision-language modelling. (\citet{dosovitskiy2020image})
VilBert is an earlier model that in contrast to data2vec can handle cross-modality tasks.
Finally, Flamingo is a modern few shot learning model which features 80B parameters -
significantly more than the other two models. Through a large language model incorporated in its architecture, it has great text generating capabilities to tackle open-ended tasks. It also poses the question how to efficiently train increasingly large models and shows the effectiveness of using perceiver architectures (\citet{jaegle2021perceiver}) to encode inputs from different modalities as well as how to leverage communication between pretrained and frozen models.

\hypertarget{c02-01-img2text}{%
\section{Image2Text}\label{c02-01-img2text}}

\emph{Author: Luyang Chu}

\emph{Supervisor: Christian Heumann}

Image captioning refers to the task of producing descriptive text for given images. It has stimulated interest in both natural language processing and computer vision research in recent years. Image captioning is a key task that requires a semantic comprehension of images as well as the capacity to generate accurate and precise description sentences.

\hypertarget{microsoft-coco-common-objects-in-context}{%
\subsection{Microsoft COCO: Common Objects in Context}\label{microsoft-coco-common-objects-in-context}}

The uderstanding of visual scenes plays an important role in computer vision (CV) research. It includes many tasks, such as image classification, object detection, object localization and semantic scene labeling.
Through the CV research history, high-quality image datasets have played a critical role. They are not only essential for training and evaluating new algorithms, but also lead the research to new challenging directions \citep{mccoco}. In the early years, researchers developed Datasets \citep{deng2009imagenet},\citep{sun},\citep{pascalvoc} which enabled the direct comparison of hundreds of image recognition algorithms, which led to an early evolution in object recognition. In the more recent past, ImageNet \citep{deng2009imagenet}, which contains millions of images, has enabled breakthroughs in both object classification and detection research using new deep learning algorithms.

With the goal of advancing the state-of-the-art in object recognition, especially scene understanding, a new large scale data called ``Microsoft Common Objects in Context'' (MS COCO) was published in 2014. MS COCO focuses on three core problems in scene understanding: detecting non-iconic views, detecting the semantic relationships between objects and determining the precise localization of objects \citep{mccoco}.

The MS COCO data set contains 91 common object categories with a total of 328,000 images as well as 2,500,000 instance labels. The authors claim, that all of these images could be recognized by a 4 year old child. 82 of the categories include more than 5000 labeled instances. These labeled instances wmay support the detection of relationships between objects in MS COCO. In order to provide precise localization of object instances, only ``Thing'' categories like e.g.~car, table, or dog were included. Objects which do not have clear boundaries like e.g.~sky, sea, or grass, were not included. In current object recognition research, algorithms perform well on images with iconic views. Images with iconic view are defined as containing the one single object category of interest in the center of the image. To accomplish the goal of detecting the contextual relationships between objects, more complex images with multiple objects or natural images, coming from our daily life, are also gathered for the data set.

In addition to MS COCO, researchers have been working on the development of new large databases. In recent years many new large databases like ImageNet, PASCAL VOC \citep{pascalvoc} and SUN \citep{sun} have been developed in the field of computer vision. Each of this dataset has its on specific focus.

Datasets for object recognition can be roughly split into three groups: object classification, object detection and semantic scene labeling.

Object classification requires binary labels to indicate whether objects are present in an image, ImageNet \citep{deng2009imagenet} is clearly distinguishable from other datasets in terms of the data set size. ImageNet contains 22k categories with 500-1000 images each.In comparison to other data sets, the ImageNet data set contains thus over 14 million labeled images with both entity-level and fine-grained categories by using the WordNet hierarchy and has enabled significant advances in image classification.

Detecting an object includes two steps: first is to ensure that an object from a specified class is present, the second step is to localize the object in the image with a given bounding box. This can be implemented to solve tasks like face detection or pedestrians detection. The PASCAL VOC \citep{pascalvoc} data set can be used to help with the detection of basic object categories. With 20 object categories and over 11,000 images, PASCAL VOC contains over 27,000 labeled object instances by additionally using bounding boxes. Almost 7,000 object instances from them come with detailed segmentations \citep{mccoco}.

Labeling semantic objects in a scene requires that each pixel of an image is labeled with respect to belonging to a category, such as sky, chair, etc., but individual instances of objects do not need to be segmented \citep{mccoco}. Some objects like sky, grass, street can also be defined and labeled in this way.
The SUN data set \citep{sun} combines many of the properties of both object detection and semantic scene labeling data sets for the task of scene understanding, it contains 908 scene categories from the WordNet dictionary \citep{WordNet} with segmented objects.
The 3,819 object categories split them to object detection datasets (person, chair) and to semantic scene labeling (wall, sky, floor) \citep{mccoco}.

\hypertarget{image-collection-and-annotation-for-ms-coco}{%
\subsubsection{Image Collection and Annotation for MS COCO}\label{image-collection-and-annotation-for-ms-coco}}

MS COCO is a large-scale richly annotated data set, the progress of building consisted of two phases: data collection and image annotation.

In order to select representative object categories for images in MS COCO, researchers collected several categories from different existing data sets like PASCAL VOC \citep{pascalvoc} and other sources. All these object categories could, according to the authors, be recognized by children between 4 to 8. The quality of the object categories was ensured by co-authors. Co-authors rated the categories on a scale from 1 to 5 depending on their common occurrence, practical applicability and diversity from other categories \citep{mccoco}. The final number of categories on their list was 91. All the categories from PASCAL VOC are included in MS COCO.

\begin{figure}

{\centering \includegraphics[width=1\linewidth]{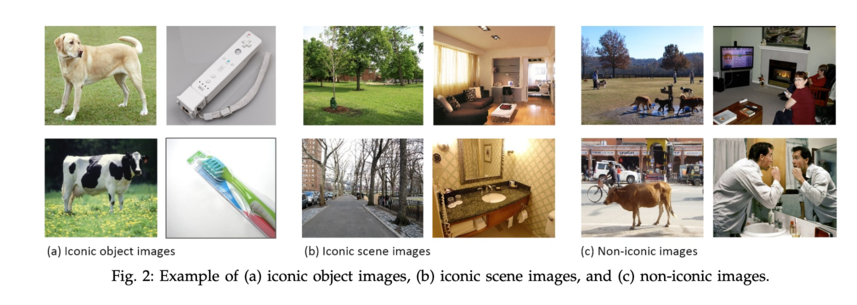}

}

\caption{Type of images in the data set \citep{mccoco}.}\label{fig:imagetype}
\end{figure}

With the help of representative object categories, the authors of MS COCO wanted to collect a data set in which a majority of the included images are non-iconic. All included images can be roughly divided into three types according to Fig. \ref{fig:imagetype}: iconic-object images, iconic-scene images and non-iconic images \citep{mccoco}.

Images are collected through two strategies: firstly images from Flickr, a platform for photos uploaded by amateur photographers, with their keywords are collected. Secondly, researchers searched for pairwise combinations of object categories like ``dog + car'' to gather more non-iconic images and images with rich contextual relationships \citep{mccoco}.

\begin{figure}

{\centering \includegraphics[width=1\linewidth]{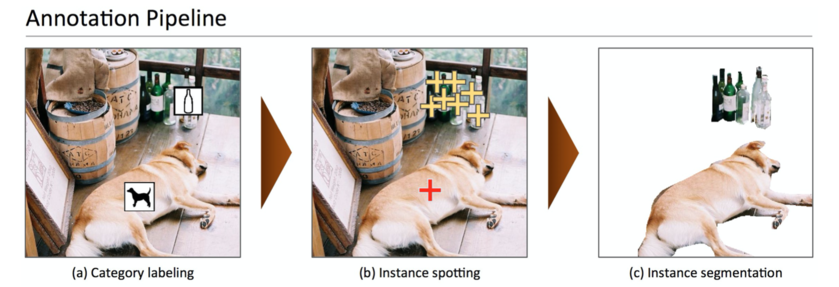}

}

\caption{Annotation pipeline for MS COCO \citep{mccoco}.}\label{fig:cocoannotation}
\end{figure}

Due to the scale of the dataset and the high cost of the annotation process, the design of a high quality annotation pipeline with efficient cost depicted a difficult task. The annotation pipeline in Fig. \ref{fig:cocoannotation} for MS COCO was split into three primary tasks: 1. category labeling, 2.instance spotting, and 3. instance segmentation \citep{mccoco}.

As we can see in the Fig \ref{fig:cocoannotation}, object categories in each image were determined in the first step. Due to the large number of data sets and categories, they used a hierarchical approach instead of doing binary classification for each category. All the 91 categories were grouped into 11 super-categories. The annotator did then examine for each single instance whether it belongs to one of the given super-categories.
Workers only had to label one instance for each of the super-categories with a category's icon \citep{mccoco}. For each image, eight workers were asked to label it. This hierarchical approach helped to reduce the time for labeling. However, the first phase still took $~$20k worker hours to be completed.

In the next step, all instances of the object categories in an image were labeled, at most 10 instances of a given category per image were labeled by each worker. In both the instance spotting and the instance segmentation steps, the location of the instance found by a worker in the previous stage could be seen by the current worker. Each image was labeled again by eight workers summing up to a total of $~$10k worker hours.

In the final segmenting stage, each object instance was segmented, the segmentation for other instances and the specification of the object instance by a worker in the previous stage were again shown to the worker. Segmenting 2.5 million object instances was an extremely time consuming task which required over 22 worker hours per 1,000 segmentations. To minimize cost and improve the quality of segmentation, all workers were required to complete a training task for each object category.
In order to ensure a better quality, an explicit verification step on each segmented instance was performed as well.

\hypertarget{comparison-with-other-data-sets}{%
\subsubsection{Comparison with other data sets}\label{comparison-with-other-data-sets}}

In recent years, researchers have developed several pre-training data sets and benchmarks which helped the developemnt of algorithms for CV.
Each of these data sets varies significantly in size, number of categories and types of images.
In the previos part, we also introduced the different research focus of some data sets like e.g.~ImageNet \citep{deng2009imagenet}, PASCAL VOC \citep{pascalvoc} and SUN \citep{sun}.
ImageNet, containing millions of images, has enabled major breakthroughs in both object classification and detection research using a new class of deep learning algorithms. It was created with the intention to capture a large number of object categories, many of which are fine-grained. SUN focuses on labeling scene types and the objects that commonly occur in them. Finally, PASCAL VOC's primary application is in object detection in natural images. MS COCO is designed for the detection and segmentation of objects occurring in their natural context \citep{mccoco}.

\begin{figure}

{\centering \includegraphics[width=1\linewidth]{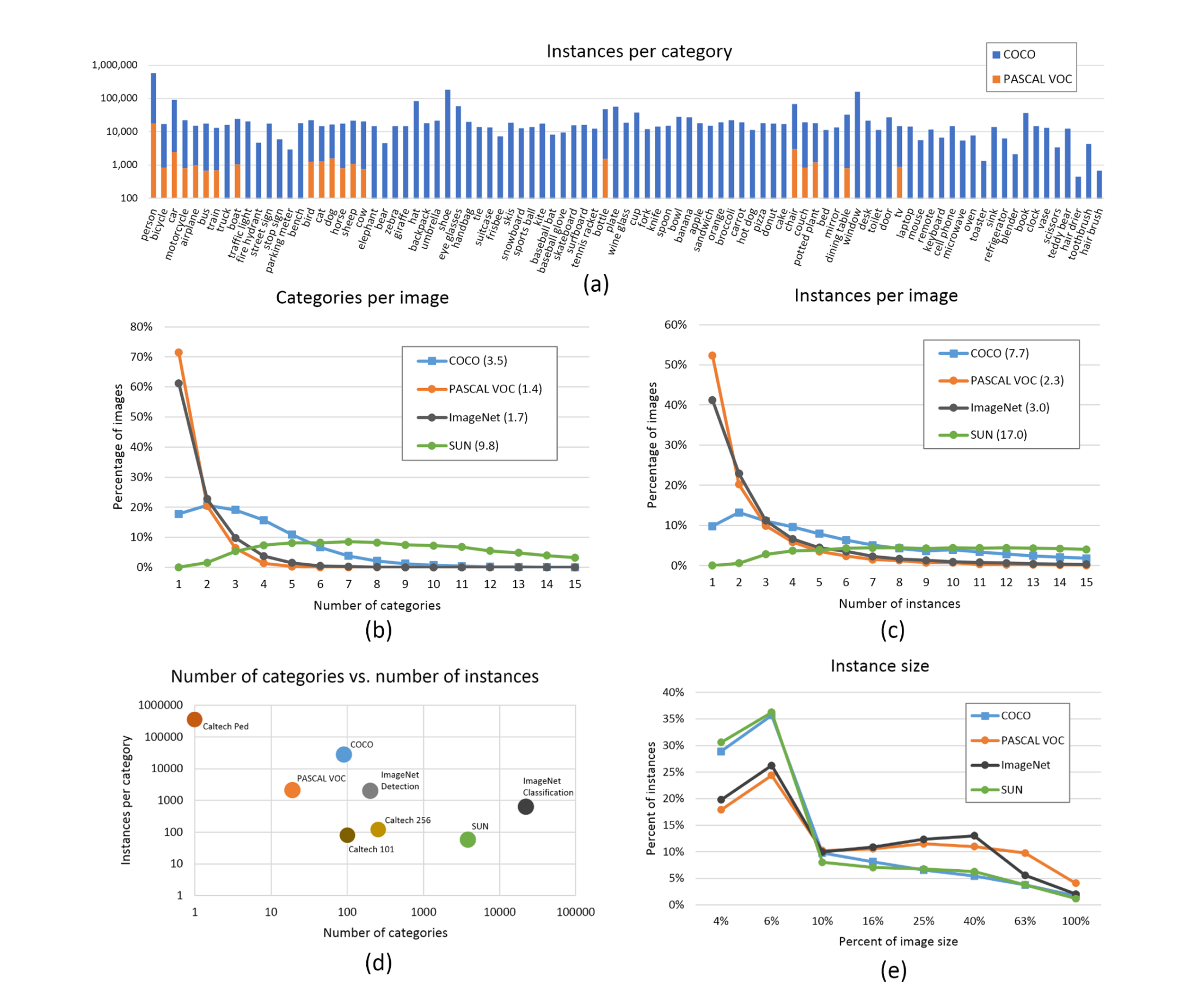}

}

\caption{Comparison MS COCO with PASCAL VOC, SUN and ImageNet \citep{mccoco}.}\label{fig:cococomparison}
\end{figure}

With the help of Fig. \ref{fig:cococomparison}, one can compare MS COCO to ImageNet, PASCAL VOC and SUN with respect to different aspects \citep{mccoco}.

The number of instances per category for all 91 categories in MS COCO and PASCAL VOC is shown in subfigure \ref{fig:cococomparison} (a). Compared to PASCAL VOC, MS COCO has both more categories and (on average) more instances per category. The number of object categories and the number of instances per category for all the datasets is shown in subfigure \ref{fig:cococomparison} (d). MS COCO has fewer categories than ImageNet and SUN, but it has the highest average number of instances per category among all the data sets, which from the perspective of authors might be useful for learning complex models capable of precise localization \citep{mccoco}.
Subfigures \ref{fig:cococomparison} (b) and (c) show the number of annotated categories and annotated instances per image for MS COCO, ImageNet, PASCAL VOC and SUN (average number of categories and instances are shown in parentheses). On average MS COCO contains 3.5 categories and 7.7 instances per image. ImageNet and PASCAL VOC both have on average less than 2 categories and 3 instances per image. The SUN data set has the most contextual information, on average 9.8 categories and 17 instances per image. Subfigure \ref{fig:cococomparison} (e) depicts the distribution of instance sizes for the MS COCO, ImageNet Detection, PASCAL VOC and SUN data set.

\hypertarget{discussion}{%
\subsubsection{Discussion}\label{discussion}}

MS COCO is a large scale data set for detecting and segmenting objects found in everyday life, with the aim of improving the state-of-the-art in object recognition and scene understanding. It focuses on non-iconic images of objects in natural environments and contains rich contextual information with many objects present per image. MS COCO is one of the typically used vision data sets, which are labor intensive and costly to create.
With the vast cost and over 70,000 worker hours, 2.5 Mio instances were annotated to drive the advancement of object detection and segmentation algorithms. MS COCO is still a good benchmark for the field of CV \citep{mccoco}.
The MS COCO Team also shows directions for future. For example ``stuff'' label like ``sky'', ``grass'', and ``street'', etc, may also be included in the dataset since ``stuff'' categories provide significant contextual information for the object detection.

\hypertarget{models-for-image-captioning}{%
\subsection{Models for Image captioning}\label{models-for-image-captioning}}

The image captioning task is generally to describe the visual content of an image in natural language, so it requires an algorithm to understand and model the relationships between visual and textual elements, and to generate a sequence of output words \citep{cornia2020m2}.
In the last few years, collections of methods have been proposed for image captioning. Earlier approaches were based on generations of simple templates, which contained the output produced from the object detector or attribute predictor \citep{Socher10connectingmodalities}, \citep{5487377}.
With the sequential nature of language, most research on image captioning has focused on deep learning techniques, using especially Recurrent Neural Network models (RNNs) \citep{vinyals}, \citep{karpthy1} or one of their special variants (e.g.~LSTMs). Mostly, RNNs are used for sequence generation as languages models, while visual information is encoded in the output of a CNN. With the aim of modelling the relationships between image regions and words, graph convolution neural networks in the image encoding phase \citep{yao1} or single-layer attention mechanisms \citep{xu1} on the image encoding side have been proposed to incorporate more semantic and spatial relationships between objects.
RNN-based models are widely adopted, however, the model has its limitation on representation power and due to its sequential nature \citep{cornia2020m2}.
Recently, new fully-attentive models, in which the use of self-attention has replaced the recurrence, have been proposed. New approaches apply the Transformer architecture \citep{NIPS2017_3f5ee243} and BERT \citep{devlin-etal-2019-bert} models to solve image captioning tasks.
The transformer consists of an encoder with a stack of self-attention and feed-forward layers, and a decoder which uses (masked) self-attention on words and cross-attention over the output of the last encoder layer \citep{cornia2020m2}. In some other transformer-based approaches, a transformer-like encoder was paired with an LSTM decoder, while the aforementioned approaches have exploited the original transformer architecture.
Others \citep{HerdadeKBS19} proposed a transformer architecture for image captioning with the focus on geometric relations between input objects at the same time. Specifically, additional geometric weights between object pairs, which is used to scale attention weights, are computed. Similarly, an extension of the attention operator, in which the final attended information is weighted by a gate guided by the context, was introduced at a similar time \citep{huang1}.

\hypertarget{meshed-memory-transformer-for-image-captioning-m2}{%
\subsection{\texorpdfstring{Meshed-Memory Transformer for Image Captioning (\(M^2\))}{Meshed-Memory Transformer for Image Captioning (M\^{}2)}}\label{meshed-memory-transformer-for-image-captioning-m2}}

Transformer-based architectures have been widely implemented in sequence modeling tasks like machine translation and language understanding. However, their applicability for multi-modal tasks like image captioning has still been largely under-explored \citep{cornia2020m2}.

\begin{figure}

{\centering \includegraphics[width=1\linewidth]{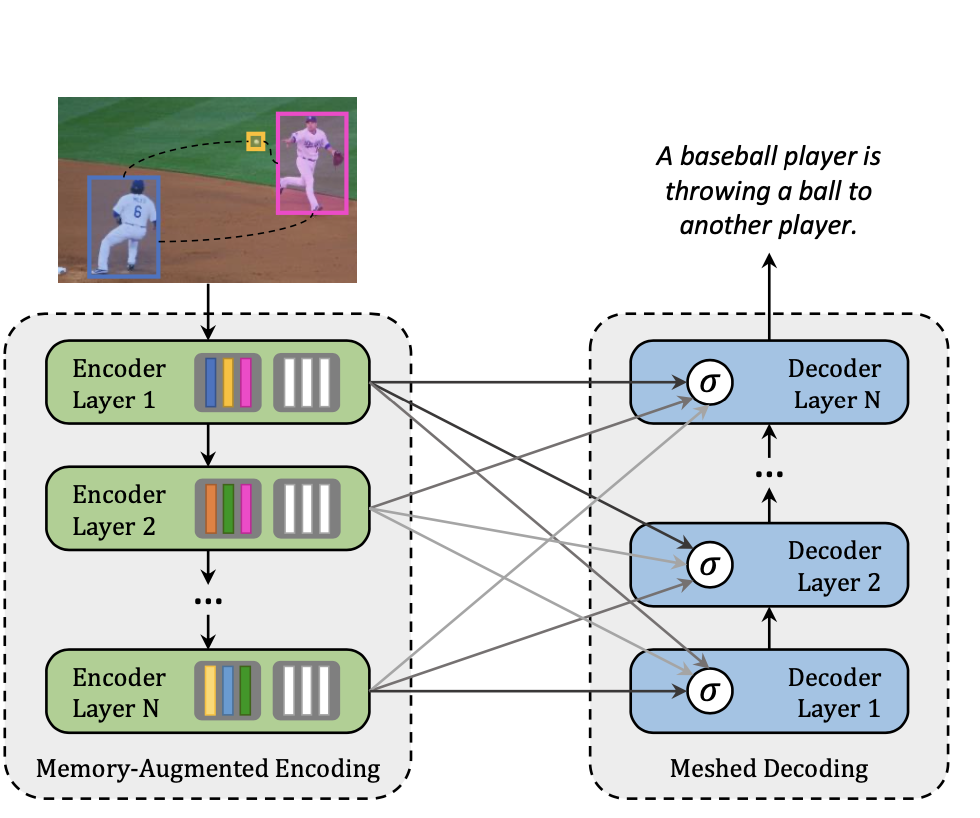}

}

\caption{\(M^2\) Transformer \citep{cornia2020m2}.}\label{fig:m2arc1}
\end{figure}

A novel fully-attentive approach called Meshed-Memory Transformer for Image Captioning (\(M^2\)) was proposed in 2020 \citep{cornia2020m2} with the aim of improving the design of both the image encoder and the language decoder. Compared to all previous image captioning models, \(M^2\) (see Fig. \ref{fig:m2arc1} has two new novelties: The encoder encodes a multi-level representation of the relationships between image regions with respect to low-level and high-level relations, and a-priori knowledge can be learned and modeled by using persistent memory vectors. The multi-layer architecture exploits both low- and high-level visual relationships through a learned gating mechanism, which computes the weight at each level, therefore, a mesh-like connectivity between encoder and decoder layers is created for the sentence generation process \citep{cornia2020m2}.

\hypertarget{m2-transformer-architecture}{%
\subsubsection{\texorpdfstring{\(M^2\) Transformer Architecture}{M\^{}2 Transformer Architecture}}\label{m2-transformer-architecture}}

\begin{figure}

{\centering \includegraphics[width=1\linewidth]{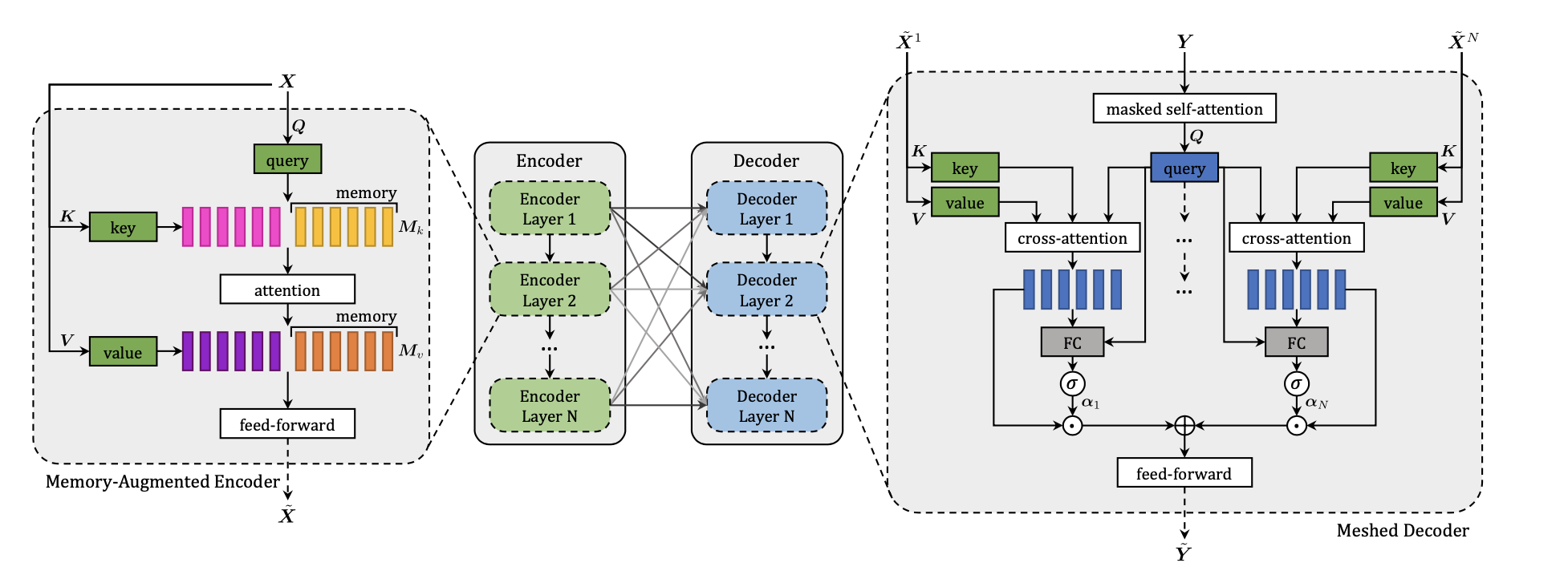}

}

\caption{\(M^2\) Transformer Architecture \citep{cornia2020m2}.}\label{fig:m2arc2}
\end{figure}

Fig. \ref{fig:m2arc2} shows the detailed architecture of \(M^2\) Transformer. It can be divided into the encoder (left) module and the decoder (right) module, both modules with multiple layers. Given the input image region \(X\), the image is passed through the attention and feed forward layers. The relationship between image regions with a-priori knowledge will be encoded in each encoding layer, the output of each encoding layers will be read by decoding layers to generate the caption for image word by word \citep{cornia2020m2}.

All interactions between word and image-level features of the input image \(X\) are modeled by using scaled dot-product attention. Attention operates on vectors of queries \(q\), keys \(k\) and values \(n\), and takes a weighted sum of the value vectors according to a similarity distribution between query and key vectors.
Attention can be defined as follows \citep{cornia2020m2}:

\begin{equation}
Attention(Q, K, V) = \operatorname{softmax}(\frac{QK^T}{\sqrt{d}}) V
\label{eq:binom}
\end{equation}

where \(Q\) is a matrix of \(n_q\) query vectors, \(K\) and \(V\) both contain \(n_k\) keys and values, all the vectors has the same dimensionality, and \(d\) is a scaling factor.

\hypertarget{memory-augmented-encoder}{%
\paragraph{Memory-Augmented Encoder}\label{memory-augmented-encoder}}

For the given image region \(X\), attention can be used to obtain a permutation invariant encoding of \(X\) through the self-attention operations, the operator from the Transformer can be defined as follows \citep{cornia2020m2}:

\begin{equation}
S(X) = Attention(W_q X, W_k X, W_vX)
\end{equation}

In this case, queries, keys, and values are linear projections of the input features, and \(W_q\), \(W_k\), \(W_v\) are their learnable weights, they depend solely on the pairwise similarities between linear projections of the input set X. The self-attention operator encodes the pairwise relationships inside the input.
But self-attention also has its limitation: a prior knowledge on relationships between image regions can not be modelled.
To overcome the limitation, the authors introduce a \textbf{Memory-Augmented Attention} operator by extending the keys and values with additional prior information, which does not depend on image region \(X\).
The additional keys and values are initialized as plain learnable vectors which can be directly updated via SGD.
The operator can be defined as follows \citep{cornia2020m2}:

\begin{align}
M_{mem}(X) &=  Attention(W_qX, K, V ) \notag \\
K &=  [W_kX, M_k]\notag \\
V &= [W_vX, M_v]
\end{align}

\(M_k\) and \(M_v\) are learnable matrices, with \(n_m\) rows, {[}·,·{]} indicates concatenation. The additional keys and value could help to retrieve a priori knowledge from input while keeping the quries unchanged \citep{cornia2020m2}.

For the \textbf{Encoding Layer}, a memory-augmented operator d is injected into a transformer-like layer, the output is fed into a position-wise feed-forward layer \citep{cornia2020m2}:

\begin{equation}
F(X)_i= U\sigma(V X_i + b) + c;
\end{equation}

\(X_i\) indicates the \(i\)-th vector of the input set, and \(F(X)_i\) the \(i\)-th vector of the output. Also, \(\sigma(·)\) is the ReLU activation function, \(V\) and \(U\) are learnable weight matrices, \(b\) and \(c\) are bias terms \citep{cornia2020m2}.

Each component will be complemented by a residual connection and the layer norm operation. The complete definition of an encoding layer can be finally written as \citep{cornia2020m2}:

\begin{align}
Z &= AddNorm(M_{mem}(X))\notag \\
\tilde{X}&=AddNorm(F(Z))
\end{align}

Finally the \textbf{Full Encoder} has multiple encoder layers in a sequential fashion, therefore the \(i\)-th layer uses the output set computed by layer \(i - 1\), higher encoding layers can exploit and refine relationships identified by previous layers, \(n\) encoding layers will produce the output \(\tilde{X} = (\tilde{X}^1 \dots \tilde{X}^n)\) \citep{cornia2020m2}.

\hypertarget{meshed-decoder}{%
\paragraph{Meshed Decoder}\label{meshed-decoder}}

The decoder depends on both previously generated words and image region encodings.
\textbf{Meshed Cross-Attention} can take advantage of all the encoder layers to generate captions for the image. On the right side of the Fig. \ref{fig:m2arc2} the structure of the meshed decoder is shown. The input sequence vector \(Y\) and the outputs from all encoder layers \(\tilde{X}\) are connected by the meshed attention operator gated through cross-attention. The meshed attention operator can is formally defined as \citep{cornia2020m2}:

\begin{equation}
M_{mesh}(\tilde{X}, Y) =\sum_{i = 1}^{N}\alpha_i C(\tilde{X^i}, Y)
\end{equation}

\(C(·,·)\) stands for the encoder-decoder cross-attention, it is defined with queries from decoder, while the keys and values come from the encoder \citep{cornia2020m2}.

\begin{equation}
C(\tilde{X^i}, Y) = Attention(W_q Y, W_k \tilde{X^i}, W_v \tilde{X^i})
\end{equation}

\(\alpha_i\) is a matrix of weights of the same size as the cross-attention results, \(\alpha_i\) models both single contribution of each encoder layer and the relative importance between different layers \citep{cornia2020m2}.

\begin{equation}
\alpha_i = \sigma(W_i [Y,C(\tilde{X^i}, Y)]+ b_i)
\end{equation}

The {[}·,·{]} indicates concatenation and \(\sigma(·)\) is the sigmoid activation function here, \(W_i\) is a weight matrix, and \(b_i\) is a learnable bias vector \citep{cornia2020m2}.

In decoder layers the prediction of a word should only depend on the previously generated word, so the decoder layer comprises a masked self-attention operation, which means that the operator can only make connections between queries derived from the \(t\)-th element of its input sequence Y with keys and values from left sub-sequence, i.e.~\(Y_{\leq t}\).

Simlilar as the encoder layers, the decoder layers also contain a position-wise feed-forward layer, so the decoder layer can be finally defined as \citep{cornia2020m2}:

\begin{align}
Z &= AddNorm(M_{mesh}(X,AddNorm(S_{mask}(Y ))) \notag \\
\tilde{Y} &= AddNorm(F(Z)),
\end{align}

where \(S_{mask}\) indicates a masked self-attention over time \citep{cornia2020m2}.
The full decoder with multiple decoder layers takes the input word vectors as well as the \(t\)-th element (and all elements prior to it) of its output sequence to make the prediction for the word at \(t + 1\), conditioned on \(Y_{\leq t}\). Finally the decoder takes a linear projection and a softmax operation, which can be seen as a probability distribution over all words in the vocabulary \citep{cornia2020m2}.

\hypertarget{comparison-with-other-models-on-the-ms-coco-data-sets}{%
\paragraph{Comparison with other models on the MS COCO data sets}\label{comparison-with-other-models-on-the-ms-coco-data-sets}}

The \(M^2\) Transformer was evaluated on MS COCO, which is still one of the most commonly used test data set for image captioning. Instead of using the original MS COCO dat set, \citet{cornia2020m2} follow the split of MS COCO provided by \citet{karpthy1}. Karpathy uses 5000 images for validation, 5000 images for testing and the rest for training.

For model evaluation and comparison, standard metrics for evaluating generated sequences, like BLEU \citep{papineni-etal-2002-bleu}, METEOR \citep{meteor}, ROUGE \citep{lin-2004-rouge}, CIDEr \citep{cider}, and SPICE \citep{spice}, which have been introduced in the second chapter, are used.

\begin{figure}

{\centering \includegraphics[width=1\linewidth]{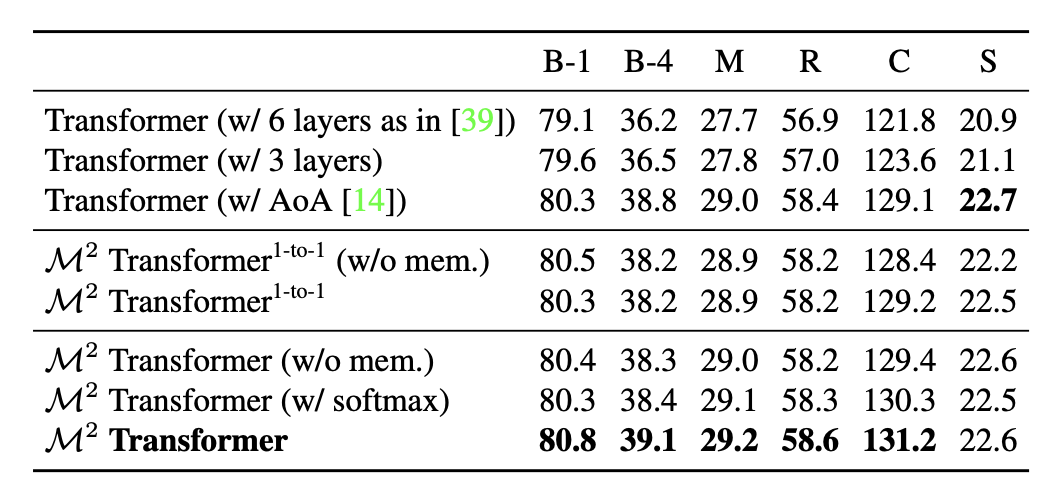}

}

\caption{Comparison of \(M^2\) with Transformer-based alternatives \citep{cornia2020m2}}\label{fig:compare1}
\end{figure}

The transformer architecture in its original configuration with six layers has been applied for captioning, researchers speculated that specific architectures might be required for captioning, so variations of the original transformer are compared with \(M^2\) Transformer. Other variations are a transformer with three layers and the ``Attention on Attention'' (AoA) approach \citep{huang1} to the attentive layers, both in the encoder and in the decoder \citep{cornia2020m2}.
The second part intends to evaluate the importance of the meshed connections between encoder and decoder layers. \(M^2\) Transformer (1 to 1) is a reduced version of the original \(M^2\) Transformer, in which one encoder layer is connected to only corresponding decoder layer instead of being connected to all the decoder layers.
As one can see from the Fig. \ref{fig:compare1}, the original Transformer has a 121.8 CIDEr score, which is lower than the reduced version of \(M^2\) Transformer, showing an improvement to 129.2 CIDEr. With respect to meshed connectivity, which helps to exploit relationships encoded at all layers and weights them with a sigmoid gating, one can observe a further improvement in CIDEr from 129.2 to 131.2. Also the role of memory vectors and the softmax gating schema for \(M^2\) Transformer are also included in the table. Eliminating the memory vector leads to a reduction of the performance by nearly 1 point in CIDEr in both the reduced \(M^2\) Transformer and the original \(M^2\) Transformer \citep{cornia2020m2}.

\begin{figure}

{\centering \includegraphics[width=1\linewidth]{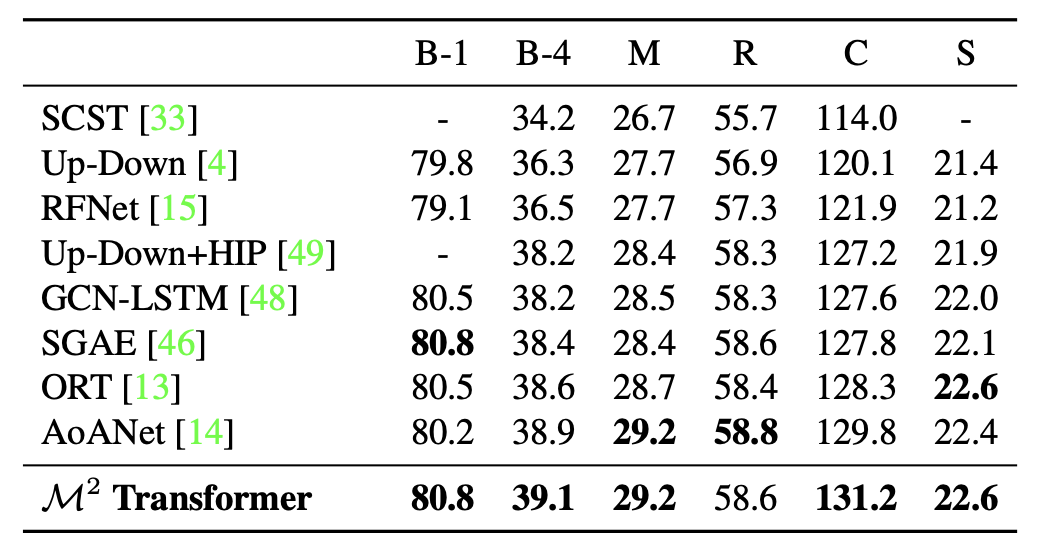}

}

\caption{Comparison with the state-of-the-art on the ``Karpathy'' test split, in single-model setting \citep{cornia2020m2}.}\label{fig:compare2}
\end{figure}

Fig \ref{fig:compare2} compares the performance of \(M^2\) Transformer with several recently proposed models for image captioning.
SCST \citep{8099614} and Up-Down \citep{8578734}, use attention over the grid of features and attention over regions. RFNet \citep{renet} uses a recurrent fusion network to merge different CNN features; GCN-LSTM \citep{GCN-LSTM} uses a Graph CNN to exploit pairwise relationships between image regions; SGAE \citep{Yang_2019_CVPR} uses scene graphs instead ofauto-encoding. The original AoA-Net \citep{huang1} approach uses attention on attention for encoding image regions and an LSTM language model. Finally, the ORT \citep{HerdadeKBS19} uses a plain transformer and weights attention scores in the region encoder with pairwise distances between detections \citep{cornia2020m2}.

In Fig. \ref{fig:compare2}, the \(M^2\) Transformer exceeds all other models on BLEU-4, METEOR, and CIDEr. The performance of the \(M^2\) Transformer was very close and competitive with SGAE on BLEU-1 and with ORT with respect to SPICE.

\begin{figure}

{\centering \includegraphics[width=1\linewidth]{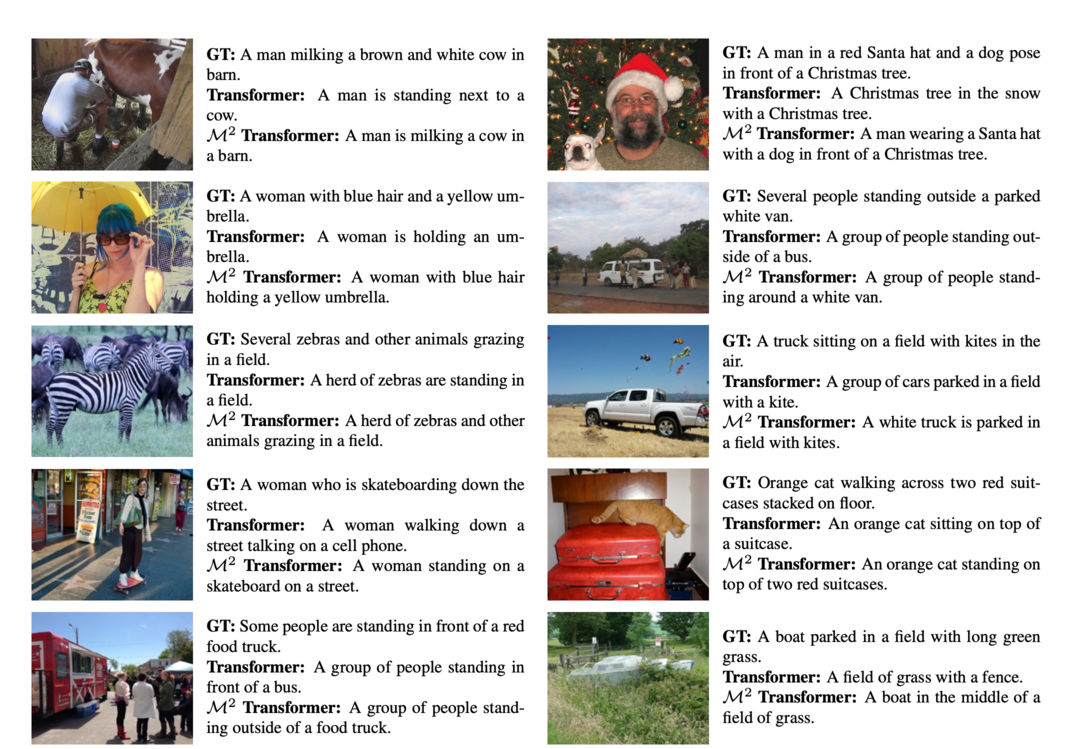}

}

\caption{Examples of captions generated by \(M^2\) Transformer and the original Transformer model, as well as the corresponding ground-truths \citep{cornia2020m2}.}\label{fig:example2}
\end{figure}

Fig. \ref{fig:example2} shows some examples of captions generated by \(M^2\) Transformer and the original transformer model, as well as the corresponding ground-truths. According to the selected examples of captions, \(M^2\) Transformer shows the ability to generate more accurate descriptions of the images, and the approach could detect the more detailed relationships between image regions \citep{cornia2020m2}.

The \(M^2\) Transformer is a new transformer-based architecture for image captioning. It improves the image encoding by learning a multi-level representation of the relationships between image regions while exploiting a priori knowledge from each encoder layer, and uses a mesh-like connectivity at decoding stage to exploit low- and high-level features at the language generation steps. The results of model evaluation with MS COCO shows that the performance of the \(M^2\) Transformer approach surpasses most of the recent approaches and achieves a new state of the art on MS COCO \citep{cornia2020m2}.

\hypertarget{c02-02-text2img}{%
\section{Text2Image}\label{c02-02-text2img}}

\emph{Author: Karol Urbańczyk}

\emph{Supervisor: Jann Goschenhofer}

Have you ever wondered what a painting artist could paint for you if you ordered \emph{a high-quality oil painting of a psychedelic hamster dragon}? Probably not. Nevertheless, one of the answers could be:

\begin{figure}

{\centering \includegraphics[width=0.4\linewidth]{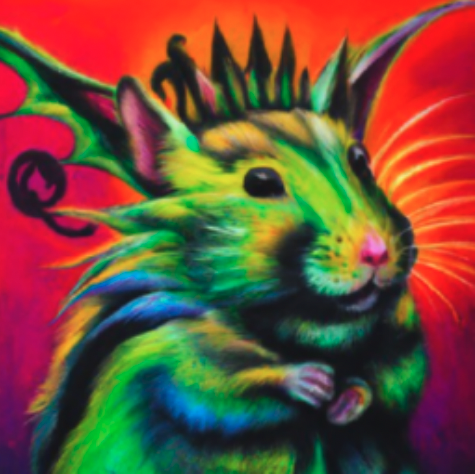}

}

\caption{Hamster dragon}\label{fig:hamsterdragon}
\end{figure}

The catch is that there is no human artist. The above picture comes from a 3.5-billion parameter model called GLIDE by OpenAI \citep{Glide2021}. Every single value of every pixel was generated from a distribution that the model had to learn in the first place. Before generating the image, GLIDE abstracted the concepts of `hamster' and `dragon' from looking at millions of training images. Only then, it was able to create and combine them successfully into a meaningful visual representation. Welcome to the world of current text-to-image modelling!

The cross-modal field of text-to-image models has developed significantly over recent years. What was considered unimaginable only a few years ago, today constitutes a new benchmark for researchers. New breakthroughs are being published every couple of months. Following these, possible business use cases are emerging, which attracts investment from the greatest players in AI research. However, a further trend of closed-source models is continuing and the text-to-image field is probably one the most obvious ones where it can be noticed. We might need to get used to the fact that the greatest capabilities will soon be monopolized by few companies.

At the same time, the general public is becoming aware of the field itself and the disruption potential it brings. Crucial questions are already emerging. What constitutes art? What does the concept of being an author mean? The result of a generative model is in a sense a combination, or variation, of the abstracts it has seen in the past. But the same stands for a human author. Therefore, is a discussion about the prejudices and biases needed? Answers to all of these will require refinement through an extensive discussion. The last section of this chapter will try to highlight the most important factors that will need to be considered.

However, the primary intention of this chapter is to present the reader with a perspective on how the field was developing chronologically. Starting with the introduction of GANs, through the first cross-domain models, and ending with state-of-the-art achievements (as of September 2022), it will also try to grasp the most important concepts without being afraid of making technical deep dives.

The author is aware that since the rapid development pace makes it nearly impossible for this section to stay up-to-date, it might very soon not be fully covering the field. However, it must be stressed that the cutting-edge capabilities of the recent models tend to come from the scale and software engineering tricks. Therefore, focusing on the core concepts should hopefully gives this chapter a universal character, at least for some time. This design choice also explains why many important works did not make it to this publication. Just to name a few of them as honorable mentions: GAWWN \citep{GAWWN2016}, MirrorGAN \citep{MirrorGAN2019}, or most recent ones: LAFITE \citep{LAFITE2021}, Make-a-Scene \citep{MakeAScene2022} or CogView \citep{CogView2021}. In one way or another, all of them pushed the research frontier one step further. Therefore, it needs to be clearly stated - the final selection of this chapter's content is a purely subjective decision of the author.

\hypertarget{seeking-objectivity}{%
\subsection{Seeking objectivity}\label{seeking-objectivity}}

Before diving into particular models, we introduce objective evaluation procedures that help assess the performance of consecutive works in comparison to their predecessors. Unfortunately, objectivity in comparing generative models is very hard to capture since there is no straight way to draw deterministic conclusions about the model's performance \citep{Evaluation2015}. However, multiple quantitative and qualitative techniques have been developed to make up for it. Unfortunately, there is no general consensus as to which measures should be used. An extensive comparison has been performed by \citet{EvaluationComparison2018}. A few of the most widely used ones in current research are presented below.

\textbf{Inception Score (IS)}

Introduced by \citet{InceptionScore2016}, Inception Score (IS) uses the Inception Net \citep{InceptionNet2015} trained on ImageNet data to classify the fake images generated by the assessed model. Then, it measures the average KL divergence between the marginal label distribution \(p(y)\) and the label distribution conditioned on the generated samples \(p(y|x)\).

\[exp(\mathop{{}\mathbb{E}}_{x}[KL(p(y|x) || p(y))])\]

\(p(y)\) is desired to have high diversity (entropy), in other words: images from the generative model should represent a wide variety of classes. On the other hand, \(p(y|x)\) is desired to have low diversity, meaning that images should represent meaningful concepts. If a range of cat images is being generated, they all should be confidently classified by Inception Net as cats. The intention behind IS is that a generative model with a higher distance (KL divergence in this case) between these distributions should have a better score. IS is considered a metric that correlates well with human judgment, hence its popularity.

\textbf{Fréchet Inception Distance (FID)}

A metric that is generally considered to improve upon Inception Score is the Fréchet Inception Distance (FID). \citet{FID2017} argue that the main drawback of IS is that it is not considering the real data at all. Therefore, FID again uses Inception Net, however this time it embeds the images (both fake and real samples) into feature space, stopping at a specific layer. In other words, some of the ultimate layers of the network are being discarded. Feature vectors are then assumed to follow a Gaussian distribution and the Fréchet distance is calculated between real and generated data distributions:

\[d^2((m, C), (m_{w}, C_{w})) = ||m-m_{w}||_{2}^2 + Tr(C+C_{w}-2(CC_{w})^{1/2})\]

where \((m, C)\) and \((m_{w}, C_{w})\) represent mean and covariance of generated and real data Gaussians respectively. Obviously, low FID levels are desired.

\begin{figure}

{\centering \includegraphics[width=1\linewidth]{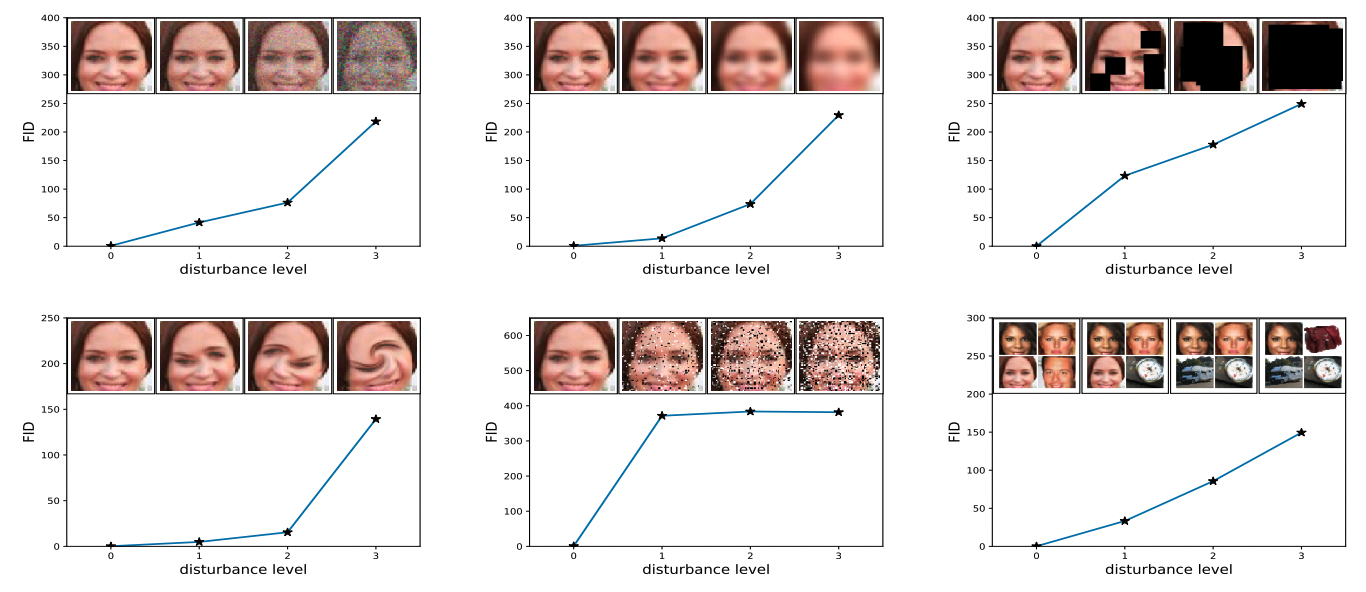}

}

\caption{FID is evaluated for different noise types. From upper left to lower right: Gaussian noise, Gaussian blur, implanted black rectangles, swirled images, salt and pepper, CelebA dataset contaminated by ImageNet images. Figure from \citet{FID2017}.}\label{fig:fiddistortions}
\end{figure}

FID is considered to be consistent with human judgement and sensitive to image distortions, which are both desired properties. Figure \ref{fig:fiddistortions} shows how FID increases (worsens) for different types of noise being added to images.

\textbf{Precision / Recall}

Precision and recall are one of the most widely used metrics in many Machine Learning problem formulations. However, their classic definition cannot be applied to generative models due to the lack of objective labels. \citet{GenerativePrecisionRecall2018} came up with a novel definition of these metrics calculated directly from distributions, which was further improved by \citet{ImprovedPrecisionRecall2019}. The argument behind the need for such an approach is that metrics such as IS or FID provide only a one-dimensional view of the model's performance, ignoring the trade-off between precision and recall. A decent FID result might very well mean high recall (large variation, i.e.~wide range of data represented by the model), high precision (realistic images), or anything in between.

Let \(P_{r}\) denote the probability distribution of the real data, and \(P_{g}\) be the distribution of the generated data. In short, recall measures to which extend \(P_{r}\) can be generated from \(P_{g}\), while precision is trying to grasp how many generated images fall within \(P_{r}\).

\begin{figure}

{\centering \includegraphics[width=0.8\linewidth]{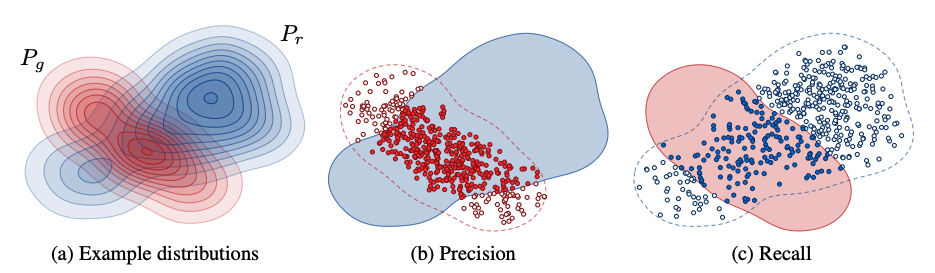}

}

\caption{Definition of precision and recall for distributions. Figure from \citet{ImprovedPrecisionRecall2019}.}\label{fig:precisionandrecall}
\end{figure}

See \citet{ImprovedPrecisionRecall2019} for a more thorough explanation.

\textbf{CLIP score}

CLIP is a model from OpenAI {[}CLIP2021{]} which is explained in detail in the chapter about \emph{text-supporting computer vision models}. In principle, CLIP is capable of assessing the semantic similarity between the text caption and the image. Following this rationale, the CLIP score can be used as metric and is defined as:

\[\mathop{{}\mathbb{E}}[s(f(image)*g(caption))]\]

where the expectation is taken over the batch of generated images and \(s\) is the CLIP logit scale \citep{Glide2021}.

\textbf{Human evaluations}

It is common that researchers report also qualitative measures. Many potential applications of the models are focused on deceiving the human spectator, which motivates reporting of metrics that are based on human evaluation. The general concept of these evaluation is to test for:

\begin{itemize}
\tightlist
\item
  photorealism
\item
  caption similarity (image-text alignment)
\end{itemize}

Usually, a set of images is presented to a human, whose task is to assess their quality with respect to the two above-mentioned criteria.

\hypertarget{generative-adversarial-networks}{%
\subsection{Generative Adversarial Networks}\label{generative-adversarial-networks}}

The appearance of Generative Adversarial Networks (GAN) was a major milestone in the development of generative models. Introduced by \citet{GAN2014}, the idea of GANs presented a novel architecture and training regime, which corresponds to a minimax two-player game between a Generator and a Discriminator (hence the word \emph{adversarial}).

GANs can be considered as an initial enabler for the field of text-to-image models and for a long time, GAN-like models were achieving state-of-the-art results, hence the presentation of their core concepts in this chapter

\hypertarget{vanilla-gan-for-image-generation}{%
\subsubsection{Vanilla GAN for Image Generation}\label{vanilla-gan-for-image-generation}}

In a vanilla GAN, the Generator model (\(G\)) and Discriminator model (\(D\)) are optimized together in a minimax game, where \(G\) aims at generating a sample so convincing, that \(D\) will not be able to distinguish whether it comes from a real or generated image distribution. On the other hand, \(D\) is being trained to discriminate between the two. Originally, a multilayer perceptron was proposed as a model architecture for both \(D\) and \(G\), although in theory any differentiable function could be used.

More formally, let \(p_{z}\) denote the prior distribution defined on the input noise vector \(z\). Then, the generator \(G(z)\) represents a function that is mapping this noisy random input to the generated image \(x\). The discriminator \(D(x)\) outputs a probability that \(x\) comes from the real data rather than generator's distribution \(p_{g}\). In this framework, \(D\) shall maximize the probability of guessing the correct label of both real and fake data. \(G\) is trained to minimize \(log(1-D(G(z)))\). Now, such representation corresponds to the following value function (optimal solution):

\[\min_{G}\min_{D}V(D,G) = \mathop{{}\mathbb{E}}_{x \sim p_{data}(x)} [log(D(x))] + \mathop{{}\mathbb{E}}_{z \sim p_{z}(z)} [log(1-D(G(z)))]\]

\begin{figure}

{\centering \includegraphics[width=0.8\linewidth]{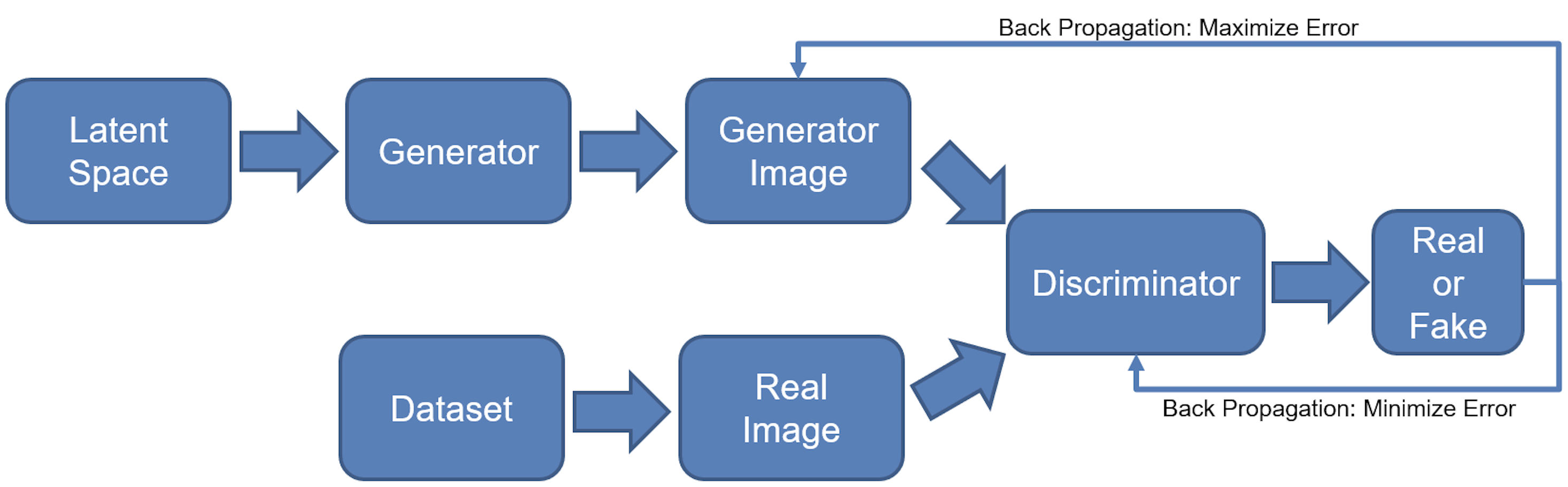}

}

\caption{GAN framework as proposed in \citet{GAN2014}.}\label{fig:vanillagan}
\end{figure}

Figure \ref{fig:vanillagan} depicts this process in a visual way.

\begin{figure}

{\centering \includegraphics[width=0.8\linewidth]{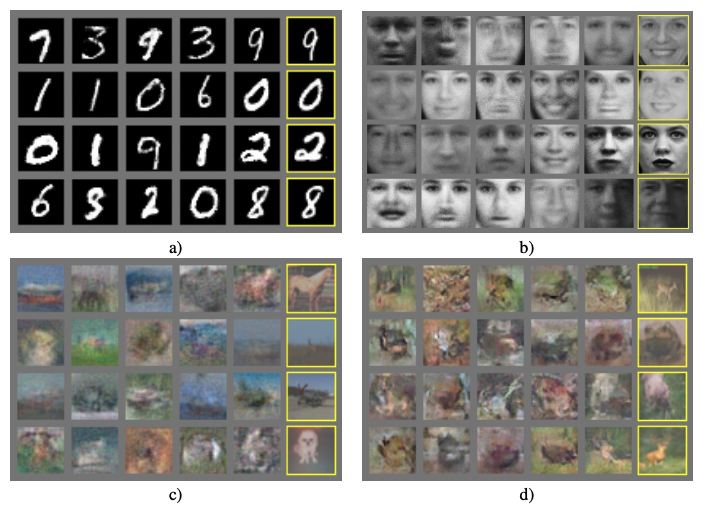}

}

\caption{Samples from generators trained on different datasets: a) MNIST b) TFD, c) CIFAR-10 (MLP used for G and D) d) CIFAR-10 (CNN used). Highlighted columns show the nearest real example of the neighbouring sample. Figure from \citet{GAN2014}.}\label{fig:vanillagansamples}
\end{figure}

Some of the generated samples that had been achieved with this architecture already in 2014 can be seen in Figure \ref{fig:vanillagansamples}.

\hypertarget{conditioning-on-text}{%
\subsubsection{Conditioning on Text}\label{conditioning-on-text}}

\begin{figure}

{\centering \includegraphics[width=0.8\linewidth]{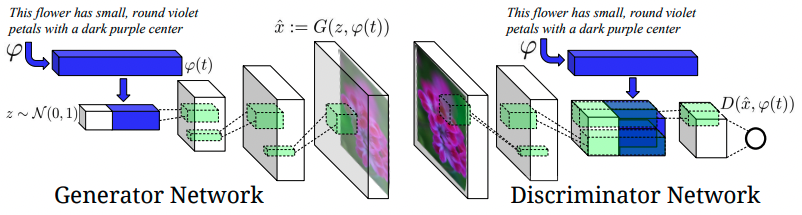}

}

\caption{The proposed architecture of the convolutional GAN that is conditioned on text. Text encoding \(\varphi(t)\) is fed into both the Generator and the Discriminator. Before further convolutional processing, it is first projected to lower dimensionality in fully-connected layers and concatenated with image feature maps. Figure from \citet{GANTextToImage2016}.}\label{fig:gancls}
\end{figure}

So far, only image generation has been covered, completely ignoring textual input. \citet{GANTextToImage2016} introduced an interesting concept of conditioning DC-GAN (GAN with CNNs as Generator and Discriminator) on textual embeddings. A separate model is being trained and used for encoding the text. Then, result embeddings are concatenated with the noise vector and fed into the Generator and the Discriminator takes embeddings as an input as well. The resulting model is referred to as GAN-INT-CLS. Both abbreviations (INT and CLS) stand for specific training choices, which are going to be explained later in the chapter. The overview of the proposed architecture can be seen in Figure \ref{fig:gancls}.

\textbf{Text embeddings}

\begin{figure}

{\centering \includegraphics[width=0.6\linewidth]{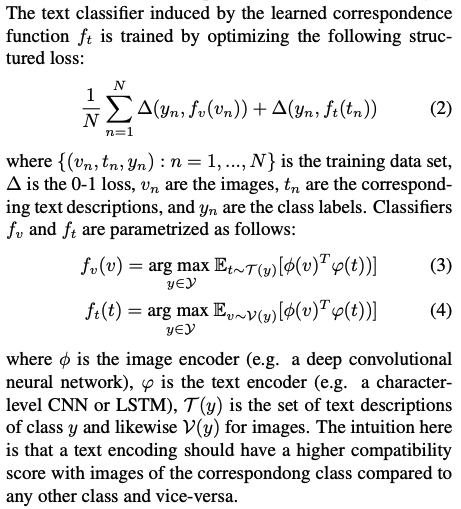}

}

\caption{Figure from \citet{GANTextToImage2016}.}\label{fig:ganclsembeddings}
\end{figure}

Since regular text embeddings are commonly trained in separation from visual modality simply by looking at the textual context, they are not well suited for capturing visual properties. This motivated \citet{JointRepresentations2016} to come up with structured joint embeddings of images and text descriptions. GAN-INT-CLS implements it in a way described in Figure \ref{fig:ganclsembeddings}.

GoogLeNet is being used as an image encoder \(\phi\). For text encoding \(\varphi(t)\), authors use a character-level CNN combined with RNN. Essentially, the objective of the training is to minimize the distance between the encoded image and text representations. The image encoder is then discarded and \(\varphi\) only is used as depicted in Figure \ref{fig:gancls}.

\textbf{GAN-CLS}

CLS stands for Conditional Latent Space, which essentially means the GAN is conditioned on the embedded text. However, in order to fully grasp how exactly the model is conditioned on the input, we need to go beyond architectural choices. It is also crucial to present a specific training regime that was introduced for GAN-CLS and the motivation behind it.

One way to train the system is to view text-image pairs as joint observations and train the discriminator to classify the entire pair as real or fake. However, in such a case the discriminator does not have an understanding of whether the image matches the meaning of the text. This is because the discriminator does not distinguish between two types of error that exist, namely when the image is unrealistic or when it is realistic but the text does not match.

A proposed solution to this problem is to present the discriminator with three observations at a time, all of which are included later in the loss function. These three are: \{real image with right text\}, \{real image with wrong text\}, \{fake image with right text\}. The intention is that the discriminator should classify them as \{true\}, \{false\}, \{false\}, respectively.

\textbf{GAN-INT}

\begin{figure}

{\centering \includegraphics[width=0.6\linewidth]{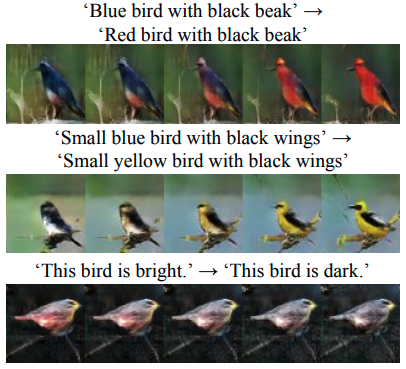}

}

\caption{Interpolating between sentences. Figure from \citet{GANTextToImage2016}.}\label{fig:interpolatingbirds}
\end{figure}

The motivation behind this concept comes from the fact that interpolating between text embeddings tends to create observation pairs that are still close to the real data manifold. Therefore, generating additional synthetic text embeddings and using them instead of real captions in the training process might help in the sense that it works as a form of data augmentation and helps regularize the training process. Figure \ref{fig:interpolatingbirds} might be helpful for developing the intuition behind the interpolation process.

\textbf{Results}

\begin{figure}

{\centering \includegraphics[width=1\linewidth]{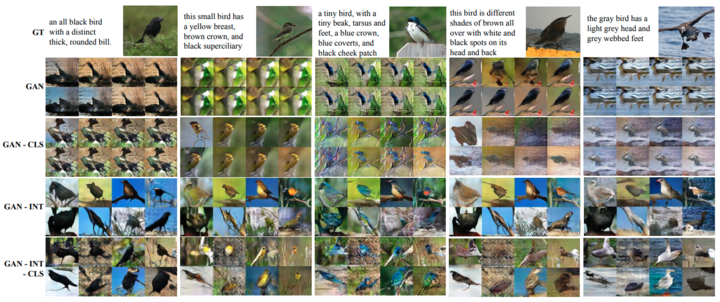}

}

\caption{Zero-shot generated birds using GAN, GAN-CLS, GAN-INT, GAN-INT-CLS. Figure from \citet{GANTextToImage2016}.}\label{fig:ganclszeroshot}
\end{figure}

\begin{figure}

{\centering \includegraphics[width=1\linewidth]{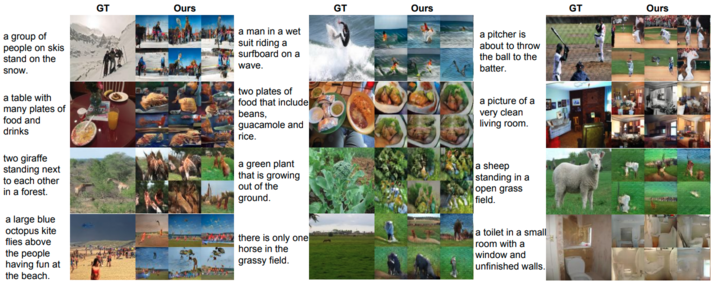}

}

\caption{Generated images using GAN-CLS on MS-COCO validation set. Figure from \citet{GANTextToImage2016}.}\label{fig:ganclsmscoco}
\end{figure}

The model achieves the best performance when both of the mentioned methods are in use (GAN-INT-CLS). Models prove to successfully transfer style (pose of the objects) and background from the training data when trained on CUB (birds) and Oxford-102 (flowers) datasets. They also show interesting zero-shot abilities, meaning they can generate observations from unseen test classes (Figure \ref{fig:ganclszeroshot}). When trained on MS-COCO, GAN-CLS proves its potential to generalize over many domains, although the results are not always coherent (Figure \ref{fig:ganclsmscoco}).

\hypertarget{further-gan-like-development}{%
\subsubsection{Further GAN-like development}\label{further-gan-like-development}}

Generative Adversarial Networks were a leading approach for text-to-image models for most of the field's short history. In the following years after the introduction of GAN-INT-CLS, new concepts were emerging, trying to push the results further. Many of them had a GAN architecture as their core part. In this section, a few such ideas are presented. The intention is to quickly skim through the most important ones. A curious reader should follow the corresponding papers.

\textbf{StackGAN}

\citet{StackGAN2016} introduced what the StackGAN. The main contribution of the paper which also found its place in other researchers' works, was the idea to \emph{stack} more than one generator-discriminator pair inside the architecture. Stage-II (second pair) generator is supposed to improve the results from Stage-I, taking into account only:

\begin{itemize}
\tightlist
\item
  text embedding (same as Stage-I)
\item
  image generated in Stage-I
\end{itemize}

without a random vector. Deliberate omission of the random vector results in the generator directly working on improving the results from Stage-I. The purpose is also to increase resolution (here from 64x64 to 256x256). Authors obtained great results already with two stages, however, in principle architecture allows for stacking many of them.

\begin{figure}

{\centering \includegraphics[width=1\linewidth]{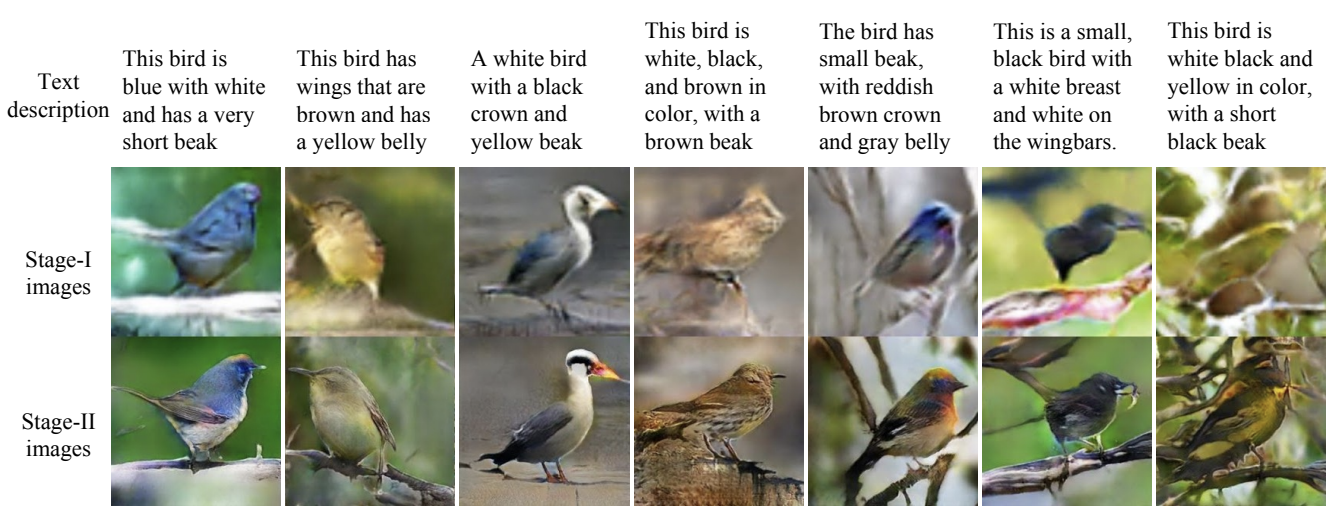}

}

\caption{(ref:stackgan)}\label{fig:stackgan}
\end{figure}

\textbf{AttnGAN}

It is 2017 and many researchers believe attention is all they need \citep{AttentionIsAllYouNeed2017}. Probably for the first time in text-to-image generation attention mechanism was used by \citet{AttnGAN2017}. The authors combined the idea with what StackGAN proposed and used three stages (generators \(G_{0}\), \(G_{1}\) and \(G_{2}\)). However, this time first layers of a particular generator are attending to word feature vectors. This mechanism not only helps control how particular areas of the image are being improved by consecutive generators but also allows for visualizing attention maps.

\begin{figure}

{\centering \includegraphics[width=0.8\linewidth]{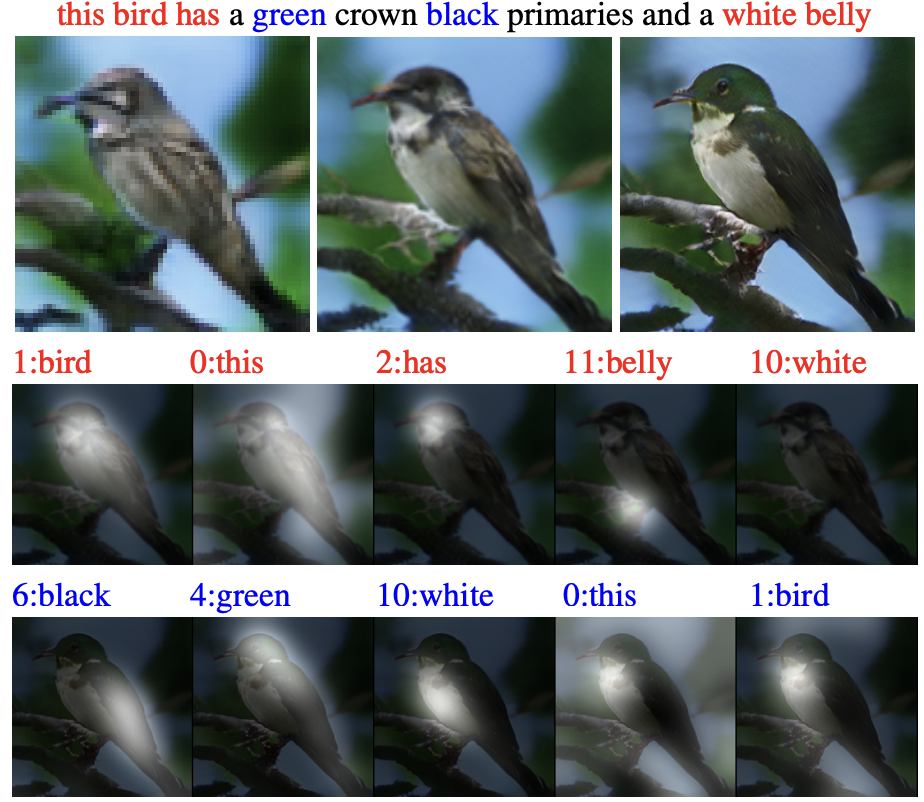}

}

\caption{Images generated by \(G_{0}\), \(G_{1}\), \(G_{2}\). Two bottom rows show 5 most attended words by \(G_{1}\) and \(G_{2}\) respectively. Figure from \citet{AttnGAN2017}.}\label{fig:attngan}
\end{figure}

\textbf{DM-GAN}

Another important milestone was DM-GAN (Dynamic Memory GAN) \citep{DMGAN2019}. At that time, models were primarily focusing on generating the initial image and then refining it to a high-resolution one (as e.g.~StackGAN does). However, such models heavily depend on the quality of the first image initialization. This problem was the main motivation for the authors to come up with a mechanism to prevent it. DM-GAN proposes a dynamic memory module, which has two main components. First, its memory writing gate helps select the most important information from the text based on the initial image. Second, a response gate merges the information from image features with the memories. Both of these help refine the initial image much more effectively.

\textbf{DF-GAN}

Last but not least, DF-GAN (Deep Fusion GAN) \citep{DFGAN2020} improves the results by proposing three concepts. One-Stage Text-to-Image Backbone focuses on providing an architecture that is capable of abandoning the idea of multiple stacked generators and using a single one instead. It achieves that by a smart combination of a couple of factors, i.a. hinge loss and the use of residual blocks. Additionally, Matching-Aware Gradient Penalty helps achieve high semantic consistency between text and image and regularizes the learning process. Finally, One-Way Output helps the process converge more effectively.

\hypertarget{dall-e-1}{%
\subsection{Dall-E 1}\label{dall-e-1}}

OpenAI's Dall-E undoubtedly took the text-to-image field to another level. For the first time, a model showed great zero-shot capabilities, comparable to previous domain-specific models. To achieve that, an unprecedented scale of the dataset and training process was needed. 250 million text-image pairs were collected for that purpose, which enabled training of a 12-billion parameter version of the model. Unfortunately, Dall-E is not publicly available and follows the most recent trend of closed-source models. Or, to put it more precisely, it started this trend, and GLIDE, Dall-E 2, Imagen, Parti and others followed. Nevertheless, Dall-E's inner workings are described in \citet{DALLE1} and this section will try to explain its most important parts. However, before that, it is crucial to understand one of the fundamental concepts that has been around in the field of generative models for already quite some time - namely Variational Autoencoders.

\textbf{Variational Autoencoder (VAE)}

\begin{figure}

{\centering \includegraphics[width=1\linewidth]{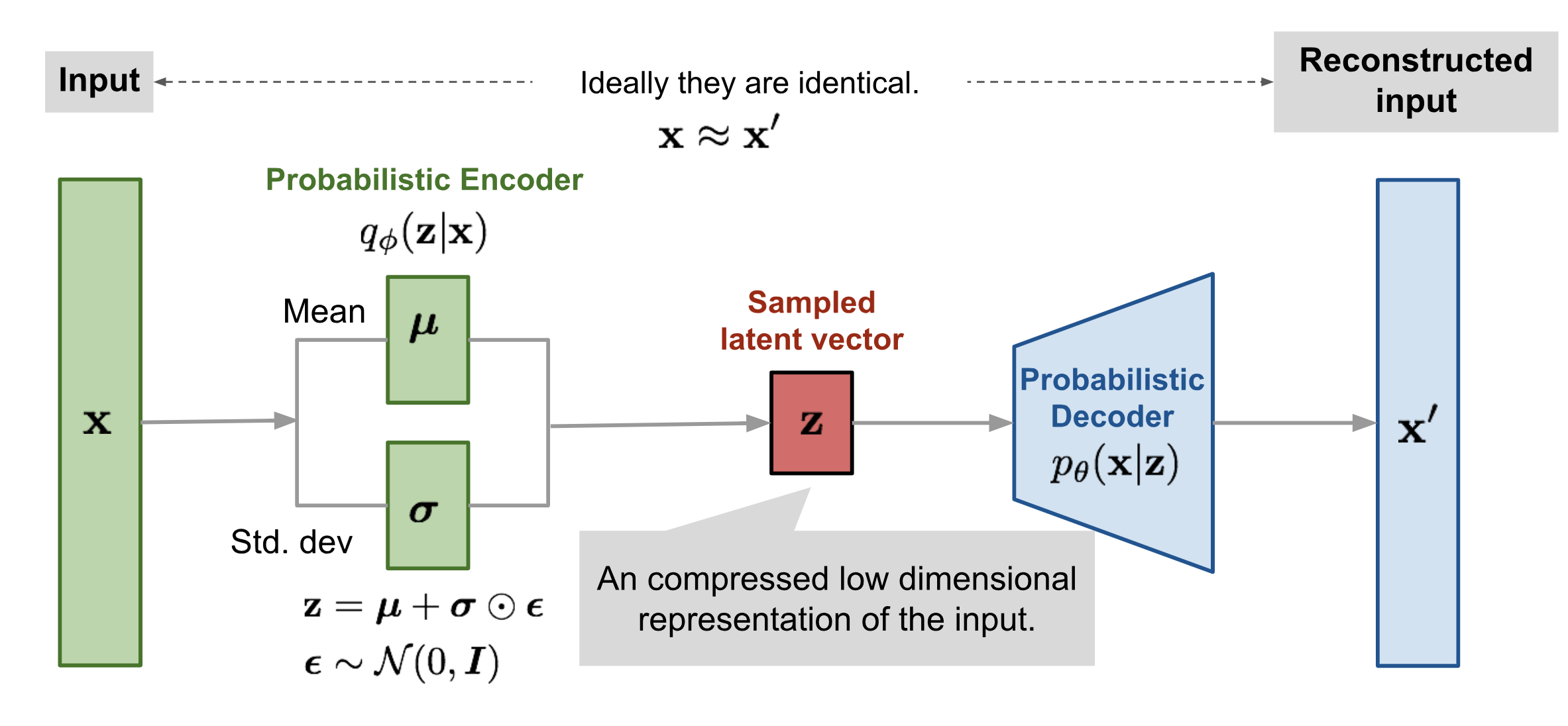}

}

\caption{Variational (probabilistic) Autoencoder architecture. Figure from \citet{weng2018VAE}.}\label{fig:vae}
\end{figure}

The regular Autoencoder architecture aims at finding an identity function that is capable of finding a meaningful representation of the data in lower-dimensional space and then reconstructing it. It is considered an unsupervised learning method for dimensionality reduction, however, trained in a supervised regime with the data itself being the label. The component performing the reduction is called an encoder, while the part responsible for the reconstruction is called a decoder. The idea behind Variational Autoencoder \citep{VAE2013} is similar, however, instead of learning the mapping to a static low-dimensional vector, the model learns its distribution. This design equips the decoder part with desired generative capabilities, as sampling from the latent low-dimensional space will result in varying data being generated. The architecture is depicted in Figure \ref{fig:vae}.

\(q_{\phi}(z|x)\) denotes the encoder under the assumption that \(z\) comes from multivariate Gaussian. \(\mu\) and \(\sigma\) are being learned. Reconstruction process is modelled by conditional probability \(p_{\theta}(x|z)\), given samples latent vector \(z\).

\textbf{VQ-VAE / dVAE}

\begin{figure}

{\centering \includegraphics[width=1\linewidth]{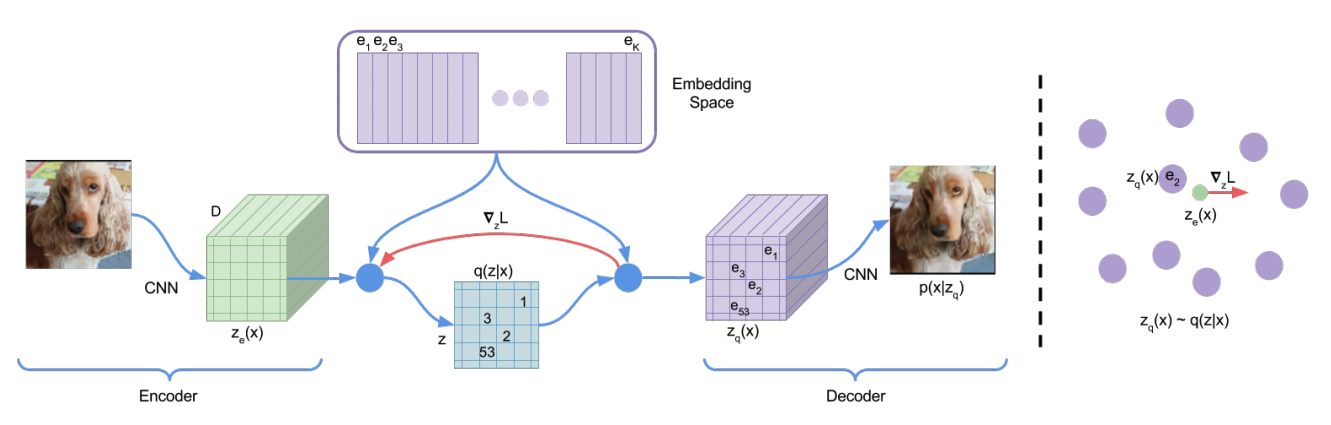}

}

\caption{VQ-VAE architecture. Figure from \citet{VQVAE2017}.}\label{fig:vqvae}
\end{figure}

The VQ-VAE (Vector Quantized VAE) \citep{VQVAE2017} differs from the regular VAE in the way it approaches encoding the latent space. Instead of mapping data into a continuous distribution, the Vector Quantized version does it in a discrete way. This is motivated by the fact that for many data modalities it is more natural to represent them in a discrete way (e.g.~speech, human language, reasoning about objects in images, etc.). VQ-VAE achieves that by using a separate codebook of vectors. The architecture is depicted in Figure \ref{fig:vqvae}.

The idea is to map the output of the encoder to one of the vectors from the \(K\)-dimensional codebook. This process is called quantization and essentially means finding the vector that is the nearest neighbour to the encoder's output (in a sense of Euclidean distance). Since this moment, this newly found vector from the codebook is going to be used instead. The codebook itself is also subject to the learning process. One could argue that passing gradients during the training through such a discrete system might be problematic. VQ-VAE overcomes this problem by simply copying gradients from the decoder's input to the encoder's output. A great explanation of the training process and further mathematical details can be found in \citet{weng2018VAE} and \citet{UnderstandingVQVAE}.

Dall-E, however, is using what is called dVAE. Essentially, it is a VQ-VAE with a couple of details changed. In short, the main difference is that instead of learning a deterministic mapping from the encoder's output to the codebook, it produces probabilities of a latent representation over all codebook vectors.

\textbf{Dall-E system}

Dall-E is composed of two stages. The above introduction of VQ-VAE was necessary to understand the first one. Essentially, it is training dVAE to compress 256x256 images into a 32x32 grid of tokens. This model will play a crucial role in the second stage.

The second stage is about learning the prior distribution of text-image pairs. First, the text is byte-pair \citep{BPE2015} encoded into a maximum of 256 tokens, where the vocabulary is of size 16384. Next, the image representation encoded by previously trained dVAE is unrolled (from 32x32 grid to 1024 tokens) and concatenated to the text tokens. This sequence (of 256+1024 tokens) is used as an input for a huge transformer-like architecture. Its goal is to autoregressively model the next token prediction.

During inference time, the text caption is again encoded into 256 tokens at most. The generation process starts with predicting all of the next 1024 image-related tokens. They are later decoded with the dVAE decoder that was trained in the first step. Its output represents the final image.

\textbf{Results}

\begin{figure}

{\centering \includegraphics[width=0.8\linewidth]{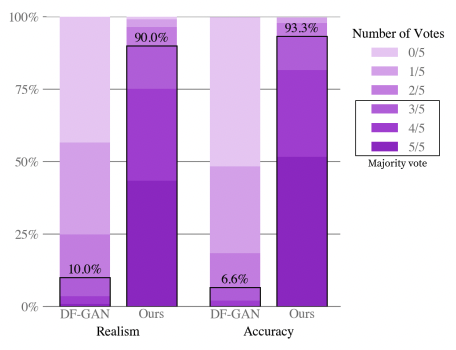}

}

\caption{Human evaluation of Dall-E vs DF-GAN on text captions from the MS-COCO dataset. When asked for realism and caption similarity, evaluators preferred Dall-E's results over 90\textbackslash\% of the time. Figure from \citet{DALLE1}.}\label{fig:dallephotorealism}
\end{figure}

Results achieved with the original Dall-E attracted so much attention mainly due to its diversity and zero-shot capabilities. Dall-E was capable of producing better results compared to previous state-of-the-art models which were trained on data coming from the same domain as data used for evaluation. One comparison can be seen in Figure \ref{fig:dallephotorealism}.

\begin{figure}

{\centering \includegraphics[width=1\linewidth]{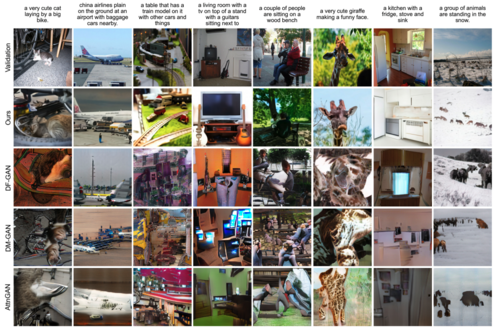}

}

\caption{Comparison of the results from Dall-E vs prior works on MS-COCO. Dall-E's outputs are chosen as the best out of 512 images, ranked by a contrastive model. Figure from \citet{DALLE1}.}\label{fig:dalleexamples}
\end{figure}

Outputs of some of the prior approaches described in this chapter compared with Dall-E can be seen in Figure \ref{fig:dalleexamples}.

\textbf{Limitations}

Although Dall-E made a huge step forward in text-to-image modelling, it still showed multiple flaws. First, photorealism of the outputs is still relatively low. In other words, when prompted for images containing realistic situations, it is rarely capable of \emph{deceiving} human evaluators. Second, the model has evident problems with understanding relatively complex abstractions, such as text inside an image, or relative object positions in the scene.

\hypertarget{glide}{%
\subsection{GLIDE}\label{glide}}

Introduced by \citet{Glide2021}, GLIDE started an era of huge-scale diffusion models. The concept of diffusion has already been used in the area of Deep Learning for some time before. However, the authors of GLIDE took a step further and combined it together with text-based guidance which is supposed to steer the learning process in the direction of the text's meaning. This powerful method was proven to achieve outstanding results which remain competitive with current state-of-the-art models at the time of writing.

\textbf{Diffusion models}

Before understanding the inner workings of GLIDE, it is important to introduce the core concept that is driving it, namely diffusion. The idea of diffusion originates from physics. In short, it corresponds to the process of diffusing particles, for example of one fluid in another. Normally it has a unidirectional character, in other words, it cannot be reversed. However, as \citet{Diffusion2015} managed to show, and \citet{DenoisingDiffusion2020} later improved, if the data diffusion process is modelled as a Markov chain with Gaussian noise being added in consecutive steps, it is possible to learn how to reverse it. This reversed process is exactly how images are generated by the model from pure random noise.

Let us construct a Markov chain, where the initial data point is denoted by \(x_{0}\). In \(t\) steps, Gaussian noise is added to the data. The distribution of the data at \(t\)-step can be characterized in the following way:

\[q(x_{t}|x_{t-1}):=N(x_{t};\sqrt{\alpha_{t}}x_{t-1},(1-\alpha_{t})I)\]

where \((1-\alpha_{t})\) parametrizes the magnitude of the noise being added at each step. Now, if \(x_{t-1}\) was to be reconstructed from \(x_{t}\), a model needs to learn to predict estimates of gradients from the previous steps. The probability distribution of previous steps can be estimated as follows:

\[p_{\theta}(x_{t-1}|x_{t})=N(x_{t-1};\mu_{\theta}(x_{t}),\Sigma_{\theta}(x_{t}))\]

where the mean function \(\mu_{\theta}\) was proposed by \citet{DenoisingDiffusion2020}. For a more detailed explanation of how this is later parametrized and trained, one could follow \citet{weng2021diffusion}.

\textbf{GLIDE system}

GLIDE can essentially be broken down into two parts. The first of them is the pretrained Transformer model, which in principle is responsible for creating the text embeddings. The last token embedding is used as a class embedding (text representation) in later stages. Additionally, all tokens from the last embedding layer are being used (\emph{attended to}) by all attention layers in the diffusion model itself. This makes the model aware of the text meaning while reconstructing the previous step in the Markov chain.

The second component of the GLIDE is the diffusion model itself. A U-Net-like architecture with multiple attention blocks is used here. This part's sole goal is to model \(p_{\theta}(x_{t-1}|x_{t},y)\), where \(y\) corresponds to last token embedding mentioned above. Or, to put it differently, to predict \(\epsilon_{\theta}(x_{t}|y)\) since the problem can be reframed as calculating the amount of noise being added at each step.

Additionally, to make the model even more aware of the text's meaning, guidance is being used at inference time. In short, the idea is to control the direction of the diffusion process. The authors test two different approaches. First, they try guidance with the use of a separate classifier, OpenAI's CLIP in this case. However, better results were in general achieved by the classifier-free guidance process. The idea is to produce two different images at each step. One is conditioned on text, while the other one is not. Distance between them is calculated and then, after significant scaling, added to the image obtained without conditioning. This way, the model speeds up the progression of the image towards the meaning of the text. This process can be written as:

\[\hat{\epsilon}_\theta(x_{t}|y)=\epsilon_{\theta}(x_{t}|\emptyset)+s*(\epsilon_{\theta}(x_{t}|y)-\epsilon_{\theta}(x_{t}|\emptyset))\]

where \(s\) denotes the parameter for scaling the difference between the mentioned images.

\textbf{Results}

\begin{figure}

{\centering \includegraphics[width=0.6\linewidth]{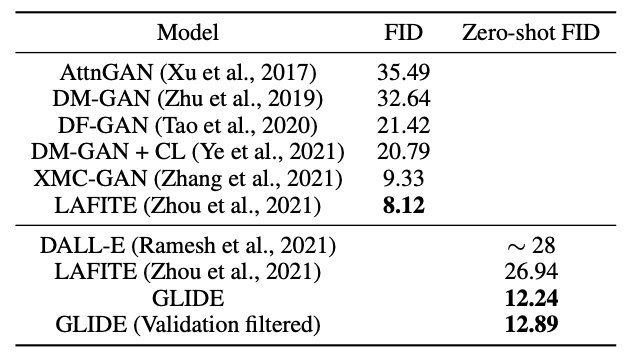}

}

\caption{Comparison of FID on MS-COCO 256×256. Figure from \citet{Glide2021}.}\label{fig:glidefid}
\end{figure}

GLIDE achieves significantly more photorealistic results compared to its predecessors. FID scores reported on the MS-COCO 256x256 dataset can be seen in Figure \ref{fig:glidefid}. It is worth noting that GLIDE was not trained on this dataset, hence its zero-shot capabilities are even more impressing.

\begin{figure}

{\centering \includegraphics[width=0.6\linewidth]{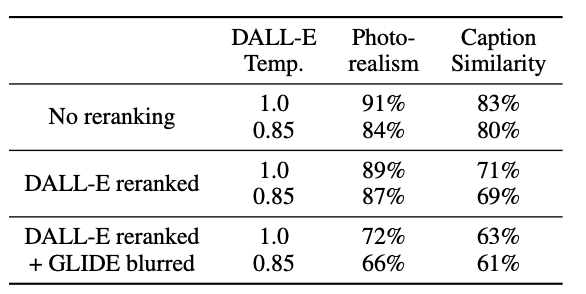}

}

\caption{Win probabilities of GLIDE vs DALL-E. Figure from \citet{Glide2021}.}\label{fig:gliderealism}
\end{figure}

Results are also preferred by human evaluators in terms of photorealism and the similarity of the image to its caption. A comparison to DALL-E 1 results can be seen in Figure \ref{fig:gliderealism}

\begin{figure}

{\centering \includegraphics[width=1\linewidth]{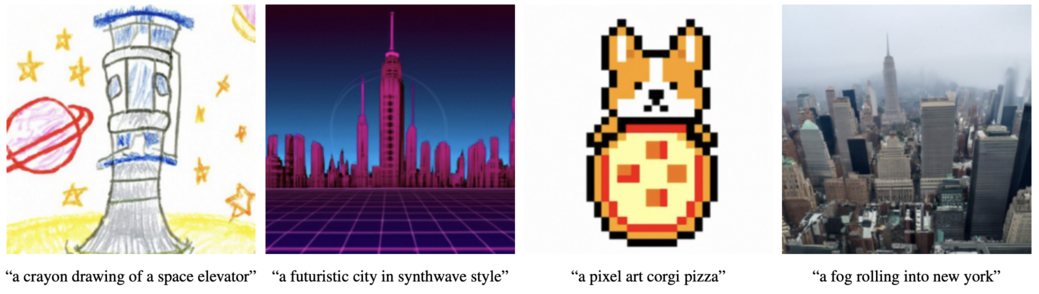}

}

\caption{Samples from GLIDE with classifier-free-guidance and s=3. Figure from \citet{Glide2021}.}\label{fig:glideresults}
\end{figure}

Finally, some of the cherry-picked images together with their corresponding captions can be seen in Figure \ref{fig:glideresults}.

\textbf{Limitations}

\begin{figure}

{\centering \includegraphics[width=1\linewidth]{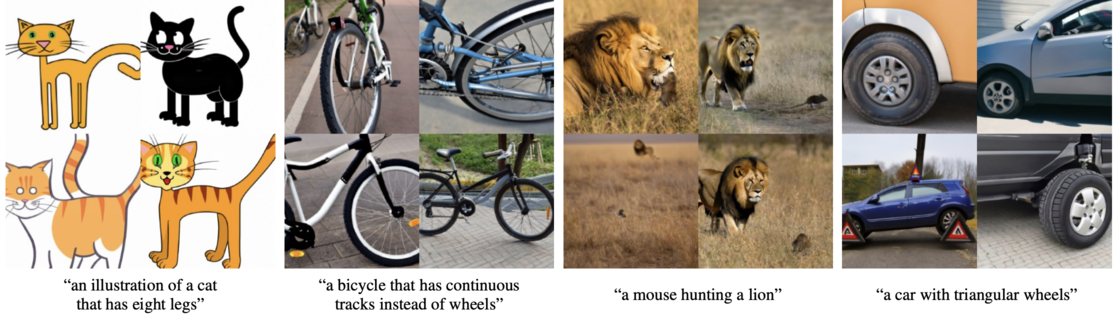}

}

\caption{Failures happen mostly for unusual prompts. Figure from \citet{Glide2021}.}\label{fig:glidefails}
\end{figure}

GLIDE suffers from two problems. First, it fails when being presented with a complex or unusual text prompt. A few examples can be seen in Figure \ref{fig:glidefails}. Also, the model is relatively slow at inference time (much slower than GANs). This is caused by the sequential character of the architecture, where consecutive steps in Markov chain reconstruction cannot be simply parallelized.

\hypertarget{dall-e-2-unclip}{%
\subsection{Dall-E 2 / unCLIP}\label{dall-e-2-unclip}}

The contribution that probably attracted the most attention in the field is known under the name Dall-E 2 \citep{DALLE2}. For the first time, the wider public had picked interest in its potential applications. This might be due to a great PR that could be seen from the authors, namely OpenAI. Dall-E 2, also known as just Dall-E, or unCLIP, has been advertised as a successor of Dall-E 1, on which results it significantly improved. In reality, the architecture and the results it achieved are much more similar to that of GLIDE. Additionally, social media has been flooded with images generated by the model. This was possible thanks to OpenAI giving access to it to everybody who was interested and patient enough to get through a waiting list. However, the model itself again remains unpublished. Another factor that might have contributed to Dall-E's success were its inpainting and outpainting capabilities. Although, it is worth mentioning they were already also possible with GLIDE.

In essence, UnCLIP is a very smart combination of pior work from OpenAI that was re-engineered and applied in a novel way. Nevertheless, the model represents a significant leap forward, which is why it cannot be omitted in this chapter.

\textbf{Dall-E 2 system}

UnCLIP consists of two components: prior and decoder. Let \(x\) be the image and \(y\) its caption. \(z_{i}\) and \(z_{t}\) are CLIP image and text embedding of this \((x, y)\) pair. Then, \emph{prior} \(P(z_{i}|y)\) is responsible for producing CLIP image embeddings conditioned on the text caption. A decoder \(P(x|z_{i},y)\) outputs an image conditioned on the CLIP image embedding and, again, the text caption itself.

For the \emph{prior} authors try two different approaches, namely autoregressive and diffusion models. The latter ended up yielding slightly better results. The diffusion prior isa Transformer taking as an input a special sequence of an encoded text prompt, CLIP text embedding, embedding for the diffusion step, and a noised CLIP image embedding.

The decoder consists of diffusion models again. Firstly, a GLIDE-like model takes a CLIP image embedding as its \(x_{t}\) instead of the pure noise that was used in its original version. Similarly to the original GLIDE, classifier-free guidance is applied, however with slight differences. Lastly, two diffusion upsampler models are trained to bring images first from 64x64 to 256x256, and then from 256x256 to 1024x1024 resolution. The authors found no benefit in conditioning these models on text captions. Finally, unCLIP can be summarized as a mixture of GLIDE and CLIP with a lot of engineering behind it.

\textbf{Results}

\begin{figure}

{\centering \includegraphics[width=0.8\linewidth]{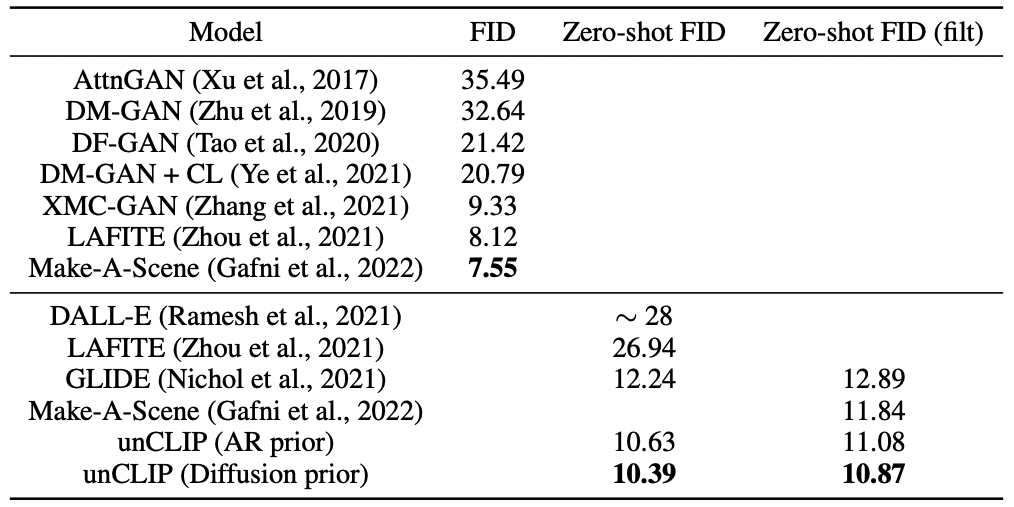}

}

\caption{Comparison of FID on MS-COCO. The best results for unCLIP were reported with the guidance scale of 1.25. Figure from \citet{DALLE2}.}\label{fig:uncliptable}
\end{figure}

\begin{figure}

{\centering \includegraphics[width=1\linewidth]{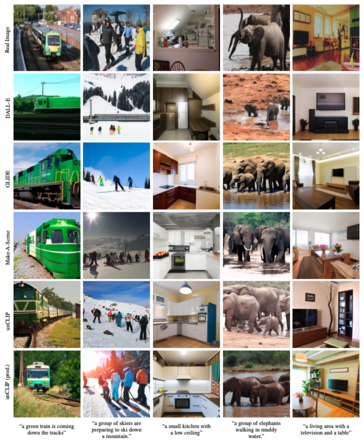}

}

\caption{Image samples on MS-COCO text prompts. Figure from \citet{DALLE2}.}\label{fig:unclipimages}
\end{figure}

When compared to GLIDE, unCLIP shows it is capable of representing a wider diversity of the data, while achieving a similar level of photorealism and caption similarity. Comparison to previous works on the MS-COCO dataset shows that unCLIP achieves unprecedented FID (Figure \ref{fig:uncliptable}). A few output examples calculated on MS-COCO captions can be found in Figure \ref{fig:unclipimages}.

\textbf{Limitations}

\begin{figure}

{\centering \includegraphics[width=1\linewidth]{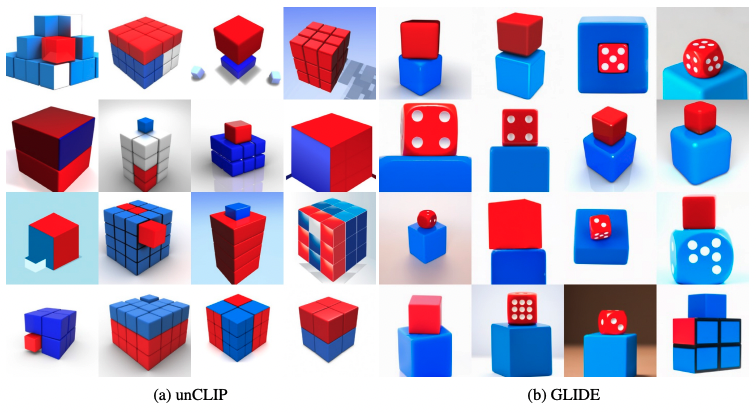}

}

\caption{`a red cube on top of a blue cube' Figure from \citet{DALLE2}.}\label{fig:cube}
\end{figure}

\begin{figure}

{\centering \includegraphics[width=1\linewidth]{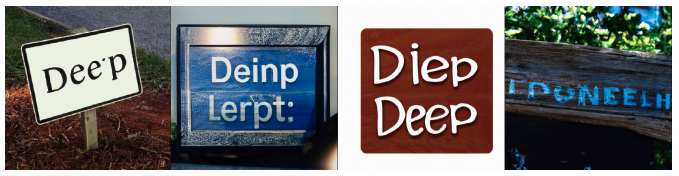}

}

\caption{`A sign that says deep learning.' Figure from \citet{DALLE2}.}\label{fig:sign}
\end{figure}

\begin{figure}

{\centering \includegraphics[width=1\linewidth]{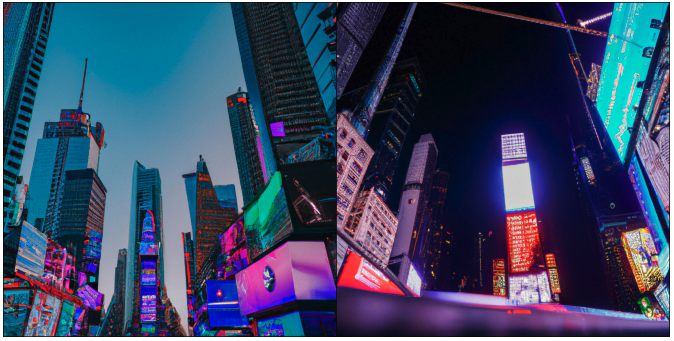}

}

\caption{`A high quality photo of Times Square.' Figure from \citet{DALLE2}.}\label{fig:timessquare}
\end{figure}

UnCLIP suffers from very similar problems as its predecessor GLIDE. First, compositionality in the images tends to sometimes be confused by the model. Failure cases can be seen in Figure \ref{fig:cube}. Second, UnCLIP struggles with generating coherent text inside an image (Figure \ref{fig:sign}). The authors hypothesize that using CLIP embeddings, although improving diversity, might be responsible for making these problems more evident than in GLIDE. Lastly, UnCLIP often fails with delivering details in highly complex scenes (Figure \ref{fig:timessquare}). Again, according to the authors, this might be a result of the fact that the decoder is producing only 64x64 images which are later upsampled.

\hypertarget{imagen-parti}{%
\subsection{Imagen \& Parti}\label{imagen-parti}}

Only a few months after unCLIP was released by OpenAI, for the first time Google came into play with its new autoregressive model called Imagen \citep{Imagen2022}. Another one followed just two months later - Parti \citep{Parti2022}. Both of these models pushed the boundaries even further, although they take entirely different approaches. None of them is introducing a completely new way of looking at the problem of text-to-image generation. Their advancements come from engineering and further scaling existing solutions. However, it must be stressed that currently (September 2022) they are delivering the most outstanding results.

Imagen is a diffusion model. Its main contribution is that instead of using a text encoder trained on image captions, it actually uses a huge pretrained NLP model called T5-XXL \citep{T5XXL2019} that is taken off the shelf and frozen. Authors argue that this helps the model understand language much more deeply, as it has seen more diverse and complex texts than just image captions.

On the other hand, Parti takes an autoregressive approach. Similarly to the first version of Dall-E, it consists of two stages, namely the image tokenizer and sequence-to-sequence autoregressive part which is responsible for generating image tokens from a set of text tokens. In this case, ViT-VQGAN \citep{VitVQGAN2021} is used as a tokenizer and the autoregressive component is again Transformer-like.

\textbf{Results}

\begin{figure}

{\centering \includegraphics[width=0.8\linewidth]{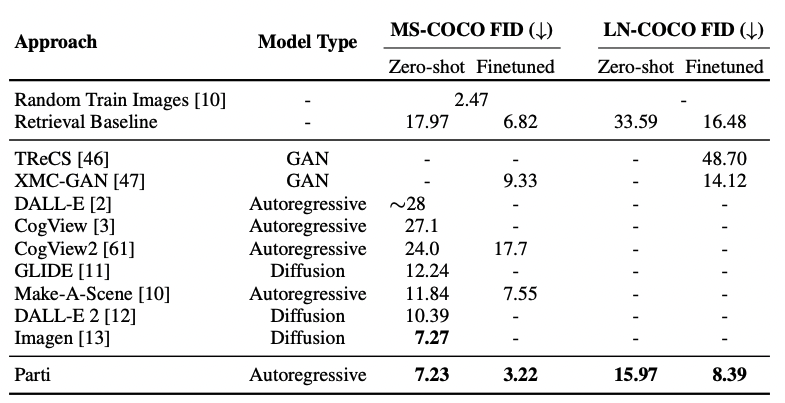}

}

\caption{Comparison of FID on MS-COCO. Figure from \citet{Parti2022}.}\label{fig:partiresults}
\end{figure}

Both of the models improved the FID significantly compared to the previous works. Figure \ref{fig:partiresults} shows the comparison.

\begin{figure}

{\centering \includegraphics[width=1\linewidth]{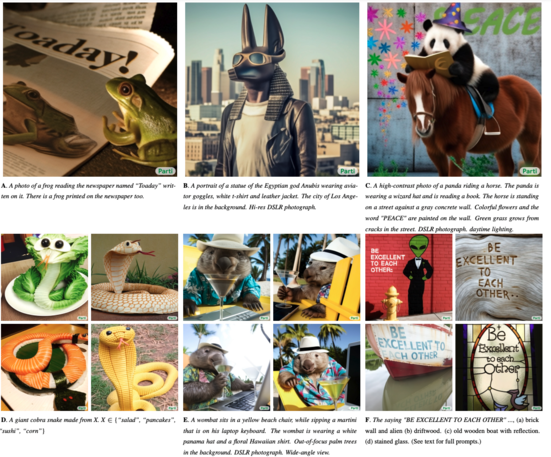}

}

\caption{Selected samples from Parti. Figure from \citet{Parti2022}.}\label{fig:partiimages}
\end{figure}

Samples from Parti can be seen in Figure \ref{fig:partiimages}. They are included here on purpose - this is the current state-of-the-art as of the moment of writing!

\textbf{Limitations}

\citet{Parti2022} mention an extensive list of problems, with which Parti still struggles. At this point, all of them can be treated as a set that is common to almost all available models. Among others, they touch:

\begin{itemize}
\tightlist
\item
  feature blending (where features of two different objects are missed)
\item
  omission or duplicating details
\item
  displaced positioning of objects
\item
  counting
\item
  negation in text prompts
\end{itemize}

and many many more. These flaws pose a challenge for future research and undoubtedly they are the ones that need to be addressed first to enable another leap forward in the field of text-to-image generation.

\hypertarget{discussion-1}{%
\subsection{Discussion}\label{discussion-1}}

Lastly, it is important to mention a couple of different topics, or trends, which are intrinsically linked with text-to-image generation. Together with previous sections, they should give the reader a holistic view of where research currently stands (again, as of September 2022).

\textbf{Open- vs closed-source}

The first trend that has emerged only recently is AI labs to not open-source their state-of-the-art models and training data. This is in clear opposition to how the entire AI community was behaving from the very beginning of the recent Deep Learning era Apparently, possible commercial opportunities that come along with owning the software are too big to be ignored. The trend is very disruptive - it is clear that the community is currently witnessing the maturation of AI business models. Needless to say, it is followed by all the greatest AI labs, just to name a few: OpenAI, DeepMind, Google Brain, Meta AI, and many others. As long as commercial achievements will have an edge over academic community research, it is highly doubtful that the trend will be reversed. However, it needs to be stressed that all of them are still issuing more or less detailed technical specifications of their work in the form of scientific papers, which is definitely a positive factor. We, as a community, can only hope it will not change in the future.

\textbf{Open-Source Community}

As the trend of closed-sourceness is clearly visible across many Deep Learning areas, the text-to-image research is actually well represented by an open-source community. The most important milestones of the recent years indeed come from OpenAI, however, new approaches can be seen across a wide community of researchers. Many of these models are public, meaning that any user with minimal coding experience can play with them. Although we decided not to go into details of particular works, it is important to name a few that became the most popular:

\begin{itemize}
\tightlist
\item
  VQGAN-CLIP \citep{VQGANCLIP2022}
\item
  Midjourney \citep{Midjourney}
\item
  Latent Diffusion \citep{LatentDiffusion2021}
\item
  Stable Diffusion \citep{StableDiffusion2022}
\end{itemize}

\textbf{Potential applications}

Image generation that can be done in a controllable manner has undoubtedly huge potential for commercialization. Although the field is currently still very immature, hypotheses about which industries might be disrupted are emerging. Essentially, every branch that has to do with generating visual art, be it static images or videos, should observe the trend closely. Graphic design, movie making, stock photos - just to name a few that might be interested. Currently, experimental use cases in the area of texture synthesis, product design, or building virtual reality worlds can already be observed. AI, even if still incapable of generating the final product, can help automate a significant part of the production chain, which essentially means time and money savings. The inpainting and outpainting capabilities of recent models play a significant role in this trend. Although it is still very hard to judge which direction it takes in the future, it will definitely be a very interesting and disruptive change. Who wouldn't like to see movies being soon generated directly from a book's text, pixel value by pixel value?

\textbf{Ethics / Conclusion}

Automated image generation poses an array of serious questions of ethical character. Fortunately, many of them are already very well recognized by the community. For example, OpenAI elaborates extensively on the risks and limitations of their Dall-E 2 in this blog post by \citet{mishkin2022risks}. A few of the most important topics are presented here.

The first and very significant risk is the potential misuse of the models. Fake image generation can easily be used for harassment and disinformation. Especially combined with inpainting, which is capable of erasing or adding objects to real scenes, it poses a non-trivial challenge for researchers on how to responsibly share their work.

\begin{figure}

{\centering \includegraphics[width=1\linewidth]{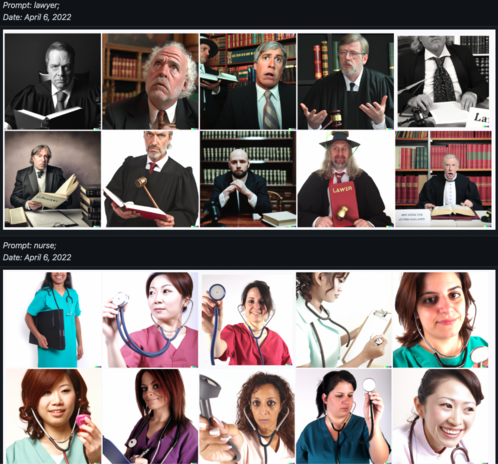}

}

\caption{Biased samples from Dall-E 2. Figure from \citet{mishkin2022risks}.}\label{fig:bias}
\end{figure}

Another important area touches on biases and stereotypes which are intrinsically built into the technology. Obviously, a model combines concepts from the data it has seen. However, if this area is to be commercialized, it needs to ensure broader diversity. An interesting example of Dall-E 2 samples can be seen in Figure \ref{fig:bias}.

In order to fully enable AI generation, the problem of copyrights needs to be solved in the first place. It is definitely not clear who is the author of generated images. Is it the person who came up with a text prompt and ran the model? Is it a model engineer? The author of the model's architecture? The owner of the data it has been trained on? Or maybe the model itself? Another question is what really is a creative contribution and eventually should result in copyright being granted. These and many others definitely require extensive debate and hopefully, legal solutions following it.

\hypertarget{c02-03-img-support-text}{%
\section{Images supporting Language Models}\label{c02-03-img-support-text}}

\emph{Author: Giacomo Loss}

\emph{Supervisor: Matthias Aßenmacher}

\hypertarget{words-in-non-symbolic-contexts}{%
\subsection{Words In (Non-Symbolic) Contexts}\label{words-in-non-symbolic-contexts}}

Imagine you were alone in a foreign country, you could not speak the language and the only resource you had were a dictionary in the foreign language. You see a word written on a sign but you cannot understand its meaning. What could you do? One idea would be do open the dictionary and look the word up. The problem is that the word is defined by using other words in the foreign language. As a second step you would thus look these new words up and continue like that in further steps to the ``infinity and beyond'' (cit. Buzz Lightyear). But even after looking every single word in the dictionary up, you would still not be able to understand the meaning of the word written on the sign. If on that sign, next to the unknown word, something else was instead depicted, for example an image of a fork and a knife, you might speculate that the word indicates something which has to do with food, like a restaurant. And this without explicitly knowing the meaning of the word. This example is inspired by the work of Stevan Harnad, which formulated at the beginning of the 90's the so called \emph{Symbol Grounding Problem} (\citet{harnad1990symbol}). It asserts that it is not possible to understand the meaning (semantics) of a word by just looking at other words because words are essentially meaningless symbols. It is possible to understand the meaning only if the word is put in a context, a perceptual space, other than that of written language: the word must be \emph{grounded} in non-symbolic representations, like images, for example. Over the past 10 years there has been a whopping development of distributional semantic models (DSMs, henceforth), especially after the Word2vec (\citet{mikolov2013efficient}) revolution. This family of models assumes that the meaning of words and sentences can be inferred by the ``distribution'' of those words and sentences within a text corpus (the \emph{Distributional Hypothesis} formulated by \citet{harris1954distributional}). But the \emph{Symbol Grounding Problem} mentioned earlier suggests that DSMs do not resemble the way words are learned by humans, which is in multimodal perceptual contexts. For these reasons, models have been developed with the goal to integrate further modalities (like visual ones) in pure language models, assuming that grounding words and sentences in other perceptual contexts should lead to a better understanding of their semantics and, as a result, to better performance in pure language tasks.

\begin{figure}

{\centering \includegraphics[width=1\linewidth]{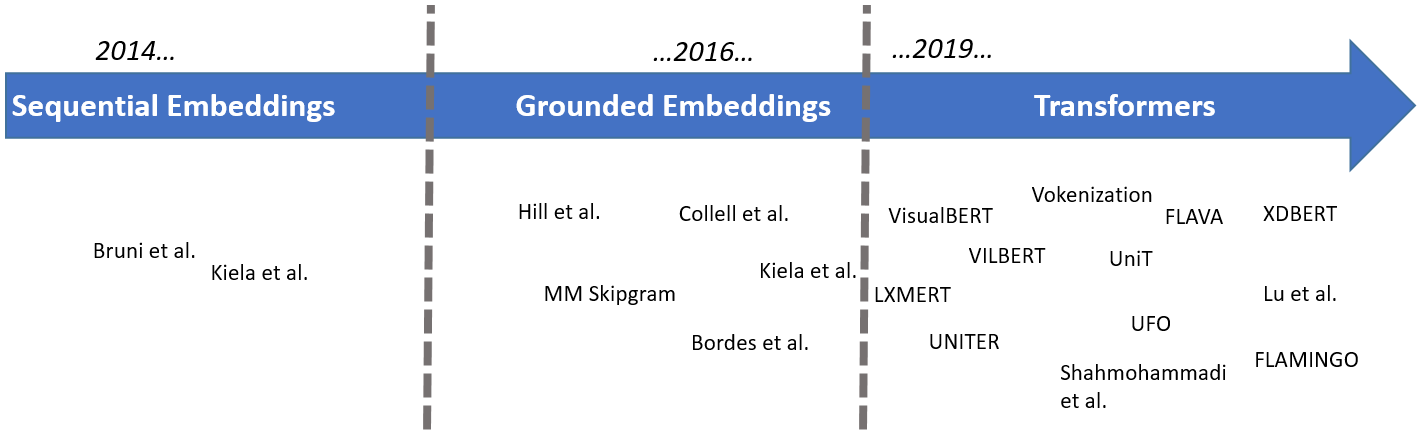}

}

\caption{Historical evolution of models which integrate visual information into pure language models.}\label{fig:img-hist}
\end{figure}

The focus of this subchapter are models which empower pure language models with visual modalities in form of images: their goal is to obtain better semantic representations (in form of embedding vectors) of words. First, a quick recap of the main pure language models will be provided. After that, the historical evolution of the integration of images as visual modalities into pure language models will be discussed: from simple concatenation of textual and visual modalities, to the projection of visual elements in a common grounded space and more recently, the use of Transformers (see figure \ref{fig:img-hist}). Eventually, a comprehensive evaluation of the different models against benchmarks will be carried out.

Again, the focus is on how to employ visual elements to obtain embeddings able to capture the semantics of words. More concrete applications, such as those in the field of machine translation are out of scope and will be only marginally addressed at the end of the subchapter.

\hypertarget{word-embeddings-survival-kit}{%
\subsection{Word-Embeddings: Survival-Kit}\label{word-embeddings-survival-kit}}

In other parts of this books, the most important NLP-models and the latest developments in the field are extensively described. In this section, some information will be provided, which might be helpful to understand some of the aspects discussed in this subchapter. As it may have been inferred in the introduction, the starting point is always a pure language model, namely a model which employs only textual inputs in order to generate word embeddings, which are representations of words in form of numerical vectors.
The most widely used pure language models in the papers presented in this subchapter are the following three:

\begin{itemize}
\tightlist
\item
  \textbf{Skipgram} (Word2vec, \citet{mikolov2013efficient}), where given a target word, the probability of the neighboring (surrounding) words in a pre-defined window has to be maximized. Trainig takes place either through a \emph{hierarchical softmax} or through \emph{negative sampling}, which involves maximizing the probability of words which are real neighbors and minimizing that of words which are not real neighbors (the ``negative samples'')
\item
  \textbf{GloVe} (\citet{pennington2014glove}), which is based on words co-occurrence across the \emph{entire} corpus, with the goal of minimizing the difference between the dot product of the embedding vectors of two words and the logarithm of the number of co-occurrences
\item
  \textbf{BERT} (\citet{devlin2018bert}): two pre-training tasks to obtain word-embeddings:

  \begin{itemize}
  \tightlist
  \item
    Masked Language Modelling (MLM): given a sentence with {[}MASK{]}ed tokens, the goal is to predict these masked tokens
  \item
    Next Sentence Prediction (NSP): given two sentences A and B, the goal is to predict if B follows from A
  \end{itemize}
\end{itemize}

Two additional remarks to conclude this section. First, Skipgram and GloVe generate embeddings which are \emph{``context-free''}: they do not take into account the context in which words occur. On the contrary, BERT is designed to represent words given the context (sentence) in which they occur: we can thus have different embeddings for the same word, depending on the context.
Second, the inputs of these models are \emph{tokens}: with the help of a \emph{tokenizer}, which can be different for different models, the text is split in ``chunks'', called \emph{tokens} (and they are not necessarily single words).

\hypertarget{the-beginning-sequential-multimodal-embeddings}{%
\subsection{The Beginning: Sequential Multimodal Embeddings}\label{the-beginning-sequential-multimodal-embeddings}}

Supposing we add linguistic and visual feature representations related to a particular word, how could we fuse them? One intuitive idea would be to \emph{concatenate} the textual and visual modalities. Let \(V_{text}\) be the textual (vectorial) representation of a word and let \(V_{img}\) be its visual (vectorial) representation, a fused representation \(F\) of a certain word \(w\) might take the following simplified form:

\[F=\gamma(V_{text})\bigoplus(1-\gamma)V_{img}\]

where \(\gamma\) is a tuning parameter which controls the relative contribution of both modalities to the final fused representation. \citet{bruni2014multimodal} propose a model where the meaning of a target word is represented in the form of a semantic vector and all vectors are collected in a \emph{text-based semantic matrix}; textual embeddings are computed based on (transformed) co-occurrence counts of words in a pre-defined window. The starting point to obtain an image-based representation of certain target word is a dataset of labeled images. For each image associated to the target word (which means that the target word is to be found in the image's caption), low-level features called ``local descriptors'' - which incorporate geometric information of specific areas of a certain picture - are extracted and then these descriptors are assigned to clusters (\emph{bags}) of ``visual words''\footnote{See for example \citet{bosch2007image} for more details on this technique, called ``bag-of-visual-words''.}. Afterwards, for each target word, visual word occurrences are summed up together to obtain the occurrence counts related to the target word. These image-based semantic vectors are then transformed and collected in an \emph{image-based semantic matrix}. The two matrices are then concatenated and projected into a common latent multimodal space with a singular value decomposition. Thanks to this process a \emph{textual \textbf{mixed} matrix} and a \emph{visual \textbf{mixed} matrix} are extracted and then combined together according to different fusion strategies to build the multimodal embeddings. In this first, relatively cumbersome (historically motivated) example, the vector representation of an image is obtained with non-trivial features engineering.

In recent years, the use of neural networks has made an ``automatic feature selection'' possible. This is what for example \citet{kiela2014learning} propose, extracting visual features from the first seven layers of a convolutional neural network (proposed by \citet{krizhevsky2012imagenet}) trained on 1.6 million images from the ImageNet database (\citet{deng2009imagenet}), which produces scores for 1,512 object categories. The linguistic part of the model relies on the Skipgram model by \citet{mikolov2013efficient} and consists of 100-dimensional vector representations. The multimodal representation is again obtained by concatenation of both modalities.

\begin{figure}

{\centering \includegraphics[width=1\linewidth]{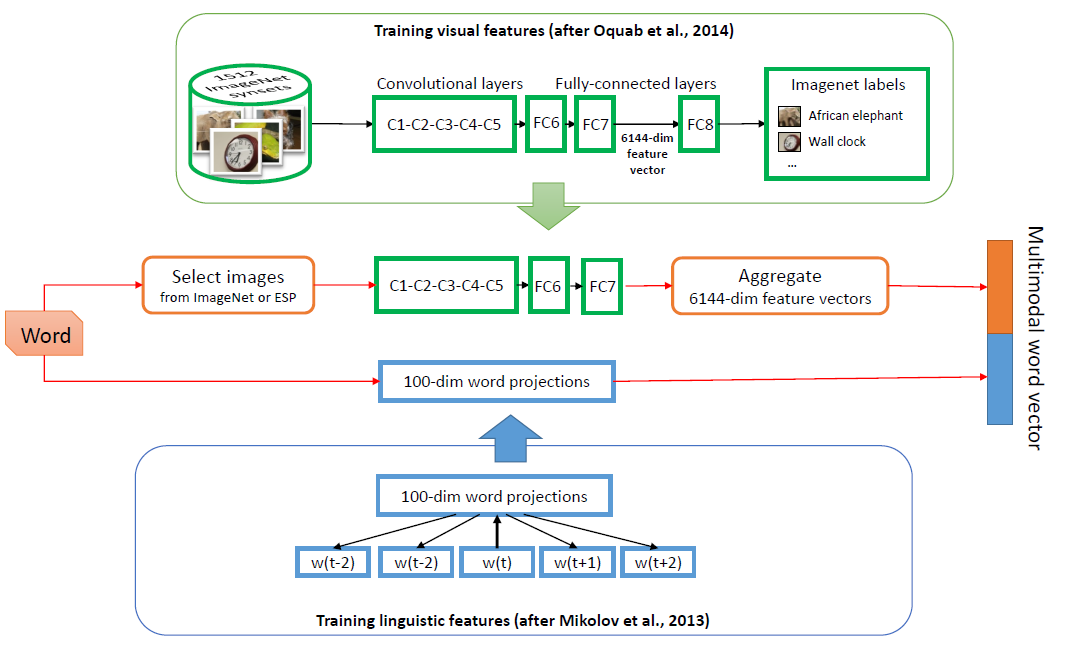}

}

\caption{From \citet{kiela2014learning}. Textual and visual features vectors are concatenated.}\label{fig:img-kiela2014-01}
\end{figure}

Another notable example of concatenation/sequential combination of textual and visual modalities is the work of \citet{silberer2014learning}: textual and visual modalities are represented by separate vectors of textual and visual attributes. During training, these textual and visual inputs vectors are separately fed to denoising (unimodal) autoencoders, the training objective of which is the reconstruction of a certain corrupted input - e.g.~through masking noise - from a latent representation. Their outputs are then jointly fed to a bimodal autoencoder to be mapped to a multimodal space, on which a softmax layer (classification layer) is added, which allows the architecture to be fine-tuned for different tasks.

\hypertarget{the-grounded-space}{%
\subsection{The Grounded Space}\label{the-grounded-space}}

The aforementioned models assume implicitly a one-to-one correspondence between text and images: a visual representation is extracted only from words which are associated to a concrete image. This is a limitation, for two partially overlapping reasons. One one hand, how can we depict words for which no image is available in our training set? Is it possible to \emph{imagine} visual representations purely from linguistic ones? On the other hand, could we hypothetically find a visual representation for each word? This might be true for concrete words but when it comes to abstract ones, it is not always possible to find suitable visual representations or, said in other terms, many words are not visually grounded. For this reasons, researches have addressed the question: could we map textual and visual elements to a grounded space and design models able to generalize images and words beyond those in the training set? Well, the answer is yes!

\citet{lazaridou2015combining} propose a multimodal Skip-gram architecture where the objective function of a Skip-gram is ``augmented'' with an additional visual objective: \[\frac{1}{T}\sum_{t=1}^{T}\left(\mathcal{L}_{ling}(w_{t})+\mathcal{L}_{vision}(w_{t})\right)\]

where \(\mathcal{L}_{ling}\) is the Skip-gram loss function and \(\mathcal{L}_{vision}\) is the additional visual loss for the target word \(w_{t}\). In particular, \(\mathcal{L}_{vision}\) has the form of a hinge loss, the goal of which is to make the (vectorial) linguistic representation of a certain word more similar to its visual representation:

\[\mathcal{L}_{vision}(w_{t})=-\sum_{w^{'}\sim P_{n}(w)}\left(max(0,\gamma-cos(z_{w_{t}},v_{w_{t}})+cos(z_{w_{t}},v_{w^{'}})\right)\]

where \(v_{w^{'}}\) is a visual representation of a randomly chosen word \(w^{'}\) (drawn from a probability distribution \(P_{n}(w)\)) used as negative sample, \(v_{w_{t}}\) is the corresponding visual vector and \(z_{w_{t}}\) is the target multimodal word representation which has to be learned by the model. It is nothing more than a linear transformation of a word representation \(u_{w_{t}}\): \(z_{w_{t}}=M^{u\rightarrow v}u_{w_{t}}\) and \(M^{u\rightarrow v}\) is a cross-modal mapping matrix from linguistic inputs to a visual representation. It is important to remark that during training, for words which do not have associated images, \(\mathcal{L}_{vision}\) gets set to zero. When this cross-modal mapping matrix is estimated, it is then possible to find a visual representation for new words, which do not have a related image in the training set: the model allows to \emph{imagine} new words. This is what is meant with grounded space: a perceptual (visual, in this case) space where a word is \emph{grounded}, put in context.

\begin{figure}

{\centering \includegraphics[width=0.8\linewidth]{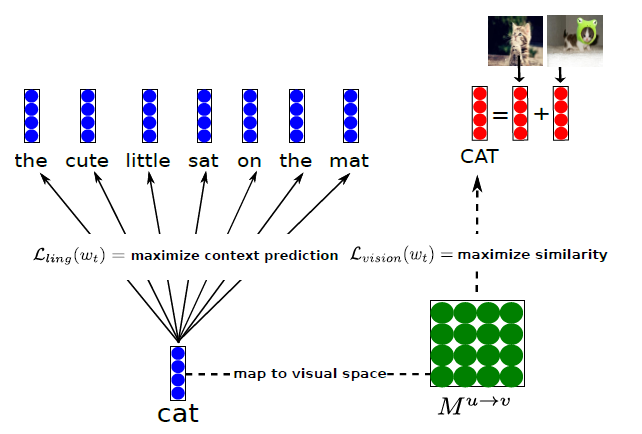}

}

\caption{From \citet{lazaridou2015combining}. The linguistic embedding of the word `cat' is mapped to a visual space, such that the similarity of vector representations of words and associated images is maximized.}\label{fig:img-lazaridou2015-01}
\end{figure}

Similar instances of a cross-modal mapping can be found for example in \citet{kottur2016visual} (a multimodal extension of the CBOW model specification of word2vec) and in \citet{collell2017imagined}, where visual features are obtained from the forward pass of a CNN, pre-trained on ImageNet (\citet{deng2009imagenet}) and a mapping function from the textual space to the visual space is obtained as a result of the training process. Also in this case it is possible to generate a visual representation from the embedding of a certain word, not necessarily present in the training set. In particular, they propose two specifications of the mapping function: a simple linear mapping and neural network with a single hidden layer. Last but not least, \citet{hill2014learning} recognize that concrete nouns are more likely to have a visual representation. For this reason, they map a set of concrete words (CSLB, \citet{devereux2014centre}) to ``bags of perceptual/visual features'' and every time one of these words is encountered during training, the Skip-gram model they are using stops training on that sentence and instead continues the training on a newly created ``pseudo-sentence'', which takes into consideration the aforementioned bag of perceptual features. This list is unfortunately not exhaustive and there are other models with similar ideas, for example \citet{ailem2018probabilistic} or \citet{kiros2018illustrative}.

The aforementioned papers and related models focus on the modeling of semantics of words. Nonetheless, there are models designed to address tasks at sentence-level, such as sentiment analysis or sentence entailment. \citet{kiela2017learning} employ a bidirectional Long Short-Term Memory (LSTM, \citet{hochreiter1997long}) architecture to model sentence representations, in order to gain information from the text in both directions. The goal is again to encode a sentence and ground it in an image. Textual embeddings are obtained with GloVe (\citet{pennington2014glove}) and they are then projected on a grounded space with a linear mapping. This grounded word vector serves as input for the bidirectional LSTM, which is trained together with the linear mapping. Their model is versatile and depending on the loss function specification, it can not only propose alternative captions to an image (which is a way to frame sentence equivalence tasks) but also predict captions from images or perform both tasks at the same time. This last point highlights an important characteristic of many of the models discussed in this subchapter: even though the focus is on the empowerment of pure language models with the addition of visual elements, some of the models discussed here can be used for purposes other than pure language tasks. The control over which task is performed is usually exercised by either specifying different loss functions (as in the last model described) or setting properly certain hyperparameters (such as in the previously described model by \citet{silberer2014learning}).

\hypertarget{the-transformers-era}{%
\subsection{The Transformers Era}\label{the-transformers-era}}

A turning point for the field of NLP was \citet{vaswani2017attention}'s paper ``Attention is all you need'', where the authors proposed for two machine translation tasks a novel architecture, the Transformer (not to be confused with the giant robots from the Michael Bay's blockbuster movies!), which leverages only the attention mechanism. Even though an exhaustive description of the Transformer architecture is beyond the scope of this subchapter, it is worth mentioning why they became so popular over the past four years in the field of NLP (among others), in comparison to Recurrent Neural Networks (RNNs) and Long Short-Term Memory networks (LSTMs).

Well, the three main properties of Transformers are the following:

\begin{itemize}
\tightlist
\item
  Self-Attention
\item
  Parallel input processing
\item
  Positional embeddings\footnote{It may be argued that this point is a necessity to be able to work on sequences rather than a strength.}
\end{itemize}

When feeding for example a textual sentence to a RNN, the network deals with one word after the other in a sequential fashion and one of the known issues is the fact that information contained in earlier parts of the sequence tend to ``fade away'' as the sentence is analyzed further: newer inputs carry a larger influence on the outputs at a given step. LSTMs try to mitigate this problem by introducing a component called ``gate'', which regulates the information flow, namely which information from the past inputs need to be ``remembered'' by the model. The goal is to capture long-term dependencies among different parts of the sentence fed into the model.\\
On the contrary, thanks to the Self-Attention mechanism, at each step Transformers can access previous steps, thus limiting to a minimum the loss of information. Moreover, inputs are processed not sequentially but all at the same time, thus allowing to capture dependencies by looking at the sentence \emph{as a whole} and this could make a fundamental difference in many downstream applications: for example in the German language, in dependent clauses (``Nebensaetze''), the verb comes at the end of the phrase but it determines the verbal case of the nouns that come \emph{before} the verb. Thus Transformer could potentially capture the dependencies between the verb coming at the end of the sentence and the words at the beginning. Lastly, Transformers encode for every input information on its position within a sentence, since it is often the case, that the importance and meaning of a certain word varies depending on its position within a sentence. These were the Transformers, in a nutshell.

But Transformers did not only bring a change of paradigm in terms of architectures. First, while for models in the pre-Transformers era described before, the focus was on the ability of word embeddings to capture similarity among words, now the focus has shifted more on downstream tasks (more on this later in the evaluation section), encompassing not only pure linguistic ones but also tasks with visual components, such as for example, visual question answering. It is now more difficult (but not impossible) to draw a line between models where ``images support pure language models'' (the object of this subchapter) and models which could be actually categorized as ``vision and language'' models but they can be employed also to solve pure linguistic tasks. This issue brings another peculiarity of many Transformers-base models, namely their ``universal vocation'': without loss of generality we could say that the idea is now to design powerful (multimodal) pre-training (mostly \emph{self-supervised}) tasks capable of generating task-agnostic representations, whose encoded knowledge can be efficaciously transferred to diverse downstream tasks, limiting the amount of labeled data necessary to fine-tune the models (this is the so-called \emph{few-shot learning}).

Let's briefly discuss two examples, Flava (\citet{singh2022flava}) and UniT (\citet{hu2021unit}). Flava has two separate encoders for images and text and a multimodal encoder, all based on the Vision Transformer (\citet{dosovitskiy2020image}). Unimodal pre-training consists of masked image modeling (where a set of image patches are to be reconstructed from other unmasked image patches) and masked language modeling. Multimodal pre-training tasks consist instead of a global contrastive loss (maximization of cosine similarities between paired images and text), a masked multimodal modeling (where image patches and text tokens are masked) and an image-text matching task. The model is pre-trained jointly on unimodal and multimodal datasets and then evaluated (fine-tuned) on 22 vision tasks, 8 pure linguistic tasks and 5 vision and language tasks.\\
UniT has an image encoder and a text encoder, a multimodal domain-agnostic decoder and task-specific heads. There is no pre-training on multimodal data and the model is trained end-to-end on 7 tasks (vision, language and vision an language) and 8 datasets, with the idea that solving different tasks across domains in a jointly fashion should prevent general knowledge from being lost due to fine-tuning over particular downstream tasks.

These two examples clearly show what it is meant by ``universal vocation'' of many modern Transformer-based models. But there are still models specifically designed to solve pure language tasks and in the following pages, two of them will be described.

\hypertarget{vokenization}{%
\subsubsection{Vokenization}\label{vokenization}}

It is often difficult for a child to describe the meaning of a certain word. A child might not be able to describe what a lion is but if he is given pictures of different animals he might be very well able to point at the picture of a lion. \emph{Visual pointing} could thus act as a form of supervision to natural language. Is it possible to build within a pure language model a form of visual supervision, which mimics the visual pointing often adopted by children? This is exactly the problem that \citet{tan2020vokenization} try to address: how to associate to each textual representation (token) a visual representation (Voken).

Let's suppose we had a dataset of word(token)-image pairs. We could integrate in the pre-training framework of pure language models the following \emph{Voken-Classification} task:

\[\mathcal{L}_{VOKEN-CLS}(s)=-\sum_{i=1}^{l}log\ p_{i}(v(w_{i};s)|s) \]
\[\textbf{h}_{1}, \textbf{h}_{2},...,\textbf{h}_{l}=languagemodel(w_{1},w_{2},...,w_{l}) \]
\[p_{i}(v|s)=softmax_{v}\{W\textbf{h}_{i}+b\}\]
where \(\{h_i\}\) is the feature representation of each token in a sentence \(s=\{w_i\}\) extracted from a language model (such as BERT) and the vokens originate from a \textbf{finite} set of images \(X\). Each \(h_i\) is then transformed into a probability distribution through a softmax layer, with the voken-classification loss defined as the negative log-likelihood of all related vokens.\\
The model architecture would then be:

\begin{figure}

{\centering \includegraphics[width=0.7\linewidth]{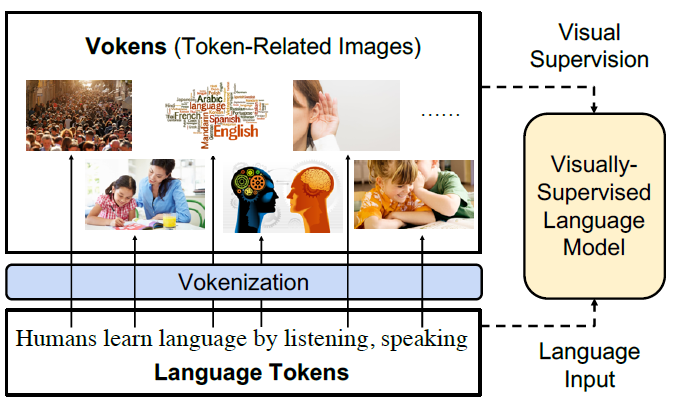}

}

\caption{From \citet{tan2020vokenization}. Visually supervised the language model with token-related images, called Vokens.}\label{fig:img-tan2020-04}
\end{figure}

Everything sounds fantastic! There is only one small pitfall: a set of \(X\) of images for all tokens does not exist! Could we find a proxy for such a set? One might consider image-captioning datasets such as MS COCO (\citet{lin2014microsoft}). But also this suboptimal solution is problematic.

\begin{figure}

{\centering \includegraphics[width=1\linewidth]{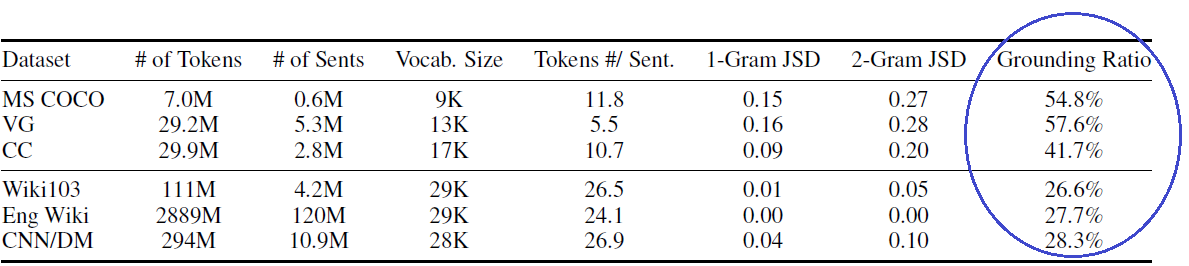}

}

\caption{From \citet{tan2020vokenization}. Statistics of image-captioning dataset and other natural language corpora. VG, CC, Eng Wiki, and CNN/DM denote Visual Genome, Conceptual Captions, English Wikipedia, and CNN/Daily Mail, respectively. JSD represents Jensen--Shannon divergence to the English Wikipedia corpus.}\label{fig:img-tan2020-01}
\end{figure}

The \emph{Grounding Ratio} is defined as the proportion of tokens in a dataset which are related to a specific visual representation (i.e.~the tokens are \emph{visually grounded}), such as ``dog'', ``table'' and the like. In figure \ref{fig:img-tan2020-01} it is striking that only around one third of tokens contained in pure language corpora such Wiki103, English Wikipedia and CNN/DM are visually grounded in image captioning datasets\footnote{From an operative point of view, the authors consider a token type ``visually grounded'' if it has more than 100 occurrences in MS COCO}. It is not possible to rely (only) on image captioning datasets to build the Voken-Classification task. But the fact that a word/token does not have a visual representation in one of these datasets, it does not mean that it is not possible to visually represent the word/token. Would it be possible to associate images to words/tokens not directly visually grounded? Well, the answer is yes!

\begin{figure}

{\centering \includegraphics[width=0.8\linewidth]{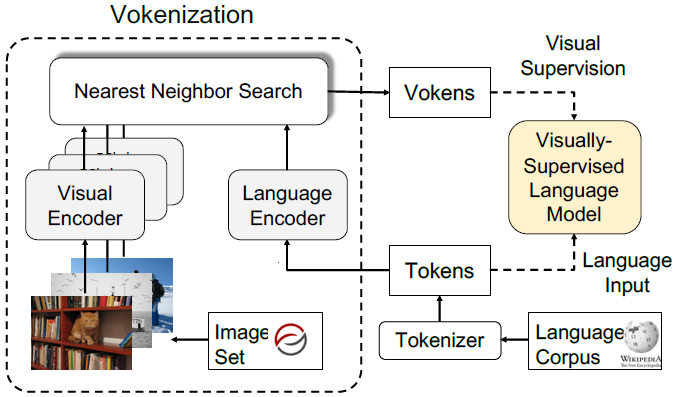}

}

\caption{From \citet{tan2020vokenization}. The Vokenization process. A contextualized image (visual token, Voken) is retrieved for every token in a sentence and with this visual token, visual supervision is performed.}\label{fig:img-tan2020-05}
\end{figure}

The \textbf{Vokenization} is a process to \emph{assign} every token \(w_i\) contained in a sentence \(s\) to a visual representation (called \emph{voken}) originating not from a generative model but rather from a finite set of images \(X=\{x_1,...,x_n\}\). The voken \(v(w_i;s)\) is the image from \(X\) which maximizes the following \emph{Relevance Score Function}:
\[v(w_i;s)=arg\ max_{x\in X}r_{\theta^{*}}(w_i,x,s)\]
This function takes into account not only the token \(w_i\) itself, but also the context (the sentence) and it is parametrized by \(\theta\) with \(\theta^{*}\) being the optimal value (which has to be estimated).

\hypertarget{the-relevance-score-function-model-training-inference}{%
\paragraph{The Relevance Score Function: Model, Training, Inference}\label{the-relevance-score-function-model-training-inference}}

The Relevance Score Function is defined as the inner product of the language feature representation \(f_{\theta}(w_i,s)\) and the visual feature representation \(g_{\theta}(x)\):

\[f_{\theta}(w_i,s)^Tg_{\theta}(x)\]

Supposing \(h_1,...,h_l\) and \(e\) are the embeddings originating from pre-trained language and visual encoders respectively (in the paper the authors use BERT and ResNeXt), the language and visual representations are obtained first by applying multi-layer perceptrons \(w\_mlp_{\theta}\) and \(x\_mlp_{\theta}\) to downproject the embeddings from the pre-trained models to a common vector space and secondly they are normalized (with L2-Norm):

\[ \textbf{f}_{\theta}(w_{i};s)= \frac{w{\_}mlp_{\theta}(\textbf{h}_{i})}{||w{\_}mlp_{\theta}(\textbf{h}_{i})||} \]
\[ \textbf{g}_{\theta}(x)= \frac{x{\_}mlp_{\theta}(\textbf{e})}{||x{\_}mlp_{\theta}(\textbf{e})||} \]

With respect to the training of the model, to estimate the optimal value for the parameter \(\theta\), image-captioning datasets, which are collections of sentence-image pairs, are employed. Operationally, for every sentence \(s_k\) associated to image \(x_k\) in the image-captioning dataset, each token \(w_i\) in \(s\) is associated to \(x_k\) and the \emph{hinge loss} is used to estimate the optimal value of \(\theta^*\):

\[ \mathcal{L}_{\theta}(s,x,x')=\sum_{i=1}^{l}max(0,M-r_{\theta}(w_{i},x,s)+r_{\theta}(w_{i},x',s))\]

The goal is to maximize the Relevance Score Function between aligned token-image pairs \((w_i,x;s)\) and to minimize the score for unaligned pairs \((w_i,x^{'};s)\) by at least a margin \(M\), with \(x^{'}\) being a randomly sampled image from the image captioning dataset \textbf{not} associated to sentence \(s\).

Once we have the language feature representation \(f_{\theta}(w_i,s)\) for each token in our language corpus and the optimal estimate of \(\theta\), how is it possible to find the image \(x\) encoded with the visual feature representation \(g_{\theta}(x)\), which maximizes the Relevance Score Function? As said earlier, the function is expressed as the inner product of the textual and visual representations and since the feature vectors have euclidean norm equal to 1, the inner product maximization problem is equivalent to a nearest neighbor search problem. It is just sufficient to find the vector \(g_{\theta}(x)\) which is the nearest neighbor of \(f_{\theta}(w_i,s)\)\footnote{The proof is straightforward. Let \(X\in \mathbb{R}^l\) and have euclidean norm equal to 1, which means \(||X||_{2}=1\). In the nearest neighbor search we need to find the vector \(Y\in \mathbb{R}^l\), also with norm equal to 1, which has minimal euclidean distance with \(X\). This is the quantity to be minimized:
  \begin{align*}
  d(X,Y) &=\sqrt{\sum_{i=1}^{l}{(x_i-y_i)^2}}
    \\&\stackrel{squared}{=} \sum_{i=1}^{l}{x_i^2}+\sum_{i=1}^{l}{y_i^2}-2\sum_{i=1}^{l}{x_iy_i}
    \\&\stackrel{}{=}||X||_{2}^2+||Y||_2^2-2X^TY
    \\&\stackrel{Norm-1}{=}1+1-2X^TY
    \\&\stackrel{}{=}2(1-X^TY)
  \end{align*}
  And through these simple algebraic manipulations, it is possible to see that minimizing the euclidean distance between \(X\) and \(Y\) is equivalent to maximize \(X^TY\), which is the inner product. This proves the equivalence between inner product maximization and nearest neighbor search.}.

With this process, it is thus possible to assign a visual representation, a voken, to any word/token in a language corpus, pooling from a finite set of images. The problem of the low Grounding Ratio outlined above is solved and the Voken-Classification task could be integrated in the pre-training framework of any pure language model. Moreover, the authors propose a method called \emph{Revokenization}, which allows to transfer vokens generated using a particular tokenizer to frameworks which employ other tokenizers.

\hypertarget{one-step-further-the-power-of-imagination}{%
\subsubsection{One Step Further: The Power Of Imagination}\label{one-step-further-the-power-of-imagination}}

Wikipedia defines \emph{imagination} as ``the production or simulation of novel objects, sensations, and ideas in the mind without any immediate input of the senses''. Indeed, humans do not only associate words with real images, but also leverage the ability to \emph{imagine} words/concepts: imagination can help the human brain solve problems with limited supervision or sample points by empowering its generalization capabilities. Until now we discussed language models supported by visual information in form of \emph{real} images (e.g.~those retrieved from image-captioning datasets). But with the recent advancements in the field of generative models for images, it is for sure worth investigating if these generative models can help pure language models to produce better representations of words. In particular, the framework proposed by \citet{lu2022imagination}, \textbf{iACE (Imagination-Augmented Cross-Modal Encoder)} will now be discussed: the idea is simply to use a generative model to obtain a visual representation of a textual input and then use these imagined representations as ``imagination supervision'' to pure language models.

\begin{figure}

{\centering \includegraphics[width=1\linewidth]{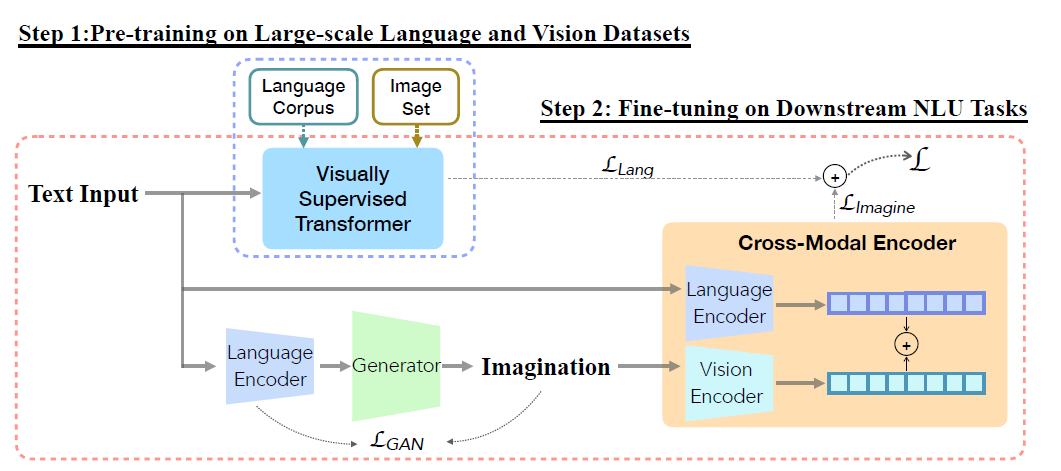}

}

\caption{From \citet{lu2022imagination}. The generator \(G\) visualize imaginations close to the encoded texts by minimizing \(\mathcal{L}_{GAN}\). The cross-modal encoder \(E_c\) learns imagination-augmented language representation. Two-step learning procedure consists of: 1) pre-train a Transformer with visual supervision from large-scale language corpus and image set, 2) fine-tune the visually supervised pre-trained Transformer and the imagination-augmented cross-modal encoder on downstream tasks.}\label{fig:img-lu2022-01}
\end{figure}

This framework has two main components:

\begin{itemize}
\tightlist
\item
  the \textbf{imagination generator \(G\)}: given an input text \(x\), VQGAN (\citet{esser2021taming}) is used to render an ``imagination'' \(i\) of \(x\) and CLIP (\citet{radford2021learning}) is used to see how well the generated image \(i\) is aligned to the input text \(x\). This generative framework is known as VQGAN+CLIP
\item
  \textbf{Cross-modal Encoder \(E_c\)}: the input text and the rendered imagination are firstly encoded with a language and a visual encoder respectively and then CLIP is employed as cross-modal encoder with inputs being text-imagination pairs
\end{itemize}

The learning procedure is composed of two main steps (depicted in figure \ref{fig:img-lu2022-01}): the first step consists in the pre-training of a visually supervised Transformer. In particular, the Voken-Classification task described before is employed, alongside a masked language modeling task. This is the baseline model, where no information from the ``imagination'' procedure comes yet into play. The second step is the \emph{imagination-augmented fine-tuning} with two downstream datasets \(D\) (GLUE, \citet{wang2018glue} and SWAG, \citet{zellers2018swag}).\\
On one side, the visually-supervised Transformer (the baseline) relies only on the textual input during the fine-tuning phase and the following loss function is employed:

\[ \mathcal{L}_{Lang}=-\sum_{j=1}^{|D|}\sum_{k=1}^{K}y_{k}\ log\ p_{k}(d_{j}(t)|D) \]

On the other hand, the \emph{iACE} is trained to minimize the following cross-entropy loss:

\[ \mathcal{L}_{Imagine}=-\sum_{j=1}^{|D|}\sum_{k=1}^{K}y_{k}\ log\ p_{k}(d_{j}(t,v)|D) \]

with \(t\) and \(v\) being the textual and imagined features representations respectively, \(j\) indicates the \(j\)-th data sample in dataset belonging to dataset \(D\), \(K\) is the number of classes and \(p_k\) is the conditional distribution of \(d_j\).
Training takes place in a jointly fashion and both losses, the imagination-augmented one \(\mathcal{L}_{Imagine}\) and the pure language loss \(\mathcal{L}_{Lang}\) are linearly combined, with \(\lambda\) being a balance factor:

\[\mathcal{L}=\lambda\mathcal{L}_{Imagine}+(1-\lambda)\mathcal{L}_{Lang} \]

To sum up, this model-agnostic framework uses \emph{generated images} for visual supervision and could be integrated on top of pure language models (such as BERT) or visually supervised models (such as the Voken model, which uses Vokens, real images for visual supervision).

\hypertarget{was-it-worth}{%
\subsection{Was It Worth?}\label{was-it-worth}}

In this subchapter we investigated how visual inputs can support pure language models in capturing the semantics of words. We started with simple concatenation of linguistic and visual features and ended up with Transformer-based models, which are able to shape different word embeddings for the same word by taking into account also the context (the sentence). But now the question arises: with the addition of visual information, do we obtain word embeddings that are better than those from pure language models? In other words, is what we all have so far discussed worth? Well, as it is often the case in scientific research, the answer is: ``it depends!''

Individual evaluation of each single model might not be ideal because each model has its peculiarities and it is impractical to make a direct comparison among them. It is more useful to capture and discuss the themes which are common to many models, in order to understand their strengths and weaknesses. This is how we will proceed and we will also differentiate between evaluation before Transformers and evaluation after Transformers.

\hypertarget{evaluation-in-the-pre-transformers-era}{%
\subsubsection{Evaluation In The Pre-Transformers Era}\label{evaluation-in-the-pre-transformers-era}}

Before the advent of Transformers, the evaluation focus was on the degree of alignment between learned semantic representations (word embeddings) and representations by human speakers, in form of correlation between model-based and human-based word-similarity judgments. Three main types of similarity are usually considered:

\begin{itemize}
\item
  Semantic similarity, e.g.~``pasta is similar to rice''
\item
  Semantic relatedness, e.g.~``Bear is related to mountain''
\item
  Visual similarity, e.g.~``cucumbers look like zucchinis''
\end{itemize}

The evaluation pipeline could be summarized as follows:

\begin{figure}

{\centering \includegraphics[width=1\linewidth]{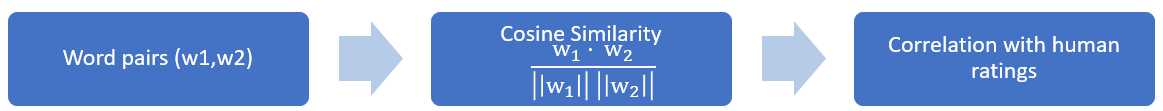}

}

\caption{Pipeline for intrisinsic evaluation of semantic representations. In the first step, the cosine similarity between two word embeddings w1 and w2 is used as similariry measure and in a second step, the correlation with human speakers'assessment is computed to gauge the quality of the embeddings. The higher the correlation, the better the embeddings.}\label{fig:img-eval01}
\end{figure}

Word embeddings are vectors and to measure the degree of similarity between two vectors, the \emph{Cosine Similarity} is often used in the literature. In an ideal setting, we would have word embeddings with the following characteristics: if two words are semantically similar, the two embedding vectors should be similar and their cosine similarity should go towards 1. If the two words are unrelated, the embedding vectors should be orthogonal to each other and as a consequence, the cosine similarity should go towards zero. Lastly, if two words are negatively related, the two embedding vectors should point at opposite directions and the cosine similarity should go towards -1.
Once these similarity measures between word pairs are computed, in order to measure the quality of the embeddings several benchmarks can be employed, such as MEN (\citet{bruni2014multimodal}), WordSim353 (\citet{agirre2009study}) and SimLex999 (\citet{hill2015simlex}). These datasets could be described as collections of word pairs and associated similarity ratings by human speakers. Operationally, this means that real people were asked if a pair of words was related or not and to which degree, on a scale between -1 (negatively related) to +1 (semantically equivalent). The higher the correlation between the cosine similarity and the similarity judgments by humans, the higher the quality of the word embeddings. Having done this methodological premise, let's discuss the performance of these pre-Transformer models!

Since the goal of these models is to enhance pure language models with the addition of visual inputs, the baseline in the evaluation is always one (or more) pure language model(s). Well, do visually grounded embeddings outperform non-grounded ones? What emerges from virtually all papers is that visual grounding can actually help get a better semantic representation of \emph{concrete} concepts, such as ``cat'', ``table'', ``bicycle'', whereas they do not help much with the representation of abstract concepts such as ``love'' and ``peace''.

\begin{figure}

{\centering \includegraphics[width=1\linewidth]{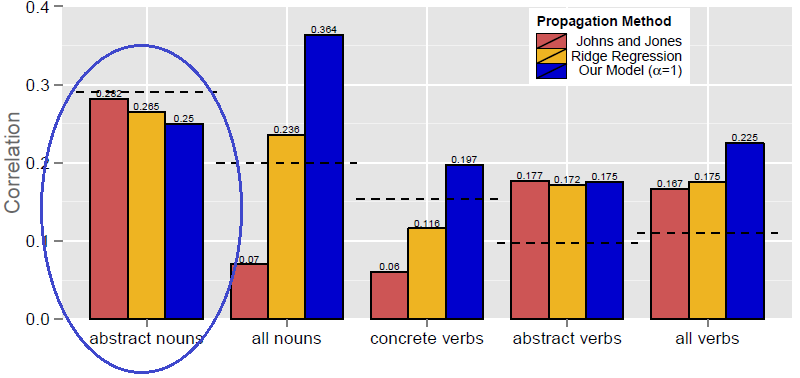}

}

\caption{From \citet{hill2014learning}: Each bar represents a different model settings and the dashed line indicates the pure linguistic benchmark model.
In figure \ref{fig:img-2014hill-01} we can see that pure language models still perform better than models with visual inputs when it comes to the representation of abstract \emph{nouns}. Another example is \citet{kiela2017learning}: they found that their models perform better when tested on datasets with a higher degree of concreteness and the same conclusion is reached by \citet{collell2017imagined}, which state that visual information can empower the representations of concepts that are to a certain extent visual. To sum up, effective semantic representation of abstract concepts constitute the main limitation common to many of the models discussed in this section.}\label{fig:img-2014hill-01}
\end{figure}

\hypertarget{evaluation-in-the-post-transformers-era}{%
\subsubsection{Evaluation In The Post-Transformers Era}\label{evaluation-in-the-post-transformers-era}}

A limitation of the \emph{intrinsic} evaluation metrics is the high degree of subjectivity: the \emph{similarity} between two concepts depends in many instances on the experience, cultural background and preferences of the human observers. This is why the evaluation focus has now shifted to a more \emph{extrinsic} dimension: how well do the models perform in downstream tasks? The problem of the ``lack of objectivity'' is thus solved because on downstream tasks there is no room for opinions. The datasets used to train the models are also different and the most widely used are:

\begin{itemize}
\tightlist
\item
  GLUE (\citet{wang2018glue}): 9 tasks, including single-sentence tasks (e.g.~sentiment analysis), similarity tasks (e.g.~paraphrasing), inference tasks (e.g.~textual entailment)
\item
  SQuAD (\citet{rajpurkar2016squad}): question/answer pairs
\item
  SWAG (\citet{zellers2018swag}): multiple choice questions about grounded situations
\end{itemize}

As previously discussed, many Transformer-based models have universal vocation: they are built to solve a heterogeneous range of tasks from the language and vision domain. If we thus consider only performance on pure language tasks, the following two tables from \citet{tan2020vokenization} are insightful:

\begin{figure}

{\centering \includegraphics[width=1\linewidth]{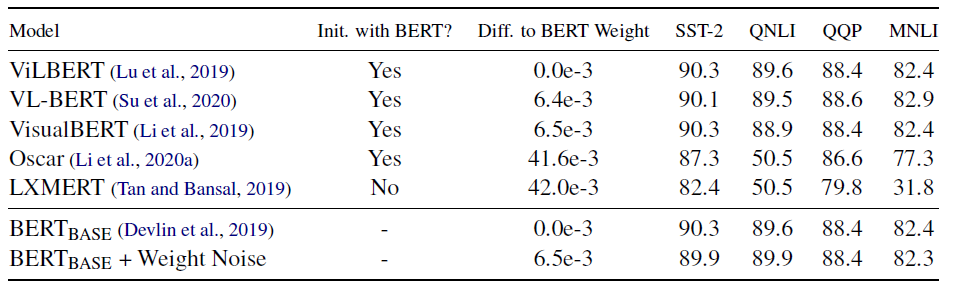}

}

\caption{From \citet{tan2020vokenization}. Statistics of image-captioning dataset and other natural language corpora. VG, CC, Eng Wiki, and CNN/DM denote Visual Genome, Conceptual Captions, English Wikipedia, and CNN/Daily Mail, respectively. JSD represents Jensen--Shannon divergence to the English Wikipedia corpus.}\label{fig:img-tan2020-02}
\end{figure}

\begin{figure}

{\centering \includegraphics[width=1\linewidth]{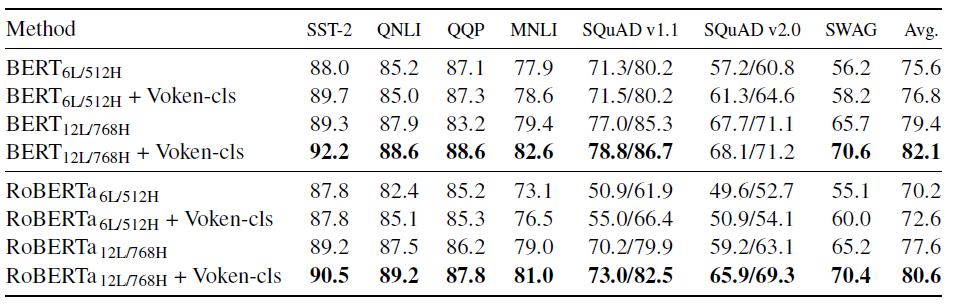}

}

\caption{From \citet{tan2020vokenization}. Fine-tuning results of different pre-trained models w/ or w/o the voken classification task (denoted as``Voken-cls'').}\label{fig:img-tan2020-03}
\end{figure}

It is straightforward: unlike in the pre-Transformers Era, where grounded word embeddings could improve performance over baselines, Transformer-based universal models \textbf{do not} outperform pure language models such as BERT or RoBERTa. Nonetheless, the addition of visual supervision (the Voken-Classification task) in the pre-training framework can boost performance above the level of pure language models.

\citet{pezzelle2021word} analyzed the \emph{intrinsic} quality of embeddings of some vision and language (``universal'') models:

\begin{figure}

{\centering \includegraphics[width=1\linewidth]{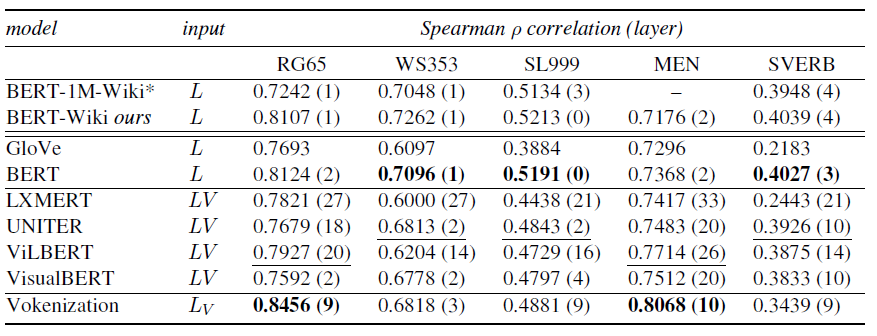}

}

\caption{From \citet{pezzelle2021word}. Spearman's rank correlation between similarities computed with representations by all tested models and human similarity judgments in the five evaluation benchmarks.}\label{fig:img-pezzele2021-01}
\end{figure}

\begin{figure}

{\centering \includegraphics[width=1\linewidth]{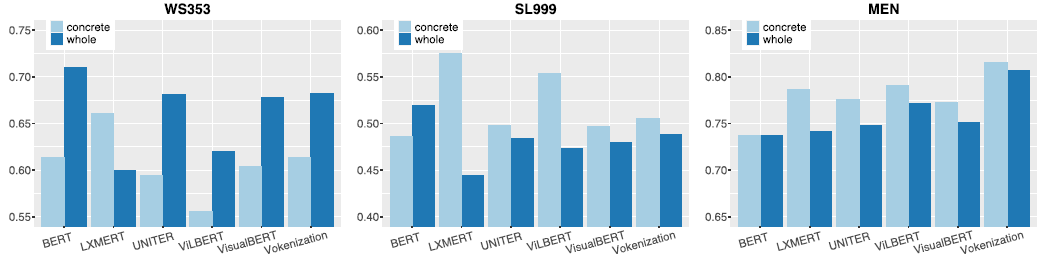}

}

\caption{From \citet{pezzelle2021word}. Correlation between model and human similarity ratings on WordSim353, SimLex999 and MEN. Each barplot reports results on both the whole benchmark and the most concrete subset of it.}\label{fig:img-pezzele2021-02}
\end{figure}

From this \emph{intrinsic} evaluation perspective (which was popular in the pre-Transformers Era), vision and language models do not generally outperform domain-specific models such as BERT and also in this case the only real competitor of pure language models is a model with visual supervision (again, Vokenization).

The bar plots depict correlation between human- and model-based similarity ratings, differentiating between the most \emph{concrete} concepts contained in a certain dataset\footnote{See \citet{brysbaert2014concreteness} for information on how \emph{concreteness} of a word can be estimated.} and the whole dataset (thus including more abstract concepts). The results confirm the trend: multimodal models are more effective than pure language models at representing concrete words but in many instances they still lag behind when it comes to more abstract concepts.

Last but not least, few words need to be spent on a topic which has been steadily gaining relevance: \textbf{Few-Shot Learning}. To train and test models, a large pool of paired images and texts is often needed and the creation of many of the datasets used in fine-tuning required a huge data collection effort, which had to be performed by human agents. This implies that the creation of such data pools can be very costly. For this reason, there is a growing interest in creating models able to cope with low-resource settings. This boils down to the question: can a model perform well on downstream tasks even with just a \emph{limited number} of training examples? The goal is actually once again, to mimic how humans learn: a person does not need to see one thousand pictures of a table, to be able to recognize a table\ldots{}

\begin{figure}

{\centering \includegraphics[width=1\linewidth]{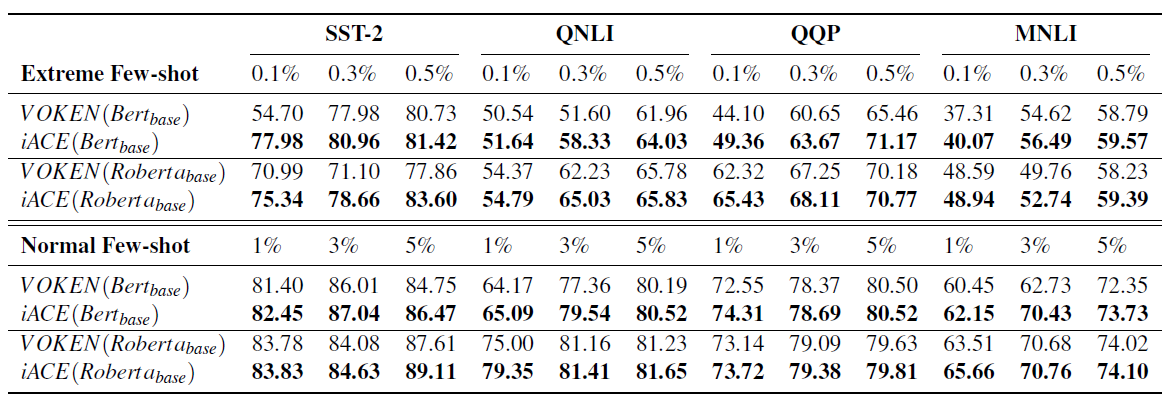}

}

\caption{From \citet{lu2022imagination}. Model-agnostic improvement in Few-shot Setting with GLUE benchmark.}\label{fig:img-lu2022-02}
\end{figure}

This table from \citet{lu2022imagination}, where models are trained using only up to 5\% of the training set, shows for example the ability for a model supervised with ``imagination'' (which was a generated visual representation of a certain textual input) to outperform models with only simple visual supervision (the Voken-model). This is just an example, but the ability to perform well in \emph{few-shot} settings has become the touchstone of the evaluation modern multimodal models.

\hypertarget{the-end-of-this-story}{%
\subsection{The End Of This Story}\label{the-end-of-this-story}}

We started this story with the \emph{Symbol Grounding Problem}, which affirms that to grasp the meaning of a word, the word has to be put in a context other than the pure linguistic one. We thus investigated some of the architectures proposed to ground words in a visual space in form of static images. The goal (hope) is to better capture the semantics of words, in form of better word embeddings, to be employed in heterogeneous tasks, from \emph{semantic-similarity} to downstream tasks, such as \emph{sentiment analysis}.\\
From this brief analysis it emerges that grounding words in images can actually improve the representation of \emph{concrete} concepts, whereas visual grounding does not seem to add value to pure language models when it comes to \emph{abstract} concepts. Nonetheless, forms of visual supervision like the \emph{Voken-Classification} task or the employment of generative models which allow to \emph{imagine} words, such as in the \emph{iACE-Framework}, might be the right way to bridge this gap.\\
The Transformers have been a revolution in the field of NLP and with their advent, the trend has now become to build models with pre-training tasks capable of generating powerful task-agnostic word representations. The knowledge gained with these tasks can be then transferred to downstream tasks with the goal to limit the amount of labeled data necessary to fine-tune models. Labeling data is indeed costly: this is why the ability of a model to generalize well when exposed to just few training examples has been steadily gaining importance as evaluation metric. This was the so called \emph{few-shot learning}. Moreover, Transformer-based models have ``universal vocation'': they tend to be multimodal and multi-task, encompassing vision, language and vision and language tasks. This idea might be appealing because humans learn by being exposed to a multitude of different inputs and tasks. But as we have seen, pure language models such as BERT tend to still outperform multimodal multi-task models. There is definitely room for improvement.\\
One might wonder whether the grounding of words in images is the right way to seek a better representation of words. Well, humans learn using all five senses and maybe the answer might be to incorporate in the models more heterogeneous perceptual information: not only static images but also videos, speech and the like. The debate is still open: the story \emph{goes on}\ldots{}

Last but not least, a mention needs to be made on concrete applications of these image-empowered word-embeddings. The use of images to support linguistic models has been experimented in several fields, from \emph{Dialogue Response Generation} (e.g. \citet{sun2021multimodal}) to \emph{Machine Translation}, where for example \citet{ive2019distilling} found images to improve the quality of translation when the textual context is generic and/or ambiguous. The number of potential applications of the models described in this subchapter is growing steadily in the scientific community. But this is yet \emph{another} story\ldots{}

\hypertarget{appendix-selected-models---summary}{%
\subsection{Appendix: Selected Models - Summary}\label{appendix-selected-models---summary}}

A table (available \href{https://github.com/slds-lmu/seminar_multimodal_dl/blob/master/Table-ch2-3-final.pdf}{here}) contains a summary of selected language models augmented with visual components. For each model, the following information are reported:

\begin{itemize}
\tightlist
\item
  Pure language model and pretraining data
\item
  Visual features and pretraining data
\item
  Fusion strategy of the two modalities
\item
  Benchmarks/baselines for evaluation
\end{itemize}

\hypertarget{c02-04-text-support-img}{%
\section{Text supporting Vision Models}\label{c02-04-text-support-img}}

\emph{Author: Max Schneider}

\emph{Supervisor: Jann Goschenhofer}

\hypertarget{introduction-2}{%
\subsection{Introduction}\label{introduction-2}}

\begin{quote}
``The biggest lesson that can be read from 70 years of AI research is that general methods that leverage computation are ultimately the most effective, and by a large margin.
{[}\ldots{]} Most AI research has been conducted as if the computation available to the agent were constant (in which case leveraging human knowledge would be one of the only ways to improve performance) but, over a slightly longer time than a typical research project, massively more computation inevitably becomes available.
Seeking an improvement that makes a difference in the shorter term, researchers seek to leverage their human knowledge of the domain, but the only thing that matters in the long run is the leveraging of computation.
{[}\ldots{]} One thing that should be learned from the bitter lesson is the great power of general purpose methods, of methods that continue to scale with increased computation even as the available computation becomes very great.''

\VA{--- \citet{sutton2019bitterlesson}}{}
\end{quote}

This insight seems to directly inspire most model choices presented in this chapter.
Each network can be seen as an attempt of its creators to employ their vast available resources on a large scale, with a particular focus on dataset sizes.
This mostly becomes feasible through the adaptation of recent findings in natural language processing (NLP; see chapter \ref{c01-01-sota-nlp}) to computer vision (CV).
On the one hand, architectural concepts firstly popularized in NLP are translated to CV \citep[e.g., self-supervised learning or the Vision Transformer;][]{ImageT} (see chapter \ref{c01-02-sota-cv}).
On the other hand, these powerful new NLP models, mostly Transformers \citep{vaswani2017attention}, support bigger models from the inside as text encoding building blocks; hence the name of this chapter.
Throughout this chapter, we will introduce recent relevant CV models CLIP \citep{radford2021learning}, ALIGN \citep{jia2021scaling} and Florence \citep{yuan2021florence} and discuss their underlying core concepts.
The strong performances confirm the potential, hinted at by the impressive GPT-3 \citep{brown2020language}, of improving CV and increasing scale with the help of NLP.

\hypertarget{concepts}{%
\subsection{Concepts}\label{concepts}}

\hypertarget{webScaleData}{%
\subsubsection{Web-scale data}\label{webScaleData}}

A core problem that troubles researchers is the lack of robustness of previous state-of-the-art CV models to distribution shifts.
I.e., when a model with good performance on its original dataset fails to generalize (transfer its knowledge) to new, more or less similar datasets.
E.g., \citet{radford2021learning} report that a ResNet101 which they trained on ImageNet to an accuracy of 76.2\% maintains only an accuracy of 32.6\% on ObjectNet.
This suggests that the model perhaps did not learn high quality latent representations, but instead overfit to the dataset-specific data-generating distribution.
A common way to tackle this would be to try out various changes on the architecture and the training algorithm of the network.
But this kind of adaptation, inscribing expert knowledge into the model, seems to repeat the mistake pointed out by \citet{sutton2019bitterlesson}; ``micromanaging'' a model is likely to thwart future scaling.

The researchers of CLIP, ALIGN and Florence follow a different approach, based on scale.
They try to increase sample size as much as possible and work with tremendous numbers of training observations:

\begin{itemize}
\tightlist
\item
  400 million \citep[CLIP;][]{radford2021learning}
\item
  900 million \citep[Florence;][]{yuan2021florence}
\item
  1.8 billion \citep[ALIGN;][]{jia2021scaling}
\end{itemize}

These large-scale dataset are generated using the vast amount of image-text pairs produced by and readily available on the internet.
Thus, error prone, cost and labor intensive (difficult to scale), manual labeling is avoided.
Unfortunately, the models trained on web data also become vulnerable to their downsides.
Because of their extremely noisy nature, still some form of pre-processing is needed, e.g., filtering for English language, excluding graphic content and, optionally, removing images with non-informative alt-texts.
This makes some degree of dataset curation, and therefore arbitrary choices, necessary.
Likewise, the social biases inherent to the internet are reproduced and furthermore, while this approach improves data efficiency to some degree (see next subsection \ref{contrObj}), the poor performance of deep learning in this area is not substantially enhanced and mainly just compensated for with a super scalable source of supervision \citep{radford2021learning}.

\hypertarget{contrObj}{%
\subsubsection{Contrastive objective}\label{contrObj}}

This source of supervision is the information contained in the co-occurrence of the image with its alt-text.
It is accessed through natural language supervision.
The architectures jointly train two sub-networks for image and text encoding, respectively.
During this, the vector encodings are aligned in the latent representation space through minimizing a variant of the contrastive loss function \eqref{eq:contrLoss} \citep{tian2020contrastive}.
Half of the first image-text pair loss

\begin{equation}
  \ell_1^{V_\text{img}, V_\text{txt}} = - \underset{\{v_\text{img}^1, v_\text{txt}^1, \ldots, v_\text{txt}^N\}}{\mathbb{E}} \left( \log \frac{h_\theta(\{v_\text{img}^1,v_\text{txt}^1\})}{h_\theta(\{v_\text{img}^1,v_\text{txt}^1\}) + \sum_{k=2}^N h_\theta(\{v_\text{img}^1, v_\text{txt}^k\})} \right),
  \label{eq:contrLoss}
\end{equation}

where \(v_\text{img}^1\) and \(v_\text{txt}^1\) are vector encodings (latent representations) of image 1 and text 1 and \(h_\theta(\cdot)\) is a similarity measure.
In order to guarantee symmetry, the total loss is formed by the sum of \(\ell_1^{V_\text{img}, V_\text{txt}}\) and \(\ell_1^{V_\text{txt}, V_\text{img}}\), where the pairwise similarities of one text and every image is calculated instead of the other way around.

\begin{figure}

{\centering \includegraphics[width=1\linewidth]{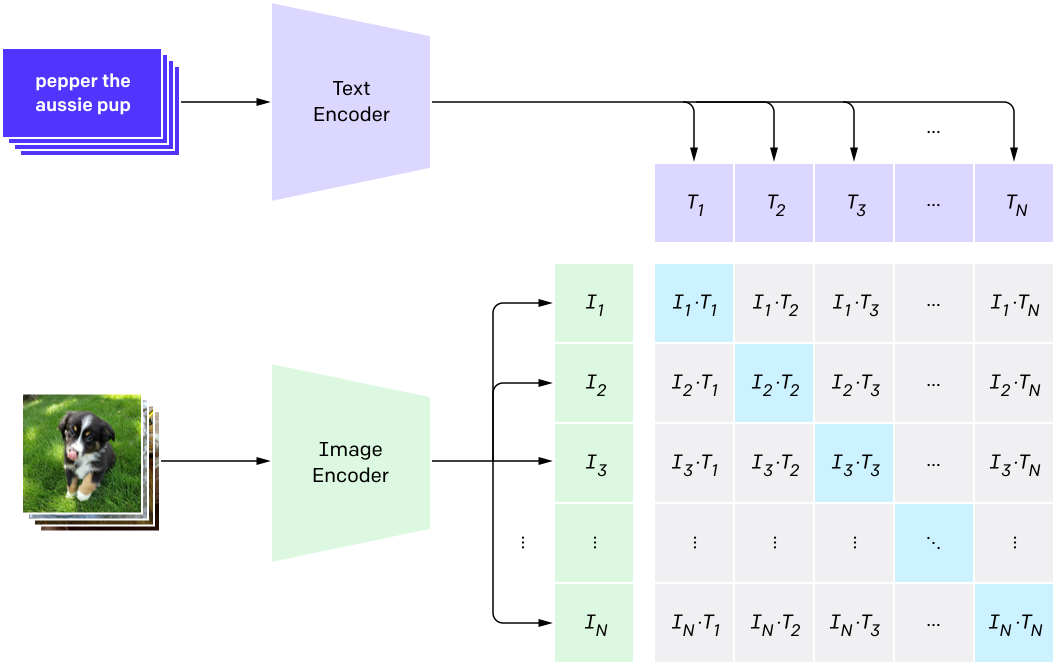}

}

\caption{Visualization of a contrastive objective \citep{radford2021learning}. After encoding the data, a similarity matrix for the images and texts is computed. The aim is that the N true image-text pairs score high in terms of similarity, while the \(\text{N}^2 - \text{N}\) other possible combinations score low.}\label{fig:contr-viz}
\end{figure}

Figure \ref{fig:contr-viz} visualizes this.
Initially all images and texts in the training data are encoded by the responsible sub-network.
Using the resulting encodings, a similarity matrix with elements \(h_\theta(\{v_\text{img}^i,v_\text{txt}^j\})\) can be calculated.
Loosely speaking, the contrastive objective is to maximize elements on the diagonal and minimize the others.

\begin{figure}

{\centering \includegraphics[width=1\linewidth]{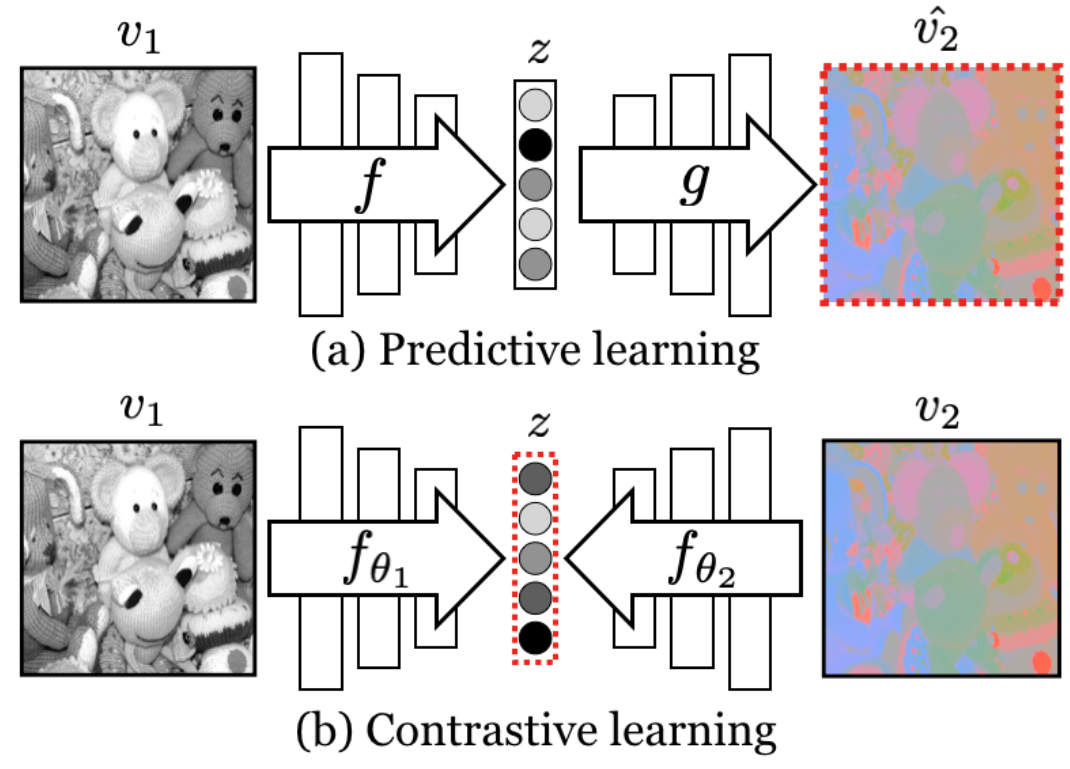}

}

\caption{Predictive vs.~contrastive learning: Predictive losses are measured in the output space while contrastive losses are measured is in the representation space, indicated by red dotted boxes \citep{tian2020contrastive}.}\label{fig:contr-vs-pred-learn}
\end{figure}

Contrastive learning can be contrasted with classical predictive learning.
Figure \ref{fig:contr-vs-pred-learn} gives an interesting insight into the choice of space, where goodness of fit is measured.
The exemplary task is to color an image given its B/W version.
Approach (a) first encodes the B/W image and then decodes the interim latent representation to fitting colors.
The goodness of this fit is measured in the output space, meaning the estimated colors are compared to the true colors.
Conversely, approach (b) measures the loss in the representation space.\footnote{Note that contrastive learning easily works with other combinations of modalities than text and image; here B/W and colors.}
A reason for the good performance of contrastive learning could be that, while common prediction losses (e.g., the \(\mathcal{L}_2\) loss) penalize each prediction output dimension independently, approach (b) implies measurement in the intertwined representation space \citep{tian2020contrastive}.

\begin{figure}

{\centering \includegraphics[width=1\linewidth]{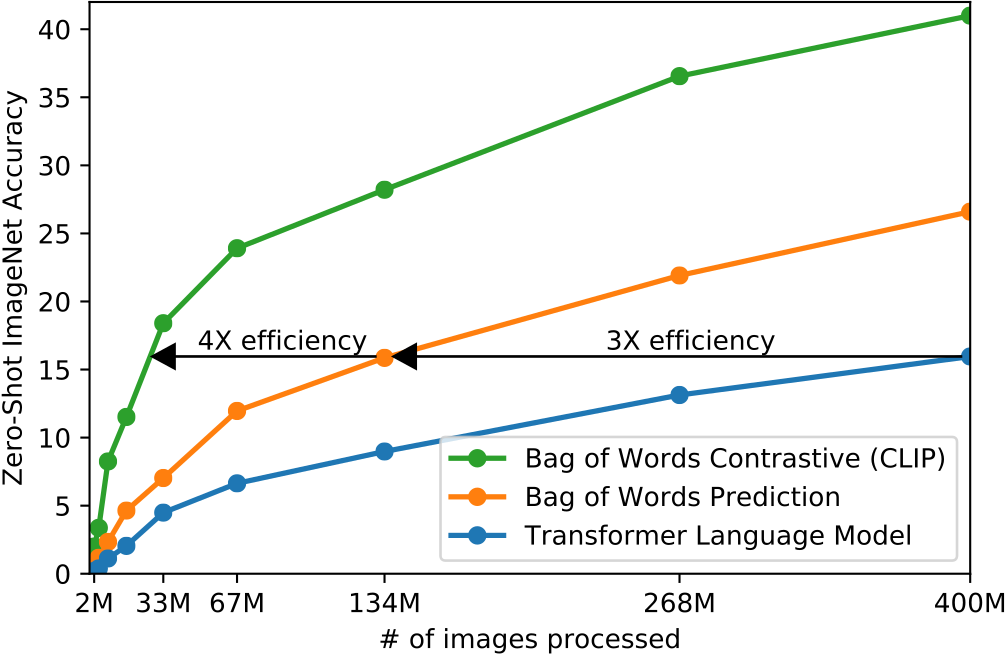}

}

\caption{Data efficiency of contrastive objective. Development of zero-shot accuracy (see next subsection \ref{foundMod}) on ImageNet with increasing number of instances of training data processed by the models. The contrastive objective reaches similar accuracy scores as the generative approach with only a seventh of the amount of data \citep{radford2021learning}.}\label{fig:data-efficiency}
\end{figure}

But in the end, rather than theoretical considerations, the driving factor for using this objective is data efficiency.
As can be seen in figure \ref{fig:data-efficiency}, \citet{radford2021learning} start their search for an adequate pre-trained model (more on this in subsection \ref{foundMod}) by experimenting with a Transformer-based language model predicting the exact captions of an image.
It turns out that this approach trains three times slower, in terms of data efficiency, compared to a simpler baseline of predicting a bag-of-words text encoding.
Additionally, switching to the contrastive objective of CLIP improves data efficiency by a factor of four.

Nonetheless, the switch to contrastive learning leads to some limitations.
Its rigidity demands certain extra steps and forfeits the high flexibility of generative models.
In particular, this means contrastive models similar to CLIP are limited to choose from available options and cannot freely generate texts or images.
To extend the capabilities of those models additional network building blocks are necessary.

\hypertarget{foundMod}{%
\subsubsection{Foundation models and zero-shooting}\label{foundMod}}

The first models which are considered foundation models today began to appear in NLP.
The term, later coined by \citet{bommasani2021opportunities}, refers to models that are noteworthy due to their large scale and ability to adapt to a wide variety of downstream tasks.
An early example is BERT \citep{Devlin2018}.
Often, foundation models have an unfinished touch to them and the true scope of their capabilities cannot be sketched out clearly.
This generally is the case because the desired abilities of neural networks are not designed for explicitly, but rather emerge during their implementation and usage on downstream tasks.
\citet{bommasani2021opportunities} cite GPT-3's ability to perform certain types of new tasks solely by confronting it with the right natural language prompt.
E.g., it is possible to get GPT-3 to summarize a paragraph by appending ``TL;DR'' (too long, didn't read) to the prompt, which is a common pattern on the internet to signal a following summery.
This is referred to as ``in-context learning'' \citep{brown2020language}.
It is apparent that one can make up plenty of unexpected ways to employ these models and it remains unknown whether there is a further way no one thought of yet.
This means possibly saving computational and data collection costs down the line, which ineptly is true for malicious use cases, e.g., surveillance, too.

Foundation models build on the concept of transfer-learning, i.e., pre-training a model on a feasible source task and applying it to the desired downstream task.
In the context of this chapter this means pre-training on web-scale data (see subsection \ref{webScaleData}) and evaluating performance on various common classification datasets.
E.g., \citet{radford2021learning} name the SVHN dataset as a proxy for the task ``street number transcription'' with the caveat ``on the distribution of Google Street View photos'', but they remark that a lot of datasets have no obvious, specific task associated, e.g., CIFAR-10.
They use these kind of datasets for measuring the ``robustness to distribution shift and domain generation'' of their model, which still is a topic of great interest as mentioned in subsection \ref{webScaleData}.
When there is no further fine-tuning on the downstream task, i.e., no resuming of training on the new target dataset, this is referred to as zero-shooting.
Zero-shooting has the clear advantage of evaluating performance more unbiased, as processes like overfitting to the data-generating distribution will not distort results.

\begin{figure}

{\centering \includegraphics[width=1\linewidth]{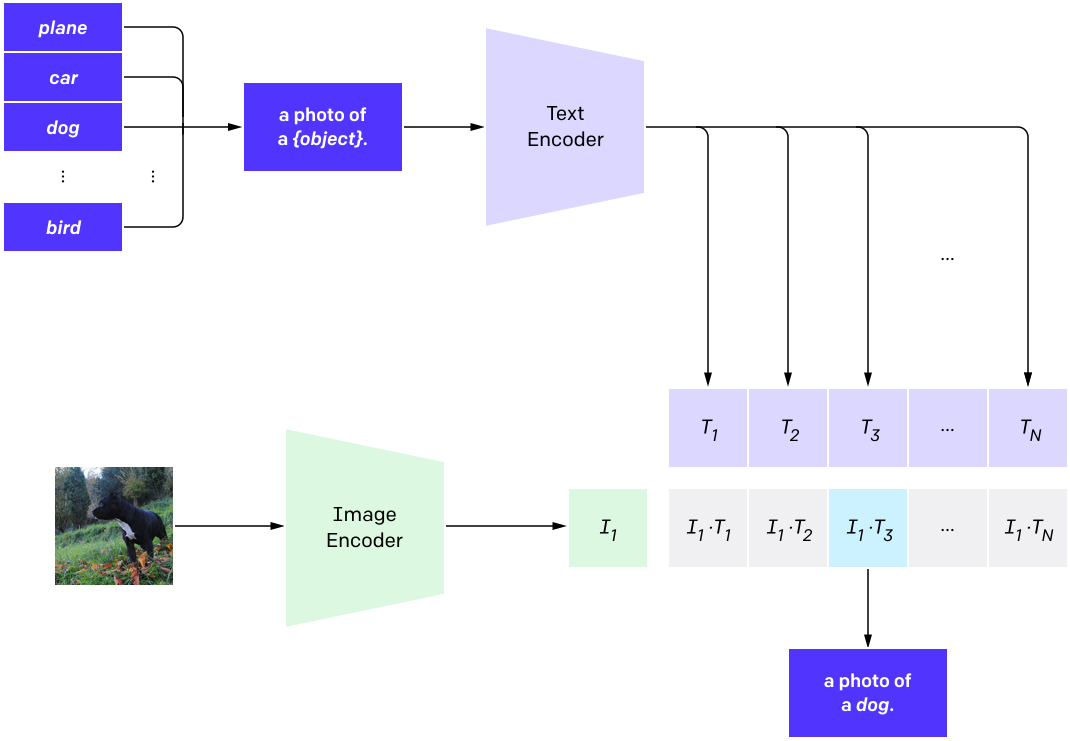}

}

\caption{Visualization of zero-shooting \citep{radford2021learning}.}\label{fig:zero-shooting}
\end{figure}

Figure \ref{fig:zero-shooting} shows how contrastive models perform zero-shot transfer.
In the case of image classification all available classes are encoded by the language model.
Afterwards, the CV sub-network computes the encoding of the image to be classified and all pair-wise similarity scores are returned.
The pair with the best score can be retrieved as the decision.
Image retrieval works the other way around:
After an initial encoding of all images, the ones most similar to the encoded natural language text prompt in the representation space can be returned.

\hypertarget{architectures}{%
\subsection{Architectures}\label{architectures}}

\hypertarget{clip}{%
\subsubsection{CLIP}\label{clip}}

The first of the large scale contrastive CV models that were published is CLIP, short for Contrastive Language-Image Pre-training \citep{radford2021learning}.
The components of its name are explained in previous subsections \ref{contrObj}, \ref{webScaleData} and \ref{foundMod} and are the crucial concepts of ALIGN and Florence as well.
CLIP is a product of OpenAI, but its code is freely available and the different versions can be accessed as \href{https://github.com/openai/CLIP}{python modules}.
The dataset used for training is not released though.

A lot of preliminary work stems from \citet{zhang2020contrastive}, who introduced contrastive representation learning using image-text pairs.
Their implementation of the contrastive loss function \eqref{eq:contrLoss} follows

\begin{equation}
  \ell_1^{V_\text{img}, V_\text{txt}} = - \log \frac{\exp(\langle v_\text{img}^1, v_\text{txt}^1 \rangle / \tau)}{\sum_{k=1}^{N} \exp(\langle v_\text{img}^1, v_\text{txt}^k \rangle / \tau)},
  \label{eq:contrLossCLIP}
\end{equation}

where \(\langle v_\text{img}^1, v_\text{txt}^1 \rangle\) represents the cosine similarity, i.e., \(v_\text{img}^{1 \top} v_\text{txt}^1 / (\|v_\text{img}^1\| \|v_\text{txt}^1\|)\), and \(\tau \in \mathbb{R}^+\) is a temperature parameter, which is directly learned during training \citep{zhang2020contrastive}.
CLIP adopts this.
\(\ell_1^{V_\text{txt}, V_\text{img}}\), the counterpart to \(\ell_1^{V_\text{img}, V_\text{txt}}\) for the total loss, is function \eqref{eq:contrLossCLIP} with switched arguments.
This can be viewed as a symmetric cross entropy loss over the cosine similarity of the embeddings \citep{radford2021learning}.

\textbf{Architecture}

The text encoder for CLIP (see figure \ref{fig:contr-vs-pred-learn}) is a modified Transformer \citep{vaswani2017attention}, which was also used for GPT-2 \citep{radford2019language}.
For the image encoder multiple sub-networks are evaluated:

\begin{itemize}
\tightlist
\item
  ResNets: ResNet-50, ResNet-101
\item
  ResNets which follow EfficientNet-style model scaling: RN50x4, RN50x16, RN50x64
\item
  Vision Transformers: ViT-B/32, ViT-B/16, ViT-L/14
\end{itemize}

The best performing sub-network was the ViT-L/14.
In turn, they trained it for an additional epoch with higher resolution images (336px), denoting this version \href{mailto:ViT-L/14@336px}{\nolinkurl{ViT-L/14@336px}}.
If not indicated otherwise, the performances of this version of CLIP are displayed.
The EfficientNet-style ResNets use x4, x16 and x64 of the compute of a ResNet-50 and the largest model (the RN50x64) trained for 18 days on 592 V100 GPUs, while the ViT-L/14 only took 12 days on 256 GPUs.
The high parallelization capabilities of Transformers seem to pay off.

When explaining zero-shooting initially (see subsection \ref{foundMod}), a text processing step was skipped.
As can be seen in figure \ref{fig:zero-shooting}, there is an additional operation before the labels are fed into the text encoder.
In order to help the model understand the context of the words, the class labels are embedded in a sentence, e.g., ``A photo of a \{label\}.''.
This increases the models zero-shot accuracy on ImageNet by 1.3 percentage points (pp).
When ensembling 80 different context prompts\footnote{Prompts like: ``A photo of a big \{label\}.'', ``A photo of a small \{label\}.'' \citep{radford2021learning}} \citet{radford2021learning} improve ImageNet accuracy by an additional 3.5pp, which adds up to a total of nearly 5pp.
The average performance gain across 36 datasets is reported to be 5pp.
It is similarly possible to directly communicate visual concepts like ``picture'', ``macro'', ``drawing'' or even ``dog'' to the model.

\textbf{Robustness}

\begin{figure}

{\centering \includegraphics[width=1\linewidth]{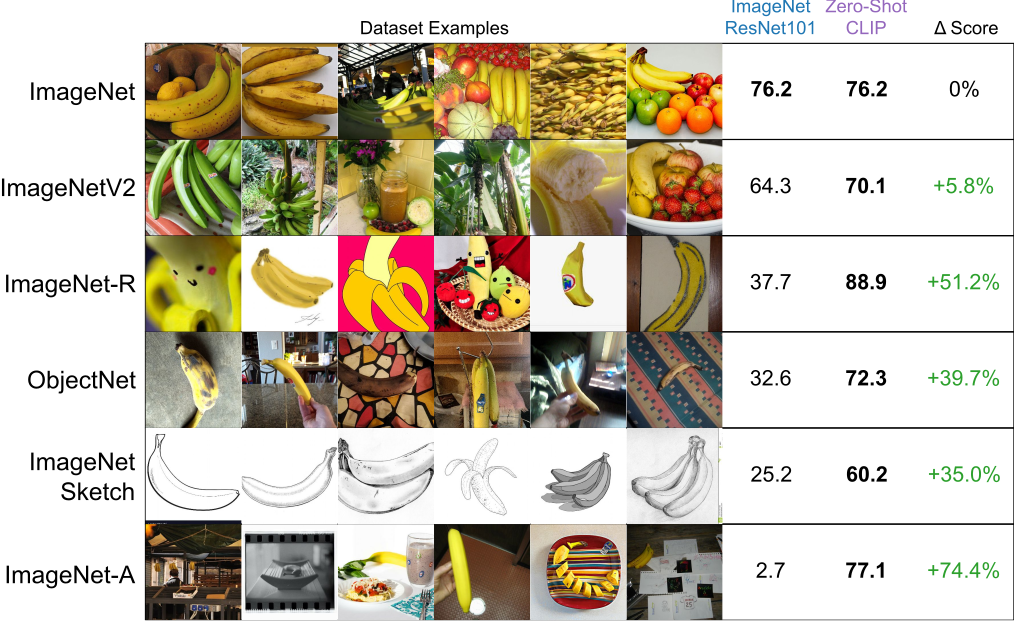}

}

\caption{Robustness of zero-shot CLIP to distribution shifts \citep{radford2021learning}.}\label{fig:performance-clip}
\end{figure}

Figure \ref{fig:performance-clip} illustrates the performance of CLIP and a ResNet101, whose training on ImageNet was stopped at the point it reached the same accuracy as zero-shot CLIP.
It can be deduced that the methods studied in the paper of \citet{radford2021learning} constitute an important step towards closing the robustness gap mentioned earlier (see subsection \ref{webScaleData}).
While the performance of the ResNet101 deteriorates with datasets generated from more and more different data distributions compared to ImageNet, CLIP remains fairly accurate.
Note that these findings have to be taken with a grain of salt.
Because OpenAI does not grant public access to their training data, independent parties cannot investigate these claims on their own.
E.g., it has to be relied on the conclusions of their overlap analysis to rule out that CLIP has not seen biasing amounts of future test data during training.

\textbf{CLIP as a building block}

\begin{figure}

{\centering \includegraphics[width=1\linewidth]{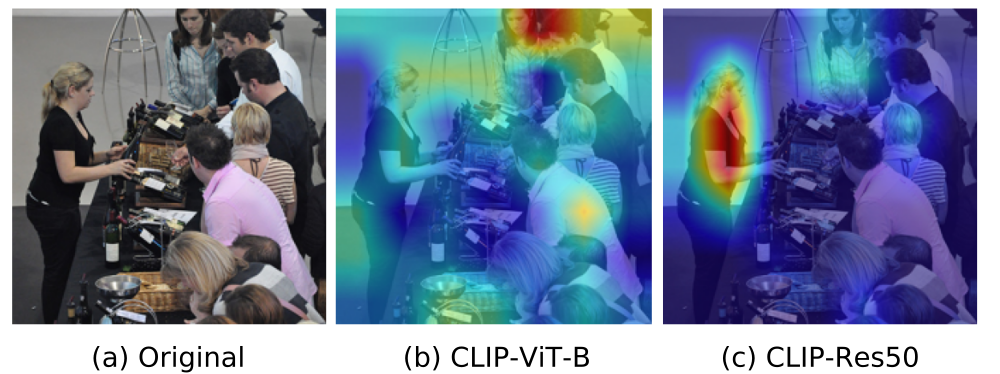}

}

\caption{Grad-CAM Visualizations for the prompt ``What color is the woman's shirt on the left?''.}\label{fig:attention-ViT}
\end{figure}

\citet{shen2021much} study how the performance of Vision-and-Language (V\&L) models improves, when the visual encoder is switched to CLIP's strong image encoder.
They discover that in this field of CV the ViT-B scores significantly worse than the ResNets.
E.g., tests on image captioning reveal that the V\&L model using ViT-B often performs only half as strong as the version using the RN50x4 (the largest network used in this study).
This is possibly due to the pooling strategies of ViT-B, which result in a lack of visual localization abilities.
\citet{shen2021much} test their hypothesis and generate, e.g., figure \ref{fig:attention-ViT} which depicts Grad-CAM Visualizations for a V\&L model with a ViT-B backbone and a ResNet-50 backbone and the question ``What color is the woman's shirt on the left?''.
The red area indicates relevant pixels and appears much more focused for CLIP-Res50 than for CLIP-ViT-B.

\hypertarget{align}{%
\subsubsection{ALIGN}\label{align}}

The approach of \citet{jia2021scaling} is largely similar to CLIP.
They reiterate the necessity of large-scale vision datasets, but assert that even CLIP's data collection process still involves a non-trivial amount of data curation.
They propose that the amount of additional observations obtained through minimizing the amount of filtering makes up for the increased noise.
Following this rationale, they create a training dataset with 1.8 billion image-text pairs.
The corresponding model is named ALIGN, short for ``A Large-scale ImaGe and Noisy-text embedding'', whose acronym hints at the contrastive loss, which aligns vector encodings in the representation space (see subsection \ref{contrObj}).

\textbf{Architecture}

ALIGN follows the dual encoder architecture employed by \citet{zhang2020contrastive} and \citet{radford2021learning}, but uses a part of BERT-Large as the text and EfficientNet-L2 as the image encoder, which they jointly train from scratch.
The model has around 800 million parameters \citep{alford2021alignparams}.
Subsection \ref{performanceComp} goes into more detail about the performance of ALIGN and compares all three models discussed in this subsection.

\textbf{Connecting image and text representations}

\begin{figure}

{\centering \includegraphics[width=1\linewidth]{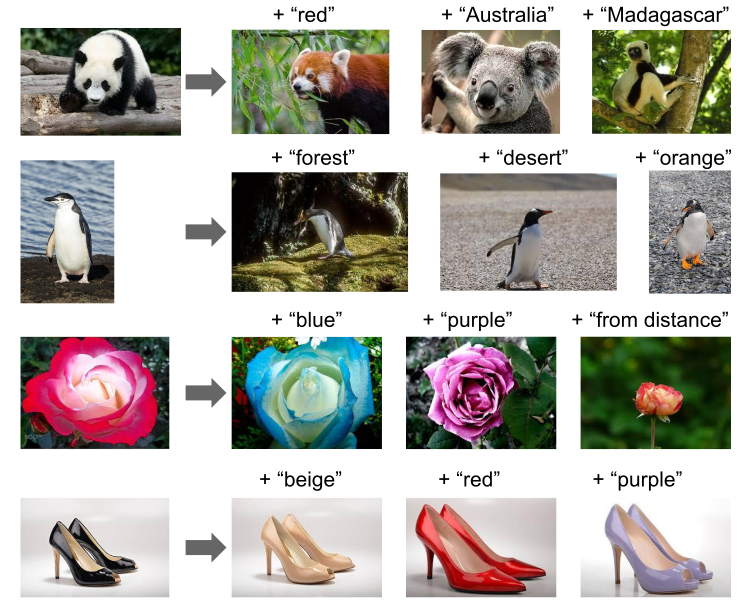}

}

\caption{Multimodal image retrieval via arithmetic operations on word and image embeddings.}\label{fig:img-txt-addition}
\end{figure}

The contrastive loss function aligns the latent representations of the different modalities.
In other words, the explicit objective is that similar vector encodings implicate similar inputs.
This means arithmetic operations like the ones mentioned in chapter \ref{c01-01-sota-nlp} are not only meaningful on encodings belonging to the same modality, but to different modalities.
E.g., one can add up the image encoding of a picture of the Eiffel tower and the text encoding of the word ``snow'' and retrieve pictures with high cosine similarity as a result, see figure \ref{fig:img-txt-addition} for an illustration.

\hypertarget{florence}{%
\subsubsection{Florence}\label{florence}}

\begin{figure}

{\centering \includegraphics[width=1\linewidth]{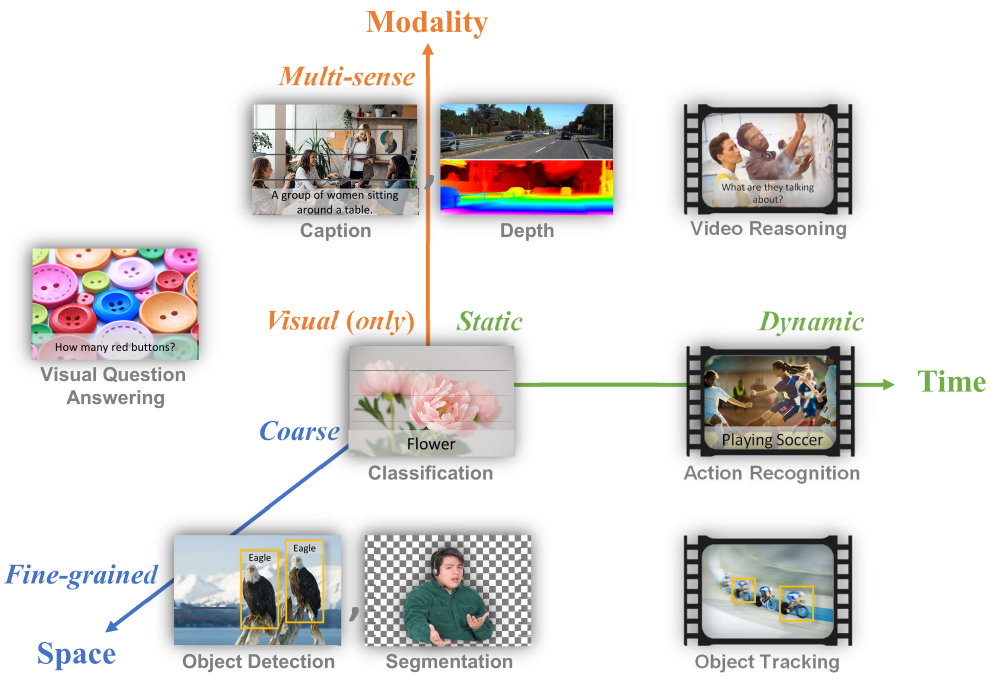}

}

\caption{Florence' approach to foundation models: A general purpose vision system for all tasks.}\label{fig:florence-dimensions}
\end{figure}

While in principle the approach of \citet{yuan2021florence} does not largely differ from the others, the focus of this paper is more about creating a true foundation model.
In order to achieve this, they propose a map of possible vision applications which the try to cover via extending the core model with modules.
As figure \ref{fig:florence-dimensions} depicts, they want to advance into the dimensions of fine-grained object detection, dynamic action recognition and true multimodal tasks.
Due to their big ambitions, they name their model Florence after ``the birthplace of Renaissance'' \citep{yuan2021florence}.

\textbf{Architecture}

\begin{figure}

{\centering \includegraphics[width=1\linewidth]{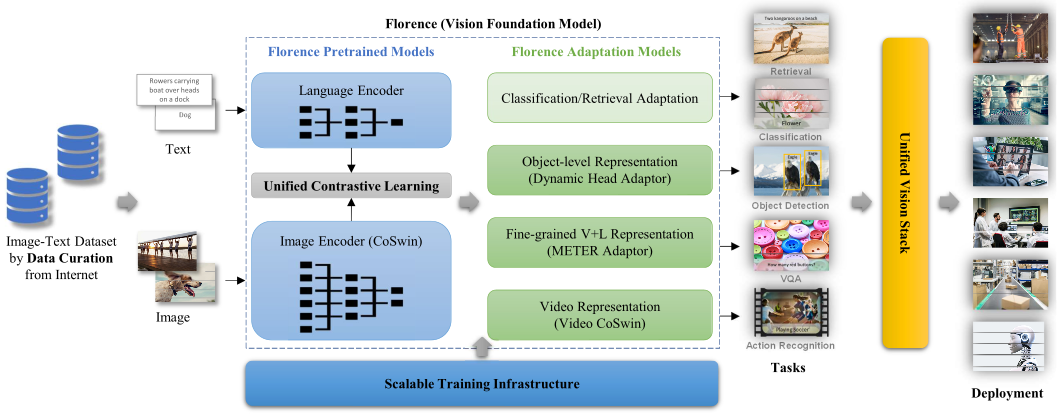}

}

\caption{Modular architecture of Florence.}\label{fig:florence-architecture}
\end{figure}

As the two encoders for the pre-trained core they use a hierarchical Vision Transformer (CoSwin Transformer) for images and a Transformer similar to CLIP's for text.
Their 893 million parameters are also jointly trained from scratch on 900 million image-text pairs.
The alignment happens in the so called image-label-description space which is encoded through a special version of the contrastive loss function which regards all image-text pairs with the same label as positive instances.
Figure \ref{fig:florence-architecture} depicts their version of figure \ref{fig:contr-viz} where one can schematically see how they flexibly add modules to the pre-trained core in order to adapt to various downstream tasks.

\hypertarget{performanceComp}{%
\subsection{Performance comparison}\label{performanceComp}}

Throughout the papers of \citet{radford2021learning}, \citet{jia2021scaling} and \citet{yuan2021florence} we were able to collect three tables with reported performance measures to compare these approaches.

\begin{figure}

{\centering \includegraphics[width=1\linewidth]{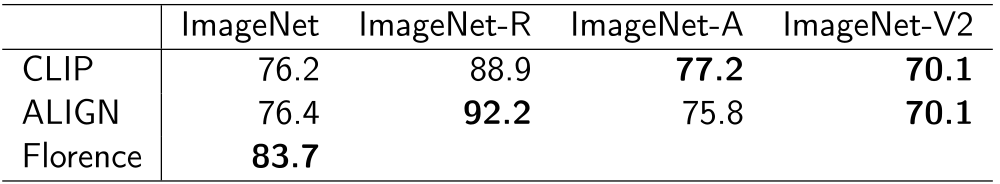}

}

\caption{Top-1 Accuracy of zero-shot transfer of models to image classification on ImageNet and its variants.}\label{fig:table1}
\end{figure}

Table \ref{fig:table1} summarizes the zero-shot accuracies on four different ImageNet variants.
Unfortunately \citet{yuan2021florence} only stated their performance on the original ImageNet, where they beat CLIP and ALIGN by a margin of 7.3pp.
The results on the other three ImageNet pendants are mixed and there is no clear winner between CLIP and ALIGN.

\begin{figure}

{\centering \includegraphics[width=1\linewidth]{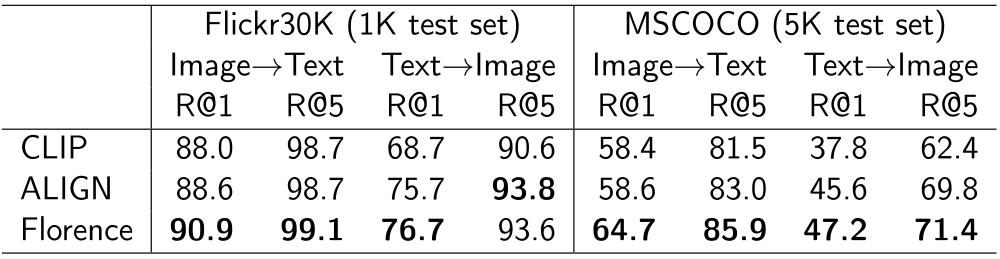}

}

\caption{Zero-shot image and text retrieval \citep{yuan2021florence}.}\label{fig:table2}
\end{figure}

Table \ref{fig:table2} concerns zero-shot image retrieval on the Flickr30K and the MSCOCO dataset (see chapter \ref{c01-03-benchmarks}).
Even though there are not many major score differences, there is a clear ranking with CLIP on third, ALIGN on second and Florence on the first place.

\begin{figure}

{\centering \includegraphics[width=1\linewidth]{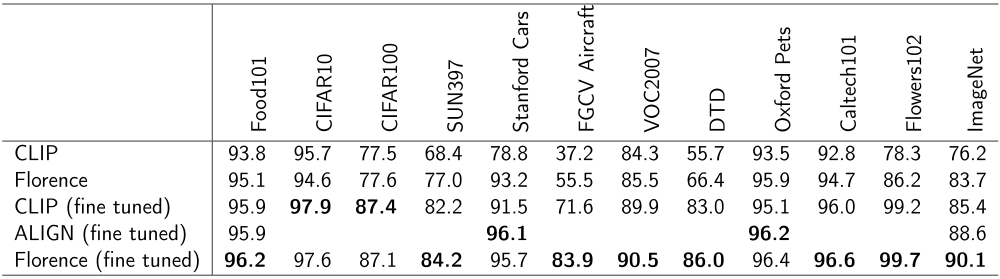}

}

\caption{Top-1 Accuracy of CLIP, Florence and ALIGN on various datasets.}\label{fig:table3}
\end{figure}

The most comprehensive comparison is shown in table \ref{fig:table3}.
It depicts the accuracy of zero-shot CLIP and Florence on various datasets as well as the scores of all three models fine tuned to the respective datasets.
Florence beats CLIP in nearly all evaluations, for the zero-shot setting as well as for fine tuned performance.
\citet{jia2021scaling} only report on four of these twelve datasets, where they win half of the time.

Summing up, ALIGN achieves its goal of replicating CLIP's impressive performance while dramatically reducing the required data curation effort and Florence has the overall top performance.
This could be attributed to its custom loss, maybe to \citet{yuan2021florence} striking the best balance between sample size and data curation or to Florence having the best sub-networks; or a combination of all three.

Once again note that none of the training datasets were made publicly available.
It cannot be guaranteed that all benchmarks were evaluated on unseen datasets.

\hypertarget{resources}{%
\subsection{Resources}\label{resources}}

One can access the pre-trained CLIP models on \href{https://github.com/openai/CLIP}{Github} and they even found their way into simple command line tools already.
For example there is a CLI named \href{https://github.com/yurijmikhalevich/rclip}{rclip}, which can be used for personal image retrieval, wrapping the \emph{ViT-B/32} CLIP architecture.
On a (mid-range, regular) laptop, we were able to find seemingly good matches for search terms which we tried out inside a folder containing about 100 different pictures.
After an initial caching one request took about ten seconds.
Furthermore CLIP continues to be used inside new models, e.g., DALL\(\cdot\)E 2, where it is used for the image embedding \citep{ramesh2022hierarchical}.
Also, there is a crowd-sourcing effort to replicate CLIP's training dataset called LAION-400M \citep{schuhmann2022laion}.
To validate the image-text pairs collected for this, their cosine similarity is computed using CLIP and instances with a value too low are discarded.
To our knowledge no resources were open-sourced as part of the other two papers ALIGN and FLORENCE.

\hypertarget{c02-05-text-plus-img}{%
\section{Models for both modalities}\label{c02-05-text-plus-img}}

\emph{Author: Steffen Jauch-Walser }

\emph{Supervisor: Daniel Schalk}

Data is naturally at the heart of every data scientific issue. While there have been many advances made in machine learning in recent years, many promising research areas remain, as do a multitude of problems associated with them. One such promising area are multi-modal machine learning models. Combining different input data is a key aspect towards making models more sophisticated. When thinking about teaching robots specific tasks, detecting hateful memes or deep fakes, it is apparent that only through the combination of multiple modalities, success might be achieved. Context is key.

However, learning context requires increasingly complex models. While early machine learning models built their success upon the possibility to analyze the big pool of available, often unstructured data, modern machine learning models are so demanding that there is often not enough data or training time available. Obtaining data is a major issue for multi-modal machine learning. Since labelling data in vast amounts is prohibitively expensive, larger models have to come up with specific strategies to move forward such as self-supervised training or automatically scraped web datasets. Nevertheless, when models become so large that billions of parameters have to be learned, even scraping the whole web starts to show its limits. Another natural issue is the transformation of different types of data into usable model inputs.

There is no shortage of different single modality machine learning models. On the contrary, when every new hyperparameter configuration might be seen a new model, it becomes hard to keep track. More importantly, it is often not clear how a model from one area transfers to another. Did we learn some modality specific bias or a general principle? Consolidating different models into a unifying framework is a key prospect of multimodal machine learning. While the grand dream of a single unifying model might be out of reach, consolidating different areas is well in sight. In the following, we will have a look at the challenges and prospects of multimodal machine learning against the background of visual language models. Visual Language Models are models which can deal with both language and images as input data. Specifically, we will have a closer look at three different models: Data2vec, VilBert and Flamingo. Data2vec is an unsupervised model that can handle different modalities, but not their interaction, using a single unifying training framework. VilBert is an early visual-language model that can handle interactions between images and text through its innovative concept of cross-attention. Flamingo is a recent few shot visual language model that features large expressive text capabilities through the use of a large language model. With 80B parameters, it particularly highlights how to leverage the communication between frozen models when further scaling up the model size.

\begin{figure}

{\centering \includegraphics[width=0.7\linewidth]{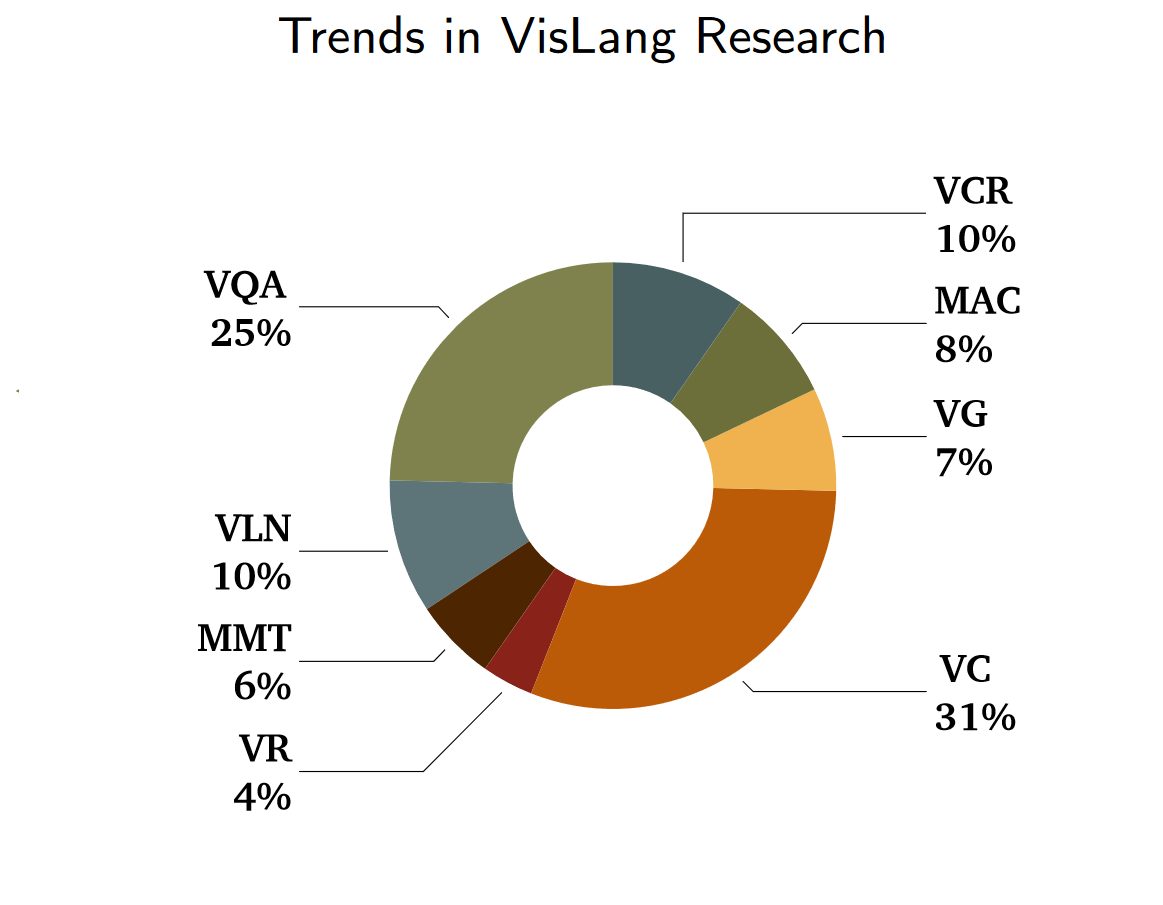}

}

\caption{\citet{uppal2022multimodal}: VisLang Paper Trends (previous 2 years)}\label{fig:vltasks}
\end{figure}

An overview across the popularity of current research fields in visual language modelling is provided in figure \ref{fig:vltasks}. A detailed list of trends for each of those fields can be found in \citet{uppal2022multimodal}. Most research is done in the areas of visual question answering (VQA) and visual captioning (VC), but also for example visual commonsense reasoning (VCR), vision-language navigation (VLN) or multimodal affective computing (MAC). MAC uses images and text to infer sentiment, for example through facial expressions. VCR as an extension of VQA is particularly interesting in the realm of making models more interpretable. After all, we would like to know why machine learning models do what they do. Finally, VLN has many promising practical applications in the field of robotics, particularly the interaction of humans and robots.

\hypertarget{data2vec}{%
\subsection{Data2vec}\label{data2vec}}

With data2vec \citep{baevski2022data2vec}, data scientists at Meta, formerly Facebook, developed an architecture that addresses some of the mentioned issues while highlighting the importance of sophisticated training schemes. Their algorithmic structure is able to work with either text, image or speech data. On top of that, the model is self-supervised based on a teacher-student relationship which reduces the need for human labelling. It is not a universal model in the sense that it works with any input, nor is it even a general model in the sense that the algorithm is exactly the same for each modality. However, the overall model structure remains the same for either text, speech or image input data, while only the specific encoding, normalization and masking strategies are modality-specific. In that regard, it is a step towards a more general way of dealing with different modalities and it is very effective at doing so given the benchmark results on typical data sets. Particularly noteworthy is also the way they implement the self-supervised learning. Data2vec predicts contextualized and continuous representations rather than typically used discrete tokens such as sub-words. Working with latent representations of the input space has two advantages: not only is the number of prediction targets not a-priori limited, but they are also richer in information.

\begin{figure}

{\centering \includegraphics[width=1\linewidth]{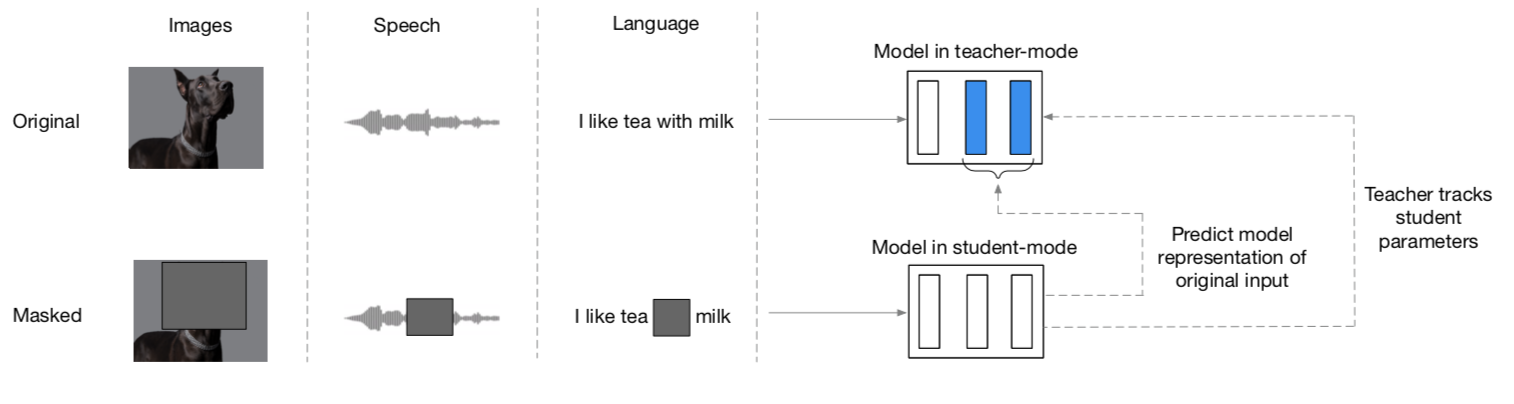}

}

\caption{\citet{baevski2022data2vec}: Data2vec Architecture - a teacher model creates contextualized latent targets on the basis of its top K layers (blue) as prediction task to train the student model}\label{fig:data2vecoverview}
\end{figure}

Figure \ref{fig:data2vecoverview} depicts the general model architecture. The two main components are a teacher and a student model which only differ in one aspect, the weights of the teacher model are an exponentially decaying average of the student's weights. The purpose of the teacher model is to create training targets for the student model. In a first step, a modality is chosen and inputs are encoded according to the specific encoding scheme for that modality. A masked version is given to the student model, but notably, the teacher model has access to an unmasked, complete view of the input data. Hence, the resulting training targets will be fully contextualized using a self-attention mechanism over the whole input data. The training targets are based on the top K layers of the teacher model depicted in blue in Figure \ref{fig:data2vecoverview}. More specifically, denoted by \(y_t\), the training target at time \(t\) and by \(a_t^l\) the outputs of the \(l\)-th block, then \(y_t = \frac{1}{K}\sum_{l=L-K+1}^{L} \hat{a}_t^l\), i.e.~the training targets are the average of the outputs of the top K layers of the teacher network after a normalization has been applied. Normalization helps to stabilize the training process and prevent model collapse which can be an issue with models that learn their own representation.

From the authors point of view, working with a latent representation of the actual learner as training target is a simplification of many commonly used modality-specific designs despite the caveat that this paper still uses modality-specific encoding strategies. Compared to other models, there is no cross-modality training. The specific loss function used to regress the targets is a smooth L1 loss.

\begin{align*}
       L(y_t,f_t(x)) =
    \begin{cases}
        \frac{(y_t - f_t(x))^2}{\beta}    \quad &\text{if} \quad | (y_t - f_t(x)) | \leq \beta \\
        | (y_t - f_t(x) | - \frac{\beta}{2} \quad &\text{otherwise}
    \end{cases}
\end{align*}

Using a smooth L1 loss has the advantage of being continuous, yet sensitive to outliers, however the \(\beta\) parameter needs tuning. As far as the general model architecture is concerned, the underlying architecture is a standard transformer architecture \citep{vaswani2017attention}.

How does the modality specific input handling work?

\begin{figure}

{\centering \includegraphics[width=1\linewidth]{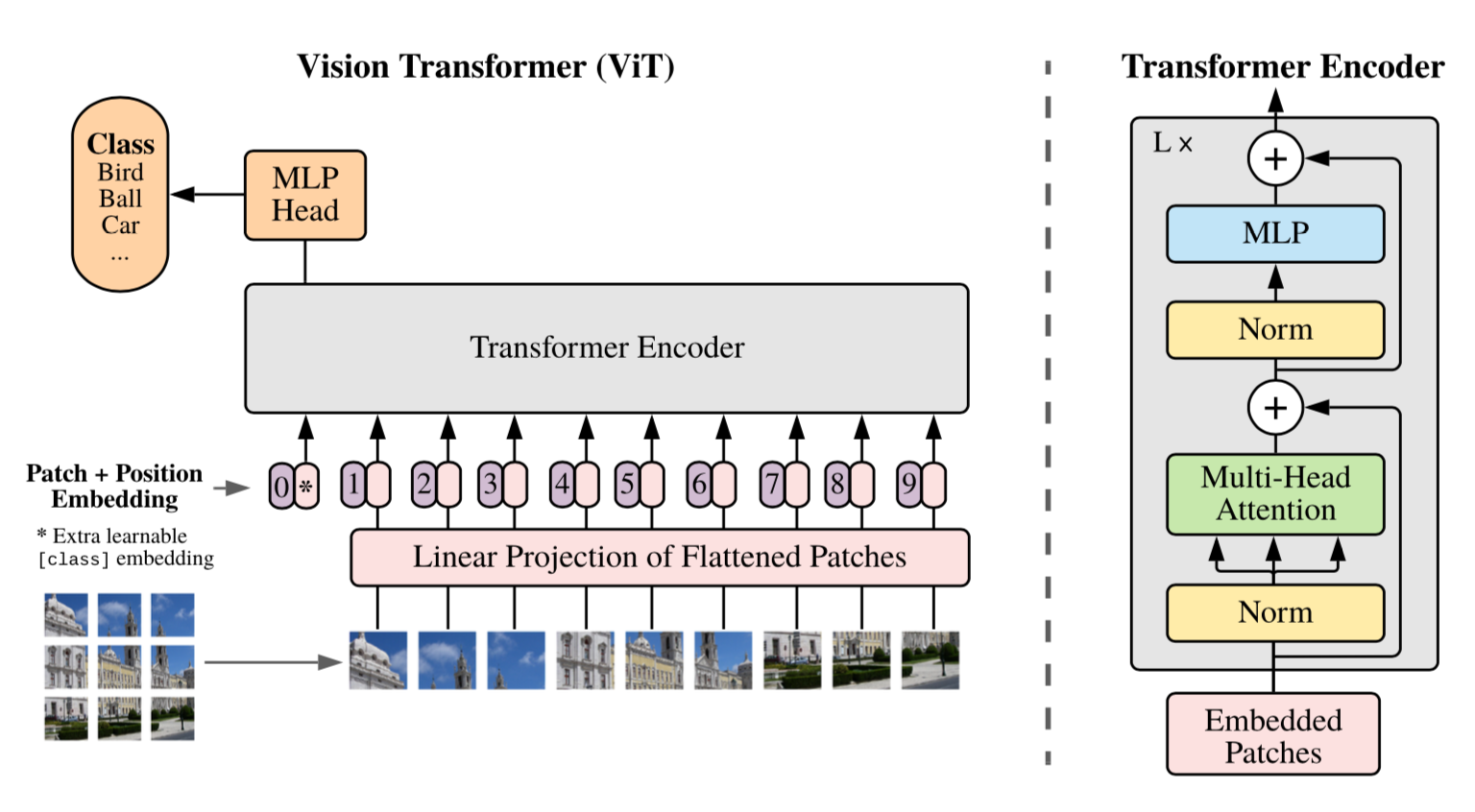}

}

\caption{\citet{DosovitskiyB0WZ21}}\label{fig:visiontransformer}
\end{figure}

In many ways, in this work the authors combine the strategies developed in multiple previous works and add a unifying framework on top of it. For images, the typical Vision Transformer (ViT) strategy (\ref{fig:visiontransformer}) to transform images with a size of 224x224 pixels into 16x16 pixel patches is employed. Every patch is then linearly transformed into a sequence of 196 flattened representations including a learn-able positional encoding that serve as input to the vision transformer. A classification token is used to produce the final categorization. The contextualization is produced in the multi-head attention blocks as explained in earlier chapters. In short, multi-head attention first projects the keys, queries and values with learned linear projections which are then evaluated in parallel to create more expressive attention maps. Attention itself is calculated as scaled dot-product-attention using a softmax over the scaled product of keys, queries and values \citep{vaswani2017attention}. As far as the vision transformer itself is concerned, datav2vec tests two different model sizes, a base model size of 12 and a large model of 24 transformer blocks. The masking strategy for images follows the Bert pre-training approach of image transformers, BEiT, proposed by \citet{bao2021beit}. In particular, multiple adjacent blocks are being masked with random aspect ratio. The minimum size of a masked block is 16 patches. In total, 60\% of patches were masked in the data2vec algorithm, which is an increase over the original 40\% used by BEiT. However, the authors note that they found increased masking to be more accurate. The augmentation strategies are similar, as well. Resizing crops, horizontal flipping and colour jittering were used. Naturally, the student and teacher model are the given the same modified image. Finally, for image data, the model is measured on a classification task. Hence, the authors use a mean-pooling over all patches in the last transformer block and input that into a softmax-normalized projection that conducts the classification, which is again based on the BEiT model.

The natural language processing model is implemented with a PyTorch toolkit named fairseq and based on the RoBERTa \citep{liu2019roberta} architecture which redesigned the standard Bert model training procedure to make it more robust and effective. In particular, it increases hyperparameters such as the learning rate and the batch size. It also removes the next sentence prediction task to improve on the masked language modelling performance. In this case they follow \citet{sennrich2015neural} and encode sub-words as 50k byte-pairs. A separate embedding vector is learned for each type. For the masking, the Bert masking is being used. 15\% of the embedded tokens are replaced, thereof 80 percent are learned masks, 10\% are unchanged and the remaining 10\% are replaced with random tokens in the vocabulary. Another strategy that the authors also consider is the wave2vec masking strategy' to mask four consecutive tokens with a probability of 0.35 while only using learned tokens \citep{baevski2020wav2vec}. As it turns out, the later strategy further improves the results. The natural language processing model is evaluated on the General Language Understanding Evaluation (GLUE) benchmark \citep{wang2018glue} which includes for example includes NLP inference, sentence similarity and sentiment analysis tasks.

The speech category is also implemented in fairseq. The feature encoder for speech is based on the wave2vec framework and uses 16 kHz inputs. It is built upon seven temporal convolutions intertwined with normalization layers and a GELU activation function such that the output of the encoder is 50 kHz.

As far as the results are concerend, data2vec achieved state-of-the-art performance in vision and language tasks among similar self-supervised models.

\begin{figure}

{\centering \includegraphics[width=0.5\linewidth]{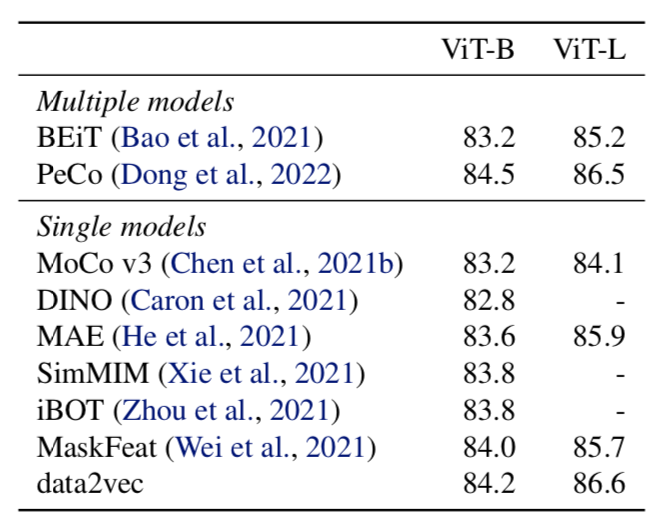}

}

\caption{\citet{baevski2022data2vec}: data2vec performance (vision)}\label{fig:data2vecresults1}
\end{figure}

Figure \ref{fig:data2vecresults1} shows the model's performance in computer vision. Pre-trained and fine-tuned simply on the data of the well known ImageNet-1K dataset, data2vec was evaluated using top1-accuracy, the standard notion of accuracy, on the task to predict single labels for images. The base model ViT-B comprises 86M parameters and ViT-L 307M parameters. The results show that predicting contextualized latent representations in a masked prediction setup can work well as model training compared to classical local methods such as predicting visual tokens. MoCov3 \citep{chen2021empirical} is a self-supervised model trained on a contrastive loss. The most similar model is DINO \citep{caron2021emerging}, which also uses a self-distillation setup to predict teacher outputs using a cross-entropy loss. However, their prediction target was the final layer rather than averaged layers while using differing images for teacher and student network. The well performing MAE model \citep{he2022masked} is a masked autoencoder which is trained on reconstructing masked pixels using an asymmetric encoder-decoder architecture. In contrast, MaskFeat \citep{wei2022masked} uses masked feature prediction. Notably, data2vec outperforms all of them although trained for the same amount or less. Particularly, MAE and MaskFeat use 1600 epochs rather than 800 like data2vec.

\begin{figure}

{\centering \includegraphics[width=1\linewidth]{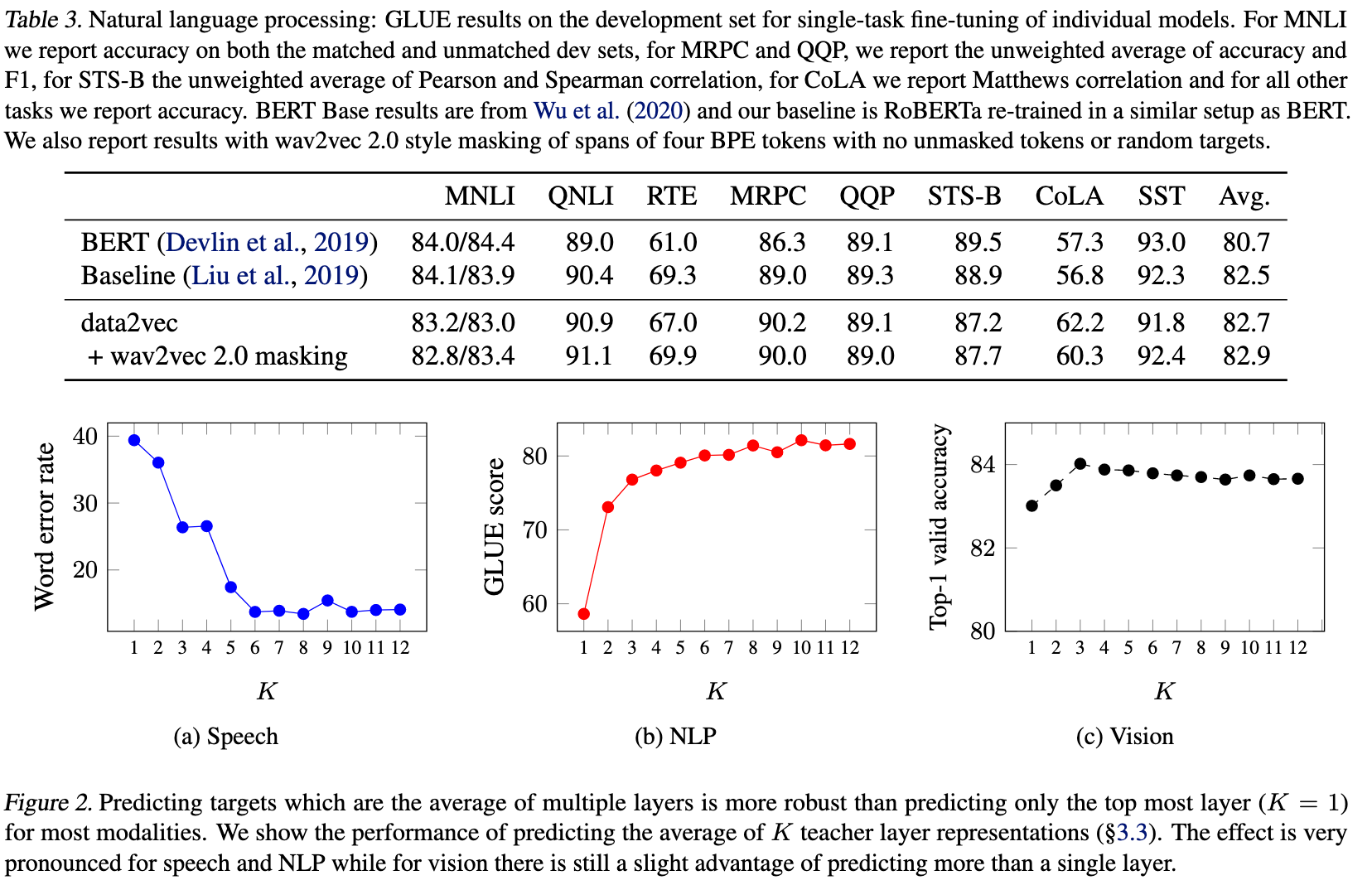}

}

\caption{(res:data2vecresults2)}\label{fig:data2vecresults2}
\end{figure}

(\url{res:data2vecresults2}) \citet{baevski2022data2vec}: data2vec results (language)

Figure \ref{fig:data2vecresults2} shows the performance in the language domain. For the language domain, the model is evaluated on the GLUE benchmark \citep{wang2018glue}. The model is pre-trained and fine-tuned separately on the labelled data from each task. Accuracy is reported as the average across 5 tuning cycles. While data2vec achieves a higher average performance than the baseline model, there are tasks where the baseline model prevails. A large portion of the performance difference seems to be driven by the CoLA task. The Corpus of Linguistic Acceptability (CoLA) consists of 10657 sentences from 23 linguistics publications and the task is to judge whether they are grammatically correct. Hence, it is distinctly different from the other tasks. The Stanford Sentiment Treebank (SST) analyzes sentiment in language through movie reviews. The Multi-Genre Natural Language Inference (MultiNLI) corpus contains sentence pairs and focusses on textual entailment across genres. Similar tasks are used in the Recognizing Textual Entailment (RTE) dataset which focuses on text from news and Wikipedia. The QNLI (Question-answering NLI) dataset is a Natural Language Inference dataset that contains answers from Wikipedia to corresponding questions posed by an annotator. The task for the model is to find out whether the sentence contains the answer to the question. QQP stands for Quora Question Pairs, which analyzes paraphrases. Finally, the Microsoft Research Paraphrase Corpus (MRPC) also consists of sentence pairs from newswires which may or may not be paraphrases of each other.

As a suitable baseline model, the authors retrain RoBERTa in the respective setup. On top of the heterogeneous performance across language tasks, the evaluation also clearly shows that averaging over multiple layers to create prediction targets improves performance across all three domains. The effects seem to be most pronounced on NLP tasks whereas CV does not benefit from averaging more than three layers. In the speech domain, six layers seems to be enough to reach peak performance. In any case, performance loss while following the strategy to simply average the maximum amount of layers, rather than fine-tuning K, seems small enough to be potentially acceptable.

To sum it up, data2vec is a self-supervised model that can work with either text, speech or image data, but not across modalities. It aims at unifying the learning framework through a teacher-student-setup that allows for contextualized latent target prediction. The teacher model is based on a complete view of the input data, which introduces contextualization, while the student model only sees a masked version of the input. Compared to previous work, the authors average the top K layers rather than only the final layer of the model, which has a notable effect as shown in \ref{fig:data2vecresults2}. As there are different layers in the transformer network, the authors also investigate which layers work best for prediction. They conclude that the output of the feedforward layer works best. Built on a transformer architecture, self-attention is the main driver that creates contextualized targets in the teacher model and hence performance. The authors also show that contextualization through the teacher model works best with the complete view of the input rather than a partial view. On top of not being able to work across modalities, one drawback is that the model's structure still uses modality specific encoding and masking schemes. In that regard, the perceiver architecture \citep{jaegle2021perceiver} for example used in the Flamingo model is a complementary approach worth exploring. An earlier model that works across modalities is VilBert.

\hypertarget{vision-and-language-bert-vilbert}{%
\subsection{Vision-and-Language Bert (VilBert)}\label{vision-and-language-bert-vilbert}}

As seen in the previous section, data2vec can handle text, image or speech as input data. However, it cannot do so at the same time. The model's focus is on unifying the training approach rather than working across modalities. However, when we think about multimodal models, we usually think of working with different modalities at the same time. VilBert \citep{lu2019vilbert} is a natural extension of the iconic Bert architecture \citep{devlin2018bert} to vision-and-language modelling. An immediate question is whether vision and language inputs should be handled together in a single stream or in parallel. As we will see, it turns out that encoding inputs in parallel and working with parallel streams increases performance. At heart of that architecture is a co-attention mechanism which enables information exchange between both modalities.

\begin{figure}

{\centering \includegraphics[width=1\linewidth]{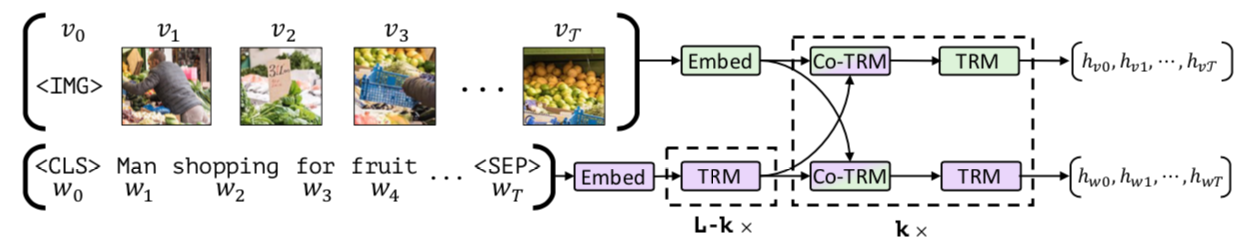}

}

\caption{\citet{lu2019vilbert}: VilBert's Dual Stream Architecture: dashed transformer modules can be repeated, co-attention modules allow sparse interaction between modalities.}\label{fig:vilbertarc}
\end{figure}

Figure \ref{fig:vilbertarc} shows the employed parallel stream architecture. Each modality is handled separately and fed into two Bert-style transformer models. This allows for both modalities to be handled according to their respective needs while co-attention layers allow for communication between the streams. For the language stream, the encoding uses the vocabulary plus a special classification token (cls), a sentence separation token (sep) and a masking token (mask). For the vision stream, image region features are extracted via a Faster R-CNN \citep{ren2015faster} model which was pre-trained on the Visual Genome Dataset \citep{krishnavisualgenome}. Since image regions lack a natural ordering, its spatial location has to be encoded, as well. VilBert achieves that through a five dimensional vector that encapsulates the image coordinates and the fraction of the covered image area. Through projection, the dimensions of the positional encoding and visual features are matched and then summed. The image token marks the beginning of such an image region sequence while representing the whole image.

Through the dual stream architecture, the complexity of the model can be adjusted separately for each modality. An alternative approach would have to discretize the visual space via clustering and then use the resulting tokens in the same way as text tokens. The drawbacks of that approach are the potential loss of detail at the discretization stage and the loss of flexibility across modalities as a result of the same processing. Finally, a single stream architecture can interfere with the pre-training of the language models. The model will have to be fine-tuned based on the created visual tokens. As those might be very different from the text tokens, there is potential for the pre-trained language model to be become `damaged' in the process and lose capabilities - and idea that is also central to the Flamingo model presented later on.

\begin{figure}

{\centering \includegraphics[width=1\linewidth]{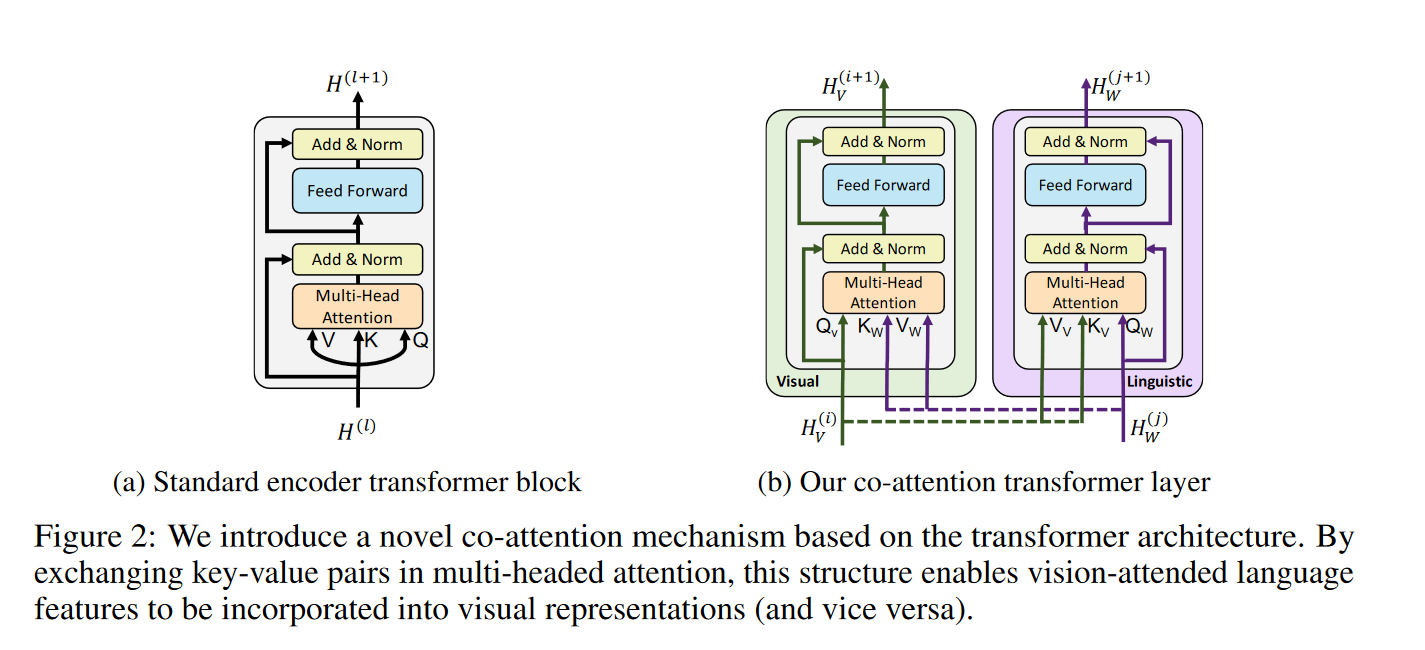}

}

\caption{\citet{lu2019vilbert}: Cross-Attention in VilBert}\label{fig:vilbertattention}
\end{figure}

The key innovation in the Vilbert paper \citep{lu2019vilbert} is the use of co-attention layers. In figure \ref{fig:vilbertattention}, the basic architecture is depicted. The co-attention module computes query, key and value matrices in a standard transformer attention fashion. However, it then feeds the keys and values from each modality into the other modalities multi-head-attention block. As a result, the visual attention will be conditioned on text whereas the language attention will be image-conditioned. This communication between streams only occurs at specific sections in the model, denoted by co-trm in figure \ref{fig:vilbertarc}. Notably, the language stream features a lot more preprocessing before the first co-attention layer than the image stream.

\begin{figure}

{\centering \includegraphics[width=1\linewidth]{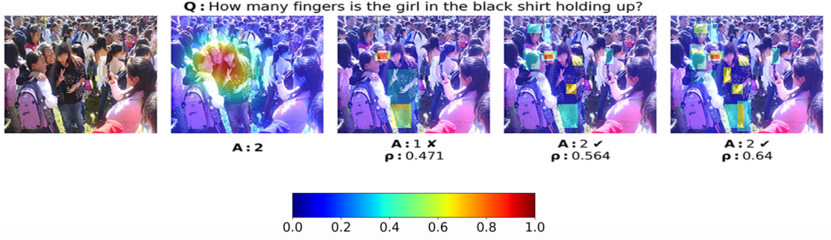}

}

\caption{\citet{sikarwar2022efficacy}: (Left to Right) Picture, Human Attention, 36 Regions, 72 Regions, 108 Regions. Similarity between human and model attention is measured using rank correlation.}\label{fig:vilbertmaps}
\end{figure}

An interesting question to ask is what is actually learned in those attention layers and how they correspond to human attention maps. \citep{sikarwar2022efficacy} analyze the efficacy of co-attention layers for VQA tasks in a VilBert network. Specifically, they compute the question conditioned image attention scores and compare them to human attention maps created in experiments. In those experiments, humans are tasked with unblurring specific image regions to answer the same questions one would expect the machine learning model to answer. Such human attention maps are collected in the VQA-HAT dataset \citep{das2017human}. Rank correlation is used to compare attention maps. \citet{sikarwar2022efficacy} find that in a 6 layer network rank correlation plateaus at layer 4 and increases in the number of image regions proposed while encoding the images. Perhaps more surprisingly, they find a minimal influence of semantics on the generation of the attention maps. Randomly shuffling words in a sentence when testing the model performance barely changes the attention output, which suggests that keywords rather than sentence structures drive the attention output. Note however that while attention maps remained similar, the model's actual performance on answering the questions dropped notably by approximately 15\% such that it seems clear that coherent sentences are important for the overall VQA task, but not for the attention creation process. What are the keyword that drive cross-attention in VilBert? The evidence provided by the authors clearly shows that nouns are the most influential parts-of-speech when considering attention maps. On top of that, prepositions can sometimes help identify spatial relations. There is also some support for the hypothesis that removing Wh-words such as ``who'' and ``where'' can improve fine-grained attention maps in the final layer which might be worth exploring further as preprocessing for deeper networks. Another approach would be to search for ways to improve the way attention maps are generated by finding ways to include more of the available sentence information. Most notably, however, using object-based region proposals to process images can lead to bottlenecks that can prevent the model from learning sufficiently fine-grained attention maps as shown in figure \ref{fig:vilbertmaps}. Overall, humans are naturally good at VQA tasks. Hence, it is not surprising that attention maps which correlate well with human attention maps also improve model performance.

Figure \ref{fig:vilbertmaps} shows that the number of region proposals fed into the model after processing an image affects the ability of the model to produce adequate attention maps. In this particular case the question ``How many fingers is the girl in the black shirt holding up?'' was correctly answered by humans, as well as a VilBert model using 72 or 108 region proposals. It was answered incorrectly when using only 36 region proposals. Note however that in either case, the machine learning model captured the face of the wrong girl. The model using 72 regions also identified the wrong hand despite answering the question correctly. While the 108 region model identifies the correct hand holding up the fingers, it does not seem to prioritize it over the other identified hands in the picture. Hence, the attention maps are sufficiently different from the human attention map which highlights the need to look closer not only at how models are performing, but also into how their performance has been achieved.

As far as the model training is concerned, VilBert is pre-trained and fine-tuned. The pre-training tasks comprise masked-multi-modal modelling and multi-modal alignment prediction performed on the Conceptual Captions dataset. That dataset contains about 3,1 million usable aligned image-caption pairs, which have been automatically scraped from web images. For the alignment task, the authors create unaligned images by randomly mismatching captions and images. For the masking task, 15\% of the both the visual and language tokens are masked. The task is to reconstruct the mask from the remaining input in a classical Bert fashion. While the text masks are directly regressed like in Bert, the model predicts distributions over semantic classes for the image regions. This is achieved through minimizing the KL divergence, a measure for the similarity of distributions, between the output distribution of the pre-trained model used in feature extraction and the VilBert predictions.

\begin{figure}

{\centering \includegraphics[width=1\linewidth]{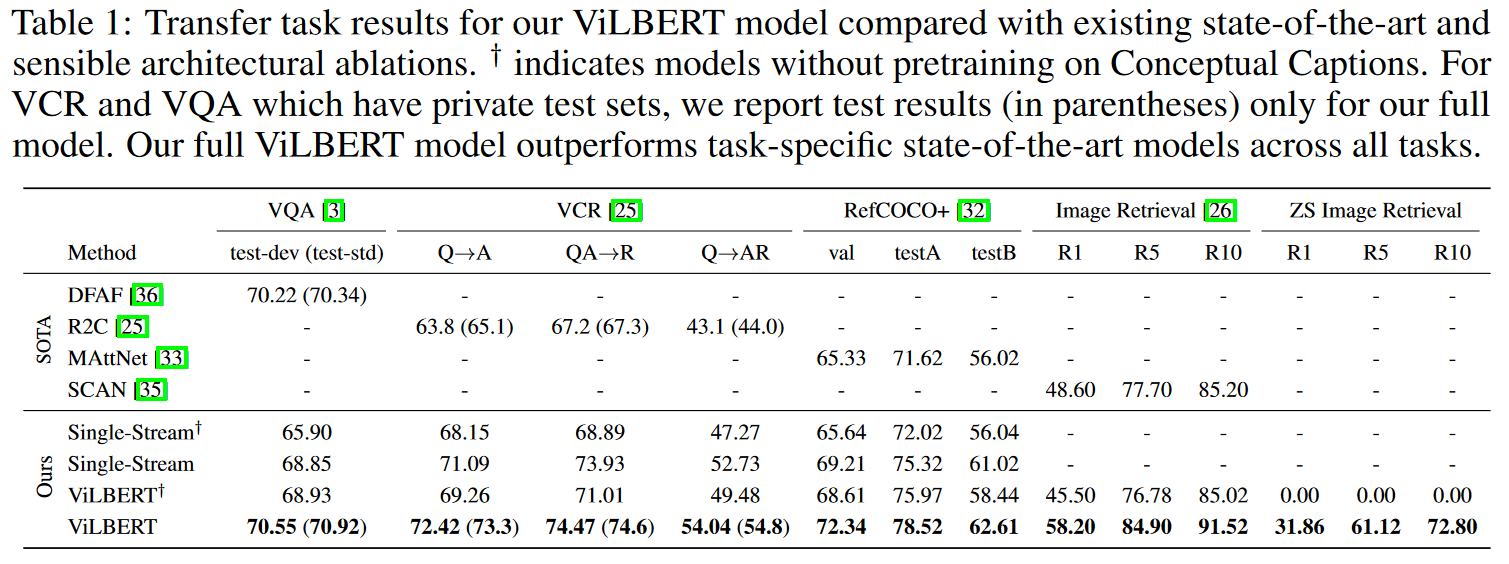}

}

\caption{\citet{lu2019vilbert}: VilBert Performance}\label{fig:vilbertresults}
\end{figure}

The performance results are depicted in figure \ref{fig:vilbertresults}.

As mentioned before, the dual stream architecture outperforms the single stream architecture. Furthermore, pre-training considerably boosts performance, as does fine-tuning. Interestingly, the authors also study the effect of the size of the dataset and effect of the architecture depth. Performance increases monotonically with dataset size, suggesting that performance can be further improved with more data. The results on the optimal layer depth are task dependent. VQA and Image Retrieval reach peak performance at 6 layers, where a layer denotes a repeatable block as depicted in figure \ref{fig:vilbertarc}. Zero Shot Image retrieval greatly benefits from even deeper depth. However, the VCR and RefCOCO+ tasks seemingly benefit from shallower models. The VQA task is based on the VQA 2.0 dataset. Each image must be matched to one of ten answers. Hence, the VQA task is not open-ended, but treated like a classification task. To achieve that, the model is amended by two MLP layers which use the element-wise product of the model-generated img and cls tokens. The VCR task is also posed as a multiple choice problem with images from movie scenes. To fine-tune for the task, questions and answers are concatenated into four different text input and given as model input together with the image. In the end, four scores are generated accordingly and selected through softmax. The RefCoCO+ task is a grounding task. An image region has to be selected according to a natural language reference. Caption-Based Image Retrieval requires the model to find an image that corresponds to a selected caption. The dataset used is the Flickr30k dataset which contains 30 000 pictures with five captions that are of higher quality than the automatically generated captions from web data.

\hypertarget{flamingo}{%
\subsection{Flamingo}\label{flamingo}}

The VilBert model showed one way how to actually combine visual and language inputs. In contrast, data2vec showed how to design an unsupervised model and how influential the actual training process as well as contextualization can be. A natural question to ask is then is whether we can build a truly multimodal architecture like VilBert that is self-supervised like data2vec or at little task-specific training and how to optimized its training procedure. In particular, both VilBert and data2vec were tested on multiple tasks, but each task needs slight re-adjustments to the model as well as additional fine-tuning. Ideally, a multimodal architecture would not only be efficient in its initial training, but also easily adaptable to different tasks. Finding ways to not only work with different input modalities, but also with different task is crucial towards building a more general AI. A promising approach in that direction is few shot learning. The following section presents Flamingo \citep{alayrac2022flamingo}, a few shot multimodal architecture developed by Google which comprises key innovations such as handling arbitrarily interleaved vislang sequences as inputs, as well as ways to effectively combine pre-trained vision-only and language-only models. As such, it is a visually conditioned autoregressive text generation model.

\begin{figure}

{\centering \includegraphics[width=1\linewidth]{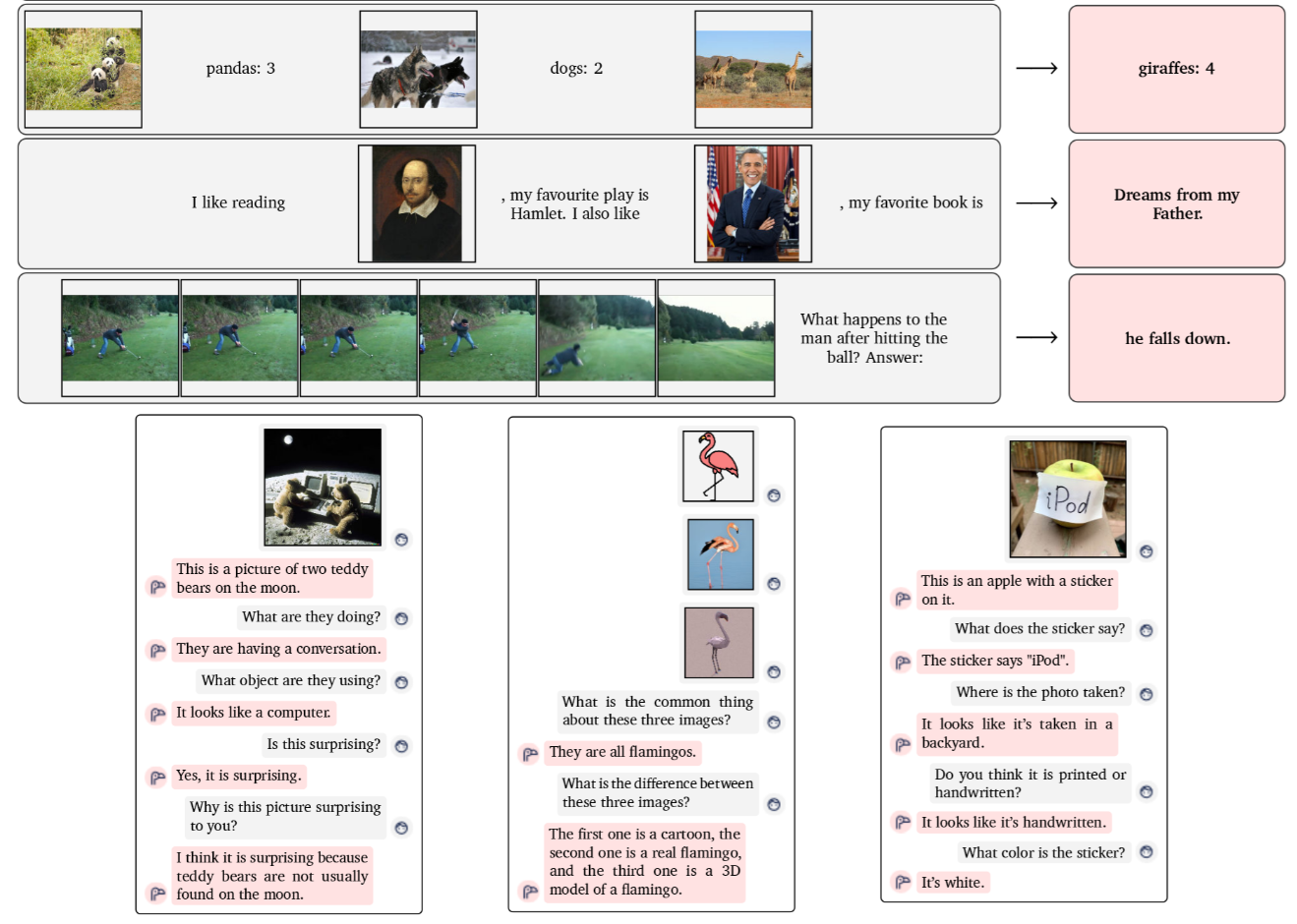}

}

\caption{\citet{alayrac2022flamingo}: Flamingo Prompt-Output-Examples}\label{fig:flamingoexamples}
\end{figure}

Figure \ref{fig:flamingoexamples} demonstrates Flamingos capabilities. It can function as chat bot, describe pictures, work with image sequences (videos) and in doing so, simply needs a few prompts.

At the heart of the model is a large language model, Chinchilla \citep{hoffmann2022training}, with 70B parameters. Large language models such as GPT-3 \citep{brown2020language}, as their name suggests, can be trained on a large amount of text data which gives them impressive text generative capabilities. However, multimodal generative modelling presents some specific challenges not present in language-only modelling. First of all, training large language models is expensive. Hence, it is paramount to work with a pre-trained version, but trying to teach a large language model the means to work with visual inputs, as well, has the potential to deteriorate or destabilize the pre-trained model. Second, large language models can suffer from memory constraints that are potentially severely aggravated by simply adding high-dimensional visual data into an input sequence. Third, good generalist capabilities typically require a huge amount of heterogeneous training data. There might not exist enough labelled image-caption-pair data to successfully accomplish training a capable few shot learning model in the vision-and-language domain. To train Flamingo, the authors solve these challenges by foremost exploring ways to generate their own web-scraped multimodal data set similar to existing ones in the language-only domain. Furthermore, they use a perceiver architecture \citep{jaegle2021perceiver} that resamples inputs into a fixed amount of visual tokens. Finally, the self-attention layers of the language model are kept frozen during training while cross-attention layers are interleaved. A gating mechanism ensures that those new cross-attention layers do not interfere at model initialization, thereby improving stability and final performance.

\begin{figure}

{\centering \includegraphics[width=1\linewidth]{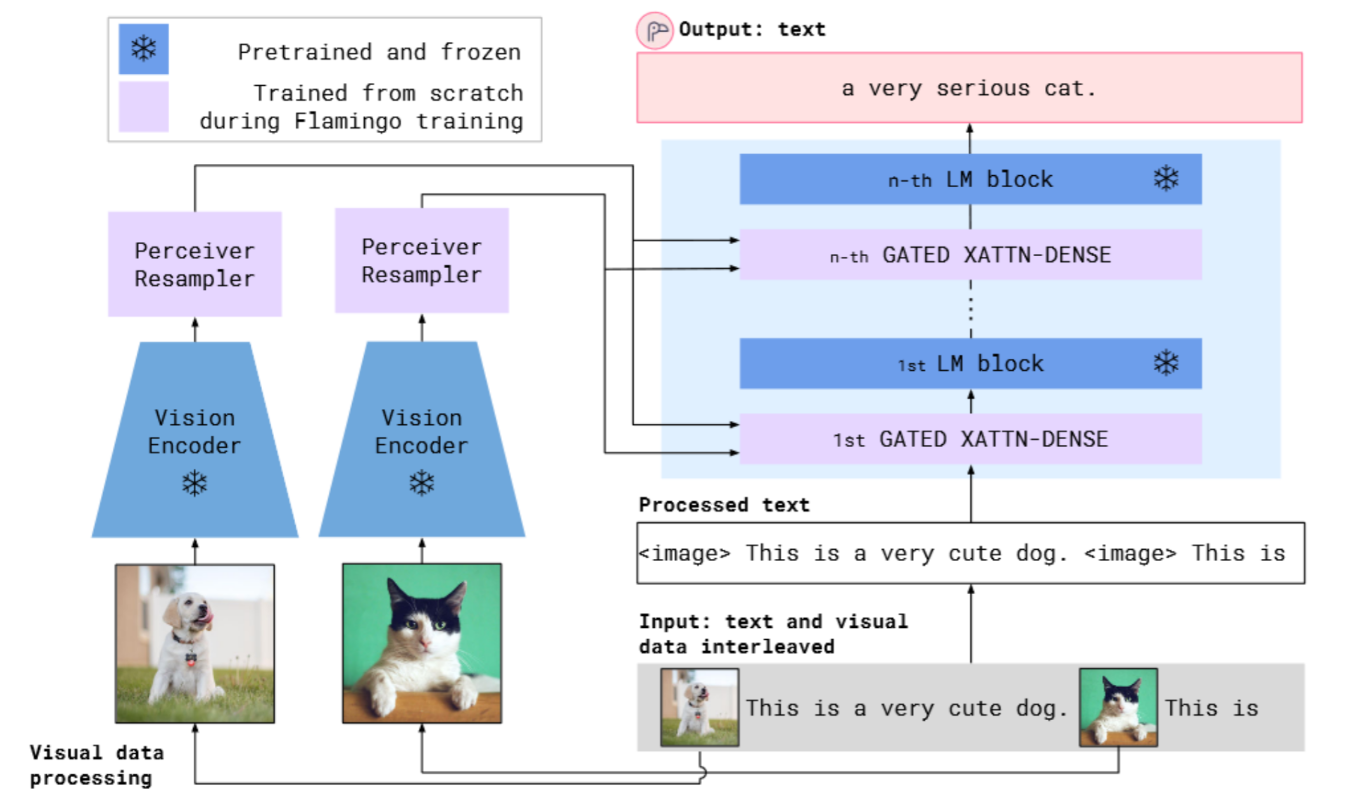}

}

\caption{\citet{alayrac2022flamingo}: Flamingo Model Structure}\label{fig:flamingoarc}
\end{figure}

Figure \ref{fig:flamingoarc} shows the fundamental architecture of Flamingo. A pre-trained vision model as well as a pre-trained language model are frozen. Together they built the cornerstones of the model. The vision model is pre-trained using a contrastive text-image approach. Its role is to extract features such as colour, shape, nature and the position of objects - typical semantic spatial features that one would use in querying. The language model is an existing pre-trained language model. On top of those frozen parts, the authors add a perceiver-resampler and gated cross-attention layers as learnable architectures. The perceiver-resampler turns the outputs of the vision model into a fix set of visual tokens. Those visual tokens are then used to create cross-attention layers which are interleaved into the frozen language model. As a result, Flamingo can model the likelihood of some text y interleaved with a sequence of images or videos x as

\[p(y|x) = \prod_{l=1}^{L} p(y_l | y_{<l}, x_{<l}).\]

Here, \(y_l\) denotes the language token associated with the input text and \((y,x)_{<l}\) is the set of preceding tokens. Parameterized by the model is \(p\). As shown in the initial figure, one conditions the Flamingo model simply by giving it an alternating image and text sequence. This is because the attention mechanism only allows text tokens to attend to the previous image, which turned out to work better than alternative schemes. In particular, this means that the model can generalize to an arbitrary amount of images regardless of the amount of images used in training.

The training was based on multiple different datasets. The most important one is a Multimodal MassiveWeb datatset (M3M). To generate it, the authors collect images and text from HTML of approximately 43 million webpages. In the process, the position of images relative to the surrounding text is also identified. Image tags (image) added in plain text signal the original location of images to the model. In addition to that, end of chunk (EoC) tokens before images separate image-text sequences. The embedding of that token is added to the vocabulary of the language model with a random initialization that can later be learnt. Then, it is possible to infer the length of an image-text sequence, which is another piece of derived information. To give an idea of the scope of the dataset, M3M contains about 182GB of text as well as roughly 185 million images. The authors pay special attention not to include traditional task-specific datasets curated particularly for machine learning purposes to guarantee the generality of their modelling approach. As a second important dataset, aligned image text pairs are used. In particular, the ALIGN dataset \citep{jia2021scaling}. The dataset is further augmented with Long Text and Image Pairs (LTIP) as well as Video and Text Pairs (VTP). The later datasets contain more descriptive captions than ALIGN. Together the process ensures that the available training datasets are sufficiently large and heterogeneous - two key properties necessary to hopefully achieve good few shot performance.

The training objective is to minimize the weighted sum of dataset specific expected negative log likelihood.

\begin{center}\includegraphics[width=0.5\linewidth]{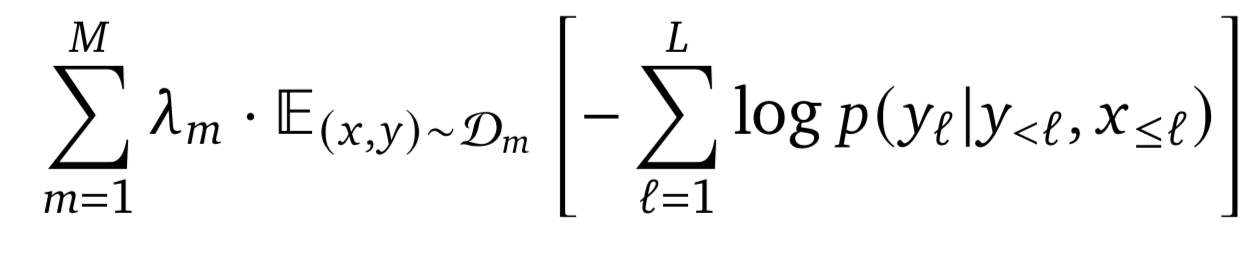} \end{center}

Each dataset is weighted with a scalar \(\lambda\) as datasets can be of different quality or feature different properties. Hence, it might be preferable to pay different attention to different datasets. According to the authors, tuning these weights was essential to overall performance. In practice, the optimization works as follows: a sample batch with visual language sequences from each dataset is used to compute the gradient of the loss in accordance to the weight of the dataset. Importantly, the authors find that it is beneficial to accumulate the gradients of all datasets before triggering an updating process. Naturally, the actual datasets used to train the models are extremely crucial, as well. In their ablation studies, the authors find that removing their web-scraped multimodal dataset from the training pool drops model performance as measured across all selected tasks from a score of 68.4 to 46.9. Removing the dataset containing aligned captions and images drops performance to a score of 56.5 and not accumulating gradients before the updating process decreases performance to 59.7.

\begin{figure}

{\centering \includegraphics[width=0.5\linewidth]{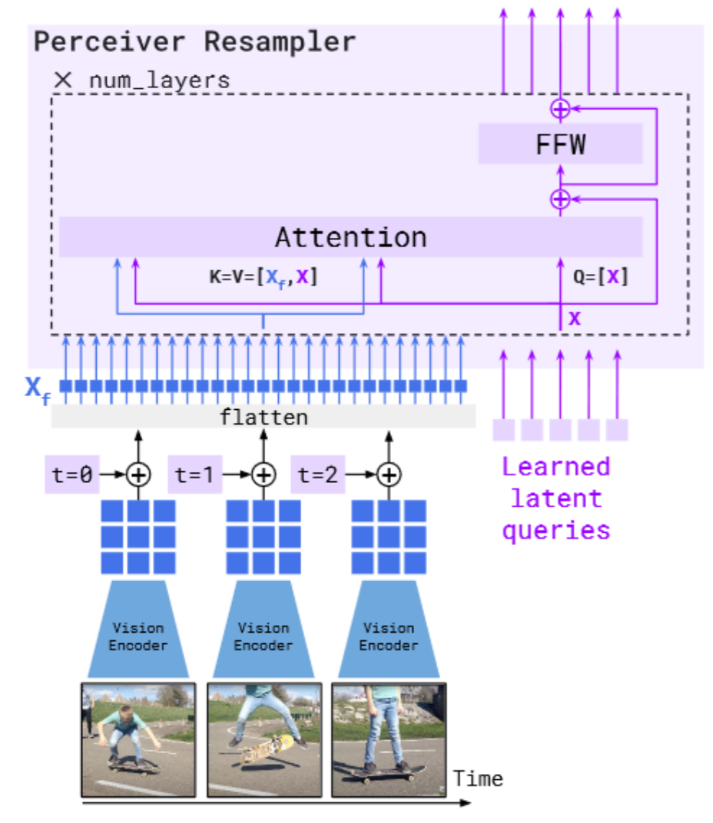}

}

\caption{\citet{alayrac2022flamingo}: Flamingo Perceiver-Resampler}\label{fig:perceiver}
\end{figure}

Taking a closer look at the model architecture, the two key structures are the perceiver resampler and the attention layers. Figure \ref{fig:perceiver} shows the architecture of the perceiver \citep{jaegle2021perceiver}. Before the input data reaches the perceiver, it is processed by the vision encoder - a Normalizer-Free ResNet which is trained with a constrastive loss similar to the well known Clip model \citep{radford2021learning} and yields a good trade-off between performance and efficiency. The output of the vision encoder is a 2D grid which is than flatted before being fed into the perceiver that connects the vision encoder with the frozen language model. The resampling performed by the perceiver-resampler is crucial to reduce the complexity of vision-text cross-attention in the next step. This is particularly notable for video inputs. Inside the perceiver, a set of learned latent queries cross attend to the flattened vision encoder output. The number of outputs generated by the perceiver is equal to the number of learned latent queries. A change the authors make compared to previous work is to concatenate the keys and values from the latent queries with the keys and values from the flattened features. The ablation studies show that a medium sized perceiver architecture works best with improvements around two to three score points. Furthermore, a too large architecture can lead to unstable trainings in conjunction with large frozen language model. The authors also test the perceiver against a transformer or MLP, which showed the perceiver to improve performance scores by around three to five points.

\begin{figure}

{\centering \includegraphics[width=1\linewidth]{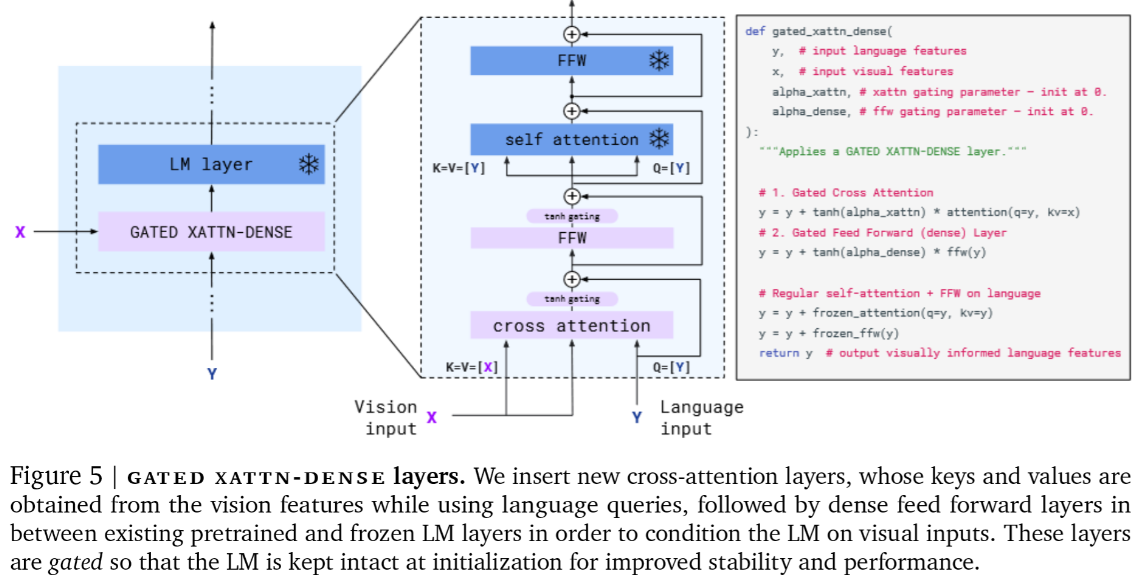}

}

\caption{\citet{alayrac2022flamingo}: Flamingo Gated Cross-Attention}\label{fig:flamingoattention}
\end{figure}

The main component that turns the frozen large language model into a functioning visual language model are the cross-attention layers depicted in figure \ref{fig:flamingoattention}. The number of layers added controls the number of free parameters and hence the complexity and expressiveness of the model. Keys and values of those layers are obtained from the visual features output by the perceiver while using language queries. Specifically, gated cross-attention dense layers are used. The layers are dense because the cross-attention layer is followed by a feed forward layer. They are gated because a \(\tanh(\alpha)\) gating mechanism is applied between layers. The gating mechanism ensures that the frozen large language model remains stable at initialization through introducing a learnable scalar parameter \(\alpha\) initialized at 0. Without that initialization, training instabilities can occur. A key question is how many cross-attention layers should be used. For the small Flamingo model with 3B parameters, the ablation studies show that adding cross-attention between every self-attention layer of the frozen model yields the best results. However, adding further cross-attention layers does notably scale the parameter count of the model. A clear performance trade-off exists. After making hardware considerations, the authors settled for adding \(\tanh(\alpha)\) gated cross-attention layers every 7th layer in the frozen large language model. The practical implementation of those attention layers works as follows: recall that the authors found that attending only to the nearest image improves performance by approximately 8 score points. To achieve this, while all text tokens attend to all visual tokens, a masking strategy is applied which ensures that in effect, language tokens only see a specific amount of visual tokens. Note however, that while the model can, unless specified otherwise, only attend to one image at a time, there is still a causal dependency to all previous images through the self-attention in the text-decoder.

\begin{figure}

{\centering \includegraphics[width=1\linewidth]{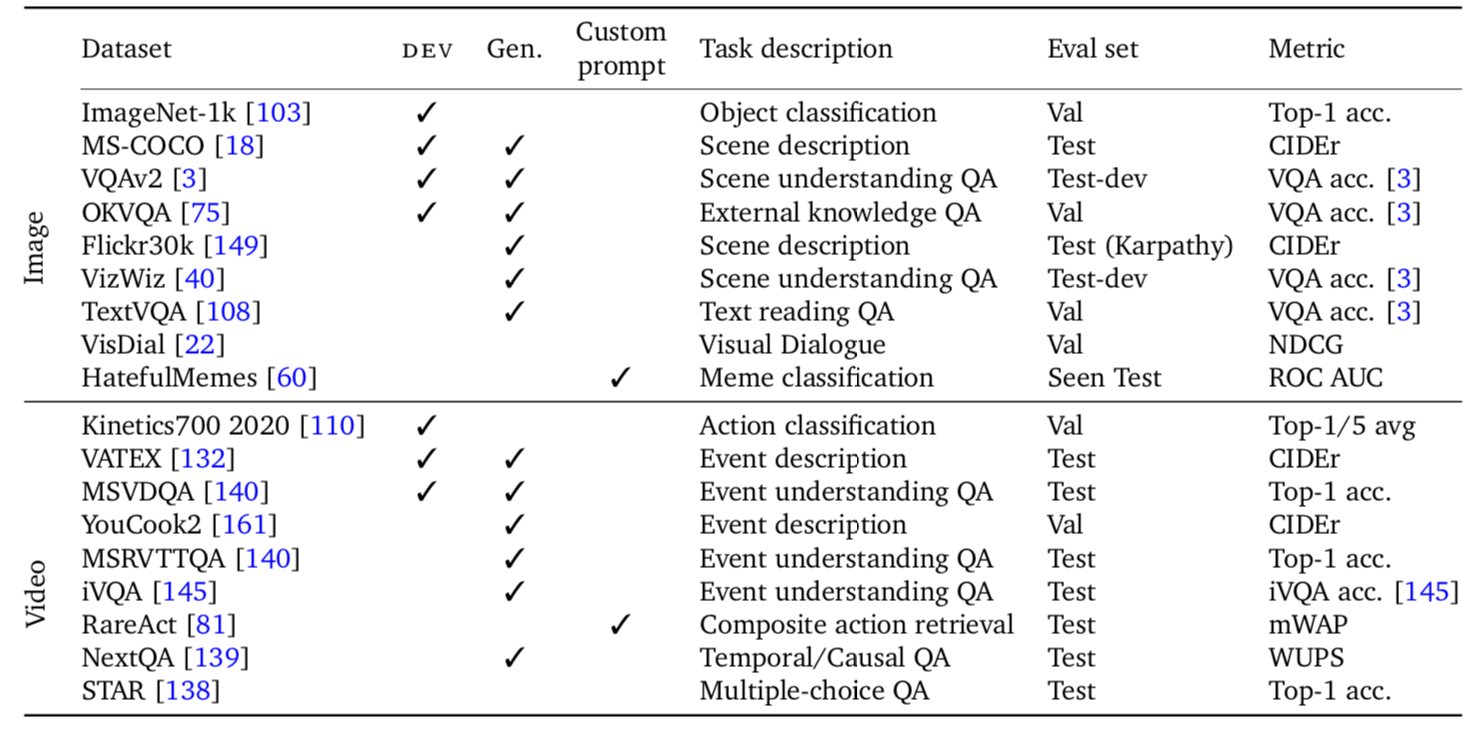}

}

\caption{\citet{alayrac2022flamingo}: Flamingo Datasets (Table2, p.19)}\label{fig:flamingodatasets}
\end{figure}

To evaluate the model, the authors chose 18 different vison-language benchmarks including video benchmarks as shown in \ref{fig:flamingodatasets}.

Note that seven benchmarks are used to validate design choices of the architecture. They are part of the development (dev) set. As those datasets could potentially report biased performance results, the remaining eleven datasets are solely used to estimate unbiased performance scores. Unfortunately, unbiased estimation in few-shot learning is not ubiquitous. Since hyperpameter tuning requires more prompts, it is easy to forget about them when counting how many shots in effect have been used, which can in turn lead to overestimation of performance \citep{perez2021true}. As far as the Flamingo model is concerned, the authors take great care to evaluate it in a true few-shot fashion as coined by \citet{perez2021true}. Furthermore, most of the tasks require a generative answer (gen) which encompasses open-ended, more interesting tasks.

\begin{figure}

{\centering \includegraphics[width=1\linewidth]{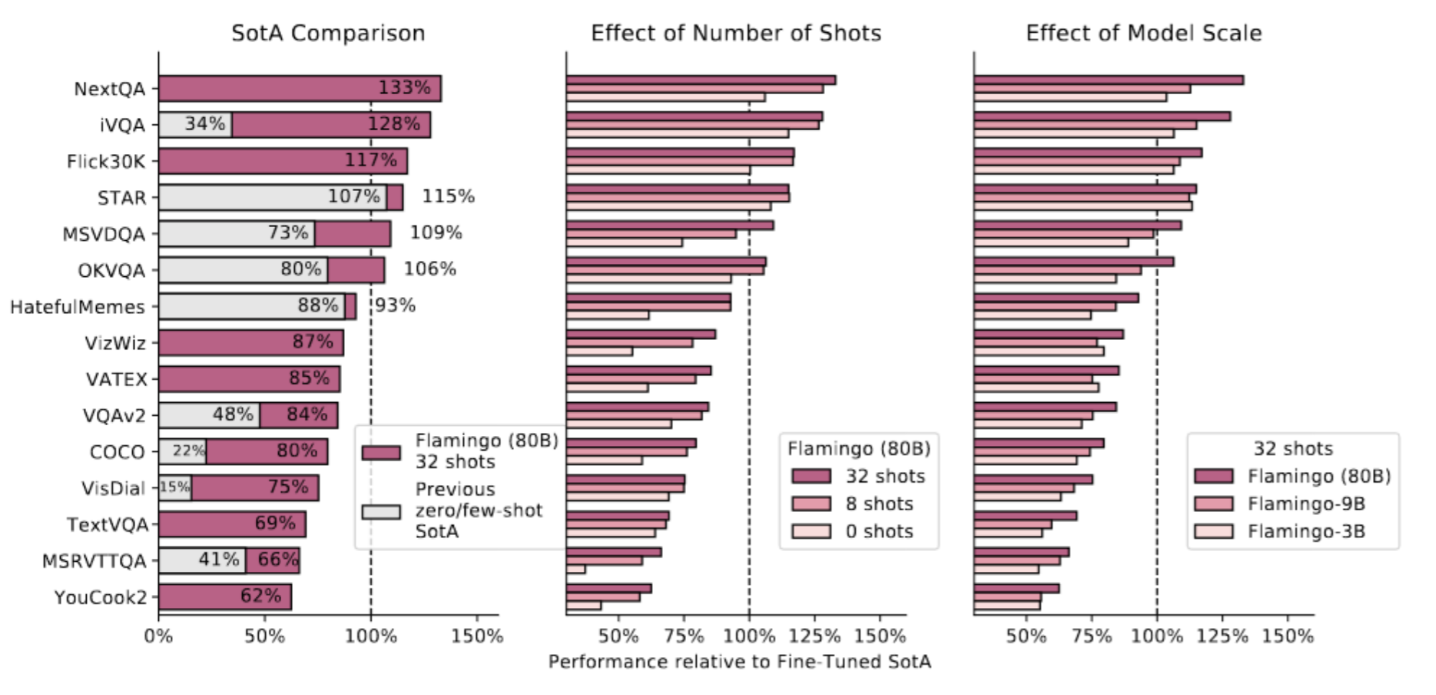}

}

\caption{\citet{alayrac2022flamingo}: Flamingo Results without Fine-Tuning}\label{fig:flamingoresult}
\end{figure}

The results portayed in figure \ref{fig:flamingoresult} show that Flamingo does not only outperform all previous few shot models, but also the general state-of-the art on six tasks. It is also, not too surprisingly, evident that more shots and larger models lead to better performance. The in-context learning works analogous to GPT-3 \citep{brown2020language}. Given a set of supporting examples (image, text), where text is the expected response based on the supporting visual input, a multimodal prompt is built by concatenating the examples in random order and adding the selected query image for the prediction. Interestingly, rather than using in-context learning with prompts, the authors also explore fine-tuning the model on the tasks which achieved state-of-the-art performance with few-shot learning. Fine-tuning the model is very expensive and requires additional hyperparameter tuning, but substantially improves results even further.

\begin{figure}

{\centering \includegraphics[width=1\linewidth]{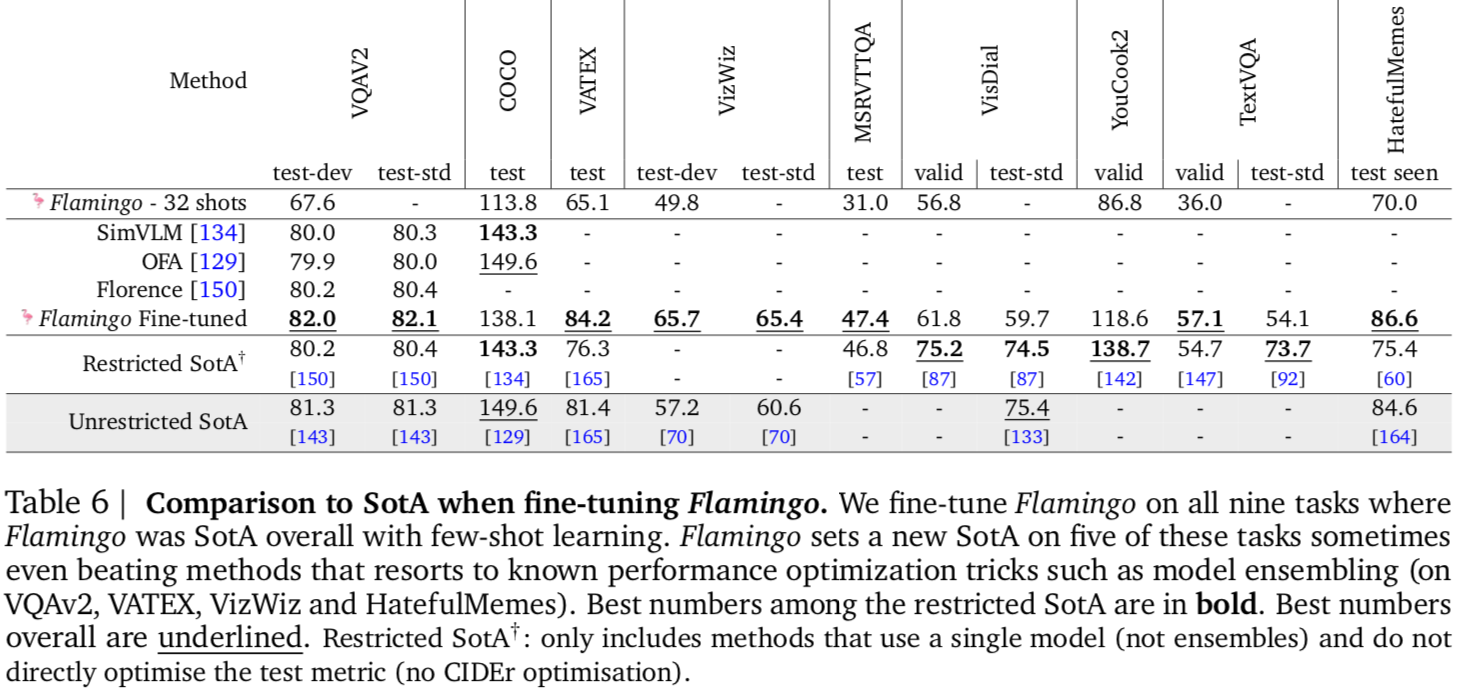}

}

\caption{\citet{alayrac2022flamingo}: Flamingo Results with Fine-Tuning}\label{fig:flamingfinetune}
\end{figure}

One notable exception that the authors remark upon is the classification performance of the model. In that realm, contrastive models outperform Flamingo. A working hypothesis is that training for text-image retrieval is particularly important on those tasks. Nevertheless, few-shot learning and open-ended generative capabilities provide great advantages over contrastive models both in terms of model training as well as the range of applications. Flamingo shows how to leverage a unifying structure between models. Communication between models is key to reducing training overload and to enable multitasking models.

\hypertarget{discussion-2}{%
\subsection{Discussion}\label{discussion-2}}

In the previous sections, we've analyzed three different models with a common goal, to work towards a more general model capable of succeeding on multiple tasks across multiple modalities, specifically image and text. Which lessons can be learned from that?

Context is key. Across all three models, it became apparent that \textbf{larger models have an edge}. A clever architecture can of course lead to great results - after all Flamingo outperformed many fined-tuned models, but ceteris paribus, larger models will deliver stronger performance. While Flamingo with 80B parameters, mostly made up by its large language model component, was far larger than data2vec or VilBert, it is also far from the largest model in existence, as the following chapter will show. More tasks, more generalizability also means larger models.

Larger models in turn require either incredible resources or clever designs to be trained - likely both. It is not without reason that the three models presented have been developed by Meta and Google. At the high end, the amount of people with access to the necessary resources is naturally limited and there is little reason to expect change. On top of resources constraints limiting access to mostly a few private companies, resulting models are often also simply not publicly available to shield intellectual property from competitors. For example, while the codebase for the somewhat older \href{https://github.com/facebookresearch/vilbert-multi-task}{VilBert} model is publicly available, data2vec and Flamingo are not generally accessible.

At any rate, even for the companies with the greatest resources, \textbf{a performance-cost trade-off exists}. The question of how to cut down on required training time is essential. The general approach is to pre-train models and then fine-tune them on specific tasks. However, \textbf{few shot in-context learning} provides a resource friendly alternative although fine-tuning likely still leads to better absolute results. \textbf{Freezing models}, particularly large language models, \textbf{is a key idea} on top of pre-training. In some cases, it is paramount to avoid the loss of capabilities that can go along with retraining a model. This could already be seen when VilBerts dual-stream architecture outperformed a single-stream design, but becomes more notable in the Flamingo architecture, where retaining the full expressiveness of the large language model is key which prompted the authors to introduce a gating mechanism into the cross-attention layers to stabilize the training process. In general, model collapse is always a concern, in particular when working with latent representations such as data2vec. In essence, rather than building single models from scratch, reusing models and leveraging communication between models is a new, promising approach. In that regard, \href{https://socraticmodels.github.io/}{Socratic models} \citep{zeng2022socratic} also show that the knowledge stored in different models is symbiotic which they used for exciting tasks such as multimodal assisted dialogue with people or robot perception. Finally, \textbf{data matters}. Not only is the amount of data important, but also its composition. Heterogeneous data are important, but so is the optimization across datasets. The Flamingo model was specifically trained with a weighted loss across datasets and it was possible to quantify the performance contribution of each of them. Particularly in few shot learning settings, it is thereby important to be careful about unbiased performance estimation as \citet{perez2021true} noted. Otherwise, it is easy to overestimate performance.

In any case, the quest towards more general, more unified models is far from completed. The common theme is to combine larger and larger models while employing resource friendly training regimes. For example \href{https://blog.google/technology/ai/introducing-pathways-next-generation-ai-architecture/}{Pathway} models \citep{chowdhery2022palm}, which will be further discussed in the upcoming chapter, use sparse activation which reduces the energy consumption to less than 10\% of what would be expected from similar dense models.

\hypertarget{c03-00-further}{%
\chapter{Further Topics}\label{c03-00-further}}

\emph{Authors: Marco Moldovan, Rickmer Schulte, Philipp Koch}

\emph{Supervisor: Rasmus Hvingelby}

So far we have learned about multimodal models for text and 2D images. Text and images can be seen as merely snapshots of the sensory stimulus that we humans perceive constantly. If we view the research field of multimodal deep learning as a means to approach human-level capabilities of perceiving and processing real-world signals then we have to consider lots of other modalities in a trainable model other than textual representation of language or static images. Besides introducing further modalities that are frequently encountered in multi-modal deep learning, the following chapter will also aim to bridge the gap between the two fundamental sources of data, namely structured and unstructured data. Investigating modeling approaches from both classical statistics and more recent deep learning we will examine the strengths and weaknesses of those and will discover that a combination of both may be a promising path for future research. Going from multiple modalities to multiple tasks, the last section will then broaden our view of multi-modal deep learning by examining multi-purpose modals. Discussing cutting-edge research topics such as the newly proposed Pathways, we will discuss current achievements and limitations of the new modeling that might lead our way towards the ultimate goal of AGI in multi-modal deep learning.

\hypertarget{c03-01-further-modalities}{%
\section{Including Further Modalities}\label{c03-01-further-modalities}}

\emph{Author: Marco Moldovan}

\emph{Supervisor: Rasmus Hvingelby}

Over the course of the previous chapters, we have introduced the basics of computer vision (CV) and natural language processing (NLP), after that we have learned about several directions of how we can combine these two subfields in machine learning. In the most general sense, we have explored ways in which we can process more than just one modality with our machine learning models.

So far, we have established the basics of multimodal deep learning by examining the intersection of these two most well-understood subfields of deep learning. These fields provide us with easy-to-handle data as seen in the corresponding previous chapter as well as a plethora of established and thoroughly examined models.

In reality though, text and images can be seen as only discrete snapshots of our continuous highly multimodal world. While text and images serve as an important foundation for us to develop concepts and algorithms for multimodal learning, they only represent a small part of what we as humans can perceive. First and foremost, we perceive reality in a temporal direction too, for a machine this could mean receiving video as input instead of just still images \citep{iv2021multimodal}. In fact, as videos are some of the most abundant types of data, we will later see that self-supervised learning on raw video is one of the major subtasks of multimodal deep learning. Clearly our reality is not just a sequence of RGB images though: just like in most videos we experience sound and speech which we would also like our models to process. Furthermore, we have different senses that can perceive depth, temperature, smell, touch, and balance among others. We also have sensors that can detect these signals and translate them to a digital signal so it is reasonable to want to have a machine learning algorithm detect and understand the underlying structure of these sensory inputs as well.

Now it might be tentative to simply list all different types of signals that we have developed sensors for and give a few examples of a state of the art (SOTA) deep neural network for each that tops some arbitrary benchmark. Since we are talking about multimodal learning, we would also have to talk about how these different modalities can be combined, and what the current SOTA research is, on all of these permutations of modalities. Quickly we would see that this list would get extremely convoluted and that we would not see the end of it. Instead of basing our understanding simply on a list of modalities we need a different, more intuitive system that lets us understand the multimodal research landscape. In the first part of this chapter we will attempt to introduce such a taxonomy based on challenges rather than modalities \citep{baltrušaitis2017multimodal}.

If we consider multimodal deep learning as the task to learn models that can perceive our continuous reality just as precisely (if not more) than us humans \citep{lecun2022path}, we have to ask ourselves how we can generalize our learnings from image-text multimodal learning to more types of signals. We have to ask what constitutes a different type of signal for us versus for a machine. What types of representation spaces we can learn if we are faced with having to process different signal types (modalities) and what are the strategies to learn these representation spaces. Here we will see that in large we can have two ways of processing modalities together, where defining their togetherness during training and inference will play the central role. After formalizing the types of multimodal representation learning we will move on and elaborate what the fundamental strategies are that allow is to learn these representation spaces. Then again, we can ask what we can practically do with these representation spaces: Here the notion of sampling and retrieving from our learnt representation spaces will play a major role. In fact we will see that almost all practical multimodal tasks can be generalized to what we call multimodal translation, where given a signal in one modality we want to return a semantically related signal in another modality.

The ideas that were just introduced are in fact what we consider to be the central challenges of multimodal learning, these challenges constitute the main pillars of our taxonomy of multimodal deep learning. Every problem in multimodal learning will have to solve at least one of these challenges. By viewing multimodal deep learning through these lens we can easily come across a new modality and understand immediately how to approach this problem without breaking our taxonomy.

After understanding these challenges the reader will hopefully take home a new way of thinking about how to solve and understand multimodal problems. Hopefully, when coming across a new research paper and tackling a new research project the reader will identify the challenges that the paper is trying to solve or which challenge requires solving for the research project and immediately know where to look.

Looking at the broader spectrum of the AI research landscape, as Yann LeCun has done in his recent paper \citep{lecun2022path}, we can see that multimodal perception through deep learning is one particularly important building block for creating autonomous agents capable of displaying reason.

After having thoroughly introduced these central multimodal learning challenges we will look at some of the current research trends of multimodal deep learning from the point of view of our challenge taxonomy. In order to solve these challenges a system must implement two major building blocks: a multimodal model architecture and a training paradigm. In this part of the chapter we will introduce examples for both and successively generalize these concepts. By introducing more and more universal and problem- as well as modality-agnostic systems from current research we will lead into a research project that we ourselves are undertaking to merge a general multimodal model with a problem-agnostic training paradigm which will form the conclusion of this chapter. Hopefully by then two major concepts have transpired: 1) Introduce models and training paradigms that are general enough as to give a conclusion to this chapter's very title: learning from any and including an arbitrary amount of further modalities in our learner and 2) sticking to the analogy of the human perceptive system and presenting models and training paradigms that can learn from any type of input signal just like we humans can. In the spirit of Yann LeCun's JEPA paper the perceptive aspect of artificial intelligence is only one aspect of the system. Looking at the broader spectrum of the AI research landscape -- as Yann LeCun has done in his recent paper, we can identify that multimodal perception through deep learning is one particularly important building block for creating autonomous agents capable of displaying reason. Other aspects such as reasoning and especially multi-tasking and scaling will be elaborated in {[}this{]} following chapter.

\hypertarget{taxonomy-of-multimodal-challenges}{%
\subsection{Taxonomy of Multimodal Challenges}\label{taxonomy-of-multimodal-challenges}}

In this part we will introduce a taxonomy based on challenges within multimodal learning \citep{baltrušaitis2017multimodal}.

\hypertarget{multimodal-representation-learning}{%
\subsubsection{Multimodal Representation Learning}\label{multimodal-representation-learning}}

At the core of most deep learning problems lies representation learning: learning an expressive vector space of distributed embedding vectors in which we can define a distance function that informs us about the semantic relatedness of two data points in this learnt vector space. For the sake of simplicity, we will assume that these vector spaces are learnt via deep neural networks trained with backpropagation. Normally we will have to apply some preprocessing to our raw data in order to transform it into a format that a neural network can read, usually in the form of a 2-dimensional matrix. As output the neural network will return some high-dimensional vector. But what if we are presented with more than one signal type (i.e., multimodal input)? How do we structure our input so that our models can sensibly learn from this multimodal input?

\begin{figure}

{\centering \includegraphics[width=1\linewidth]{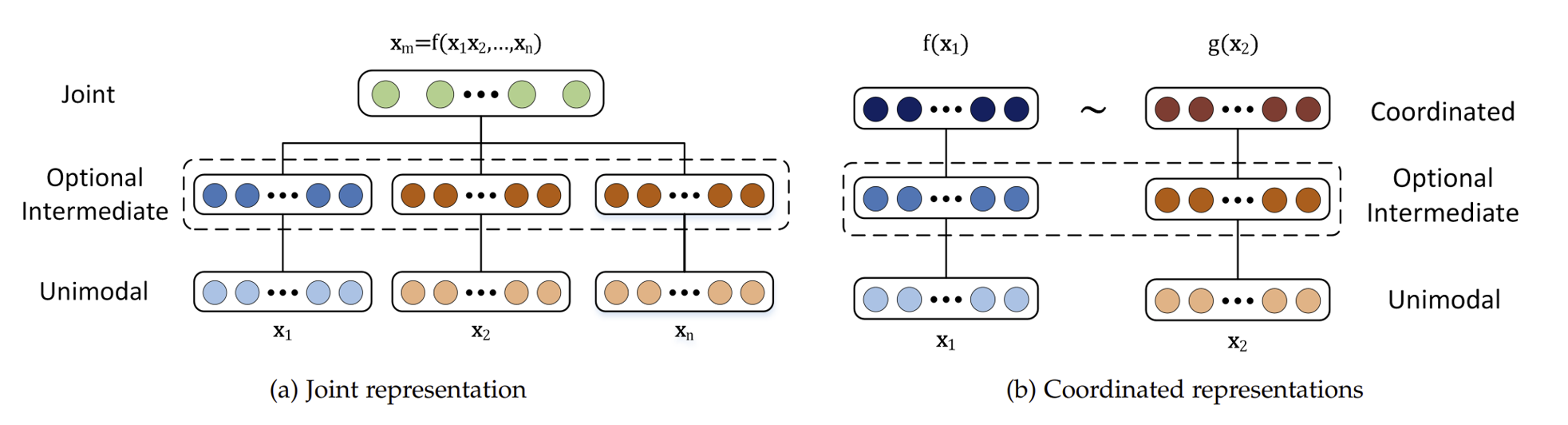}

}

\caption{Joint and coordinated multimodal representations[@baltrušaitis2017multimodal].}\label{fig:joint-coordinated}
\end{figure}

In the introduction for this chapter, we briefly mentioned the togetherness of multimodal signals during training and inference \citep{DBLP:journals/pami/BengioCV13}. By virtue of having more than one modality present as input into our learner -- whether it be during training or inference -- we want to relate these modalities somehow, this is the essence of multimodal learning. If we consider that our input signals from different modalities are somehow semantically related, we would like to leverage this relatedness across modalities and either have our learner share information between modalities and leverage this relatedness. Therefore cross-modal information has to come together at some point in our training and/or inference pipeline. How and when this happens is the central question of multimodal representation learning which we describe in this subchapter.

First, we have to specify that what is meant by their togetherness during training and inference. Togetherness loosely means that unside our learner we ``merge'' the information of the modalities.

To make this more concrete: on one side we could think of concatenating the input from different modalities together to form one single input matrix. This joint input then represents a new entity that consists of multiple modalities but is treated as one coherent input. The model then learns one representation for the joint multimodal signal. On the other hand, we could think of the input always as strictly unimodal for one specific model. Each model would be trained on one modality and then the different modalities are brought together only in the loss function in such a way as to relate semantically similar inputs across modalities. To formalize what we just introduced, joint representation learning refers to projecting a concatenated multimodal input into one representation space while coordinated representation learning will learn different representation spaces for each modality and coordinate them such that we can sensibly align these representation spaces and apply a common distance function that can relate points across modalities to each other.

\hypertarget{joint-representations}{%
\paragraph{Joint Representations}\label{joint-representations}}

Given for example a video that consist of a stream of RGB images and a stream of audio signals as a waveform we would like our model to learn a representation of this whole input video as how it appears ``in the wild.'' Considering the entirety of the available input means that our model could leverage cross-modal information flow to learn better representations for our data: this means the model learns to relate elements from one modality to elements of the other. Of course, one could imagine concatenating all sorts of modalities together to feed into a model, such as audio and text, RGB image and depth maps, or text and semantic maps. The underlying assumption simply has to be that there is something to relate between the modalities -- in other words there has to be a sensible semantic relationship between the modalities.

\hypertarget{coordinated-representation}{%
\paragraph{Coordinated Representation}\label{coordinated-representation}}

When we are given data in multiple modalities, for learning coordinated representations, the underlying assumption will be that there exists some semantic relation between a signal in modality m and modality n.~This relation can be equivalence -- as in a video dataset where the audio at a given timestep t is directly intertwined with the sequence of RGB images at that timestep: they both are stand-ins for conceptually the same entity. The relation can also be a different function such as in the problem of cross-modal speech segment retrieval: here we want to return a relevant passage from an audio or speech file given a textual query. The text query is not the exact transcript of the desired speech segment, but they do relate to each other semantically, for this our model would have to learn this complex relationship across modalities \citep{baltrušaitis2017multimodal}.

To do this we learn a class of models where each model will learn to project one modality into its own representation space. We then have to design a loss function in such a way as to transfer information from one representation to another: we essentially want to make semantically similar data points sit close together in representation space while having semantically dissimilar points sit far away from each other. Since each modality lives in its own representation space our loss function serves to align -- or coordinate -- these vector spaces as to fulfill this desired quality.

After having introduced what representation spaces we want to learn in the sections \protect\hyperlink{multimodal-fusion}{multimodal fusion} and \protect\hyperlink{multimodal-alignment}{multimodal alignment} we will elaborate further on exactly how we can learn joint and coordinate multimodal representation spaces respectively.

\hypertarget{multimodal-alignment}{%
\subsubsection{Multimodal Alignment}\label{multimodal-alignment}}

Alignment occurs when two or more modalities need to be synchronized, such as matching audio and video. It deals with the how rather than the what of learning coordinated representation spaces. Here, the goal is to learn separate representation spaces for each present modality, given that a dataset of corresponding data n-tuples exist. The embedding spaces are technically separate but through a carefully chosen learning strategy they are rotated and scaled such that their data points can be compared and queried across representation spaces. Currently the most common learning paradigm for alignment is contrastive learning. Contrastive learning was described extensively in a previous chapter, so in short: given a pair of semantically equivalent samples in different modalities we would want these data points to be as close as possible in embedding space while being far apart from semantically dissimilar samples\citep{baltrušaitis2017multimodal}.

\hypertarget{multimodal-fusion}{%
\subsubsection{Multimodal Fusion}\label{multimodal-fusion}}

\begin{figure}

{\centering \includegraphics[width=1\linewidth]{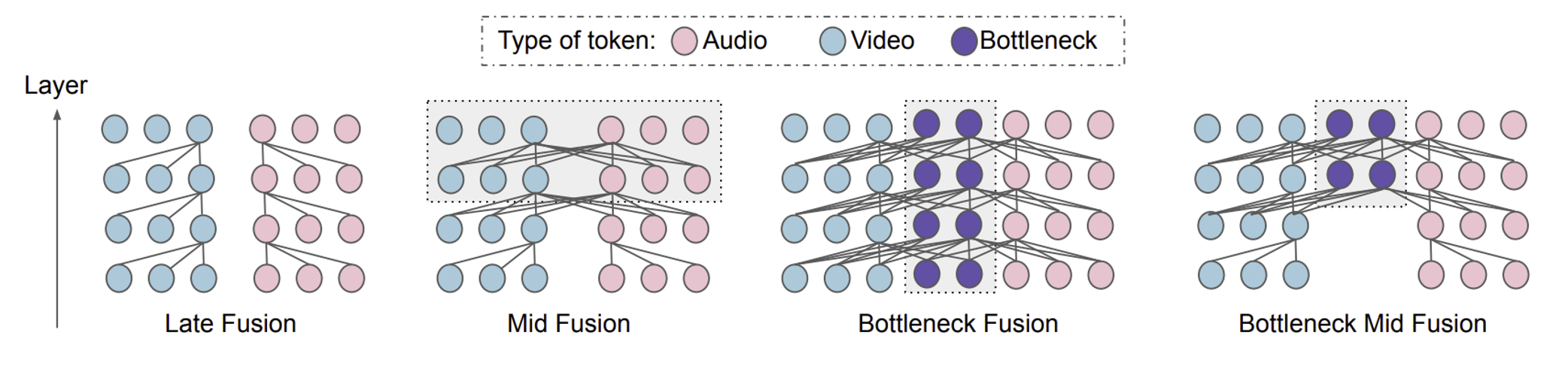}

}

\caption{Different types of multimodal fusion[@baltrušaitis2017multimodal].}\label{fig:fusion}
\end{figure}

Analogous to alignment, multimodal fusion describes how joint representations are learnt. Fusion describes the process of merging modalities inside the model, usually a concatenated and tokenized or patched multimodal input is fed into the model as a 2D matrix. The information from the separate modalities have to combine somehow inside the model to learn from one another to produce a more meaningful, semantically rich output. In the context of Transformer \citep{DBLP:conf/nips/VaswaniSPUJGKP17} based models this usually means where the different inputs start attending to one another cross-modally. This can happen either early on in the model, somewhere in the middle, close to the output in the last layer(s) or based on a hybrid approach. These techniques are usually either based on heuristics, the researcher's intuition, biological plausibility, experimental evidence, or a combination of all {[}\citet{DBLP:conf/nips/NagraniYAJSS21}{]}{[}@ DBLP:journals/jstsp/ZhangYHD20{]}{[}@ shvetsova2021everything{]}.

\hypertarget{multimodal-translation}{%
\subsubsection{Multimodal Translation}\label{multimodal-translation}}

\begin{figure}

{\centering \includegraphics[width=1\linewidth]{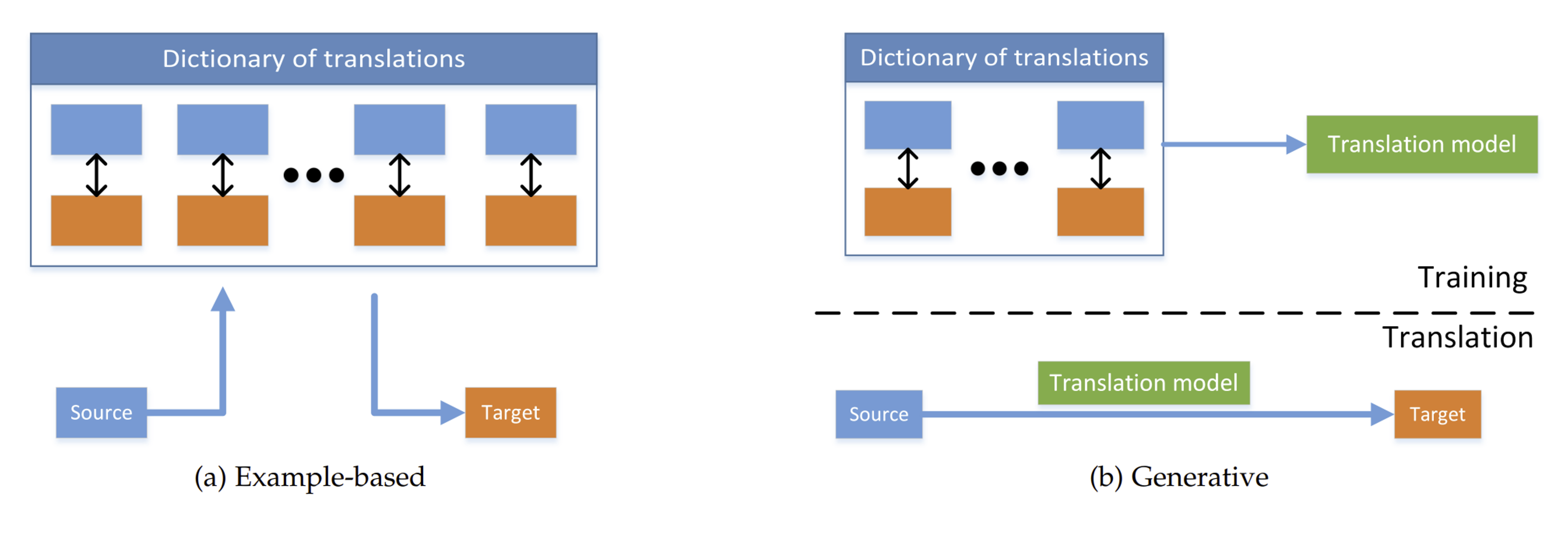}

}

\caption{Different types of multimodal translation [@baltrušaitis2017multimodal].}\label{fig:translation}
\end{figure}

In many practical multimodal use-cases we actually want to map from one modality to another: As previously mentioned we might want to return a relevant speech segment from an audio file given a text query, we want to return a depth map or a semantic map given an RGB image or we want to return a description of an image to read out for visually impaired people\citep{bachmann2022multimae}. In any way we are presented with a datapoint in one modality and want to translate it to a different modality. This another one of the main challenges of the multimodal deep learning landscape and it is what this subsection will be about \citep{DBLP:journals/mt/SulubacakCGREST20}.

\hypertarget{retrieval}{%
\paragraph{Retrieval}\label{retrieval}}

In order to perform cross-modal retrieval one essentially has to learn a mapping that maps items of one modality to items of another modality. Practically this means aligning separate unimodal representation spaces so that the neighborhood of one datapoint contains and equivalent datapoint of a different modality when its representation space is queried at that point {[}\citet{shvetsova2021everything}{]}{[}@ DBLP:conf/eccv/Gabeur0AS20{]}.

Currently cross-modal retrieval is almost exclusively learnt via contrastive learning which we described previously {[}\citet{DBLP:conf/icml/ChenK0H20}{]}{[}@ oord2018representation{]}{[}@ DBLP:conf/icml/ZbontarJMLD21{]}.

\hypertarget{generation}{%
\paragraph{Generation}\label{generation}}

We might also be presented with the case where we have a query in one modality but a corresponding datapoint in a different modality simply does not exist. In this case we can train generative multimodal translation models that learn to decode samples from a vector space into an output of a different modality. This requires us to learn models with a deep understanding of the structure of our data: when sampling datapoint from our cross-modal representation space and applying a decoder to produce the intended output we need to sample from a relatively smooth distribution \citep{DBLP:journals/jstsp/ZhangYHD20}. Since we are actually doing interpolation between known points of our data distribution, we want to produce sensible outputs from ``in between'' our original data. Learning this smooth distribution often requires careful regularization and appropriate evaluation poses another challenge\citep{baltrušaitis2017multimodal}.

With the hype around generative multimodal models created mostly by models such as Dall-E \citep{DBLP:conf/icml/RameshPGGVRCS21} came a huge spike in research around this area {[}\citet{saharia2022photorealistic}{]}{[}@ wu2022nuwainfinity{]}. Currently lots of models generate photorealistic outputs through diffusion \citep{DBLP:conf/nips/HoJA20}, yet they still employ models such as a pretrained CLIP \citep{DBLP:conf/icml/RadfordKHRGASAM21} module as the backbone.

\hypertarget{current-research-trends-generalized-self-supervised-multimodal-perception}{%
\subsection{Current Research Trends: Generalized Self-Supervised Multimodal Perception}\label{current-research-trends-generalized-self-supervised-multimodal-perception}}

So far, we have understood the challenges we are faced with when trying to solve multimodal learning problems. We have understood that from a theoretical perspective we need to learn one or several semantic representation spaces and what the overarching constraints are for learning these vector spaces. Moreso, we have seen that given a coordinated representation space we can translate between modalities and decode our vector space into new data points. For joint representation spaces we can apply traditional downstream tasks such as classification or regression to better solve real world problems leveraging the interplay of all modalities at hand.
Going forward we will explore the two major building blocks for realizing these challenges from a more practical perspective:

\begin{itemize}
\tightlist
\item
  Multimodal Architectures
\item
  Multimodal Training Paradigms
\end{itemize}

A combination of carefully chosen model architecture and training scheme is necessary to solve the challenges we have described on a high level. Throughout the rest of this subchapter, we will look at more and more general concepts for each of these components. In this subchapter we will also connect back to one central principal that we have introduced earlier in this chapter: approaching human-level of multimodal perception. This means that we will follow one of the major lines of research within multimodal deep learning: building more general and problem-agnostic solutions. We pose the question: Why apply hyper-specific solutions when we can simplify and generalize our methods while retaining (or even improving) on experimental results.
Towards the end of the chapter, we will also briefly introduce our own research in which we attempt to combine a modality agnostic model architecture with a generalized non-contrastive training paradigm for uni- and multi-modal self-supervised learning.

\hypertarget{general-multimodal-architectures}{%
\subsubsection{General Multimodal Architectures}\label{general-multimodal-architectures}}

First, we want to establish some desirable characteristics that our generalized multimodal model architectures should have:

\begin{itemize}
\tightlist
\item
  Input-Agnosticism: Whether our input consist of 1-dimensional sequences of audio waveforms or text or 3-dimensional inputs such as video we want out model to process all kinds of modalities equally with as little adjustments as possible.
\item
  Multimodal Fusion: Ideally, we would also like to feed a concatenated input of several modalities into the model to learn joint representations.
\item
  Preservation of Locality
\item
  Respect Compositionality
\item
  Flexible outputs: The model produces not only scalar or vector outputs but can ideally decode into any arbitrary output, thereby essentially having the capability for multimodal translation integrated.
\end{itemize}

We have not explicitly listed multimodal alignment as a desirable characteristic because the capability to perform alignment becomes trivial since we included the point about flexible outputs: to do alignment we need our model to output vectors that we can predict or regress over via a loss function.
To illustrate the state of current research we will briefly introduce three multimodal model architecture that fulfill some, if not all of the above-mentioned criteria.

\hypertarget{nuxfcwa}{%
\paragraph{NÜWA}\label{nuxfcwa}}

\begin{figure}

{\centering \includegraphics[width=1\linewidth]{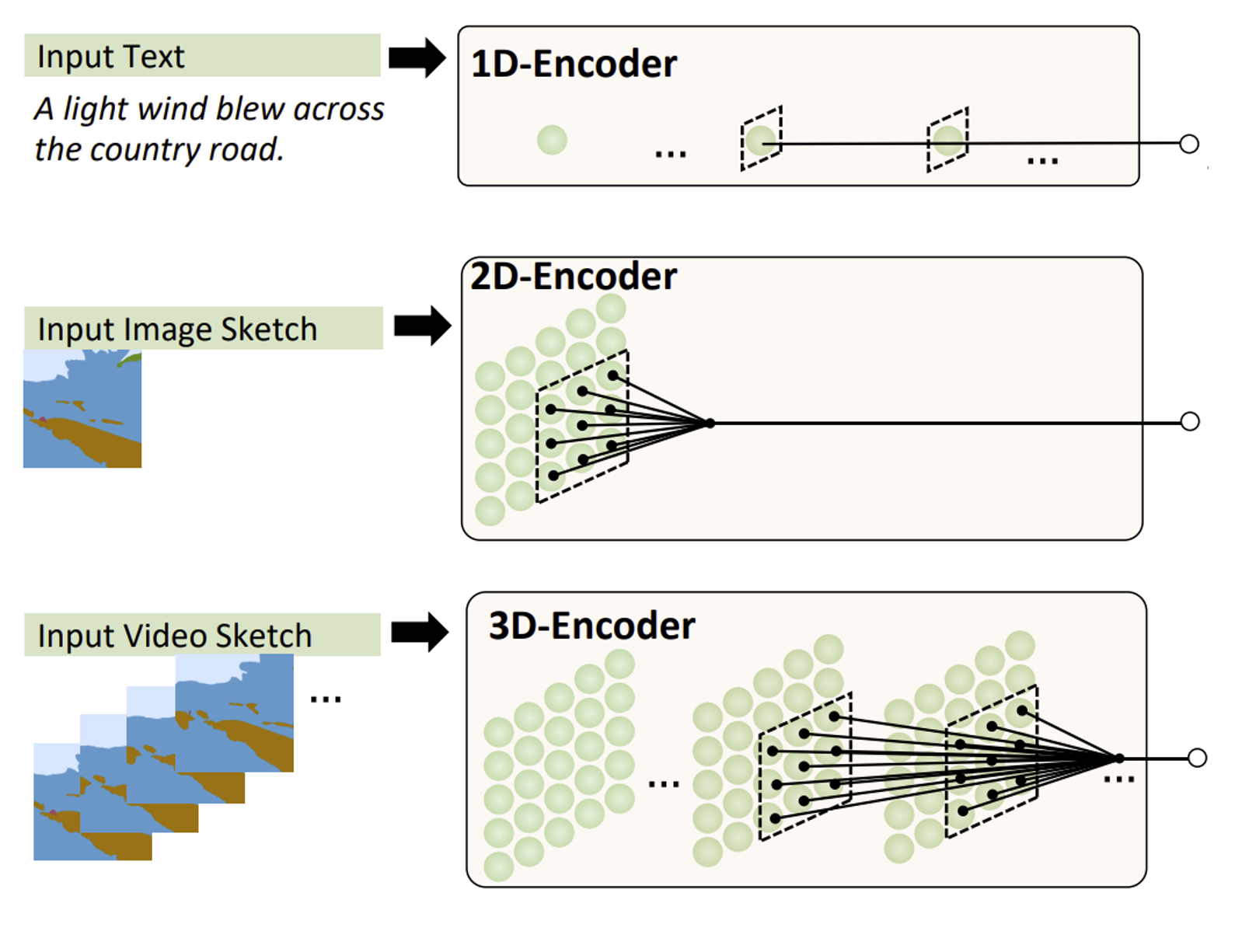}

}

\caption{Data being represented in the NÜWA-imposed 3D format[@wu2021nwa].}\label{fig:nuwa}
\end{figure}

Initially conceived as a generative multimodal translation model here we are especially interested in the 3D-Nearby-Attention encoder-decoder stack at the center of NÜWA (Neural visUal World creAtion). We assume that our input, whether it be 1-dimensional sequences like text (or audio, although it was not used in the paper), 2-dimensional matrices like RGB images, sketches or semantic maps or 3-dimensional matrices like video (or possibly other modalities, e.g., depth maps) is already tokenized or divided into vectorized patches in case of 2D or 3D inputs. This is simply because we are using Transformer based encoder/decoder modules, therefore we need to reduce the input size beforehand. In the case of images or video this is done by a VQ-VAE. In principle any input is in the shape of a 3D matrix where one axis represents the temporal dimensions and the other axis representing the height and width. Clearly videos would fill out this input format in every direction, for still images the temporal dimension is simply one and for sequences such as text or audio they have height and width of one respectively and only stretch along the temporal dimension. Then for every token or patch a local neighborhood is defined amongst which the self-attention mechanism is applied. This saves on computational costs as for larger inputs global self-attention between all tokens or patches can become expensive. By imposing this 3D input format on all inputs, the model preserves the geometric and temporal structure of the original inputs, together with the locality respecting 3DNA mechanism the model introduces valuable and efficient inductive biases that make the model agnostic to input modality (if it is represented in the correct format), respects locality and allows for flexible outputs as it is intended to translate from any input to arbitrary outputs. Depending on how patching is performed for input data one could imagine a setup where the 3D encoder-decoder could also implement a hierarchical structure which would also respect compositionality in data\citep{kahatapitiya2021swat}, but this was not studied in this paper, although a similar idea was implemented in the follow-up paper {[}\citet{wu2021nwa}{]}{[}@ wu2022nuwainfinity{]}.

\hypertarget{perceiver-io}{%
\paragraph{Perceiver IO}\label{perceiver-io}}

\begin{figure}

{\centering \includegraphics[width=1\linewidth]{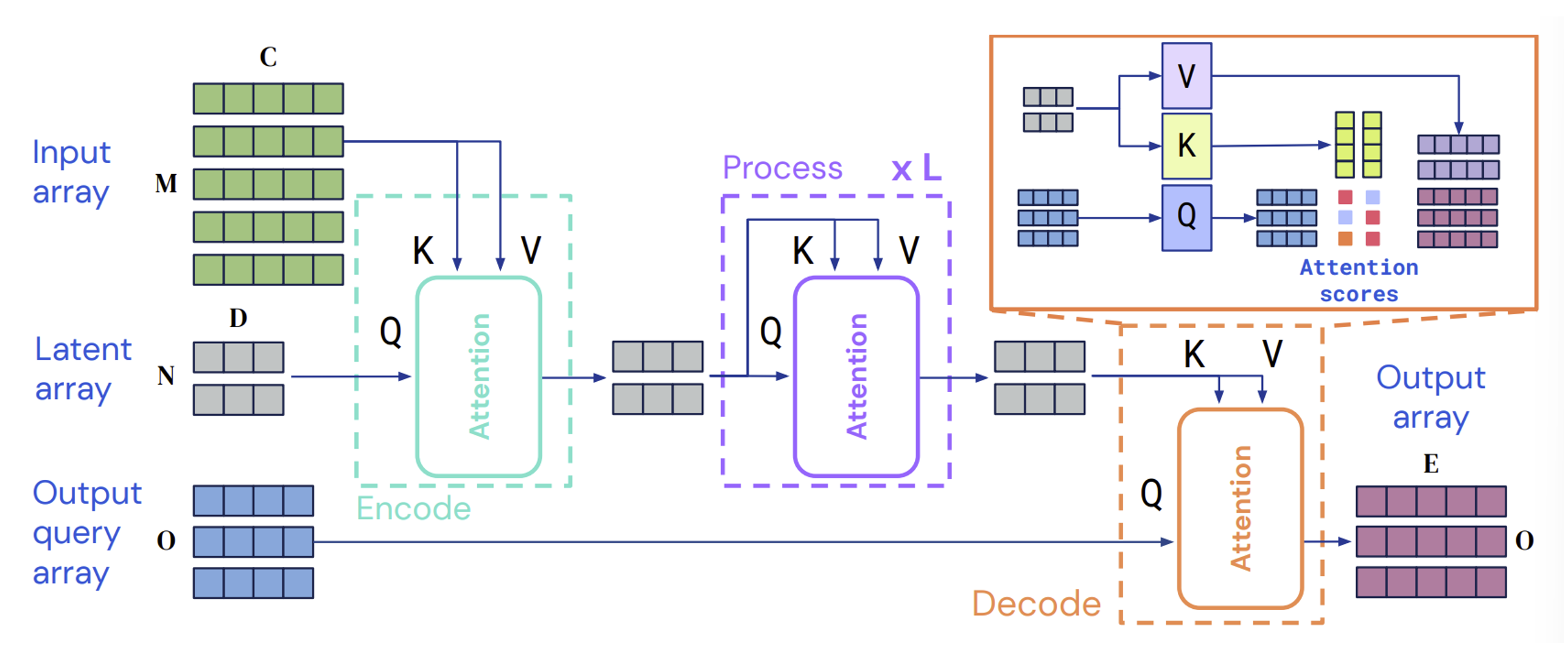}

}

\caption{Perceiver encoder stack shows how the cross-attention mechanism transforms large inputs into smaller ones that can be processed by a vanilla Transformer encoder [@jaegle2021perceiver].}\label{fig:perceiver-io}
\end{figure}

The Perceiver consists of a vanilla Transformer encoder block, with the follow-up paper, Perceiver IO, adding an analogous Transformer decoder in order to produce arbitrary multimodal outputs. The trick the Perceiver introduces in order to process nearly any sort of (concatenated) multimodal input is a specific cross-attention operation. Given a (very long) input in of size \(M \times C\) a so-called latent array of size \(N\times D\) is introduced, where \(N\ll M\). With the input array acting as key and value and the latent array as the query a cross-attention block is applied between the two, this transforms the original input to a much smaller size, achieving higher than 300x compression. Perceiver IO is currently likely the most flexible model when it comes to processing and outputting arbitrary multimodal outputs, it also easily handles the learning of joint representation spaces at it can process very large input array of concatenated multimodal data such as long videos with audio or optical flow maps{[}\citet{DBLP:conf/icml/JaegleGBVZC21}{]}{[}@ jaegle2021perceiver{]}.

\hypertarget{hierarchical-perceiver}{%
\paragraph{Hierarchical Perceiver}\label{hierarchical-perceiver}}

\begin{figure}

{\centering \includegraphics[width=1\linewidth]{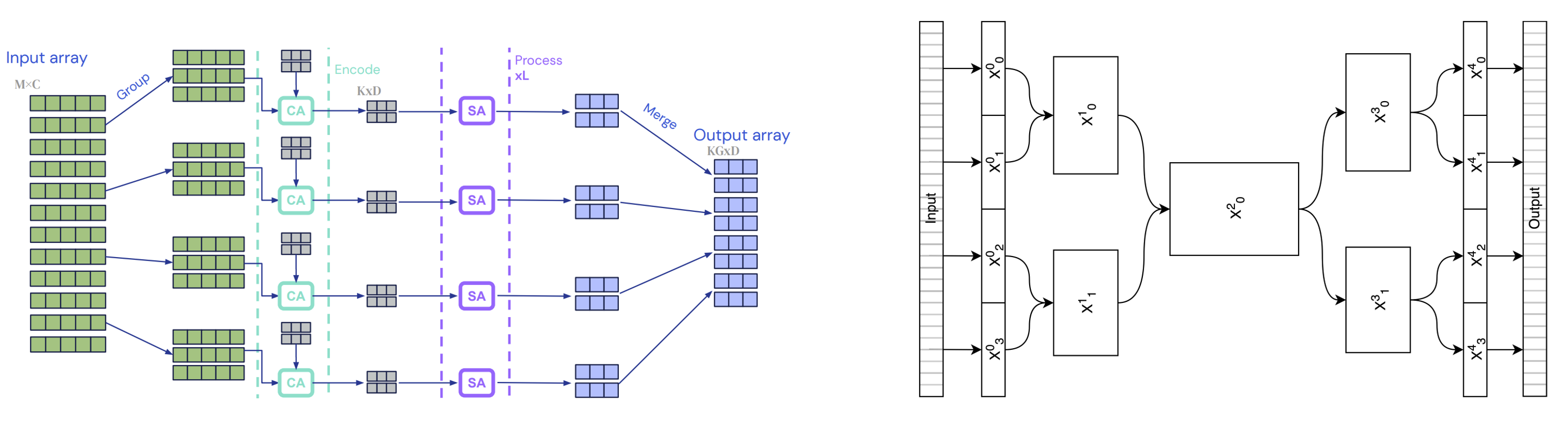}

}

\caption{The hourglass structure of the Hierarchical Perceiver (HiP)[@carreira2022hierarchical].}\label{fig:hierarchical-perceiver}
\end{figure}

With information about locality and compositionality being mostly lost in the Perceiver IO encoder this follow up paper imposes a hierarchical hourglass-like structure on the encoder-decoder stack. An input matrix with length M tokens is broken down into G groups, each M/G in size. For each group a separate latent array of size \(K \times Z\) is initialized, and a cross-attention operation is applied between each group and its respective latent array, followed by a number of self-attention + MLP blocks. The set of output vectors of each group is then merged to form an intermediary matrix consisting of KG tokens. This intermediary matrix can be used as input to the next block, forming the hierarchical structure of the encoder. Besides embedding more of the locality and compositionality of data into the model, this architecture also improves upon computational costs on comparison to Perceiver IO \citep{carreira2022hierarchical}.

\hypertarget{multimodal-training-paradigms}{%
\subsubsection{Multimodal Training Paradigms}\label{multimodal-training-paradigms}}

Research of the past years has shown that in deep learning usually it is best to perform some sort of generalized task-agnostic pretraining routine. During self-supervised training that does not rely on any labeled data our model is trained to find underlying structures in the given unlabeled data. Given the right modeling and training scheme the model is able to approximate the true data distribution of a given dataset. This is extremely helpful as unlabeled data is extremely abundant so self-supervised learning lends itself as a way to infuse knowledge about the data into our model that it can leverage either directly for a downstream task (zero-shot) or helps as a running start during fine-tuning.

Since the conceptual part of pre-training was shown to have such an immense influence on downstream task performance, we will mainly focus on the self-supervised learning aspect of multimodal deep learning in this subchapter.

For self-supervised multimodal training paradigms, we can devise two major subcategories: those training paradigms that are agnostic to the input modality but operate only on a unimodal input and those that are both agnostic to input modalities but are truly multimodal.

\hypertarget{uni-modal-modality-agnostic-self-supervised-learning}{%
\paragraph{Uni-Modal Modality-Agnostic Self-Supervised Learning}\label{uni-modal-modality-agnostic-self-supervised-learning}}

BYOL \citep{grill2020bootstrap} has introduced a paradigm shift for uni-modal self-supervised learning with its latent prediction mechanism. Its core idea is that the model to be trained is present in a student state and a teacher state where the teacher is a copy of the student with its weights updated by an exponentially moving average (EMA) of the student. Initially, BYOL was trained only on images: two augmentations would be applied to a base image and are then fed to the student and teacher network. The student would predict the latent states of last layers the teacher network via a simple regression loss. Data2vec extends this idea by generalizing it to other modalities: instead of applying specific augmentations to a base image a masking strategy is designed for each modality in order to augment the inputs, i.e., construct a semantically equivalent altered input. In the paper each modality has its own specific masking strategy and encoder backbone but in principle the paper showed that latent prediction SSL can be applied to other modalities such as text and audio just as well. Later we will introduce our own line of research where we try to generalize and simplify this even further and apply this concept to joint and coordinated representation problems.

Data2vec \citep{baevski2022data2vec} has already been extensively introduced in a previous chapter, because of that we would like to focus here on the importance of this relatively new line of SSL strategy that we call latent prediction SSL and why we think it is especially suitable for multimodal problems.

\begin{figure}

{\centering \includegraphics[width=1\linewidth]{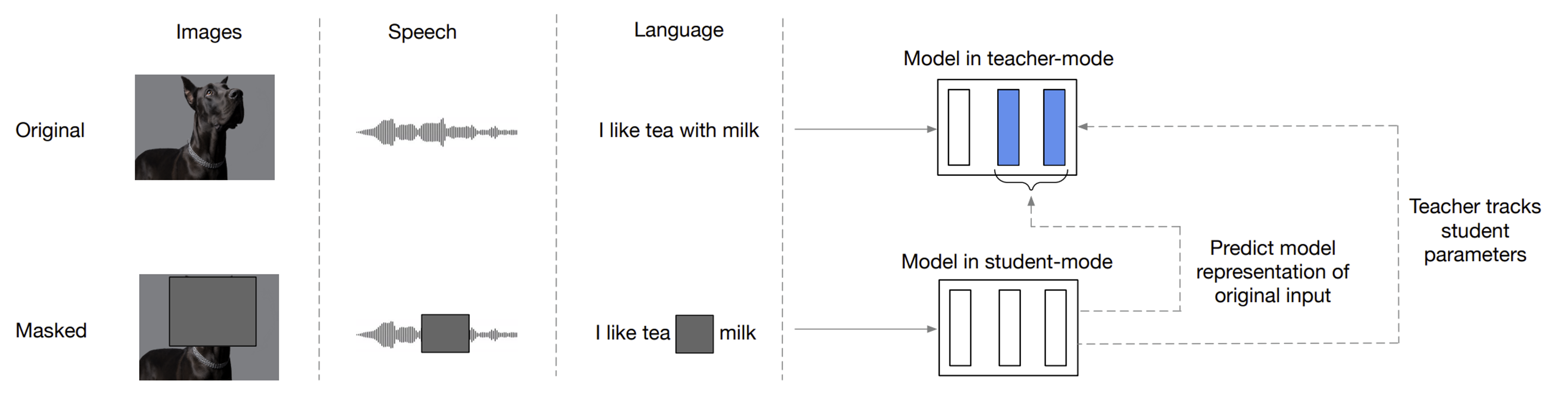}

}

\caption{Illustration of data2vec[@baevski2022data2vec].}\label{fig:data2vec}
\end{figure}

First off, how is latent prediction training different from other major SSL strategies like contrastive learning, masked auto-encoding (MAE) or masked token prediction? Whereas MAE and masked token prediction predict or decode into data space -- meaning they need to employ a decoder or predict a data distribution -- latent space prediction operates directly on the latent space. This means that the model has the predict the entire embedded context of an augmented or masked input, this forces the model to learn a better contextual embedding space and (hopefully) learn more meaningful semantic representations. Compared to contrastive approaches latent prediction methods do not require any hard negatives to contrast with. This alleviates us from the problem of producing hard negatives in the first place. Usually, they are sampled in-batch at training time with nothing guaranteeing us they are even semantically dissimilar (or how much so) from the anchor. The loss function of latent prediction SSL is usually L1 or L2 regression loss which is easy and straight-forward and without the need to predict in data space or mine hard negatives we avoid many of the disadvantages of the other SSL training schemes while also improving upon contextuality of our embeddings by virtue of prediction in latent space \citep{baevski2022data2vec}.

Since this training paradigm is generalizable to multimodal problems and avoids common points of failure of other major SSL strategies it is in line with the principle that we follow here, namely: Why solve a harder problem when we can simplify it and retain performance and generality?

\hypertarget{multimodal-self-supervised-learning}{%
\paragraph{Multimodal Self-Supervised Learning}\label{multimodal-self-supervised-learning}}

As we have elaborated extensively, we are faced with learning either joint or coordinated representations in virtually any multimodal problem. Currently no learning framework or paradigm covers both joint and coordinated learning at the same time. Joint representations are usually learnt via contrastive methods whereas coordinated representations usually employ some variant of masked input prediction or denoising autoencoding.

As an example, for multimodal contrastive learning for joint representations we will first look at VATT \citep{DBLP:conf/nips/AkbariYQCCCG21}, short for Video-Audio-Text Transformer. In this paper the authors propose a simple framework for learning cross-modal embedding spaces across multiple (\textgreater= 2) modalities at once. They do so by introducing an extension to InfoNCE loss called Multiple-Instance-Learning-NCE (MIL-NCE). The model first linearly projects each modality into a feature vector and feeds it through a vanilla Transformer backbone. If the model is only used to contrast two modalities, then a normal InfoNCE loss is being used, for a video-audio-text triplet a semantically hierarchical coordinated space is learnt that enables us to compare video-audio and video-text by the cosine similarity. First a coordinated representation between the video and audio modality is constructed via the InfoNCE \citep{DBLP:conf/icml/ChenK0H20} loss. Then a coordinated representation between the text modality and the now joint video-audio modality is also constructed similarly as shown in this figure. This hierarchy in these coordinated representations is motivated by the different levels of semantic granularity of the modalities, therefore this granularity is introduced into the training as an inductive bias. The aggregation of the several InfoNCE at different levels serves as the central loss function for this training strategy. It quickly becomes evident how this principle of learning (hierarchical) coordinated embeddings spaces can serve to learn between any n-tuples of different modalities if the respective n-tuples exist in a dataset.

MultiMAE \citep{bachmann2022multimae} is a different kind of paper in which the authors learn a joint representation of a concatenated input consisting of RGB images, depth, and semantic maps. The input is partitioned into patches with some of them randomly selected for masking. The flattened masked input is then fed into a multimodal ViT \citep{DosovitskiyB0WZ21} backbone. The authors then use different decoder blocks that act upon only the unimodal segments of the input. They hypothesize that the multimodal transformer can leverage cross-modal information in the input well enough as to embed multimodal semantic information in the output states. An additional global token that can access all input modalities is added for learning the joint representation. The task-specific decoders reconstruct their respective modality also by using one cross-attention module that can access information from the whole (multimodal) input. The aggregate reconstruction loss of all decoders serves as the model's loss function. This training strategy thereby produces a joint representation of an arbitrary ensemble of patched 2D modalities and can simultaneously learn to perform unimodal tasks as well.

\begin{figure}

{\centering \includegraphics[width=1\linewidth]{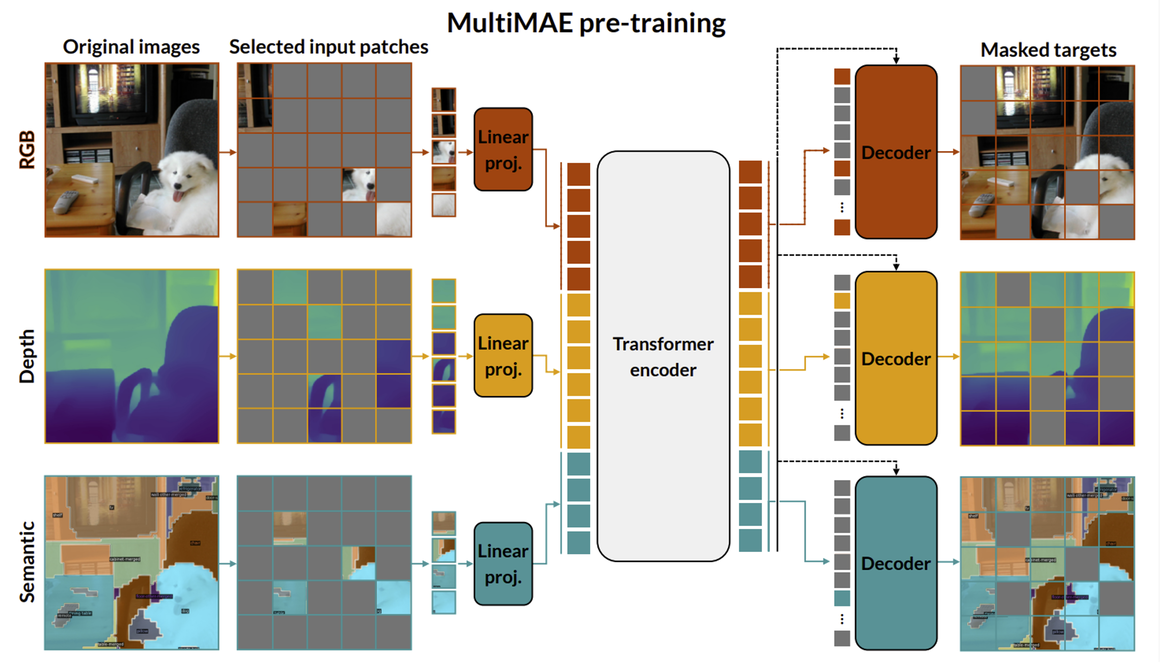}

}

\caption{Masking and cross-modal prediction visualized for MultiMAE[@bachmann2022multimae].}\label{fig:multimae}
\end{figure}

So far, we have seen that for multimodal self-supervised learning we have a class of strategy that revolves around learning coordinated representations using contrastive learning. After having met contrastive multimodal models such as CLIP in earlier chapters we have shown that we can extend the same principles to include further modalities such as audio and video. In fact, if we are provided, with collections of multimodal data that pairs one modality to another, this principle can be applied to any set of given modalities.

Also, we have met training strategies that aim to generalize joint representation learning to multiple modalities. While the presented MultiMAE focuses on 2-dimensional modalities such as RGB, depth and semantic maps we could easily imagine applying the same principle to include other modalities as well -- given we can process our raw signals to represent them in the appropriate format a model can read.

We have omitted any specific learning strategies that pretrain specifically for translation tasks. For retrieval tasks it is evident that contrastive methods would offer zero-shot cross-modal retrieval capabilities. For generative tasks, the interested reader is invited to study the NÜWA paper whose 3D multimodal encoder we have introduced earlier: in it the authors leverage an identical 3D decoder to translate modalities one into another in a self-supervised manner. While the NÜWA 3D Attention encoder-decoder stack is not technically a multimodal model they do apply cross-modal cross-attention in to transfer semantic information from an encoded prompt to the decoder.

\hypertarget{personal-research-general-non-contrastive-multimodal-representation-learning}{%
\paragraph{Personal Research: General Non-Contrastive Multimodal Representation Learning}\label{personal-research-general-non-contrastive-multimodal-representation-learning}}

So far, we have looked at unimodal and multimodal SSL as two separate categories. Research so far has not married the two concepts into a training paradigm that can learn both multimodal joint representations as well as cross-modal coordinated representations.

Let us consider a concatenated multimodal input signal. This concatenated input array would only really differ from a unimodal signal in that it already contains modality specific encodings added to the raw input -- similar to those seen in the Perceiver. In fact, let us consider a multimodal input of the exact format seen in the Perceiver. In principle we could apply some masking strategy to this input array to mask out consecutive chunks of the input matrix and apply the same latent prediction training paradigm as seen in data2vec. We craft this masking strategy in such a way as to account for highly correlated nearby signals. If we were to randomly mask single rows of the input the task of predicting the mask input for very long inputs such as in videos or audio files becomes too trivial.

By representing all inputs, whether they are unimodal or multimodal in this unified format inspired by the Perceiver and applying a generic masking strategy we have essentially generalized data2vec to any arbitrary uni- or multimodal input. A Perceiver backbone model ensures that the handling and encoding of exceptionally large input arrays becomes efficient and effective.

Similarly let us consider a multimodal input for coordinated representations. Let us also assume that our model shares its weights across the separate modality-specific representation spaces (similar to VATT). Latent prediction training schemes such as BYOL and data2vec feed separate augmentations of the same input into the model which can be either in student or teacher (or online and offline) mode. The assumption is that both inputs should be roughly semantically equivalent so the model can learn to ignore the augmentations or masks to catch the essential structure within the data. We pose the question: Are different modalities of the same thing also not just as semantically equivalent? Can we view different modalities simply as augmentations of one another and leverage the same training paradigm as in BYOL and data2vec, feeding one modality into the student model while feeding another into the teacher model? Would this learner be able to catch the essence of semantic equivalence of these two input signals? In our research project we try to answer these questions as well, our unified generalized multimodal learning framework is the first of its kind to be applicable to both joint as well as coordinated representations without any adjustments.

We propose this unified multimodal self-supervised learning framework as a novel and first-of-its-kind training paradigm that generalizes the unimodal self-supervised latent prediction training scheme inspired by BYOL and data2vec to an arbitrary number of input modalities for joint representation learning as well as cross-modal coordinated representation learning without the use of contrastive methods. Our method requires data to be presented in a generic format proposed by the Perceiver and requires just one single masking strategy.

This resolves the need for modality-specific masking strategies and models like in data2vec. For the cross-modal use-case we eliminate the need for hard negatives which are usually required for contrastive learning.

\hypertarget{c03-02-structured-unstructured}{%
\section{Structured + Unstructured Data}\label{c03-02-structured-unstructured}}

\emph{Author: Rickmer Schulte}

\emph{Supervisor: Daniel Schalk}

\hypertarget{intro}{%
\subsection{Intro}\label{intro}}

While the previous chapter has extended the range of modalities considered in multimodal deep learning beyond image and text data, the focus remained on other sorts of unstructured data. This has neglected the broad class of structured data, which has been the basis for research in pre-deep learning eras and which has given rise to many fundamental modeling approaches in statistics and classical machine learning. Hence, the following chapter aims to give an overview of both data sources and will outline the respective ways how these have been used for modeling purposes as well as more recent attempts to model them jointly.

Generally, structured and unstructured data substantially differ in certain aspects such as dimensionality and interpretability. This has led to various modeling approaches that are particularly designed for the special characteristics of the data types, respectively. As shown in previous chapters, deep learning models such as neural networks are known to work well on unstructured data. This is due to their ability to extract latent representation and to learn complex dependencies from unstructured data sources to achieve state-of-the art performance on many classification and prediction tasks. By contrast, classical statistical models are mostly applied on tabular data due the advantage of interpretability inherent to these models, which is commonly of great interest in many research fields. However, as more and more data has become available to researchers today, they often do not only have one sort of data modality at hand but both structured and unstructured data at the same time. Discarding one or the other data modality makes it likely to miss out on valuable insights and potential performance improvements.

Therefore, in the following sections we will investigate different proposed methods to model both data types jointly and examine similarities and differences between those. Different fusion strategies to integrate both types of modalities into common deep learning architectures are analyzed and evaluated, thereby touching upon the concept of end-to-end learning and its advantages compared to separated multi-step procedures. The different methods will be explored in detail by referring to numerous examples from survival analysis, finance and economics.
Finally, the chapter will conclude with a critical assessment of recent research for combining structured and unstructured data in multimodal DL, highlighting limitations and weaknesses of past research as well as giving an outlook on future developments in the field.

\hypertarget{taxonomy-structured-vs.-unstructured-data}{%
\subsection{Taxonomy: Structured vs.~Unstructured Data}\label{taxonomy-structured-vs.-unstructured-data}}

In order to have a clear setup for the remaining chapter, we will start off with a brief taxonomy of data types that will be encountered. Structured data, normally stored in a tabular form, has been the main research object in classical scientific fields. Whenever there was unstructured data involved, this was normally transformed into structured data in an informed manner. Typically, doing so by applying expert knowledge or data reduction techniques such as PCA prior to further statistical analysis. However, DL has enabled unsupervised feature extraction from unstructured data and thus to feed it to the models directly. Classical examples of unstructured data are image, text, video, and audio data as shown in the figure below. Of these, image data in combination with tabular data is the most frequently encountered. Hence, this combination will be examined along various examples later in the chapter. While previously mentioned data types allow for a clear distinction, lines can become increasingly blurred. For example, the record of a few selected biomarkers or genes from patients would be regarded as structured data and normally be analyzed with classical statistical models. On the contrary, having the records of multiple thousand biomarkers or genes would rather be regarded as unstructured data and usually be analyzed using DL techniques. Thus, the distinction between structured and unstructured data does not only follow along the line of dimensionality but also concerns the interpretability of single features within the data.

\begin{figure}

{\centering \includegraphics[width=1\linewidth]{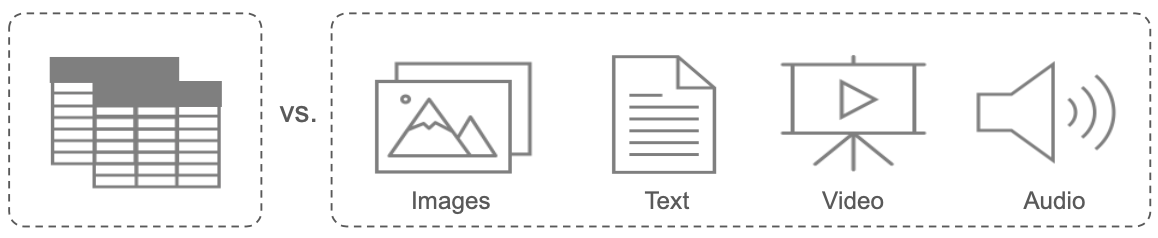}

}

\caption{Structured vs. Unstructured Data}\label{fig:struc-vs-unstrc}
\end{figure}

\hypertarget{fusion-strategies}{%
\subsection{Fusion Strategies}\label{fusion-strategies}}

After we have classified the different data types that we will be dealing with, we will now discuss different fusion strategies that are used to merge data modalities into a single model. While there are potentially many ways to fuse data modalities, a distinction between three different strategies, namely early, joint and late fusion has been made in the literature. Here we follow along the taxonomy laid out by \citet{HuangFusion2020} with a few generalizations as those are sufficient in our context.

\textbf{Early fusion} refers to the procedure of merging data modalities into a common feature vector already at the input layer. The data that is being fused can be raw or preprocessed. The step of preprocessing usually involves dimensionality reduction to align dimensions of the input data. This can be done by either training a separate DNN (Deep Neural Network), using data driven transformations such as PCA or directly via expert knowledge.

\textbf{Joint fusion} offers the flexibility to merge the modalities at different depths of the model and thereby to learn latent feature representations from the input data (within the model) before fusing the different modalities into a common layer. Thus, the key difference to early fusion is that latent feature representation learning is not separated from the subsequent model. This allows backpropagation of the loss to guide the process of feature extraction from raw data. The process is also called end-to-end learning. Depending on the task, CNNs or LSTMs are usually utilized to learn latent feature representations. As depicted in the figure below, it is not required to learn lower dimensional feature representations for all modalities and is often only done for unstructured data. A further distinction between models can be made regarding their model head, which can be a FCNN (Fully Connected Neural Network) or a classical statistical model (linear, logistic, GAM). While the former can be desirable to capture possible interactions between modalities, the latter is still frequently used as it preserves interpretability.

\textbf{Late fusion} or sometimes also called decision level fusion is the procedure of fusing the predictions of multiple models that have been trained on each data modality separately. The idea originates from ensemble classifiers, where each model is assumed to inform the final prediction separately. Outcomes from the models can be aggregated in various ways such as averaging or majority voting.

\begin{figure}

{\centering \includegraphics[width=1\linewidth]{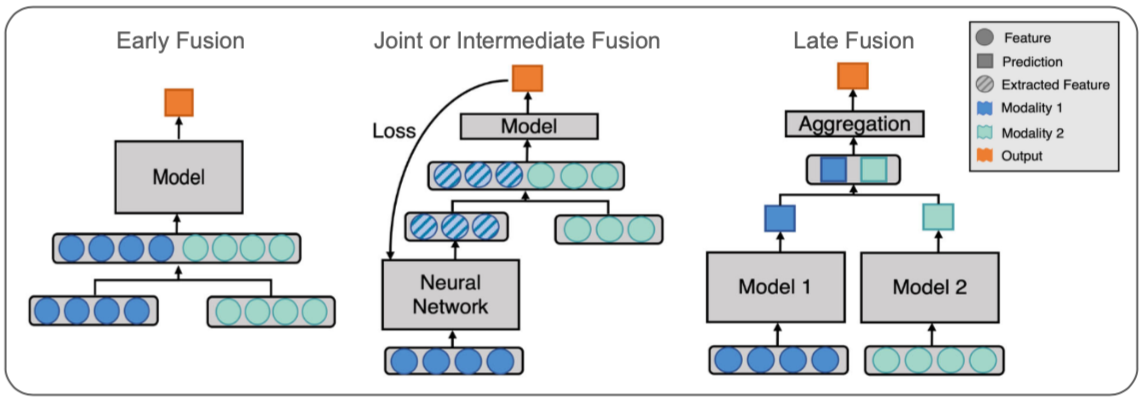}

}

\caption{Data Modality Fusion Strategies \citep[Adopted from][]{HuangFusion2020}.}\label{fig:fusion-strategies}
\end{figure}

We will refer to numerous examples of both early and joint fusion in the following sections. While the former two are frequently applied and easily comparable, late fusion is less common and different in nature and thus not further investigated here. As a general note, for the sake of simplicity we will refer to the special kind of multimodal DL including both structured and unstructured data when we speak about multimodal DL in the rest of the chapter.

\hypertarget{applications}{%
\subsection{Applications}\label{applications}}

The following section will discuss various examples of this kind of multimodal DL by referring to different publications and their proposed methods. The publications originate from very different scientific fields, which is why methods are targeted for their respective use case. Hence, allowing the reader to follow along the development of methods as well as the progress in the field. Thereby, obtaining a good overview of current and potential areas of applications. As there are various publications related to this kind of multimodal DL, the investigation is narrowed down to publications which either introduce new methodical approaches or did pioneering work in their field by applying multimodal DL.

\hypertarget{multimodal-dl-in-survival}{%
\subsubsection{Multimodal DL in Survival}\label{multimodal-dl-in-survival}}

Especially in the field of survival analysis, many interesting ideas were proposed with regards to multimodal DL. While clinical patient data such as electronic health records (EHR) were traditionally used for modeling hazard functions in survival analysis, recent research has started to incorporate image data such as CT scans and other modalities such as gene expression data in the modeling framework. Before examining these procedures in detail, we will briefly revisit the classical modeling setup of survival analysis by discussing the well-known Cox Proportional Hazard Model (CPH).

\hypertarget{traditional-survival-analysis-cph-model}{%
\subsubsection{Traditional Survival Analysis (CPH Model)}\label{traditional-survival-analysis-cph-model}}

Survival Analysis generally studies the time duration until a certain event occurs. While many methods have been developed to analyze the effect
of certain variables on the survival time, the Cox Proportional Hazard Model (CPH) remains the most prominent one. The CPH model models the hazard rate which is the conditional probability of a certain event occurring in the next moment given that it has not so far:

\[
h(t|x) = h_0(t) * e^{x\beta}
\]
where \(h_0(t)\) denotes the baseline hazard rate and \(\beta\) the linear effects of the covariates \(x\) on which the probability is conditioned on. The fundamental assumption underlying the traditional CPH is that covariates influence the hazard rate proportionally and multiplicatively. This stems from the fact that the effects in the so-called risk function \(f(x) = x\beta\) are assumed to be linear. Although this has the advantage of being easily interpretable, it does limit the flexibility of the model and thus also the ability to capture the full dynamics at hand.

\hypertarget{multimodal-dl-survival-analysis}{%
\subsubsection{Multimodal DL Survival Analysis}\label{multimodal-dl-survival-analysis}}

Overcoming the limitations of the classical CPH model, \citet{Katzman2018} were among the first to incorporate neural networks into the CPH and thereby replacing the linear effect assumption. While their so-called DeepSurv model helped to capture interactions and non-linearities of covariates, it only allowed modeling of structured data. This gave rise to the model DeepConvSurv of \citet{DeepConvSurv}, who apply CNNs to extract information from pathological images in order to predict risk of patients subsequently. They showed that learning features from images via CNNs in an end-to-end fashion outperforms methods that relied on hand-crafting features from these images. Building on the idea of DeepConvSurv, \citet{DeepCorrSurv} extended the model by adding further modalities. Besides pathological images, their proposed DeepCorrSurv model also includes molecular data of cancer patients. The name of the model stems from the fact that separate subnetworks are applied to each modality and that the correlation between the output of these modality specific subnetworks are maximized before fine-tuning the learned feature embedding to perform well on the survival task. The correlation maximization procedure aims to remove the discrepancy between modalities. It is argued that the procedure is beneficial in small sample settings as it may reduce the impact of noise inherent to a single modality that is unrelated to the survival prediction task.

The general idea is that the different modalities of multimodal data may contain both complementary information contributed by individual modalities as well as common information shared by all modalities. The idea was further explored by subsequent research. \citet{TongAE} for example introduced the usage of auto encoders (AE) in this context by proposing models that extract the lower dimensional hidden features of the AE applied to each modality. While their first model trains AEs on each modality separately before concatenating the learned features (ConcatAE), their second model obtains cross-modality AEs that are trained to recover both modalities from each modality respectively (CrossAE). Here, the concept of complementary information of modalities informing survival prediction separately gives rise to the first model, whereas the concept of retrieving common information inherent across modalities gives rise to the latter. Although, theoretically both models could also handle classical tabular EHR data, they were only applied to multi-omics data such as gene expressions of breast cancer patients.

Similar to \citet{TongAE}, \citet{Cheerla2019} also derive their model from the idea of common information that is shared by all modalities. Besides, having specialized subnetworks for each modality to learn latent feature embeddings, they also introduce a similarity loss that is added to the classical cox loss from the survival prediction. This similarity loss is applied to each subnetwork output and aims to learn modality invariant latent feature embeddings. This is desirable not only for noise reduction but also in cases of missing data. While previous research often applied their models only on subsets of the large cancer genome atlas program (TCGA), \citet{Cheerla2019} analyze 20 different cancer types of the TCGA using four different data modalities. As expanding the scope of the study increases the problem of data missingness, they specifically target the problem by introducing a variation of regular dropout, which they refer to as multimodal dropout. Instead of dropping certain nodes, multimodal dropout drops entire modalities during training in order to make models less dependent on one single data source. This enables the model to better cope with missing data during inference time. Opposed to \citet{TongAE}, the model is trained in an end-to-end manner and thus allows latent feature learning to be guided by the survival prediction loss. More impressive than their overall prediction performances are the results of T-SNE-mappings that are obtained from the learned latent feature embeddings. One sample mapping is displayed in the figure below, which nicely shows the clustering of patients with regards to cancer types. This is particularly interesting regarding the fact that the model was not trained on this variable. Besides being useful for accurate survival prediction, such feature mappings can directly be used for patient profiling and are thus pointed out as a contribution to the research on their own.

\begin{figure}

{\centering \includegraphics[width=1\linewidth]{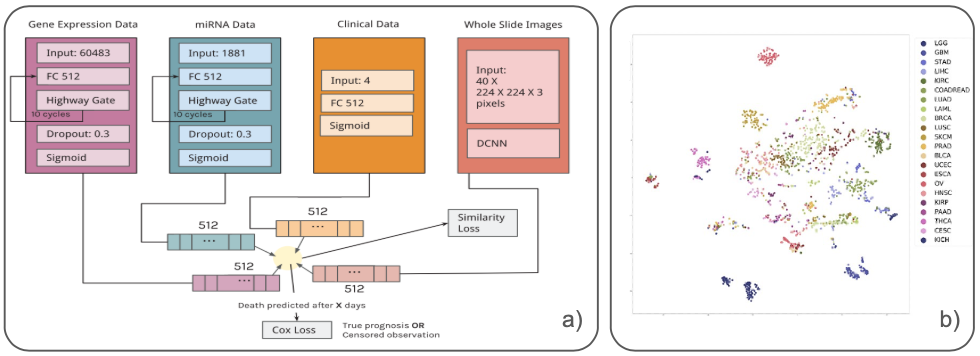}

}

\caption{a) Architecture with Similarity Loss b) T-SNE-Mapped Representations of Latent Features (Colored by Cancer Type) \citep{Cheerla2019}.}\label{fig:cheerla-model}
\end{figure}

\citet{MultiSurv2021} extend the previous work by enlarging the scope of study, analyzing up to six different data modalities and 33 cancer types of the TCGA dataset. Their so-called MultiSurv model obtains a straightforward architecture, applying separate subnetworks to each modality and a subsequent FCNN (model head) to yield the final survival prediction. Testing their modular model on different combinations of the six data modalities, they find the best model performance for the combination of structured clinical and mRNA data. Interestingly, including further modalities lead to slight performance reductions. Conducting some benchmarking, they provide evidence for their best performing model (structured clinical + mRNA) to outperform all single modality models. However, it is worthwhile mentioning that their largest model, including all six modalities, is not able to beat the classical CPH model, which is based on structured clinical data only. While this already may raise concerns about the usefulness of including so many modalities in the study, high variability of model performance between the 33 cancer types is also found by the authors and may indicate a serious data issue. The finding may seem less surprising, considering the fact that tissue appearances can differ vastly between cancer types. This is particularly problematic as for some of these cancer types only very few samples were present in the training data. For some there were only about 20 observations in the training data. Although state-of-the-art performance is claimed by the authors, the previously mentioned aspects do raise concerns about the robustness of their results. Besides, facing serious data quantity issues for some cancer types, results could simply be driven by the setup of their analysis by testing the model repeatedly on different combinations of data modalities. Thereby increasing the chances to achieve better results at least for some combinations of data modalities. Moreover, the study nicely showcases that the most relevant information can often be retrieved from classical structured clinical data and that including further modalities can by contrast even distort model training when sample sizes are low compared to the variability within the data. While these concerns could certainly have been raised for the other studies as well, they simply become more apparent in \citet{MultiSurv2021} due their comprehensive and transparent analysis.

In the last part of this section we will refer to a different set of survival models by introducing the concept of Wide \& Deep NN. The idea for Wide \& Deep NN was first introduced by \citet{WideDeepNN2016}, who proposed to not only feed data inputs to either a linear or FCNN model part, but both at the same time. Applying it in the context of Recommender Systems, the initial assumption was that models need to be able to memorize as well as generalize for prediction tasks and that these aspects could be handled by the linear and FCNN part, respectively.

\begin{figure}

{\centering \includegraphics[width=1\linewidth]{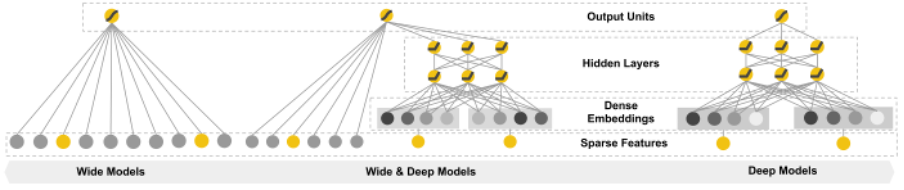}

}

\caption{Illustration of Wide \& Deep Neural Networks \citep{WideDeepNN2016}.}\label{fig:wide-deep-nn}
\end{figure}

The idea of Wide \& Deep NN is applied in the context of multimodal DL survival by \citet{Poelsterl2020} and \citet{DeepPAMM2022}. Similar to previous studies \citet{Poelsterl2020} make use of the CPH model and integrate Wide \& Deep NN in these. By contrast, \citet{DeepPAMM2022} integrate them in a different set of survival models, namely the piecewise exponential additive mixed model (PAMM). The general purpose of this model class is not only to overcome the linearity but also the proportionality constraint in the classical CPH. By dropping the proportionality assumption, these models yield piecewise constant hazard rates for predetermined time intervals. Although the two studies differ in their model setup, both studies leverage structured as well as visual data and additionally make use of a linear model head. The latter is particularly interesting as it is this additive structure in the last layer of the models which preserves interpretability. Thus, they obtain models that not only have the flexibility for accurate predictions themselves but which are also able to recover the contributions of single variables to these predictions.

Although, Wide \& Deep NN are advantageous due to their flexibility and interpretability, special care needs to be taken regarding a possible feature overlap between the linear and NN part as it can lead to an identifiability problem. This can be illustrated by considering the case that a certain feature \(x\) is fed to the linear as well as the FCNN model part. Because of the Universal Approximation Theorem for Neural Networks, it is known that the FCNN part could potentially model any arbitrary relation between the dependent and independent variable (\(d(x)\)). However, this is what raises the identifiability issue as the coefficients (\(\beta\)) of the linear part could theoretically be altered arbitrarily (\(\widetilde{\beta}\)) without changing the overall prediction when the weights of the NN (\(\widetilde{d}(x)\)) are adjusted accordingly.

\[
x\beta + d(x) = x\widetilde{\beta} + d(x) + f(x) = x\widetilde{\beta} + \widetilde{d}(x)
\]
Generally, there are two ways to deal with this identifiability problem. The first possibility would be to apply a two-stage procedure by first estimating only the linear effects and then applying the DL model part only on the obtained residuals. An alternative way would be to incorporate orthogonalization within the model, thereby performing the procedure in one step and allowing for efficient end-to-end training. The latter was proposed by \citet{SSDDR2020} and utilized in the DeepPAMM model by \citet{DeepPAMM2022}. The next section will go into more detail about the two possibilities to solve the described identifiability issue and proceed by discussing further applications of multimodal DL in other scientific fields.

\hypertarget{multimodal-dl-in-other-scientific-fields}{%
\subsubsection{Multimodal DL in Other Scientific Fields}\label{multimodal-dl-in-other-scientific-fields}}

After having seen multiple applications of multimodal DL in survival analysis which predominantly occurs in the biomedical context, we will now extend the scope of the chapter by discussing further applications of multimodal DL related to the field of economics and finance. While structured data has traditionally been the main studied data source in these fields, recent research has not only focused on combining both structured and unstructured data, but also on ways to replace costly collected and sometimes scarcely available structured data with freely available and up-to-date unstructured data sources using remote sensing data. Before examining these approaches, we will first go into more detail about the model proposed by \citet{SSDDR2020}, which not only introduced a new model class in the context of multimodal DL but also offered a method to efficiently solve the above mentioned identifiability problem.

As previous research exclusively focused on mean prediction, uncertainty quantification has often received less attention. \citet{SSDDR2020} approach this by extending structured additive distributional regression (SADR) to the DL context. Instead of learning a single parameter e.g.~the mean, SADR provides the flexibility to directly learn multiple distributional parameters and thereby natively includes uncertainty quantification. It is nevertheless possible to only model the mean of the distribution, which is why SADR can be regarded as a generalization of classical mean prediction. \citet{SSDDR2020} now extend this model class by introducing a framework that can model these distributional parameters as a function of covariates via a linear, generalized additive (GAM) or NN model. All distributional parameters are resembled in a final distributional layer (output layer). An illustration of their so-called Semi-Structured Deep Distributional Regression (SSDDR) is given in the figure below.

\begin{figure}

{\centering \includegraphics[width=1\linewidth]{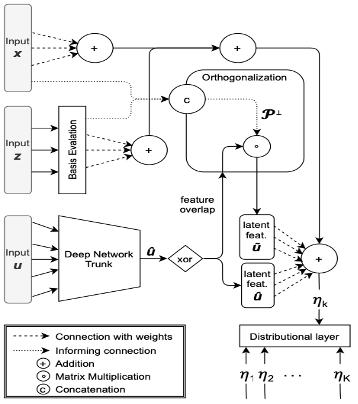}

}

\caption{Architecture of SSDDR (X+Z (Struct.) and U (Unstruct.) Data) \citep{SSDDR2020}.}\label{fig:SSDDR}
\end{figure}

If the mean is now modeled by both a linear and DNN part and the same feature inputs are fed to both model parts, we are in the setting of Wide \& Deep NN. As illustrated above, such feature overlaps give rise to an identifiability issue. The key idea to mitigate this problem from \citet{SSDDR2020} was to integrate an orthogonalization cell in the model, that orthogonalizes the latent features of the deep network part with respect to the coefficients of the linear and GAM part if feature overlaps are present. More precise, in case \(\boldsymbol{X}\) contains the inputs, that are part of the feature overlap, the projection matrix \(\boldsymbol{\mathcal{P}^{\perp}}\) projects into the respective orthogonal complement of the linear projection which is on the column space spanned by \(\boldsymbol{X}\). This allows backpropagation of the loss through the orthogonalization cell and therefore enables end-to-end learning. As the linear and GAM effect channels are directly connected to the distributional layer, the orthogonalization cell is therefore able to preserve the interpretability of the model.

Another way of orthogonalizing feature representations is by applying a two-stage procedure as described above. \citet{Law2019} utilize this procedure to make their latent feature representations retrieved from unstructured data orthogonal to their linear effect estimates from structured data. More specifically, they try to accurately predict house prices in London using multimodal DL on street and aerial view images as well as tabular housing attributes. Applying the two-stage procedure they aim at learning latent feature representations from the image data which only incorporate features that are orthogonal to the housing attributes. Thereby, they limit the chances of confounding in order to obtain interpretable housing attribute effects. Conducting a series of experiments, they find that including image data next to the tabular housing data does improve the prediction performance over single modality models albeit structured data remains the most relevant single data source. As a next step, they test their models with different model heads as depicted in the figure below to explore their respective potentials. Although fully nonlinear models with a DNN as model head generally offer larger modeling flexibility, as they can incorporate interactions, they achieved only slight performance gains over the semi-interpretable models with additive linear model heads. This is particularly interesting as the latter additionally preserve the often desired interpretability of effects. As the semi-interpretable models perform reasonably well, the authors argue that it is indeed possible to obtain interpretable models without losing too much on the performance side.

\begin{figure}

{\centering \includegraphics[width=1\linewidth]{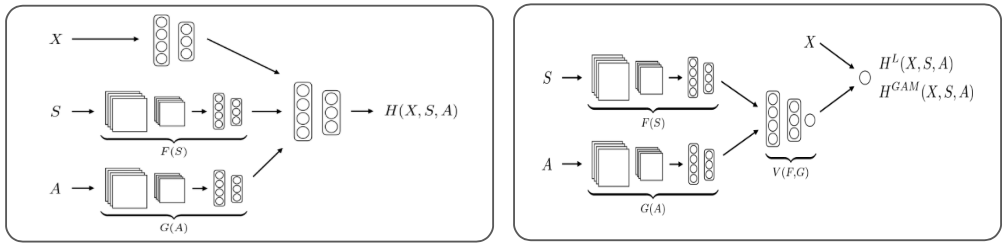}

}

\caption{Fully Nonlinear and Semi-Interpretable Models (X (Struct.) and S+A (Unstruct.) Data) \citep{Law2019}.}\label{fig:model-heads}
\end{figure}

In the last part of this section, we will allude to several other promising approaches that did pioneering work related to multimodal DL. While most of them use unstructured data sources such as remote sensing data, some do not specifically include structured data. They are still covered in this chapter to give the reader a broad overview of current research in the field. Moreover, structured data could easily be added to each of these models, but often studies intentionally avoid the use of structured data sources as they are sometimes scarcely available due to the cost of data collection. Besides availability, structured data such as household surveys is often irregularly collected and differs vastly between countries, making large scale studies impossible. Therefore, different studies have tried to provide alternatives to classical surveys by applying DL methods on freely available unstructured data sources. While \citet{Jean2016} use night and daylight satellite images to predict poverty in several African countries, \citet{Gebru2017} use Google Street View images to estimate socioeconomic attributes in the US. Both deploy the classical DL framework such as CNNs to retrieve relevant information from image data for the prediction task. Achieving reasonable prediction results while keeping analysis costs at low levels, both studies outline the potential of their proposed methods as being serious alternatives to current survey based analysis.

Other studies such as \citet{DeepGPYou2017} and \citet{Sirko2021} proposed DL frameworks for satellite imagery in contexts where labelled data is normally scarce. While \citet{DeepGPYou2017} use Deep Gaussian Processes to predict corn yield in the US, \citet{Sirko2021} apply CNNs to detect and map about 516 million buildings across multiple African countries (around 64\% of the African continent). Besides being of great importance for applications such as commodity price predictions or financial aid distribution, the results of the two studies could easily be combined with other structured data sources and thereby could constitute a form of multimodal DL with high potential.

\hypertarget{conclusion-and-outlook}{%
\subsection{Conclusion and Outlook}\label{conclusion-and-outlook}}

In the previous sections we have come across various methods of multimodal DL that can deal with both structured and unstructured data. While these often differed substantially in their approach, all of them had in common that they tried to overcome limitations of classical modeling approaches. Examining several of them in detail, we have seen applications of different fusion strategies of data modalities and thereby touched upon related concepts such as end-to-end learning. The issue of interpretability was raised along several examples by discussing the advantages of different model heads as well as ways to solve identifiability problems using orthogonalization techniques.

It was indeed shown that it is possible to obtain interpretable models that are still capable of achieving high prediction performances. Another finding of past research was that end-to-end learning frequently showed to be superior compared to methods which learn feature representation via independent models or simply retrieve information via expert knowledge. Furthermore, research that actually conducted a comparison between their proposed multimodal DL and single modality models, almost always found their proposed multimodal model to outperform all models which were based on single modalities only. Nevertheless, within the class of single modality models, those using only structured data usually performed best. This leads to the conclusion that structured data often incorporates the most relevant information for most prediction tasks. By contrast, unstructured data sources may be able to add supplementary information and thereby partially improve performances.

While there certainly has been a lot of progress in the field of multimodal DL, conducted analyses still have their limitations which is why results need to be considered with care. Although most research finds their proposed multimodal DL models to achieve excellent performances, not all of them conduct benchmarking with regard to single modality models. Thereby, they limit the possibility to properly evaluate actual improvements over classical modeling approaches. Another aspect that may raise concerns regarding the reliability of results is that multimodal DL models such as most deep learning models have multiple hyperparameters. Together with the flexibility of choosing from a wide variety of data modalites, it opens up the possibility to tune the multimodal models in various ways. Thereby making it possible that actual performance improvements may only be existent for certain configurations of the model as well as combinations of data modalities. This problem is likely to be empathized for studies using only small datasets. Small datasets are especially common in the biomedical context where image data of certain diseases is normally scarce. On top of the previously mentioned aspects, publication bias may be a large problem in the field as multimodal DL models that do not show improvements over single modality or other existing benchmark models, are likely to not be published.

Although there might be concerns regarding the robustness and reliability of some results, past research has surely shown promising achievements that could be extended by future research. While small sample sizes especially for unstructured data such as clinical images were outlined as a great limitation of past research, more of such data will certainly become available in the future. As deep learning methods usually require large amounts of training data to uncover their full potential, the field will probably see further improvements once sufficiently large datasets are available. Hence, including only an increasing number of modalities with limited samples in the models will likely be insufficient. Instead, the most promising approach seems to be incorporating sufficiently large data amounts of certain unstructured and structured data modalities that contain relevant information for the problem at hand.

\hypertarget{c03-03-multi-purpose}{%
\section{Multipurpose Models}\label{c03-03-multi-purpose}}

\emph{Author: Philipp Koch}

\emph{Supervisor: Rasmus Hvingelby}

In this chapter, we will broaden the focus to include multitask learning additionally to multimodal learning. We will call this approach multipurpose models.
Many multipurpose models have been introduced in recent years (\citet{Kaiser2017}, \citet{Hu2021}, \citet{Wang2022}, \citet{Reed2022}), and the field gained attention.
First, we will provide an in-depth overview of existing multipurpose models and compare them. In the second part, challenges in the field will also be
discussed by reviewing the Pathways proposal \citep{Dean21} and promising work addressing current issues for the progress of multipurpose models.

\hypertarget{prerequisites}{%
\subsection{Prerequisites}\label{prerequisites}}

At first, we will define the concept of multipurpose models and lay out the necessary prerequisites to make the later described models more accessible.
We will introduce the definition of multipurpose models and further concepts that this book has not covered so far.

\hypertarget{multitask-learning}{%
\subsubsection{Multitask Learning}\label{multitask-learning}}

After the extensive overview of multimodal learning \protect\hyperlink{c02-00-multimodal}{in the previous chapter}, we now need to introduce multitask learning as another concept to define
multipurpose models.

Multitask learning (\citet{Caruana1997}, \citet{Crawshaw2020}) is a paradigm in machine learning in which models are trained on multiple tasks simultaneously.
Tasks are the specific problems a model is trained to solve, like object recognition, machine translation, or image captioning. Usually, this happens using a single model,
which does not leverage helpful knowledge gained from solving other tasks. It is assumed that different tasks include similar patterns that the model can exploit and use
to solve other tasks more efficiently. The equivalent in human intelligence is the transfer of knowledge for new tasks since humans do not need to learn each task from scratch
but recall previous knowledge that can be reused in the new situation. However, this assumption only sometimes holds since some tasks may require opposing resources, so
performance decreases.

Multitask learning thus aims to achieve better generalization by teaching the model how to solve different tasks so that the model learns relationships that can be used
further on. For a more in-depth overview of multitask learning, we refer to \citep{Caruana1997} and \citep{Crawshaw2020}.

\hypertarget{c03-03-MoE}{%
\subsubsection{Mixture-of-Experts}\label{c03-03-MoE}}

Another prerequisite to this chapter is the mixture-of-expert (MoE) (\citet{Jacobs1991}, \citet{Jordan1994}, \citet{Shaazer2017}) architecture, which is aimed at increasing the overall model
size while still keeping inference time reasonably low. In an MoE, not all parts of the net are used but just a subset. The \emph{experts} are best suited to deal with the input
allowing the model to be sparse.

MoE is an ensemble of different neural networks inside the layer. MoEs allow for being more computationally efficient while still keeping or even improving performance. The
neural networks are not used for every forward pass but only if the data is well suited to be dealt with by a specific expert. Training MoEs usually requires balancing the
experts so that routing does not collapse into one or a few experts. An additional gating network decides which of the experts is called. Gating can be implemented so that
only \emph{K} experts are used, which reduces the computational costs for inference by allowing the model to be more sparse.

\hypertarget{c03-03-EA}{%
\subsubsection{Evolutionary Algorithms}\label{c03-03-EA}}

An evolutionary algorithm is used to optimize a problem over a discrete space where derivative-based algorithms cannot be applied to. The algorithm is based on a population
(in the domain to be optimized) and a fitness function that can be used to evaluate how close a member of the population is to the optimum. Parts of the population are chosen
to create offspring either by mutation or recombination. The resulting population is then evaluated with respect to their fitness function, and only the best-suited individuals are kept.
The same procedure is repeated based on the resulting population until a specific criterion is met (e.g., convergence). While evolving the population, it is necessary to balance exploration
and exploitation to find the desired outcome. Since EAs are research topics themselves and may vary heavily, we refer to \citep{Baeck1993} and, more recently, to \citep{Doerr2021} for further insights.

\hypertarget{multipurpose-models}{%
\subsubsection{Multipurpose Models}\label{multipurpose-models}}

Now multipurpose models can be defined as multimodal-multitask models. Akin to the underlying assumptions of both learning paradigms, it can also be deduced
that multipurpose models mimic human intelligence by marrying the concepts of multiple perceptions and transferring knowledge about different tasks for better generalization.

\hypertarget{overview-of-mulitpurpose-models}{%
\subsection{Overview of Mulitpurpose Models}\label{overview-of-mulitpurpose-models}}

In this section, we will closely examine existing multipurpose models. The main focus will be on how combining different modalities and tasks is achieved. At the end of
this section, all models will be compared to provide a comprehensive overview of promising research directions.

\hypertarget{multimodel}{%
\subsubsection{MultiModel}\label{multimodel}}

The first prominent multipurpose model is the so-called MultiModel \citep{Kaiser2017}. This model, from the pre-transformer era, combines multiple architectural approaches
from different fields to tackle both multimodal and multiple tasks. The model consists of four essential modules: The so-called modality nets, the encoder, the I/O Mixer, and the decoder.

Modality nets function as translators between real world data and a suitable representation for the inner modules. They also follow the purpose of back-translating,
from the representation to the real world, to create output. For language tasks, the modality net is a tokenizer that outputs the appropriate embeddings, while for vision tasks,
convolution operations transform the images into the proper representation. Furthermore, there are also nets for audio and categorical modalities. The modality nets embed the input
into a unifying vector space which can be passed to the encoder. To produce the output, the representations from the decoder are fed into another modality net to produce the output.
Language and categories are the only target modalities that have respective modality nets.

The core model consists of the encoder, the I/O mixer, and the decoder. Input is passed from the modality nets to the encoder first. Subsequently, the encoder passes its output further
to the I/O mixer and the decoder. The decoder produces the output sequence. However, producing an autoregressive sequence requires knowledge of the previously generated sequence. Thus
the output of the decoder is also read by the I/O mixer, which provides the decoder with the necessary information about the previous sequence. The I/O mixer passes its output back to the
decoder to provide the necessary information. The decoder and I/O mixer require modality nets to read and write in the target modality. The encoder consists of multiple convolution operations
and a \protect\hyperlink{c03-03-MoE}{mixture-of-expert} layer. The I/O mixer and the decoder combine their dual input using cross-attention. A positional encoding conceptually similar to the one in transformers \citep{vaswani2017attention}
is used for the attention mechanism.

MultiModel was trained on eight datasets, from which six were from the language modality and COCO \citep{mccoco} and ImageNet \citep{ImageNet} from vision. For training, four experts in the MoE layers
were used. The combined trained MultiModel on ImageNet and machine translation were below state-of-the-art (SOTA) models. Also, the combined model did not achieve significantly better results
than a specialist model, which is the same model but trained solely on one task. However, it was found that the combined model did perform much better on a low-resource task than the respective
specialist model.

MultiModel offers a pre-transformer approach to deal with different modalities on multiple tasks; although it is only used to generate text and clasification, the setup allows extending to
other modalities easily.

\begin{figure}

{\centering \includegraphics[width=0.8\linewidth]{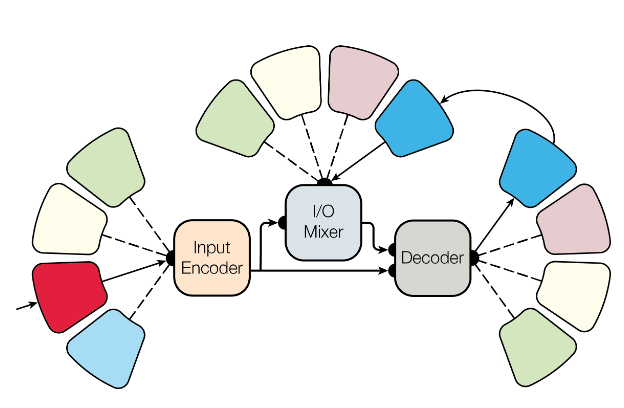}

}

\caption{Architecture of \emph{MultiModel}. The outer boxes without text are the modality nets. From \citet{Kaiser2017}.}\label{fig:multimodel}
\end{figure}

\hypertarget{unified-transformer-unit}{%
\subsubsection{Unified Transformer (UniT)}\label{unified-transformer-unit}}

A more recent multipurpose model is UniT (Unified Transformer) \citep{Hu2021}. UniT is built upon the transformer architecture, in which both encoder and decoder are used.

To account for multimodality and multitasking, the basic transformer \citep{vaswani2017attention} is enhanced. The encoder part of UniT consists of two modality-specific encoders since the initial
setup is aimed at the modalities of text and vision. However, more modality-specific encoders may be added. For the case of language, a BERT model \citep{Devlin2018} is used, while a
detection transformer (DETR) \citep{Carion2020} encoder is used for vision. DETR uses a particular approach to feed images to the encoder. At first a CNN is used to create a lower dimensional
representation of the input image, which is then reorganized as a sequence. This sequence is then fed into the encoder following \citet{vaswani2017attention}. The {[}CLS{]} token is also used in the BERT
encoder, which is also included in the output sequence of the encoder. A task-specific token is additionally added to the input of the encoders. The output of the encoders is then concatenated
to form a single sequence. The decoder is fed with this sequence and a task-specific query. Since the decoder architecture sticks to the DETR model, the decoder does not produce a sequence
autoregressively. Instead of taking the previously produced sequence (autoregressively) as input, the decoder is fed with the task-specific query vectors instead, thus producing a uniform output. On top of the
decoder are task-specific heads needed to transform the decoder output into the desired shape for the specific task.

For training, the object detection task requires bounding box loss from DETR, while the other tasks use cross-entropy loss.

In experiments, UniT was evaluated against a single-task version of itself. The
general model outperformed the specialist one on multimodal tasks but was outperformed on unimodal tasks by the specialist UniT. UniT was furthermore also outperformed by SOTA models, although
the numbers remained comparable.

Even though UniT does not achieve SOTA or consistently outperforms its specialist version, it is a powerful method to achieve a simple multipurpose model. By using available encoder models, it
is easily extendable.

\begin{figure}

{\centering \includegraphics[width=0.8\linewidth]{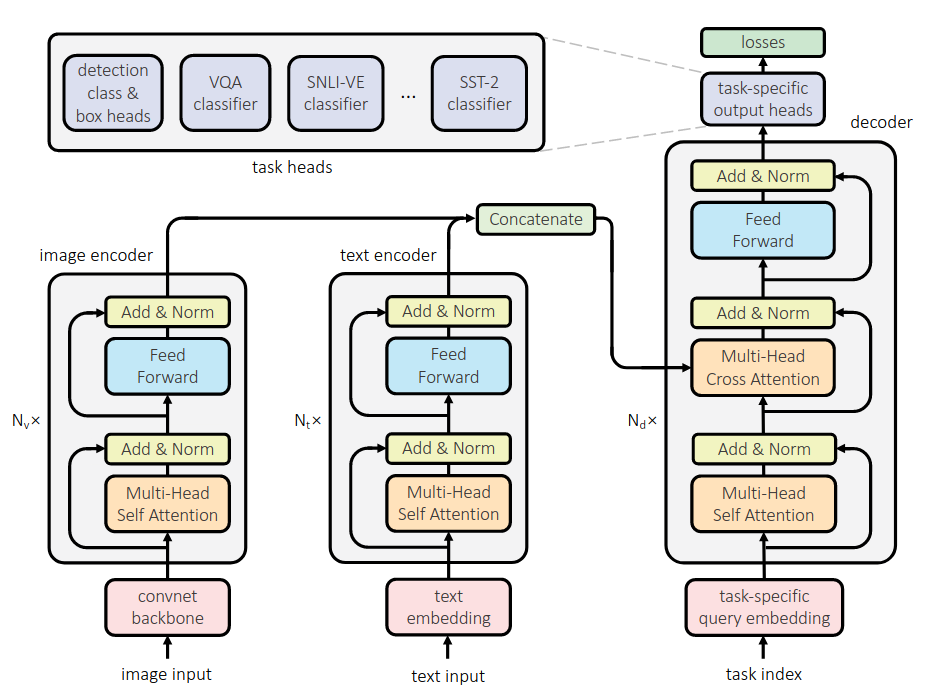}

}

\caption{Modified transformer for UniT. The decoder follows the implementation of DETR \citep{Carion2020}. From \citet{Hu2021}.}\label{fig:unit}
\end{figure}

\hypertarget{ofa---sequentialization-is-all-you-need}{%
\subsubsection{OFA - Sequentialization is All You Need}\label{ofa---sequentialization-is-all-you-need}}

Another multipurpose transformer is OFA (Once For All) \citep{Wang2022}. To utilize the sequence-to-sequence (seq2seq) architecture of the transformer, all input is transformed into a seq2seq problem.

While MultiModel and UniT use specific modules for a modality (modality nets and modality-specific encoders), a different approach is used for OFA. All input is sequentialized and embedded in a shared
representation space. Since tokenizing an image using a vocabulary is not feasible, a similar approach to ViT \citep{dosovitskiy2020image} is used (where input is flattened to 16 x 16) to
obtain a sequence of \(P\) representations. These representations are in the same dimension as the token embeddings from text input, which are tokenized using Byte-Pair-Encoding
\citep{sennrich-etal-2016-neural}. After feeding the embeddings through the encoder, the decoder produces the output as a sequence again. However, in this case, images are represented as a sequence of
tokens, similar to the image-patch vocabulary in DALL-E \citep{pmlr-v139-ramesh21a}. Furthermore, a special sequence for bounding boxes is also used for object detection and recognition. To generate
the task-specific solution, it is thus required that another model is used to generate the images based on the tokens and to visualize the bounding boxes based on the obtained coordinates.

Since OFA is an autoregressive model (the probability for the next token is predicted based on the previously produced tokens and the input provided), the objective is based on cross-entropy loss.
OFA was trained on different crossmodal tasks: visual grounding, grounded captioning, image-text matching, image captioning, and visual question answering. Further unimodal tasks for training did
include: image infilling, object detection, and text reconstruction as in BART \citep{lewis-etal-2020-bart}.

OFA outperformed SOTA models on cross-modal tasks like image captioning, visual question answering, visual entailment, and visual grounding. On uni-modal tasks, OFA performed well, although it
did not outperform SOTA models. OFA showed additional transfer capabilities to unseen tasks, which were presented with an additional description to solve the task in a few-shot manner. Although
the results were satisfactory, the model was not evaluated against a specialist baseline.

OFA proved to be a powerful model that is capable of using the entire transformer architecture by sequentializing all input and thus producing tokenized output.

\begin{figure}

{\centering \includegraphics[width=1\linewidth]{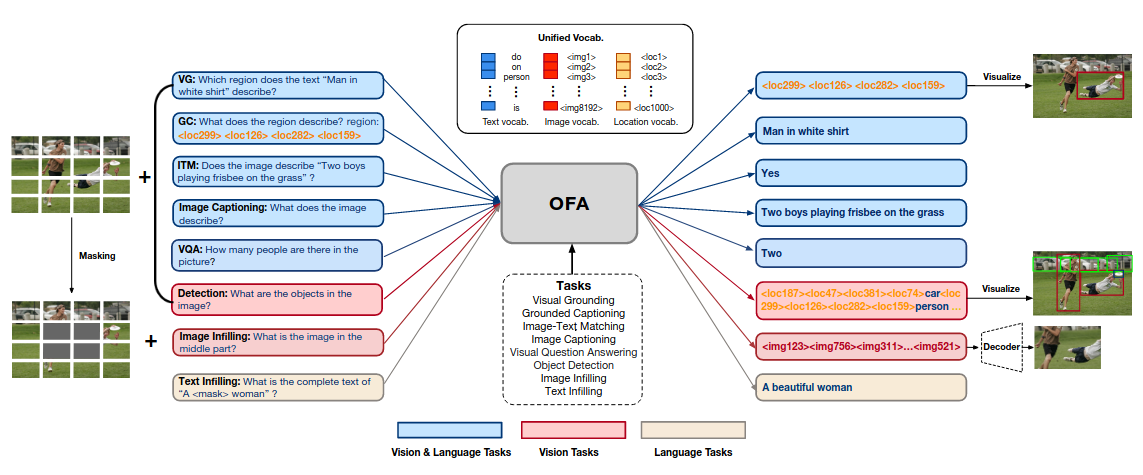}

}

\caption{\emph{OFA}, the different input and output concepts can be seen here. From \citet{Wang2022}.}\label{fig:ofa}
\end{figure}

\hypertarget{gato---a-generalist-decoder}{%
\subsubsection{Gato - A Generalist Decoder}\label{gato---a-generalist-decoder}}

Another model that utilizes the seq2seq approach in transformers is Gato \citep{Reed2022}. The model can be used as a language model, an agent to play games, and an agent to control robotics.

As in OFA, problems are transformed into a seq2seq problem, on which a transformer (decoder only) is applied. Every input from text, vision, robotics, and games is represented sequentially.
Visual input is encoded using a flattened sequence of 16x16 patches fed into a ResNet \citep{ResNet}, while text input is tokenized using SentencePiece \citep{kudo-richardson-2018-sentencepiece}.
Furthermore, discrete values like buttons in games and continuous data like movements from robotics are tokenized in a vocabulary too. To represent
all modalities sequentially, the different tokens are concatenated. A separator token ``\textbar{}'' is added to distinguish the observations from the following action, so that a sequence looks simplified as the following:

\[\left [ ... \left [ x_{\textrm{Text}}, x_{\textrm{Images}}, x_{\textrm{Discrete and Continuous Values}}, |, y_{\textrm{Action}} \right ]_i, ... \right ]\]

By using this approach, the transformer can predict the next action autoregressively since it is a sequential problem. In the case of text, the action token is also a text token. Since it is
only necessary to predict the action based on the previous values, a mask function is added to the cross-entropy loss function, which masks the previous values so that only the next action
is predicted and not the conditions for the action. The masking function is always one for text since every previous text token is necessary for language modeling.

Gato was evaluated on reinforcement-learning-based (RL) tasks against specialist RL agents, where Gato performed worse than the specialist agents. On unseen tasks, Gato required fine-tuning since
few-shot learning is not feasible due to the input length restrictions in transformers. However, the results were mixed. Some improvements were possible, and the expert was outperformed, but
in other cases, massive fine-tuning efforts only led to small gains. It was found that the generalist agent outperformed the specialist agent (particularly trained for this task) most of the time.
Only at the specific Atari Boxing \citep{atari} task, Gato was outperformed by the specialist Gato model. Both performed much lower than another task-specific model used as a baseline. In robotics, Gato showed comparable
behavior to the baseline SOTA model. Additionally, Gato also showed capabilities in image captioning and dialogue modeling, although these aspects were not elaborated further.

Like OFA, Gato can sequentialize all input and produce a sequential output that can be back-transformed to solve a task. It was shown that Gato could sometimes transfer knowledge on unseen tasks and outperform the specialist agent most of the time.

\hypertarget{comparison}{%
\subsubsection{Comparison}\label{comparison}}

Although many tasks and modalities lead to a curse of dimensionality for comparison, the architectures and the respective modifications of the introduced systems remain simple to compare.

A trend toward seq2seq models can be seen with MultiModel, OFA, and Gato solving tasks in a seq2seq manner. The most prominent similarity is the transformer architecture used entirely
(encoder \& decoder) in OFA and truncated (decoder only) in Gato. Another significant similarity between both architectures is the use of a particular ordering of input and output.
In Gato, the sequence is organized around predicting an action using a special token, while OFA produces a sequence as a solution which can be the bounding box or the sequence of an image
to be fed in the generator module. While Gato can solve tasks from robotics and game playing, OFA can also generate images. However, both architectures require specific modules to decode the tokens into
the respective modality.

Gato and OFA both use a shared representation space. Minor details differ, so the image tokenization process is different, and additionally, Gato can encode more modalities than the published
version of OFA (although extending OFA is theoretically simple).

MultiModel also show some familiar characteristics. The architecture is from the pre-transformer age but also brings many characteristics of the transformer architecture, like the use of
attention, positional encodings, and encoder-decoder. Since the output in the presented version only produced text or classification separately, there is no need for special orderings used
in OFA and Gato.
The necessity to produce the modality-specific output in modality nets is similar to the generator module in OFA that produces images. However, the tokens are already produced in an
intermediate step in OFA, while the modality nets are crucial to producing the final output in MultiModel. UniT follows an entirely different approach that is more pragmatic by leveraging the
contextual capabilities of the transformer decoder.
\emph{M} modalities can be encoded as a sequence on which the transformer decoder fuses the modalities and learns the relationships. The use of special tokens for each task and task-specific heads,
focus the model on the requested task yet also requires tuning the model specifically.

None of the models besides OFA achieved SOTA results. Compared to specialist models, the general models were comparable in their results (Gato, UniT, MultiModel). MultiModel, OFA, and Gato showed transferability on
low-resource or unseen tasks. However, more research in this direction is highly recommended. MultiModel was only compared on a low-resource task against a specialist model,
and OFA was not compared to another model for the unseen task. Gato performed better than a specialist model, trained from scratch on most unseen tasks, but failed against the untrained specialist
model in Atari Boxing.

\begin{longtable}[]{@{}
  >{\raggedright\arraybackslash}p{(\columnwidth - 12\tabcolsep) * \real{0.1429}}
  >{\raggedright\arraybackslash}p{(\columnwidth - 12\tabcolsep) * \real{0.1429}}
  >{\raggedright\arraybackslash}p{(\columnwidth - 12\tabcolsep) * \real{0.1429}}
  >{\raggedright\arraybackslash}p{(\columnwidth - 12\tabcolsep) * \real{0.1429}}
  >{\raggedright\arraybackslash}p{(\columnwidth - 12\tabcolsep) * \real{0.1429}}
  >{\raggedright\arraybackslash}p{(\columnwidth - 12\tabcolsep) * \real{0.1429}}
  >{\raggedright\arraybackslash}p{(\columnwidth - 12\tabcolsep) * \real{0.1429}}@{}}
\toprule()
\begin{minipage}[b]{\linewidth}\raggedright
Model
\end{minipage} & \begin{minipage}[b]{\linewidth}\raggedright
Approach
\end{minipage} & \begin{minipage}[b]{\linewidth}\raggedright
Modalities
\end{minipage} & \begin{minipage}[b]{\linewidth}\raggedright
Outperformed Specialist Model?
\end{minipage} & \begin{minipage}[b]{\linewidth}\raggedright
Unseen Tasks?
\end{minipage} & \begin{minipage}[b]{\linewidth}\raggedright
Number of Parameters
\end{minipage} & \begin{minipage}[b]{\linewidth}\raggedright
Year
\end{minipage} \\
\midrule()
\endhead
OFA & Seq2Seq & Vision, Text & & Yes & 33M-930M & 2022 \\
Gato & Seq2Seq & Vision, Text, Robotics, Discrete Entities (e.g., Buttons) & In most cases & Yes & 79M-1.18B & 2022 \\
UniT & \emph{m} Encoders, task-specific head & Vision, Text & No & No & 201M & 2021 \\
MultiModel & Different \emph{modality nets} for Seq2Seq & Vision, Text, Audio, Categorical & Comparable & Excelled on low resource task & Unknown & 2017 \\
\bottomrule()
\end{longtable}

Comparing the models among each other becomes difficult with more modalities and tasks, which is its own curse of dimensionality. For example, Gato also included robotics and RL,
which none of the other models included. MultiModel also has a modality net for sound, while UniT and OFA only worked for vision and text. Further research into the comparability of
multipurpose models becomes essential.

\hypertarget{pathways-and-promising-works}{%
\subsection{Pathways and Promising Works}\label{pathways-and-promising-works}}

Although models have become more capable of solving complex tasks, significant limitations remain.
A persisting issue in current deep learning is the necessity to train from scratch and disregard already obtained knowledge, which is highly ineffective compared to human intelligence. Another
issue arises from the evergrowing, dense networks that requires more and more resources.

In this section, we will review the Pathways proposal \citep{Dean21} and promising techniques to address these issues. Overcoming these problems would be especially beneficial for multipurpose
models. Reusability of knowledge is crucial for the multitask perspective, and improving the performance of potentially billion-parameter-sized models will also have a significant positive impact.

\begin{figure}

{\centering \includegraphics[width=0.8\linewidth]{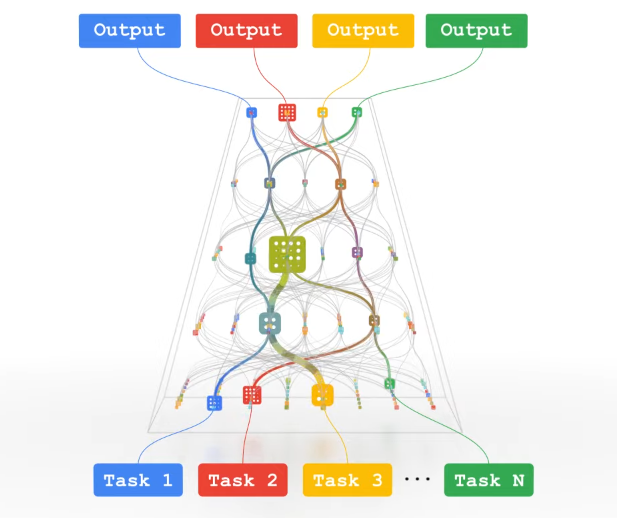}

}

\caption{Concept of Pathways. Different tasks follow different paths to different expert models. From \citet{Dean21}, \href{https://www.youtube.com/watch?v=Nf-d9CcEZ2w}{Screenshot August 31th 2022}.}\label{fig:pathways}
\end{figure}

\hypertarget{pathways-proposal}{%
\subsubsection{Pathways Proposal}\label{pathways-proposal}}

Pathways \citep{Dean21} follows a different idea than previously seen methods. The model consists of a large graph through which data can be forward passed. The nodes of the network are neural
networks themselves. A pass through this network does not include passing all nodes and thus not all neural networks, but only a few. The pass follows a specific path from one entry to the
network's exit. The underlying idea behind this is similar to the mixture-of-expert models described previously. Only the specific networks dedicated to solving a problem are to be activated during inference.

At this point, it is necessary to recall that multitask learning aims to generalize better on new tasks since the knowledge about previously learned tasks can be applied. This idea is the
foundation of Pathways too, where specialist networks (nodes) are combined in a larger network. It is assumed that the model's generalization capabilities increase significantly by finding
an appropriate path for a task to the appropriate expert nodes. In this setup, the particular task-specific problem-solving capabilities are combined. Furthermore, multimodality is also
considered as a potential extension. Adding more modalities might not be a difficult problem considering the architecture of the previously introduced transformer-based models.
Overall the approach of a sparse model combining multiple experts offers many opportunities to combine modalities and reuse task-specific capabilities. The sparsity of the model offers
decreased inference time since only few parts of the networks are activated during inference.

Another aspect of the Pathways proposal includes the improvement of current hardware limitations. It is already observable that Moore's Law (\emph{each n years, the compute capacity doubles}) has
been slowing down substantially, while deep learning research has grown exponentially in the late 2010s \citep{Dean20}. Thus, hardware also needs to be adapted to the growing demand in deep learning.
In the context of the pathway proposal, a novel framework for Google data centers has been introduced, aiming to reduce overhead during computation and access specific parts of the model to
utilize the technical advantages of sparse networks. As opposed to dense models where a whole model must be accessed, with sparse networks it is not necessary to use the whole network but only chunks of it. So far, two large pre-trained models have been introduced based on the new training framework. One is the Pathways Language Model (PaLM) {[}Chowdhery2022{]},
which is currently the largest language model using 540 billion parameters, Minerva \citep{Lewkowycz2022}. Minerva is based on PaLM, and Parti \citep{parti},

\hypertarget{pathnet}{%
\subsubsection{PathNet}\label{pathnet}}

An earlier approach for a sparse multitask network, which looks deceptively similar, is PathNet \citep{Fernando2017}. PathNet is a training concept that reuses knowledge from a previously learned
task without the risk of catastrophic forgetting (knowledge is overwritten), thus using solely the positive aspects of multitask learning. The objective of PathNet consists of a evolutionary
algorithm (EA).

Neural networks are often depicted as a graph in which the input is directed to all nodes in hidden layers, and their output is again passed to all nodes in the next hidden layer or an output layer.
In the case of PathNet, each node is itself a neural network. The training algorithm finds the best paths for a specific task through the network.

At first random paths through the network are initialized, then the paths are trained for \emph{T} epochs. After training, the paths are evaluated against each other. The winning path overwrites
the losing path. However, to achieve exploration, the overwritten path is mutated by randomly including neighbors of the winning path. Until a specific criterion to stop
(e.g., number of epochs) is reached, the current paths are frozen so that no more modifications to the parameters of the networks on this path are possible. All other parameters
are newly initialized again. Also, a different, task-specific head is initialized. The same procedure is now done again for the next task. Then, the main difference is that the previously obtained path,
including the trained networks, is frozen during training so that the model can transfer knowledge from the previous task to the new task. The model then finds appropriate paths throughout the
network until the stopping criterion is met again.

PathNet was evaluated on supervised learning tasks and RL scenarios. Learning from scratch and fine-tuning a PathNet, were chosen as a baseline. For fine-tuning, the first path was chosen as a
base model that was fine-tuned on the second task. Overall, PathNet improved training time and prediction quality for the second task compared to standard fine-tuning and learning from scratch.
PathNet has shown that different tasks can reuse the knowledge from training on previous tasks without suffering from catastrophic forgetting.

\begin{figure}

{\centering \includegraphics[width=0.8\linewidth]{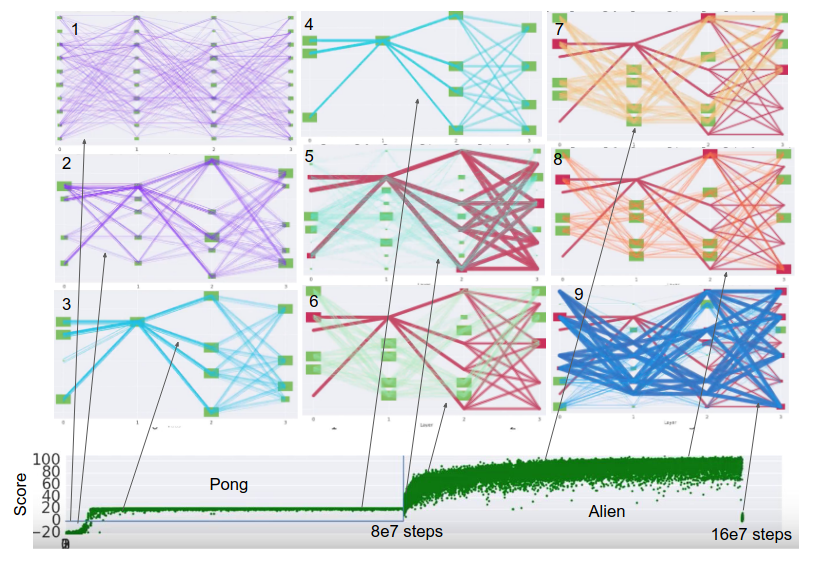}

}

\caption{Training \emph{PathNet} on two tasks. At first random paths are initialized (1), then trained (2-3) and fixed (4). The same procedure is repeated for the next paths using the previously
fixed paths and new parameters in all other nodes (5-9). From \citet{Fernando2017}.}\label{fig:pathnet}
\end{figure}

\hypertarget{limoe}{%
\subsubsection{LIMoE}\label{limoe}}

LIMoE (Multimodal Contrastive Learning with LIMoE: the Language-Image Mixture of Experts) \citep{Mustafa2022} combines text and vision input using a MoE-enhanced transformer encoder.

While previous methods used two models (two-tower) to encode modalities, LIMoE is solely based on one model, where the modalities are processed in a single modified transformer-model (one-tower).
The text data is encoded using One-Hot-SentencePiece \citep{kudo-richardson-2018-sentencepiece} encoding, while images are
tokenized in the same way as in ViT \citep{dosovitskiy2020image} (elaborated further in the previous \protect\hyperlink{c01-02-SOTA-cv}{chapter}) to provide the input appropriately. The main difference to the standard transformer is an MoE layer where the feed-forward network
usually lies. In this layer, \emph{E} experts are used, which are themselves feed-forward-networks. For each token,\emph{K} appropriate experts will map the tokens further downstream. The routing is computed
by a gating net network, which decides which \emph{K} experts are called. Another feature here is a fixed-length buffer for each expert in the MoE layer. This buffer is used to store tokens before an
expert network processes them, assuming that the allocations of tokens for each expert are balanced. If it is impossible to buffer tokens for the experts, the tokens will be dropped. To process the
more important tokens first, Batch Priority Routing \citep{Riquelme2021} is used to provide a ranking mechanism. The output of the transformer encoder is then average pooled and subsequently multiplied with
a modality-specific weight matrix, which produces the eventual output
for the token of both modalities.

\begin{figure}

{\centering \includegraphics[width=0.8\linewidth]{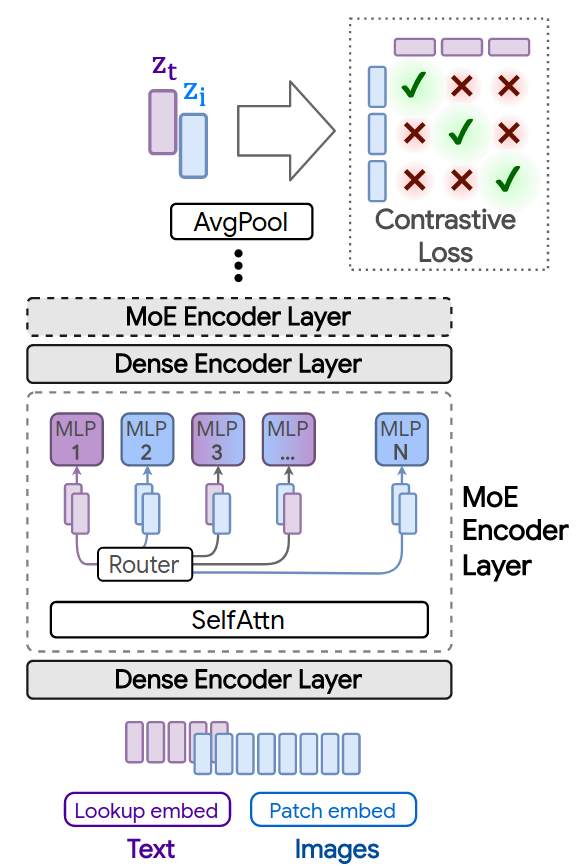}

}

\caption{Architecture of \emph{LIMoE}. From \citet{Mustafa2022}.}\label{fig:LIMoE}
\end{figure}

The model is trained using a contrastive objective. In this case, the contrastive loss aims to maximize the paired visual and textual input while minimizing all combinations of unpaired embeddings.
This objective can be achieved by using the dot-product as a similarity measure between the embeddings of both modalities, which provide a differentiable
operation through which the overall loss can be minimized.

Additionally, the pitfalls of a multimodal MoE are also considered. One challenge in MoE is the correct balancing of
routing to the experts, which is even more challenging when using unbalanced multimodal data. To address this issue, two new losses based on entropy are introduced. Entropy can be used as an appropriate
term since it provides a valuable number for the uniformity of the distribution, which is necessary to balance the expert assignments. The losses are aimed at controlling the allocation of experts
to tokens, which is also necessary to fulfill the assumptions for the implemented buffer. One loss considers the token level (local) routing distribution, and the other considers the overall
expert routing distribution (global). The local loss aims to achieve no uniform behavior in expert allocation such that each token is indeed assigned to specific experts. In contrast, the overall
global loss aims to achieve uniformity over all tokens to avoid a collapse in which tokens are solely assigned to a few experts which do not have the capacity to deal with all tokens. These losses
are computed for each modality. Furthermore, already available losses for training MoE models were also added to avoid known downsides of MoE models.

LIMoE was compared against similar models like CLIP \citep{radford2021learning}.
The test dataset was ImageNet \citep{deng2009imagenet} and COCO \citep{mccoco}. Overall, LIMoE-H/14 (largest model, 12 MoE-layers, 32 experts per layer) achieved strong performance considering that only
one model was used for two modalities against specialist models in two-tower setups. It was also possible to outperform CLIP by a significant margin while using minimal additional parameters.
Models that achieved similar results to LIMoE used at least twice the number of parameters for a forward pass.

LIMoE provides an example that an MoE-based model achieves impressive results in a multimodal model. Current language and vision encoding techniques are combined and married with the upsides of
the MoE-architecture, leading to a single model that can outperform current state-of-the-art models like CLIP.

\hypertarget{munet-multitask-network}{%
\subsubsection{muNet (Multitask Network)}\label{munet-multitask-network}}

muNet \citep{Gesmundo2022a} is an architecture that maximizes the reusability of previously learned knowledge by using an evoluationary algorithm to evolve a new model.
The authors address the current practice for fine-tuning, where a pre-trained model is copied and then explicitly trained on a task by overwriting previous knowledge.

An initial model is evolved by using an \protect\hyperlink{c03-03-EA}{evoluationary algorithm} to fit specific tasks, while keeping the previously learned knowledge.
Eventually, a set of models is obtained, which includes new neural networks, based majorly on the parameters of the initial model. The new modules can be seen as paths to task-specific
modifications of the initial network.

The EA of muNet starts with an initially proposed model that is mutated further on. All further mutations are stored so that after a set of candidates is available, the set can be split into
models trained for this task (active population) and models for other tasks (inactive population). These two sets become the sets of candidates for the following task-specific iterations.
Training a specific task follows three steps: Sampling candidate models, mutating, training, and evaluation. The best scoring model is added to the active population for further mutation.
A sampling algorithm accounts for exploration and exploitation to get a candidate model for subsequent mutation. The active population is ordered in a descending list based on the model's score.
Each list entry is then revisited, starting from the highest scoring model onward, so that the better performing models are considered first (exploitation). The draw probability is computed as:

\[\mathbb P(m|t) = 0.5 ^{ \#timesSelected(m, t)}\]

Where \(\#timesSelected(m, t)\) is the amount of previous mutations based on model \emph{m} for task \emph{t}). The more unsuccessful mutations the model has had before, the smaller the draw probability
becomes. Thus, exploration is emphasized by considering previous attempts and allowing other models to be preferred as well. However, if this method does not yield a candidate, a model is drawn from
the union of the inactive and active population.
Applying mutations is the next step in the algorithm. A random number of mutations are drawn from the set of possible mutations, which include:

\begin{itemize}
\tightlist
\item
  \textbf{Layer Cloning}: A layer is cloned for training. The layer's parameters are copied from the parent model so that training can continue using the same knowledge. The other layers are still used
  but are not updated. Additionally, the task-specific head layer is cloned to account for the underlying changes. In case of training on a new task, the head is also newly initialized.
\item
  \textbf{Layer Insertion}: Two layers are added to the model as residual adapters (\citep{Rebuffi2017}, {[}Houlsby2019@{]}). The second layer is zero-initialized to keep an identity function so that training
  can continue from the state before mutation.
\item
  \textbf{Layer Removal} is used to skip layers while still using all other layers of the parent model in a frozen state.
\item
  \textbf{Hyperparameter Change}: samples hyperparameters close to the ones of the parent model. A list of neighboring values is constructed from which a parameter is drawn.
\end{itemize}

Subsequently, the models are trained on the task and scored. If the mutated model is better than the parent model, it is also added to the task's set of active models. This routine is done for
all tasks iteratively and can be repeated several times. Ultimately, only the best scoring models are kept for each task, yielding a list of models each fit to a particular task.
muNet was evaluated for fine-tuning against a ViT instance, which was aimed at being the most generalizable one \citep{Steiner2021}. The evaluation benchmarks consisted of multiple classification problems
(to simulate multitasking). ViT was fine-tuned on all of these tasks as a baseline. In contrast, another ViT was evolved using muNet, on which the baseline model was evaluated again. The approach using
muNet outperformed the fine-tuned ViT while using significantly fewer parameters.

muNet offers a simple, evolutionary-based approach for fine-tuning and keeping all previously acquired knowledge safe, thus maximizing reusability.

\hypertarget{conclusion-pathways}{%
\subsection{Conclusion Pathways}\label{conclusion-pathways}}

The introduced models show promising novel features that might improve multipurpose models. However, these models can only be improved if research is done to combine the distinct concepts. PathNet
and muNet offer novel approaches to leverage already acquired knowledge, while LIMoE improves handling different modalities in a single, sparse model.
Furthermore, it also becomes necessary to conduct research into scaling these concepts up. Since the multitask-related models (PathNet and muNet) only included a few tasks, introducing more tasks
for training and testing might offer insights into how transfer between tasks succeeds and fails.

LIMoE offers a promising architecture with respect to performance. Due to the sparsity of the MoE-layer, LIMoE is faster, while it also outperforms previous dense models. Using MoE-layers in
transformers might also be a viable path for models like OFA and Gato. Combining the flexible encoding techniques of these models with the relative sparsity of LIMoE might result in even more capable
and efficient models. We, therefore, recommend further research in this direction.

Another potential path for future research is intelligent routing for evolving methods like muNet and PathNet. Evolutionary models offer a promising approach to leveraging previous knowledge.
However, the resulting models are tailored to a particular task. Novel routing techniques to send data to dedicated expert nodes in a complex network of models might help models generalize,
as was outlined in the Pathways proposal.

\hypertarget{discussion-3}{%
\subsection{Discussion}\label{discussion-3}}

We reviewed multipurpose models that have become capable of solving multiple tasks from different modalities. The transformer architecture also boosted the development in this field, in which
three of the four presented models were transformer-based and from recent years. Multipurpose models offers an opportunity to use one model instead of many different expert-models. Furthermore, some
multipurpose models (Gato, OFA) also outperformed expert-models. However, Gato also showed inferior performance on ATARI Boxing compared to competing models, indicating that research is still
required to explore the relationship between tasks. We also presented promising novel architectures that alleviate or may solve problems in current multipurpose models.
However, further issues remain that have not been solved by research to this day:

\begin{itemize}
\item
  A pitfall of models of these sizes is the low accessibility. Researchers need to access the model through an API since running these models on a few GPUs will likely be infeasible. It might be
  unlikely to see a BERT-like engagement with the community of researchers if the access to models remains limited. On the contrary, more open-source collaborations, as seen with
  \href{www.eleuther.ai}{EleutherAI} or \href{www.huggingface.co}{Huggingface}, might evolve as well as a countermovement and techniques like distillation \citep{Hinton2015} might become more critical.
\item
  Another issue with multipurpose models is the lack of metrics. Current metrics are not suited for multitask and multimodal models. Evaluation might also become harder since many different
  modalities can be used, as seen here with the robotics property of Gato, which was not used in any of the other reviewed models.
\item
  Eventually, it is also necessary to consider the societal impact. The bias problem will also become an issue in multipurpose models, especially since multiple datasets must be considered.
\item
  Also, the environmental impact of training large models needs to be considered since it is likely that larger models will yield better performance according to scaling laws \citep{Reed2022}
  but will also have a larger carbon footprint.
\end{itemize}

\hypertarget{c03-04-usecase}{%
\section{Generative Art}\label{c03-04-usecase}}

\emph{Author: Nadja Sauter}

\emph{Supervisor: Jann Goschenhofer}

\begin{figure}

{\centering \includegraphics[width=0.9\linewidth]{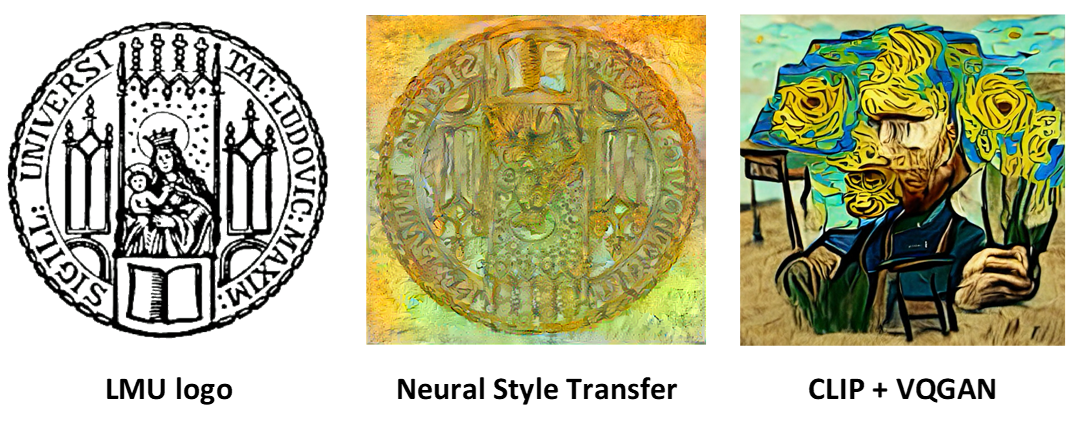}

}

\caption{LMU logo in style of Van Gogh's Sunflower painting}\label{fig:Logo}
\end{figure}

As we have seen in subsection \ref{c02-02-text2img}, computers can create images only based on text prompts via multimodal deep learning. This capability is also used in digital arts in the field of `generative art' or also known as `computer art'. The new movement comprises all artwork where the human artist cedes control to an autonomous system \citep{galanter2016generative}. In this way everyone, even artistically untrained people, can easily create pictures as the computer takes over the image generation. In some way, the computer becomes the artist with some sort of creativity, a distinct human ability. In this chapter, we want to give an overview about how computers improved over time in generating images and how this is used in the contemporary arts scene. For instance in Figure \ref{fig:Logo} we used the seal of the Ludwig Maximilians University and changed the style to Van Gogh's \href{https://wallpaperaccess.com/full/787825.jpg}{Sunflower painting} by the \href{https://www.tensorflow.org/tutorials/generative/style_transfer}{Neural Stlye Transfer Algorithm} and the method \href{https://colab.research.google.com/drive/1ZAus_gn2RhTZWzOWUpPERNC0Q8OhZRTZ\#scrollTo=FhhdWrSxQhwg}{CLIP + VQGAN} which fuses the logo with sunflowers in a Van-Gogh-style way.

\hypertarget{historical-overview}{%
\subsection{Historical Overview}\label{historical-overview}}

The first attempt to use AI to generate pictures was made by the engineer Alexander \citet{mordvintsev_2015} and his ``DeepDream'' Software. He used Convolution Neural Networks to generate very interesting and abstract images based on the activation of a layer, visualizing the patterns learned by a neural network. Below you can see a picture of a Labrador after it was processed by the DeepDream algorithm.

\begin{figure}

{\centering \includegraphics[width=0.5\linewidth]{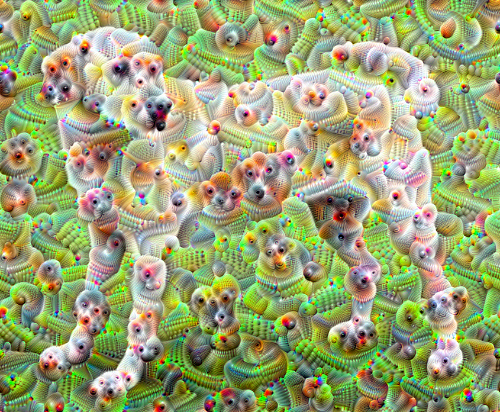}

}

\caption{Picture of a Labrador processed by DeepDream (\href{https://www.tensorflow.org/tutorials/generative/deepdream}{Google Colab})}\label{fig:DeepDream}
\end{figure}

In the following year, \citet{StyleTransfer} investigated methods to transfer the style of pictures. This method was used to transfer the style of Van Gogh's Sunflower painting to the LMU seal at the beginning of this chapter (see Figure \ref{fig:Logo}). Besides, below in Figure \ref{fig:StyleTransfer2} you can see the same Labrador picture from Figure \ref{fig:DeepDream} in \href{https://storage.googleapis.com/download.tensorflow.org/example_images/Vassily_Kandinsky\%2C_1913_-_Composition_7.jpg}{Kandinsky style}.

\begin{figure}

{\centering \includegraphics[width=0.5\linewidth]{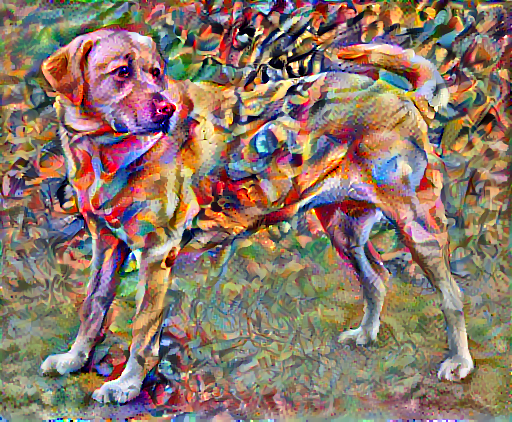}

}

\caption{Picture of a Labrador with Kandinsky style \href{https://www.tensorflow.org/tutorials/generative/style_transfer}{(Google Colab)}}\label{fig:StyleTransfer2}
\end{figure}

\begin{figure}

{\centering \includegraphics[width=0.5\linewidth]{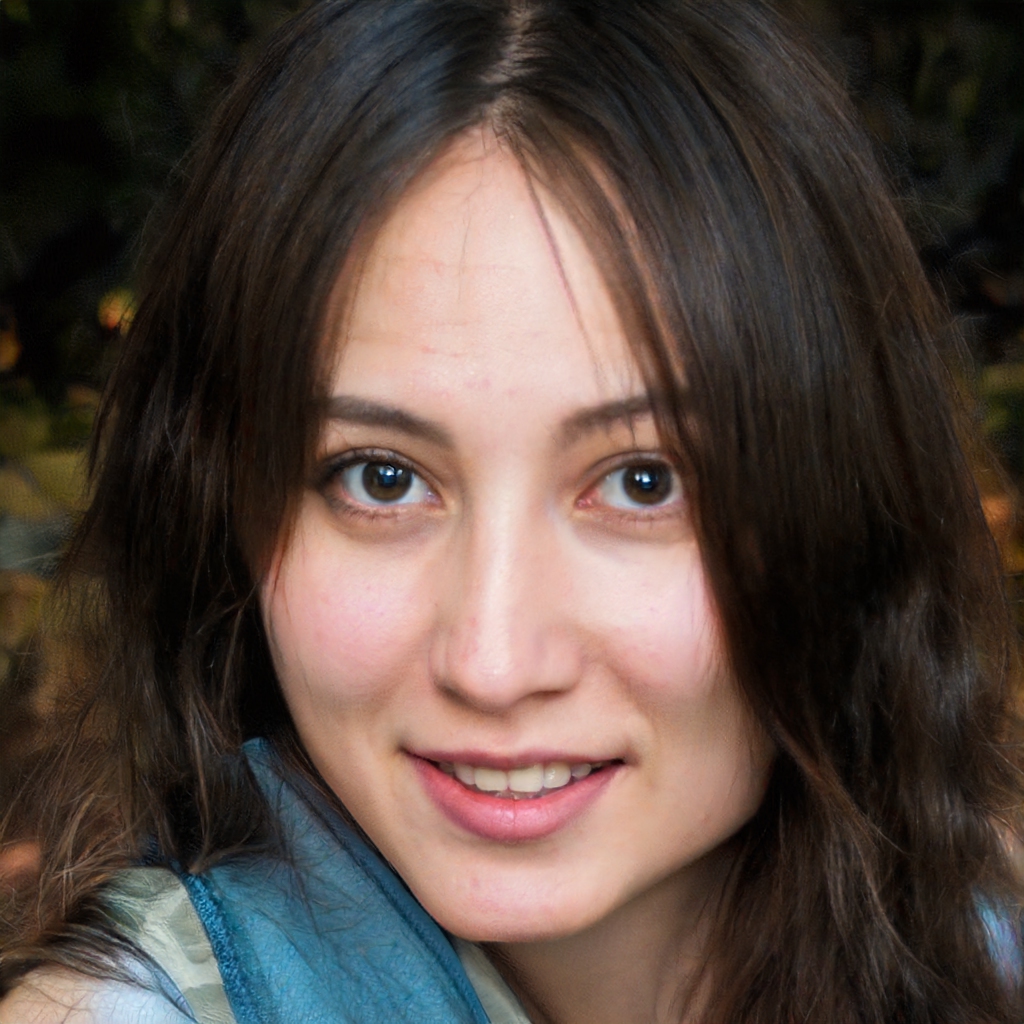}

}

\caption{Fake face generated by \href{https://thispersondoesnotexist.com/}{StyleGAN}}\label{fig:GAN}
\end{figure}

Furthermore, the architecture of Generative Adversarial Networks (GANs), which was first introduced by \citet{NIPS2014_5ca3e9b1}, was used by another research group \citet{karras2019style} to create very realistic fake images with their architecture StyleGAN. For instance, one can create pictures of people who do not exist, but look totally realistic (see Figure \ref{fig:GAN}).

Nevertheless, it was almost impossible to control the exact output of these early forms of AI art. There was no option to make specifications of how the result should look like in detail. For instance, you always get a human face with the earlier mentioned StyleGAN application, but you cannot specify to generate a blond girl with green eyes. This can be achieved by applying the artist-critic paradigm \citep{8477754}: Thereby, the computer as an artist generates a picture based on what the Neural Network learned in the training phase (e.g.~StyleGAN learns to generate pictures of human faces). Additionally, a critic is used to tell the computer if the output satisfies the concrete idea of the human artist. For this reason multimodal deep learning models emerged in the field of generative art. Here, one can control the output with the help of text prompting. In this way one can check if the generated picture matches the initial text description. Looking at the previous StyleGAN example, the multimodal architecture supervises whether the output picture is indeed a blond girl with green eyes or not. A new class of models for generating pictures evolved.

This idea was used by OpenAI for their models DALL-E \citep{DALLE} and CLIP \citep{CLIP} which were released in January 2021. Both architectures are critics for multimodal models. Only a few days after the release, Ryan Murdock combined CLIP (critic) with the already existing Neural Net ``BigGAN'' (artist) in his ``The Big Sleep'' software. Furthermore, \citet{StyleGAN} developed StyleCLIP, a combination of StyleGAN (artist) and CLIP (critic) to edit parts of images via text instructions. In the following months, Katherine Crowson combined CLIP as critic with the existing VQGAN algorithm as an artist. She also hooked up CLIP with guided diffusion models as artists to yield more fine-grained results. This approach was further investigated by OpenAI that published a paper \citep{DiffusionModels} in May 2021 about guided diffusion models. Moreover, in December 2021 they introduced GLIDE \citep{GLIDE}, a model with CLIP or classifier-free guidance as critics and diffusion models as artists. For more technical details about text2img methods like DALL-E and GLIDE refer to subsection \ref{c02-02-text2img}
or for text supporting CV models like CLIP at subsection \ref{c02-04-text-support-img}.

\hypertarget{how-to-use-these-models}{%
\subsection{How to use these models?}\label{how-to-use-these-models}}

A lot of different notebooks are publicly available to apply the different pre-trained models. In general, all notebooks work pretty similar: one only needs to enter a text prompt in the code and after running the notebook the computer generates a picture based on these instructions. It is relatively easy and no prior coding knowledge is required. Moreover, there are also some API and GUI applications (e.g.~\href{https://multimodal.art/mindseye}{MindsEye beta}) where no programming knowledge is needed at all. Using these models, it is important to think about how exactly once enters the respective text prompt. One can influence the output in a desired way with little changes in the short text instruction. This is also known as ``prompt engineering''. For instance, in the beginning of this chapter, we entered the prompt ``in the style of Van Gogh'' to change the style of the LMU seal. In this context, a special trick is to append ``unreal engine'' \citep{unrealEngine} which makes the resulting pictures more realistic with higher quality. This seems surprising at first, but the models were trained on data from the internet including pictures of the software company Epic Games that has a popular 3D video game engine called ``Unreal Engine''. This is one of the most popular prompting tricks.

\begin{figure}

{\centering \includegraphics[width=1\linewidth]{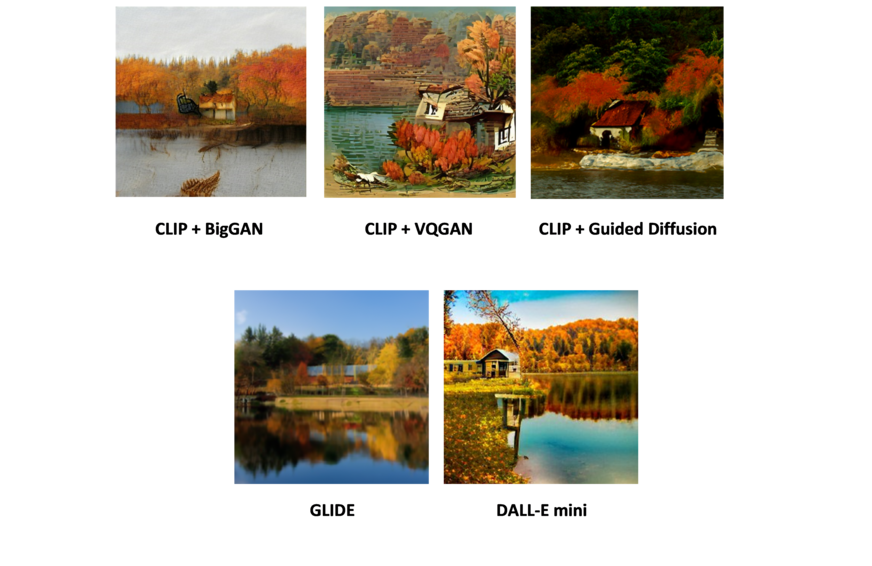}

}

\caption{Comparison of different models with prompt ``fall landscape with a small cottage next to a lake''}\label{fig:comparison1}
\end{figure}

\begin{figure}

{\centering \includegraphics[width=1\linewidth]{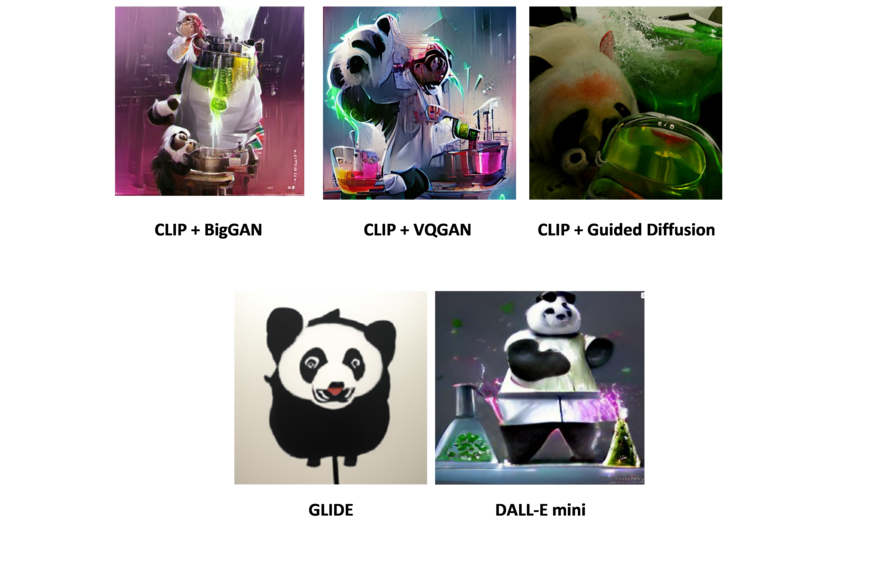}

}

\caption{Comparison of different models with prompt ``panda mad scientist mixing sparkling chemicals, artstation''}\label{fig:comparison2}
\end{figure}

Unfortunately, OpenAI has never released DALL-E. There is only an open-source version called ruDALL-E \citep{ruDALLE} that was trained on Russian language data. Besides, hugging face hosts DALL-E mini \citep{DALLEmini} where one can generate pictures, but does not have access to the model itself. PyTorch offers a replication of the DALL-E code \citep{DALLEpytorch} but no trained model. Furthermore, CLIP was released without publishing the used training data. However, there exists an open source data set with CLIP embeddings called LAION-400m \citep{LAION}. In the following, we used different publicly available notebooks to try out the different models \href{https://colab.research.google.com/drive/1NCceX2mbiKOSlAd_o7IU7nA9UskKN5WR?usp=sharing}{CLIP + BigGAN},
\href{https://colab.research.google.com/drive/1ZAus_gn2RhTZWzOWUpPERNC0Q8OhZRTZ\#scrollTo=FhhdWrSxQhwg}{CLIP + VQGAN},
\href{https://colab.research.google.com/drive/12a_Wrfi2_gwwAuN3VvMTwVMz9TfqctNj\#scrollTo=X5gODNAMEUCR}{CLIP + Guided Diffusion},
\href{https://colab.research.google.com/github/openai/glide-text2im/blob/main/notebooks/text2im.ipynb}{GLIDE}
with the text prompt \emph{``a fall landscape with a small cottage next to a lake''} (see Figure \ref{fig:comparison1}) and \emph{``panda mad scientist mixing sparkling chemicals, artstation''} (see Figure \ref{fig:comparison2}). The first prompt shows pretty realistic results, whereas the second prompt results in more different and ``crazy'' outputs. That is because the panda-prompt is more abstract than the first one and hence more difficult to illustrate. In addition, some of the notebooks run on lower resolution due to computational limitations. Besides, GLIDE is also downsized by the publisher: The released smaller model consists of 300 million parameters, whereas the unreleased model has about 3.5 billion parameters \citep{GLIDE}. So better results are possible with higher computational power and other implementations of the models.

\hypertarget{different-tasks-and-modalities}{%
\subsection{Different tasks and modalities}\label{different-tasks-and-modalities}}

So far, we concentrated on the two modalities text and image. Combining both of them, one can tackle different tasks with the models mentioned above. The main usage is to generate images based on a text prompt. Therefore, one can start from noise or but is also possible to chose a real image as starting point \citep{qiao2022initial}. This was done in the beginning with the LMU seal by CLIP + VQGAN (see Figure \ref{fig:Logo}): instead of starting from noise, the model started from the LMU seal as initialization and then used the prompt ``in style of Van Gogh''. The video captures how the model develops during fitting. In the end, the typical Van Gogh sunflowers emerge as well as what could be a part of Van Gogh's face.

\begin{figure}

{\centering \includegraphics[width=0.9\linewidth]{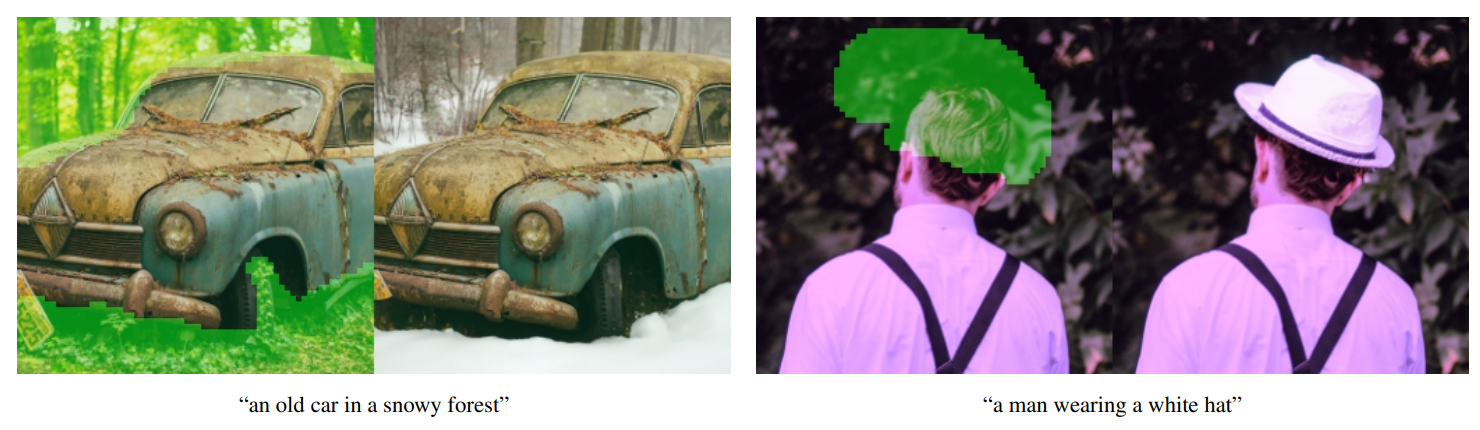}

}

\caption{Text-conditional image inpainting examples with GLIDE \citep{GLIDE}}\label{fig:inpainting}
\end{figure}

Furthermore, one can edit, extend, crop and search images with models like GLIDE \citep{GLIDE}. For instance, \citet{GLIDE} fine-tuend the model for text-conditional image inpainting (see figure \ref{fig:inpainting}). By marking some area in the pictures, here in green, and adding a text prompt, one can edit pictures very easily and precisely. This is quite impressive as the model needs to understand from the text prompt which object should be filled in and then do this in the correct style of the surrounding to produce a realistic outcome. Another idea is to use a sketch of a drawing and let the model fill in the details based on a text caption (see figure \ref{fig:sketch} below). This allows controlled changes of parts of pictures with relatively little effort. In this way, GLIDE can be used to generate pictures out of random noise, but also to edit pictures in a specific way. Furthermore, it is also possible to combine other modalities as well (see more details in subsection \ref{c03-01-further-modalities}). For instance, \citet{WZRD} accompanies custom videos with suitable audio. It is even imaginable to create sculptures with 3D-printers \citep{3D}.

\begin{figure}

{\centering \includegraphics[width=0.9\linewidth]{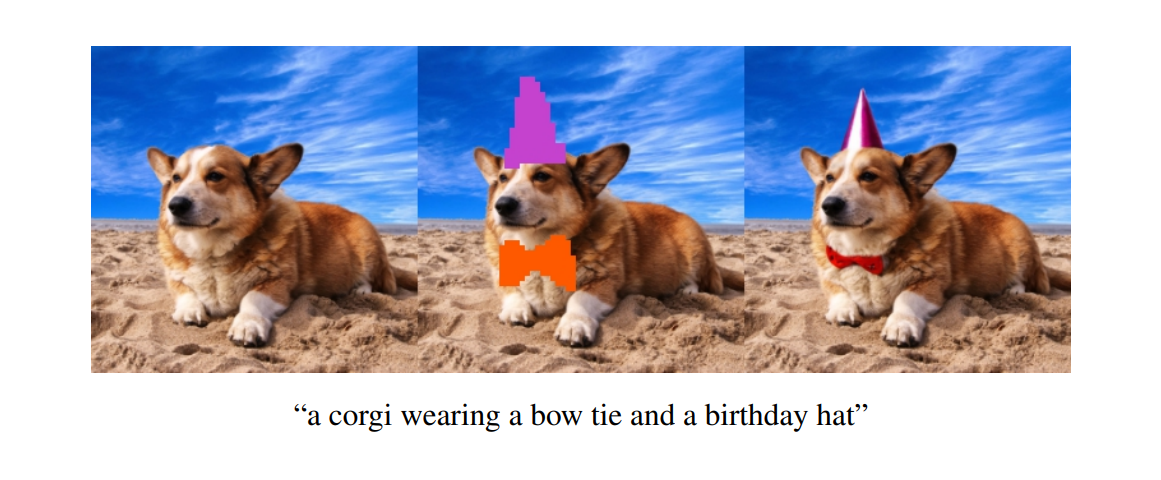}

}

\caption{Text-conditional edit from user scratch with GLIDE \citep{GLIDE}}\label{fig:sketch}
\end{figure}

\hypertarget{discussion-and-prospects}{%
\subsection{Discussion and prospects}\label{discussion-and-prospects}}

In the last years, methods to generate images via text prompting improved tremendously and a new field of art arised. It is surprising how these models are able to create images only based on a short text instruction. This is quite impressive as AI achieved some level of creativity. It is up for discussion to which extent the computer is becoming the artist in generative arts and in this way replacing the human artist. However, there is still no direct loss function that can calculate how aesthetically pleasing a picture is \citep{bias}. This is probably also quite subjective and cannot be answered for everyone in the same way. Most of the time the computer works as aid for the creative process by generating multiple images. Then, the human artist can pick the best outcome or vary the text prompt to improve the output in a desired way. However, the better the AI becomes, the less the human artist needs to intervene in this process.

Furthermore, as the output becomes more and more realistic, there is the risk that these methods are abused to facilitate plagiarism or create fake content and spread misleading information \citep{misconduct}. After all, the outputs look totally realistic, but are completely made-up and generated by the computer. For this reason, some organisations like Open-AI do not release all their models (e.g.~DALL-E) or downstream models (e.g.~CLIP). On the other hand, from a scientific point of view, it is important to get access to such models to continue research.

Moreover, similarly to most Deep Learning algorithms, these models are affected by biases in the input data \citep{bias_ML}. For instance, \citet{bias} points out that CLIP text embeddings associate a human being more with a man than with a woman. In this way it might be more likely that our models generate a man with the text prompt ``human being'' than a woman. This effect needs to be further investigated and should be removed.

After all, generative arts can be used to create Non Fungible Tokens (NFT) relatively easily. NFTs are digital artworks where a special digital signature is added making them unique and in this way non-fungible \citep{NFT}. The digital artwork is bought and sold online, often by means of cryptocurrency. That is why this field is also called Cryptoart. This provides the perfect platform to sell generative arts. However, this trading market is quite new and controversial, similar to crypotcurrency trading in general.

In conclusion, generative arts is a new and impressive field. It combines technology with arts, two rather opposite fields. The methods are already really impressive and are still getting better and better. For instance, this year Open AI already published DALLE-2 \citep{DALLE2} that outperforms DALLE-1. It remains highly interesting to follow up with the developments in this field.

\hypertarget{conclusion-1}{%
\chapter{Conclusion}\label{conclusion-1}}

\emph{Author: Nadja Sauter}

\emph{Supervisor: Matthias Aßenmacher}

It is very impressive how multimodal architectures have developed, especially over the course of the last two years. Particularly, methods to generate pictures based on text prompts, like DALL-E, became incredibly good at their ``job''. A lot of people are fascinated by the stunning results and a huge hype about these AI generated images evolved in the internet, especially on twitter. In this way, the models were not only investigated by researchers but also by the online community (e.g.~Katherine Crowson alias \href{https://twitter.com/RiversHaveWings}{Rivers Have Wings}). Even in the art scene these methods attracted a lot of attention as shown in our use case ``enerative Arts'' (subsection \ref{c03-04-usecase}). Apart from that, it is possible to deploy these methods commercially, for instance in the film production or gaming industry (e.g.~creating characters for games). However, this might also result in problems of copyright, an issue which has not yet been dealt with until now.

It is also impressive how realistic and precise outputs are achieved by such architectures. On the other hand, these methods can also be abused to spread misleading information as it is often very difficult to distinguish between a fake or a real picture by only looking at it. This can be systematically used to manipulate the public opinion by spreading AI manipulated media, also called deep fakes. That's why researchers like \citet{explainaility} demand automated tools which are capable of detecting these fabrications. Apart from that, like most deep learning models, multimodal architectures are not free from bias which also needs to be investigated further \citep{bias}. Besides, the algorithms are very complex which is why they are often called ``black-box'' models, meaning that one cannot directly retrace how the model came to a certain solution or decision. This may limit their social acceptance and usability as the underlying process is not credible and transparent enough \citep{explainaility}. For instance, in medical applications like e.g.~predicting the presence or absence of cancer, apart from the decision of the AI the reasoning and the certainty are highly relevant for doctors and patients.

Furthermore, there is a clear trend in recent years to build more and more complex architectures in order to achieve higher performance. For instance OpenAI's language model GPT-2 has had about 1.5 billion parameters \citep{Radford2019LanguageMA}, whereas its successor GPT-3 had about 175 billion parameters \citep{brown2020language}. Increasing the number of parameters often helps improving model performance, but all of these parameters need to be trained and stored which takes a lot of time, enormous computational power and storage. For example, training GPT-2 took about one week (168 hours) of training time on 32 TPUv3 chips \citep{environment}. The researchers (\citet{environment}) estimated that the cloud compute costs for training GPT-2 added up to about \$12,902--\$43,008. Apart from the enormous expenses, this also contributes to our environmental burden as this process is really energy intensive. Due to missing power draw data on GPT-2's training hardware, the researchers weren't able to calculate the CO\textsubscript{2} emission. However, for the popular BERT architecture with 110M parameters they calculated cloud compute costs of \$3,751-\$12,571, energy consumption of 1,507 kWh and a Carbon footprint of 1,438 lbs of CO\textsubscript{2}. In comparison, the footprint of flying from New York to San Francisco by plane for one passenger is about 1,984 lbs of CO\textsubscript{2}. In conclusion training BERT once results in almost the same footprint as this long-haul flight. On top of this, these numbers are only for one training run. Developing a new model or adapting it often takes several fitting and tuning phases.

Moreover, the computational power as well as the necessary hardware, technology and financial means to run these models can oftentimes only be provided by big technology companies like e.g.~Google, Facebook or OpenAI. This results in a disparate access between researchers in academia versus industry. Furthermore, the companies sometimes tend do not publishing their (best) models as they are their ``product'' and contribute to the company's intellectual property. In this way it is not possible to reproduce their work and findings independently. Besides, from an economic point of view, this may be the foundation of a monopoly which might be dangerous for economic competition and holds the possibility of abuse.

\hypertarget{epilogue}{%
\chapter{Epilogue}\label{epilogue}}

\emph{Author: Matthias Aßenmacher}

Since this project was realized in a limited time frame and accounted for about one third
of the ECTS points which should be achieved during one semester, it is obvious that this
booklet cannot provide exhaustive coverage of the vast research field of \emph{Multimodal Deep Learning}.

Furthermore this area of research is moving very rapidly at the moment, which means that
certain architectures, improvements or ideas had net yet even been published when we sat down
and came up with the chapter topics in February 2022. Yet, as you might have seen, in some cases the
students were even able to incorporate ongoing research published over the course of the seminar.
Thus, this epilogue tries to put the content of this booklet into context and relate it to what is
currently happening. Thereby we will focus on two aspects:

\begin{itemize}
\tightlist
\item
  New influential (or even state-of-the-art) architectures
\item
  Extending existing architectures to videos (instead of ``only'' images)
\end{itemize}

\hypertarget{new-influential-architectures}{%
\section{New influential architectures}\label{new-influential-architectures}}

In \href{./c02-00-multimodal.html\#c02-02-text2img}{\textbf{Chapter 3.2: ``Text2Image''}} and \href{./c03-00-further.html\#c03-04-usecase}{\textbf{Chapter 4.4: ``Generative Art}} some important models for generating images/art from free-text prompts have been presented. However, one example of an even better (at least perceived this way by many people) generative model was just published by researchers from Björn Ommer's group at LMU: \emph{``High-Resolution Image Synthesis with Latent Diffusion Models''}\\
They introduced a model called \emph{Stable Diffusion} which allows users to generate photorealisitic images. Further (as opposed to numerous other architectures, it is available open-source and can even be tried out via \href{https://huggingface.co/spaces/stabilityai/stable-diffusion}{huggingface}.

\hypertarget{creating-videos}{%
\section{Creating videos}\label{creating-videos}}

Also more recently, research has focussed on not only creating images from natural language input but also videos. The \emph{Imagen} architecture, which was developed by researchers at Google Research (Brain Team), was extended with respect to also creating videos (see \href{https://imagen.research.google/video/}{their project homepage}).
Yet, this is only on of many possible examples of research being conducted in this direction. The interested reader is refered to \href{https://imagen.research.google/video/paper.pdf}{the paper accompanying their project}.

We hope that this little outlook can adequately round off this nice piece of academic work created by extremely motivated students and we hope that you enjoyed reading.

\hypertarget{acknowledgements}{%
\chapter{Acknowledgements}\label{acknowledgements}}

The most important contributions are from the students themselves.
The success of such projects highly depends on the students.
And this book is a success, so thanks a lot to all the authors!
The other important role is the supervisor.
Thanks to all the supervisors who participated!
Special thanks to \href{https://www.misoda.statistik.uni-muenchen.de/personen/professoren/heumann/index.html}{Christian Heumann} and \href{https://www.statistik.uni-muenchen.de/personen/professoren/bischl/index.html}{Bernd Bischl} who enabled us to conduct the seminar in such an experimental way, supported us and gave valuable feedback for the seminar structure.
Thanks a lot as well to the entire \href{https://www.statistik.uni-muenchen.de/}{Department of Statistics} and the \href{http://www.en.uni-muenchen.de/index.html}{LMU Munich} for the infrastructure.

The authors of this work take full responsibilities for its content.

  \bibliography{book.bib,packages.bib}

\backmatter
\printindex

\end{document}